\documentclass[sigconf, screen, nonacm]{acmart}

\acmBooktitle{}
\acmISBN{}
\acmDOI{}

\newcommand\blfootnote[1]{%
  \begingroup
  \renewcommand\thefootnote{}\footnote{#1}%
  \addtocounter{footnote}{-1}%
  \endgroup
}

\usepackage{soul}
\usepackage[utf8]{inputenc}
\usepackage{graphicx}
\usepackage{booktabs}

\usepackage{graphicx}
\usepackage{booktabs} 

\usepackage[ruled,noend]{algorithm2e}

\SetCommentSty{mycommfont}

\usepackage{here}

\usepackage{amsmath,amssymb,amsfonts,amsbsy,amsfonts,latexsym}
\usepackage{multirow}
\usepackage{makecell}
\usepackage[labelfont=bf,textfont=it,belowskip=0pt,aboveskip=5pt,tableposition=top]{caption}
\usepackage{xcolor}
\usepackage{colortbl}

\definecolor{colorA}{RGB}{189,201,225}
\definecolor{colorB}{RGB}{103,169,207}
\definecolor{colorC}{RGB}{ 28,144,153}
\definecolor{colorD}{RGB}{  1,108, 89}

\newcolumntype{R}{>{\columncolor{gray!40}}r}
\newcolumntype{L}{>{\columncolor{gray!40}}l}
\newcolumntype{C}{>{\columncolor{gray!40}}c}

\usepackage{tabularx,colortbl,xcolor}
\usepackage{multirow}
\usepackage[normalem]{ulem}
\useunder{\uline}{\ul}{}

\usepackage{enumitem}

\usepackage{xparse}

\DeclareGraphicsExtensions{.pdf,.png}

\SetKwInput{KwInput}{Input}

\usepackage{longtable}
\usepackage{pgfplots}
\usepackage{outlines}
\usepackage[normalem]{ulem}

\usepackage[many]{tcolorbox}
\usepackage{caption}
\usepackage{subcaption}
\usepackage{nopageno}
\usepackage{graphbox} 

\tcbset{
    sharp corners,
    colback = white,
    before skip = 0.2cm,    
    after skip = 0.5cm      
}                           

\definecolor{main}{HTML}{4472C4}    
\definecolor{sub}{HTML}{EBF4FF}     

\newtcolorbox{boxA}{
    enhanced, breakable,
    boxrule = 0pt,
    colback = sub,
    borderline west = {2pt}{0pt}{main}, 
    borderline east = {2pt}{0pt}{main}, 
}

\author{Sehoon Kim$^*$}
\email{sehoonkim@berkeley.edu}
\affiliation{
  \institution{UC Berkeley}
  \city{}
  \state{}
  \country{}
}
\author{Coleman Hooper$^*$}
\email{chooper@berkeley.edu}
\affiliation{
  \institution{UC Berkeley}
  \city{}
  \state{}
  \country{}
}
\author{Thanakul Wattanawong}
\email{j.wat@berkeley.edu}
\affiliation{
  \institution{UC Berkeley}
  \city{}
  \state{}
  \country{}
}
\author{Minwoo Kang}
\email{minwoo_kang@berkeley.edu}
\affiliation{
  \institution{UC Berkeley}
  \city{}
  \state{}
  \country{}
}
\author{Ruohan Yan}
\email{yrh@berkeley.edu}
\affiliation{
  \institution{UC Berkeley}
  \city{}
  \state{}
  \country{}
}
\author{Hasan Genc}
\email{hngenc@berkeley.edu}
\affiliation{
  \institution{UC Berkeley}
  \city{}
  \state{}
  \country{}
}
\author{Grace Dinh}
\email{dinh@berkeley.edu}
\affiliation{
  \institution{UC Berkeley}
  \city{}
  \state{}
  \country{}
}
\author{Qijing Huang}
\email{jennyhuang@nvidia.com}
\affiliation{
  \institution{NVIDIA}
  \city{}
  \state{}
  \country{}
}
\author{Kurt Keutzer}
\email{keutzer@berkeley.edu}
\affiliation{
  \institution{UC Berkeley}
  \city{}
  \state{}
  \country{}
}

\author{Michael W. Mahoney}
\email{mmahoney@stat.berkeley.edu}
\affiliation{
  \institution{ICSI, LBNL, UC Berkeley}
  \city{}
  \state{}
  \country{}
}

\author{Yakun Sophia Shao}
\email{ysshao@berkeley.edu}
\affiliation{
  \institution{UC Berkeley}
  \city{}
  \state{}
  \country{}
}

\author{Amir Gholami}
\email{amirgh@berkeley.edu}
\affiliation{
  \institution{ICSI, UC Berkeley}
  \city{}
  \state{}
  \country{}
}

\begin{document}

\title{
Full Stack Optimization of Transformer Inference: a Survey
}

\begin{abstract}
\vspace{2mm}
Recent advances in state-of-the-art neural network architecture design have been moving toward Transformer models.
These models achieve superior accuracy across a wide range of applications in computer vision, natural language processing, and speech recognition.
This trend has been consistent over the past several years since Transformer models were originally introduced.
However, the amount of compute and bandwidth required for inference of recent Transformer models is growing
at a significant rate, and this has made their deployment in latency-sensitive applications challenging.
As such, there has been an increased focus on making Transformer models more efficient, with methods that range from changing the architecture design, all the way to developing dedicated domain-specific accelerators.
In this work, we survey different approaches for efficient Transformer inference, including:
(i) analysis and profiling of the bottlenecks in existing
Transformer architectures and their similarities and differences with
previous convolutional models; 
(ii) implications of Transformer architecture on hardware, including the impact of 
non-linear operations such as Layer Normalization, Softmax, and GELU,
as well as linear operations, 
on hardware design;
(iii) approaches for optimizing a fixed Transformer architecture;
(iv) challenges in finding the right mapping and scheduling
of operations for Transformer models; and
(v) approaches for optimizing Transformer models by adapting the architecture using neural architecture search.
Finally, we perform a case study by applying the surveyed optimizations on  Gemmini, the open-source, full-stack deep neural network accelerator generator, and we show how each of these approaches can yield improvements, compared to previous benchmark results on Gemmini. 
Among other things, we find that a full-stack co-design approach 
with the aforementioned methods
can result in up to 88.7$\times$ speedup with a minimal performance degradation for Transformer inference.
\blfootnote{$^{*}$Equal contribution.}
\end{abstract}

\maketitle

\thispagestyle{empty}

\section{Introduction}

Deep learning models have scaled up to billions of parameters and billions of Multiply-Accumulate (MAC) operations during both training and inference. 
As a result, there has been a growing interest in computing these models efficiently and in deploying these compute and memory-intensive workloads on resource-constrained edge devices. These edge devices have tight energy and memory constraints, and the corresponding applications that leverage deep learning models also often have real-time latency constraints.

CPUs and GPUs are both commonly used in general-performance computing platforms, and they have the advantage of being both ubiquitous and capable of supporting a wide variety of workloads and operations. However, this flexibility comes at a cost of reduced efficiency. Deep learning models are composed of a small number of distinct operations that are repeated millions or billions of times, and therefore they often do not require a high level of flexibility. 
Additionally, while modern CPUs and GPUs can perform several operations in parallel, they lack the ability to leverage the massive data reuse opportunities in deep learning models.

The combination of a need for fast, efficient computation, the use of a small number of distinct operations, and the opportunities for data reuse have all led to the use of hardware accelerators for deep learning. 
A multitude of enterprise deep learning accelerators, such as~\cite{Jouppi2017, nvdla-hotchips, tpu_edge, npu, hruska2017movidius, liao2019davinci, talpes2020compute, knowles2021graphcore, prabhakar2021sambanova, abts2022groq, lie2022cerebras}, have been developed and integrated into commodity hardware by industry in the past decade.
This parallels many research accelerators developed in academia~\cite{diannao, dadiannao, shidianno, eie, eyeriss-isca2016, chen2019eyeriss, tetris-asplos17, pei2019tianjic, gemmini-dac}. 
Together with hardware accelerator development, the software frameworks~\cite{jia2014caffe, abadi2016tensorflow, paszke2019pytorch, chen2015mxnet} and compilers~\cite{chen2018tvm, tensorrt, sabne2020xla} for deploying various deep learning algorithms have also enhanced and matured. These tools enable the execution of deep learning algorithms on accelerators, and they perform mapping optimizations to improve the performance and efficiency of the full deep learning pipeline.
Nonetheless, the fast-evolving deep learning algorithms still keep introducing new demands for hardware and software support, as well as their co-optimization, to satisfy various deployment constraints.

The recent rise in popularity of Transformers and large language models~\cite{devlin2018bert,radford2018improving,radford2019language,brown2020language, scao2022bloom,du2022glam,rae2021scaling,smith2022using,hoffmann2022training,chowdhery2022palm,raffel2019exploring} 
for solving various
natural language processing (NLP) tasks presents a brand new set of challenges 
in the design of accelerators as well as frameworks. 
There has also been an increased focus on making Transformer inference more efficient, especially due to their growing size and run-time complexity. 
However, there is still a lack of understanding regarding the workload characteristics of Transformer architectures, and thus of the design principles necessary for effectively running these models, when compared to the more well-known convolutional neural network (CNN) architectures.
For instance, compared to the conventional CNN-focused design, Transformers are mostly composed of matrix multiplications (matmuls) together with memory-intensive nonlinear operations.
In addition, the computational graph and dataflow of Transformer models are more complex than that of CNNs, with more types of operation nodes, as well as more dataflow splits and concatenations.
All these challenges require us to undertake a comprehensive analysis of the current hardware and software solutions as well as the various design trade-offs for Transformer inference.
Performing such an analysis will enable us to build a holistic and comprehensive understanding of the requirements for efficiently running Transformers.

The contribution of this work is two-fold: (1) to analyze the run-time characteristics of Transformers and to survey different approaches for efficient Transformer inference; and (2) to perform a case study by applying the surveyed methodologies on Gemmini~\cite{gemmini-dac}, the full-stack deep neural network (DNN) accelerator generator. 
The longer-term goal of this work is to characterize different factors across the hardware and software stack in order to optimize Transformer inference.

Regarding our first contribution, this paper contains a survey and analysis covering different hierarchies in end-to-end deep learning inference, with a particular focus on Transformers. 
This includes:
\begin{itemize}[leftmargin=5mm]
    \item Analysis and profiling of the runtime characteristics and bottlenecks of the Transformer architecture (Sec.~\ref{sec:architecture_and_performance}).
    \item Hardware architectures for Transformer inference, including the impact of the non-linear operations of the Transformer architecture on their design  (Sec~\ref{sec:hardware_design}).
    \item Optimization strategies such as pruning and quantization for further improving the performance of a fixed Transformer architecture (Sec~\ref{sec:optimization}).
    \item Mapping and scheduling of operations in the Transformer architecture and the associated challenges  (Sec.~\ref{sec:scheduling}).
    \item Designing and adapting Transformer architectures to be more hardware efficient through an automated neural architecture search process (Sec.~\ref{sec:adapting_nas}).
\end{itemize}

Regarding our second contribution, our case study of applying the surveyed methodologies on deploying Transformers yields several key findings, including the following:

\begin{itemize}[leftmargin=5mm]

\item 
Gemmini, which was originally designed for CNN workloads, does not yield hardware accelerator architectures that are well-suited for Transformer inference. 
The primary bottleneck for running Transformers on CNN domain-specific accelerators is not necessarily linear operations, but rather it is the time spent on floating-point non-linear operations, as well as quantization and dequantization operations. 
Unless those operations are addressed properly, this can result in less than 1\% hardware utilization
(Sec.~\ref{sec:thiswork} and Fig.~\ref{fig:bert-dac-pie}).
 
\item  
For Transformer accelerators, it is often better to have a larger accumulator size and smaller scratchpad size, while the opposite is often more optimal for CNN accelerators. Changing accelerator architecture to incorporate this observation can result in a 36\% latency improvement over the baseline optimized for CNN benchmarks (Sec.~\ref{sec:hardware:memory-hierarchy}).

\item  
Despite the fact that scheduling matmuls in Transformers only requires 3 loops, as compared to 6 for convolutions in CNNs, 
we found that it is as challenging to find performant schedules for Transformers as it is for CNNs.
The selection of appropriate scheduling decisions for Transformers involves a large number of decisions, with the best and worst solutions exhibiting performance differences of up to four orders of magnitude
(Sec.~\ref{subsec:mapspace-transformers} and 
Fig.~\ref{fig:bert_mapspace}, \ref{fig:bert_resnet_cdf}, \ref{fig:bert_synthetic}).

\item  
Fusing Batch Normalization with the neighboring convolution in CNN models is straightforward. 
However, when fusing Layer Normalization with the preceding matmul in the Transformer architecture, constraints are imposed on the mapping, particularly related to tile sizes.
This requires further consideration since the runtime cost due to the mapping constraints could outweigh the gains from operation fusion in certain circumstances (Sec.~\ref{subsec:scheduling_complexity_nonlinear} and 
Fig.~\ref{fig:mha_fusion}, \ref{fig:ffn_fusion}).

\end{itemize}

\vspace{3mm}
\section{Transformer Model Architecture and Performance Bottlenecks}
\label{sec:architecture_and_performance}

In this section, we start with a high level introduction to the
building blocks of the Transformer architecture.
We first discuss the multi-head attention and feed-forward modules, the non-linear operations
used in Transformers, and the difference between Encoder/Decoder
models, in Sec.~\ref{subsec:transforemr_high_level_overview}. 
We then analyze the impact of these different blocks on hardware performance using arithmetic, as well as the analytical modeling and direct profiling of each component, in Sec.~\ref{subsec:model-analysis}.

\subsection{High-Level Overview of Transformer Architecture}
\label{subsec:transforemr_high_level_overview}

\begin{figure*}[t]
    \centering
    \includegraphics[width=0.52\linewidth,align=c]{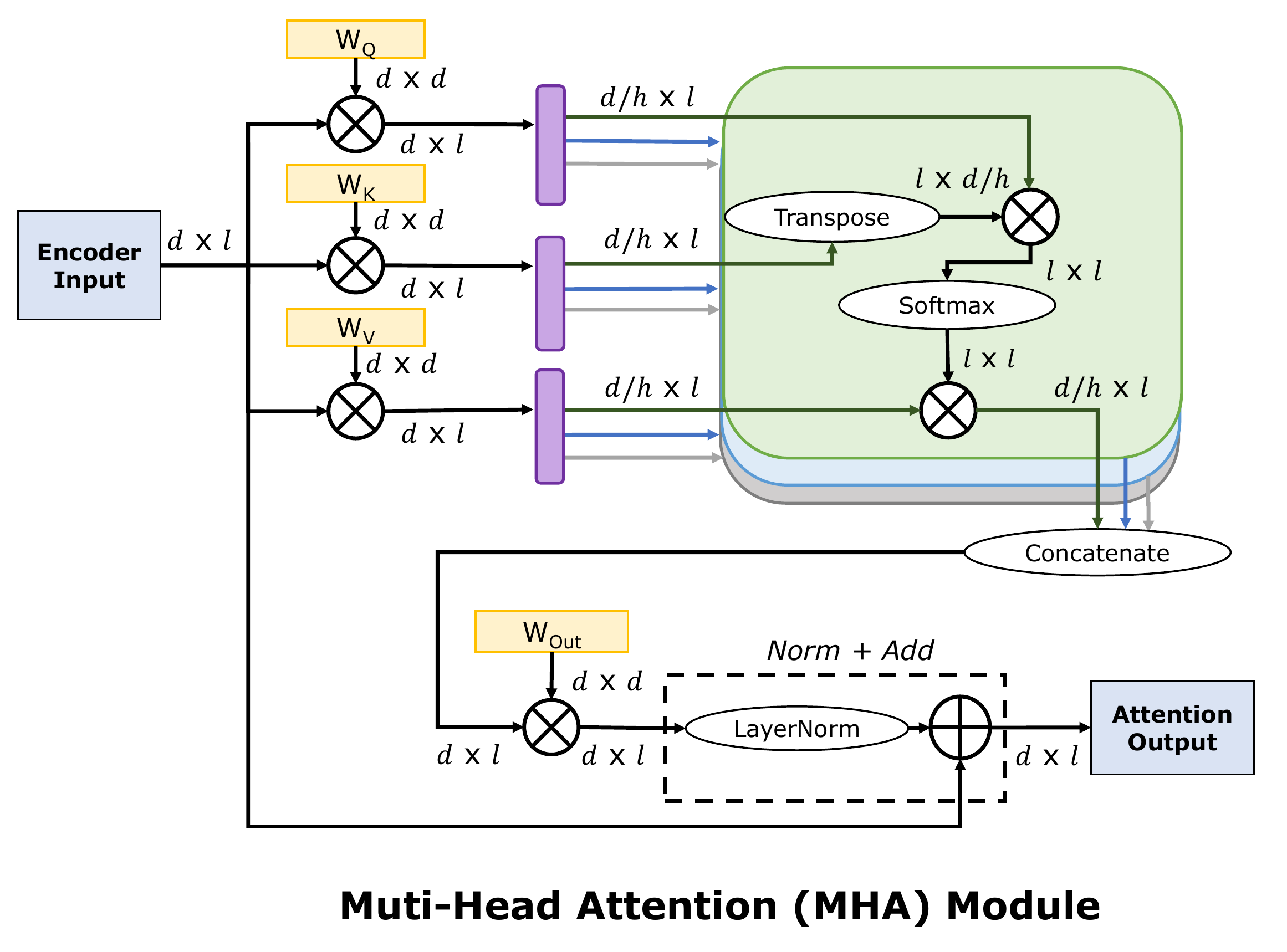}
    \includegraphics[width=0.47\linewidth,align=c]{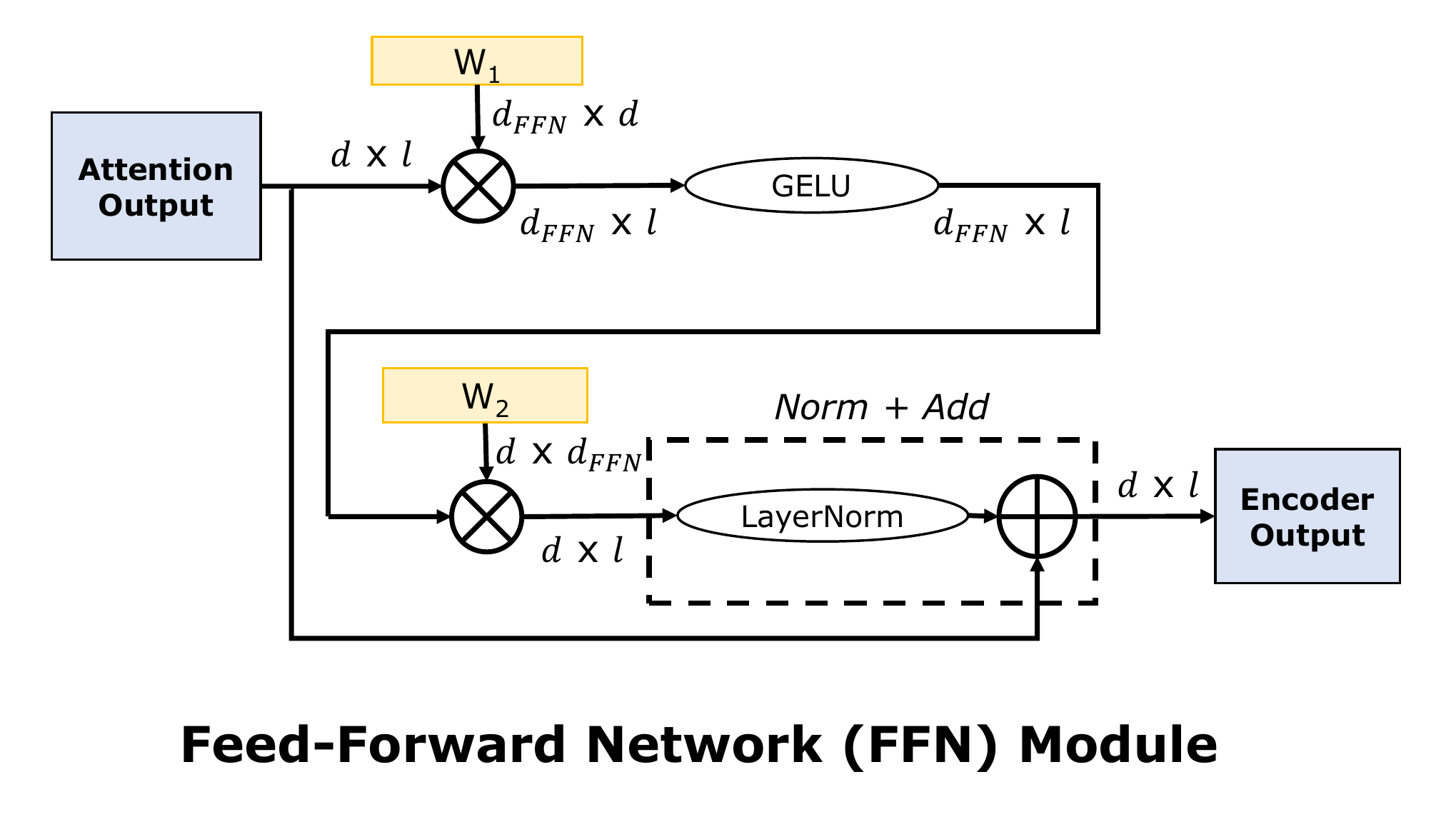}

    \caption{Map of the computations performed in (Left) the multi-head attention (MHA) module and (Right) the feed-forward network (FFN) module in the Transformer encoder block.
    }
    \label{fig:comp-map}
\end{figure*}

A Transformer architecture~\cite{vaswani2017attention} typically consists of multiple Transformer blocks, each of which includes a multi-head attention (MHA) module and a feed-forward (FFN) module, and each of which is followed by a Layer Normalization (LayerNorm) operation and a residual connection.
The detailed computations of MHA and FFN are illustrated in Fig.~\ref{fig:comp-map}, and the configuration parameters for Transformer architectures (along with the values used by BERT-Base and BERT-Large) are provided in Tab.~\ref{table:symbols}.
An input sequence to the Transformer block is composed of $l$ tokens, each represented by a vector of $d$ dimension, forming a $d \times l$ matrix.
A token is a segment of an input sequence. For example, when the input is a sentence, a token may be a word or a sentence fragment.

The MHA module (see Fig.~\ref{fig:comp-map}, Left) first projects this sequence by multiplying it with three different weight matrices: $W_Q$, $W_K$, and $W_V$ (the so-called query, key, and value matrices). 
This yields three different activations, namely the query, key, and value activations. 
The query, key, and value activations are then split into $h$ chunks, with each chunk having a hidden dimension of $d/h$.
These chunks are then forwarded to $h$ different attention heads, where the query and key chunks are multiplied along the hidden dimension, generating an activation matrix of size $l \times l$. 
This activation matrix is then passed through the Softmax operation (the output of which is often referred to as an attention score) and multiplied with the value chunk, resulting in an activation of hidden dimension $d/h$. 
Subsequently, all of the activations from the attention heads are concatenated along the hidden dimension to generate a single activation of hidden dimension $d$, which is then projected into the same dimension by the last linear layer with the weight matrix $W_{\text{out}}$. 
Finally, the output from the last linear layer in the MHA module is passed through the LayerNorm operator before being added to a residual connection to get the MHA module output.

In summary, an MHA module consists of six linear operations, 
four of which are identical weight-to-activation matmuls (i.e., the $W_Q$, $W_K$, $W_V$ and $W_{\text{out}}$ projections), 
and the remaining two of which are activation-to-activation matmuls (i.e., query $\times$ key and attention score $\times$ value). 
Throughout this paper, we refer to the first type of matmuls as \textit{projections} and the second type of matmuls as \textit{activation-to-activation matmuls} (act-to-act matmuls for short), as they have different run-time behaviors. 

\begin{table}[!t]
\caption{
Configuration parameters for Transformer architectures.
Parameters for BERT-Base, BERT-Large, and GPT-2 (smallest) are given as examples. 
Note that GPT-2 has the same parameters as BERT-Base. 
Sequence length can be any number, as long as it doesn't exceed the maximum possible sequence length.
}
\begin{center}
\small{
\begin{tabular}{c|c|ccc}
\toprule
Symbol & Parameter & BERT-Base & BERT-Large & GPT-2 \\
\midrule
$N$ & \# Layers & 12 & 24 & 12\\
$d$ & Model dimension & 768 & 1024 & 768 \\
$h$ & \# Attention Heads & 12 & 16 & 12 \\
$d_{\text{FFN}}$ & FFN dimension & 3072 & 4096 & 3072 \\
$l$ & Sequence length & - & - & -\\
\bottomrule
\end{tabular}
}
\end{center}
\label{table:symbols}
\vspace{-3mm}
\end{table}

The FFN module (see Fig.~\ref{fig:comp-map}, RIght) is a relatively simple block consisting of two linear layers. 
The input sequence is first projected from the hidden dimension $d$ to a higher FFN dimension $d_{\text{FFN}}$ via the first linear layer with the weight matrix $W_1$. 
Subsequently, the projected sequence is projected back to the original dimension $d$ through the second linear layer with the weight matrix $W_2$. 
Generally, the dimension $d_{\text{FFN}}$ is chosen to be 4$\times$ larger than $d$, resulting in the 4:1 aspect rate of $W_1$ and $W_2$ (e.g., in BERT-Base~\cite{devlin2018bert}). 
In between these two linear layers is a non-linear layer. Typically, GELU~\cite{hendrycks2016gaussian} is used for this~\cite{devlin2018bert,liu2019roberta,radford2018improving,radford2019language,brown2020language}.
Tab.~\ref{table:architecture_dims} summarizes all types of linear layers in a Transformer~block in both MHA and FFN modules.

\begin{table}[!t]
\caption{
Linear operations in Transformer models.
The last column is the matrix multiplication dimensions, i.e., $m \times n \times k$ means the input dimensions of $m \times n$ and $n \times k$, and the output dimension of $m \times k$.
Note that  act-to-act matmuls are both repeated $h$ times in the multi-headed scheme.
The entire computation graphs of MHA and FFN are illustrated in detail in Fig.~\ref{fig:comp-map}.
}
\begin{center}
\small{
\begin{tabular}{c|c|c}
\toprule
Module & operation & matmul dim \\
\midrule
MHA & $W_Q$ projection & $d \times d \times l$ \\
    & $W_K$ projection & $d \times d \times l$ \\
    & $W_V$ projection & $d \times d \times l$ \\
    & query $\times$ key & $l \times d/h \times l$ \\
    & attn. score $\times$ value & $d/h \times l \times l$ \\
    & $W_{\text{out}}$ projection & $d \times d \times l$ \\
\midrule
FFN & $W_1$ projection & $d_{\text{FFN}} \times d \times l$  \\
    & $W_2$ projection & $d \times d_{\text{FFN}} \times l$  \\

\bottomrule
\end{tabular}
}
\end{center}
\label{table:architecture_dims}
\end{table}

\begin{figure*}[!h]
  \centering
  \includegraphics[width=2.1\columnwidth]{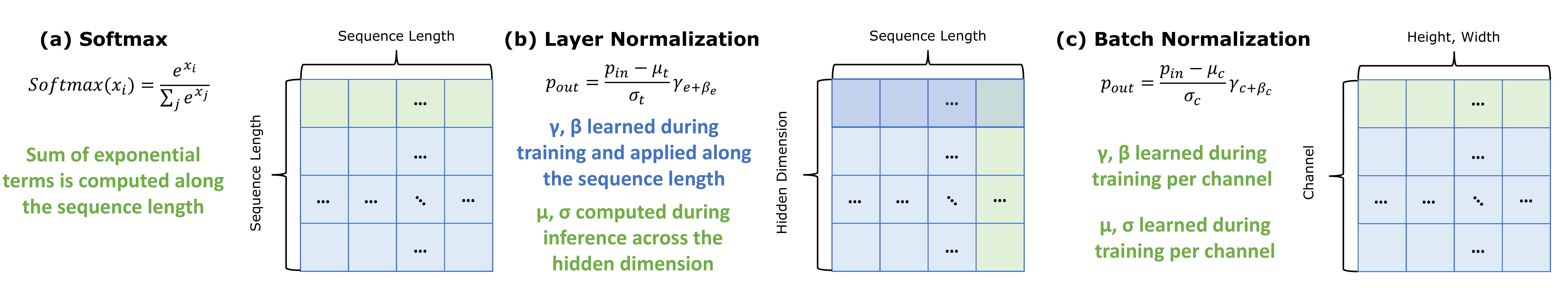}
  \caption{Diagrams outlining the Softmax, LayerNorm, and BatchNorm operations. 
  Since they rely on runtime statistics, LayerNorm and Softmax both require multiple passes over the input in order to compute the nonlinear operation. 
  In the case of Softmax, a first pass over the inputs is required to compute the denominator. For LayerNorm, three passes are required over the inputs: 
  one to compute the mean; 
  one to compute the standard deviation; and 
  one to apply the normalization. 
  Unlike LayerNorm and Softmax, BatchNorm only uses statistics which are learned during training, and therefore it only requires one pass over the inputs. 
  }
      \label{fig:nonlinear-ops}
\end{figure*}

\begin{figure*}[t!]
  \centering
  \includegraphics[width=2.1\columnwidth]{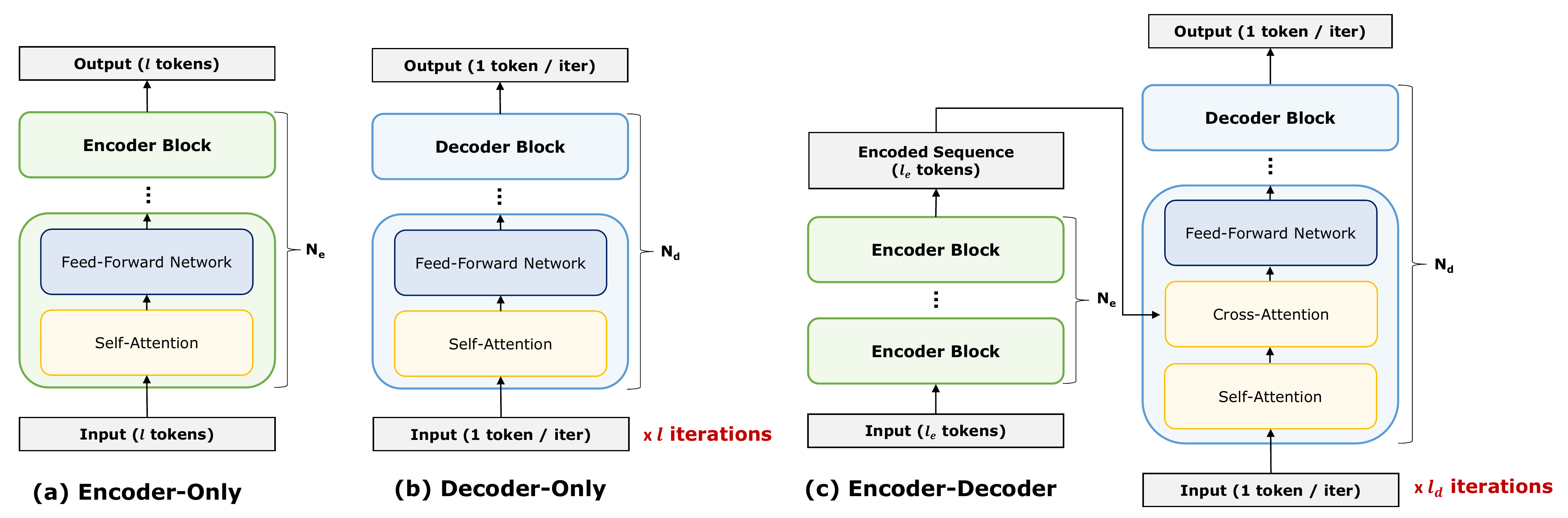}
  \caption{Variants of Transformer networks. (a) An encoder-only model, which performs inference for all tokens in parallel. (b) A decoder-only model, which performs inference in an auto-regressive manner. (c) An encoder-decoder model, which uses the output of the encoded sequence as input to a cross-attention module.}
      \label{fig:basic-architecture}
      
\end{figure*}

\subsubsection{\textbf{Nonlinear Operations}}
\label{subsec:nonlinear_ops_intro}
There are several nonlinear operations such as Softmax, LayerNorm, and GELU that require specialized support or off-chip computation. 
These nonlinear operations account for a relatively smaller portion of the overall operations, when inferring with Transformer networks, as compared to the linear operations (Sec.~\ref{sec:profiling}). 
However, they are more challenging to compute on typical hardware than matmuls, and they can incur significant overhead if not handled appropriately.

The nonlinear operations present challenges in terms of efficient utilization of temporal memory as well as efficient computation.
This is because they require multiple passes over all input values, which requires those values to be held in temporal memory.
As depicted in Fig.~\ref{fig:nonlinear-ops} (a), the Softmax operation involves (1) exponential operations, (2) summing up the results across the sequence length dimension, and (3) normalizing the input by dividing it by the summation result.
It is also well known that the exponential function is prone to numerical overflow, prompting the use of the maximum subtraction trick~\cite{max_trick} that transforms the expression $\exp(x_i)/\sum_j\exp(x_j)$ into $\exp(x_i-x_{\max})/\sum_j\exp(x_j-x_{\max})$, where $x_{\max}$ is the maximum of the $x_j$'s. 
This, however, requires an additional pass over the inputs, resulting in a three-pass numerically stable implementation.
Computing the LayerNorm function also requires multiple passes over the entire input values across the hidden dimension, as illustrated in Fig.~\ref{fig:nonlinear-ops} (b). 
In the first pass, the mean must be computed.
In the second pass, this is then used to compute the standard deviation.
Finally, in the third pass, where the normalization is actually applied, one division per input value is required.

Furthermore, the nonlinear operations entail challenges in operation fusing, which is a common technique to reduce interlayer communications by combining multiple operations into a single operation (Sec.~\ref{subsec:graph-level}).
Unlike Batch Normalization (BatchNorm) in many CNN architectures that can be seamlessly subsumed into preceding or succeeding linear operations~\cite{jacob2018quantization}, LayerNorm requires computing the mean and variance of the inputs at runtime.
Therefore, to fuse this operation with the preceding matmul operation, the entire output matrix must be accumulated in place across the reduction dimension (i.e., the dimension in which the mean and variance are computed) before writing out results.
This leads to irregular tiling dimensions and lower data reuse. 
As a result, there is a nontrivial tradeoff between fusing these operations with previous layers versus using better tiling dimensions for maximizing reuse. 
A detailed analysis of this tradeoff will be provided later in Sec.~\ref{subsec:scheduling_complexity_nonlinear}.

\vspace{5mm}
\subsubsection{\textbf{Encoder and Decoder Architectures}}
\label{subsec:encode_decoder_architectures}
The Transformer architecture was originally introduced as an encoder-decoder model for machine translation tasks~\cite{vaswani2017attention}. 
In this setting, the encoder takes the entire source language sentence as input and passes it through multiple Transformer encoder blocks, extracting the high-level features of the input sentence. 
These extracted features are then fed into the decoder, which is responsible for generating the tokens in the target language, one after another.
This is based on the source language features from the encoder as well as the tokens it has previously generated~\cite{vaswani2017attention}.
In subsequent works, encoder-only and decoder-only architectures were introduced, taking only the encoder and the decoder components, respectively, from the original encoder-decoder architecture~\cite{radford2019language,dai2019transformer} (Fig.~\ref{fig:basic-architecture}).

\paragraph{\textbf{Encoder Block.} }
In \textit{encoder-only} Transformer models~\cite{devlin2018bert,liu2019roberta,yang2019xlnet}, the input sequence is passed through the repeated encoder blocks all at once.
For this reason, the encoder-only structure is suitable for natural language understanding tasks~\cite{devlin2018bert,liu2019roberta}, such as sentiment analysis~\cite{socher2013recursive} or sentence similarity analysis~\cite{iyer2017first,cer2017semeval,dolan2005automatically}, where the entire input sequences are fed into the model.

In the encoder block, the inference is composed of matrix-matrix multiplications as well as element-wise additions and nonlinear operations. 
The cost of the projection layers in the MHA module and FFN module scales linearly with the input sequence length $l$.
However, the act-to-act matmuls in the MHA module scale quadratically with sequence length (as demonstrated in query $\times$ key and attn. score $\times$ value rows in Tab.~\ref{table:architecture_dims}).
In Sec.~\ref{sec:profiling}, we demonstrate via profiling that this depends on the sequence length: with short sequence lengths, the projection layers dominate, making the overall complexity of the encoder block $O(l)$; with long sequence lengths, however, the act-to-act matmuls dominate, making the overall complexity~$O(l^2)$.

\paragraph{\textbf{Decoder Block. }}
In contrast to encoder-only models, the \textit{decoder-only} models~\cite{radford2018improving,radford2019language,brown2020language} that consist of repeated decoder blocks are auto-regressive in nature.
This means that the output at a given time step is based on the outputs in the previous time steps.
In other words, the model predicts a token in a sentence based on the previous tokens it has generated so far,
and the inference must therefore be performed sequentially and iteratively, once for each output token. 
For instance, if the previously generated sequence is ``I am a'', 
the model takes this as input and may predict the next token ``student''. 
Then, in the next time step, the input to the model becomes ``I am a student''.
Therefore, the decoder-only structure is suitable for natural language generation tasks. It is important to note that, in decoder-only models, the input prompt tokens can be consumed in parallel before the model begins to generate subsequent tokens. For this work, we only consider open-ended generation (i.e., assuming no input prompt).

Unlike the encoder block, which operates on the entire input sequence, the decoder block is inferred one token at a time.
This results in a sequence length of one for each time step.
In the case of the projection layers, each token is independent of the previously generated token.
Thus, the projection operations are solely applied to the input token, resulting in a matrix-vector multiplication and a constant cost.
However, this does not hold for the act-to-act matmuls, as the input token is not independent of the previously generated tokens.
Instead, it is required to attend to all of them.
Consequently, these operations scale linearly with sequence length, implying that more compute is required to process a token in a larger time step than a token in a smaller time step.
A key detail to note is that the full key and value activations must be present for the input token to attend to all previously generated tokens. 
A common optimization technique for token generation is to cache and reuse the intermediate key and value of the previously generated tokens in subsequent iterations,  thus avoiding the need to recompute them for every iteration.
Taken together, the end-to-end complexity of generating the full sequence scales linearly for the projection layers 
and quadratically for the other two act-to-act matmuls.
The end-to-end computation graph of the Transformer decoder block is also provided in Fig.~\ref{fig:decoder-comp-map} of Appendix \ref{appendix:decoder}.

\begin{boxA}
\textbf{Summary (Sec.~\ref{subsec:transforemr_high_level_overview}. Transformer Overview)}

Transformers are composed of several Transformer blocks, each of which has an MHA (multi-head attention module and an FFN (feed-forward network) module (along with LayerNorm and residual addition after each module). The MHA module contains projection layers as well as act-to-act matmuls and Softmax operations. The FFN module consists of two projection layers with a nonlinear function between them. There are two types of Transformer blocks: encoder blocks and decoder blocks. Encoder blocks process the entire input sequence in parallel, making them suitable for natural language understanding tasks. Decoder blocks are autoregressive, meaning that inference must be performed once per generated output token, and are therefore typically used in generative tasks.
\end{boxA}

\subsection{Model Analysis}
\label{subsec:model-analysis}

\subsubsection{\textbf{Workload Analysis}}
\label{sec:workload-analysis}

In order to evaluate bottlenecks in Transformer, we first modelled the number of floating-point operations (FLOPs) required to compute the Transformer encoder-only and decoder-only models, as well as the arithmetic intensity of these networks. 
Arithmetic intensity is the number of floating point operations that can be performed per byte loaded from memory.
It can be computed by dividing the total number of FLOPs by the total number of bytes accessed (also referred to as MOPs, or memory operations) ~\cite{williams2009roofline}: 
\begin{equation}
    \text{Arithmetic Intensity} = \frac{\text{\# FLOPs}}{\text{\# MOPs}} .
\end{equation}
Here, we are assuming that the local memories are large enough to hold both matrices entirely in memory for a given operation, and that the computed arithmetic intensity values therefore serve as an upper bound for the achievable data reuse. We are also counting the multiplication and addition from a MAC operation separately when computing FLOPs.

\begin{figure}[t!]
  \centering
  \includegraphics[width=\columnwidth]{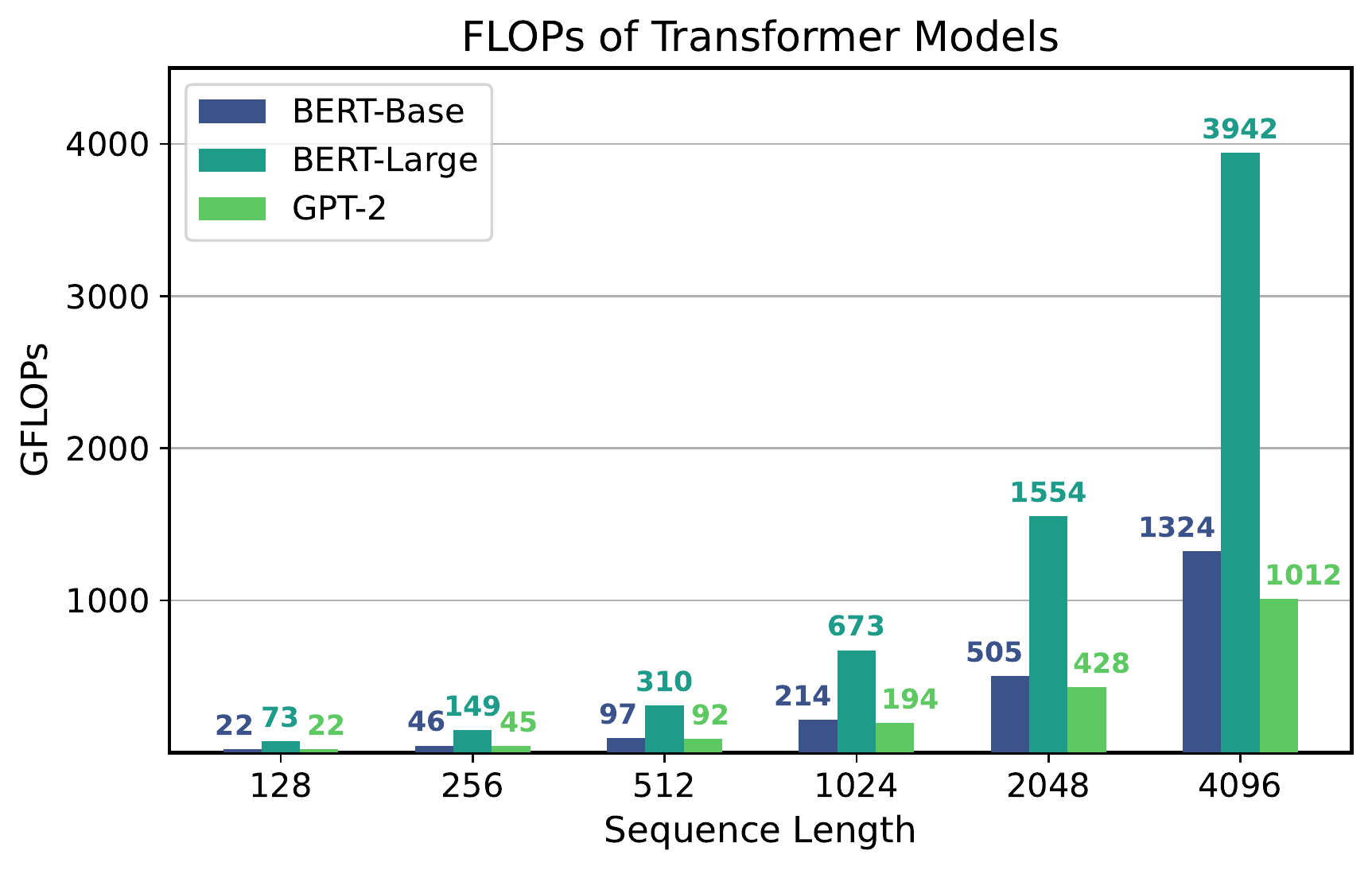}
  \caption{Plot of the FLOPs for the BERT-Base and BERT-Large encoders and the GPT-2 decoder across different sequence lengths. The FLOPs scales quadratically with sequence length due to quadratic scaling in the act-to-act matmuls as well as the Softmax function. 
  Additionally, inferring the BERT-Base encoder and the GPT-2 decoder (which have the same model architecture) requires a similar number of FLOPs for processing the same sequence length.
  }
      \label{fig:flops}
\end{figure}

\begin{figure}[t!]
  \centering
  \includegraphics[width=\columnwidth]{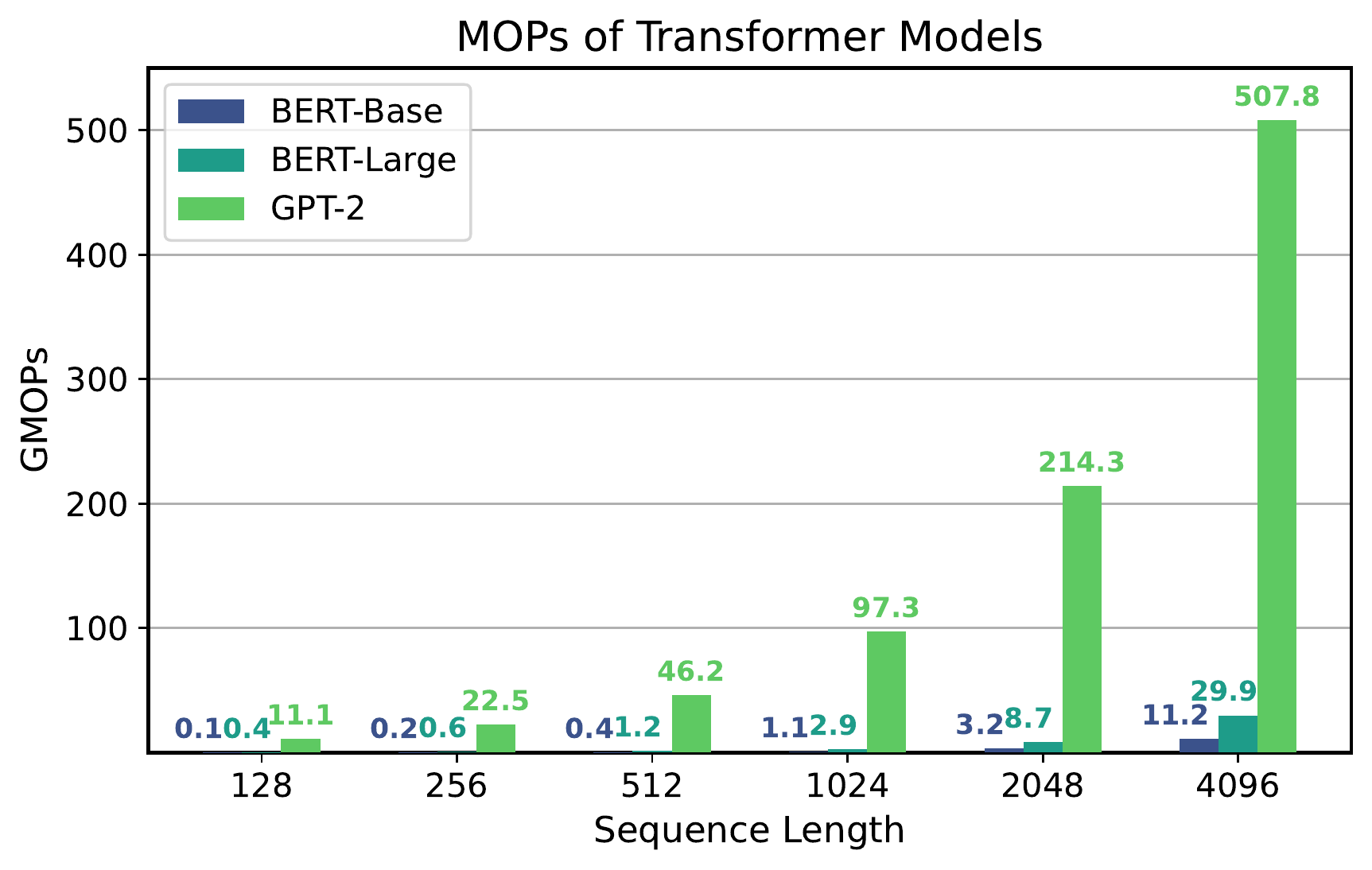}
  \caption{Plot of the MOPs for the BERT-Base and BERT-Large encoders and the GPT-2 decoder across different sequence lengths. The MOPs scale quadratically with sequence length for the encoder-only models due to quadratic scaling in the act-to-act matmuls as well as the Softmax function. 
  Additionally, the GPT-2 decoder requires a much greater number of MOPs than the BERT-Base encoder (which have the same model architecture) for processing the same sequence length as it loads weights per every token generation.
  }
      \label{fig:mops}
\end{figure}

\paragraph{\textbf{End-to-end FLOPs and MOPs.}}
For the encoder analysis, we used the 12-layer BERT-Base model and the 24-layer BERT-Large network (see Tab.~\ref{table:symbols} for model configurations); and for the decoder, we used the 12-layer GPT-2 model architecture which has the same model configuration parameters as BERT-Base.
For the purposes of analysis, we ignored the maximum input sequence lengths of 512 for standard BERT models throughout this section.
We then computed MOPs, the number of bytes that had to be accessed, when inferring these models. 
We assumed 8-bit precision for all operations, meaning that loading one parameter or activation would require loading one byte.
For the decoder model, we measured the FLOPs and MOPs as the total amount of floating point operations and memory operations needed to iteratively generate the full sequence of the given length. 
The FLOPs and MOPs for these networks for a range of sequence lengths are plotted in Fig.~\ref{fig:flops} and \ref{fig:mops}, respectively. 
As one can see, FLOPs and MOPs scale super-linearly for all models, especially in the long sequence length regime, due to the quadratic complexity with respect to sequence length in the act-to-act matmuls.

\begin{figure}[t!]
  \centering
  \includegraphics[width=\columnwidth]{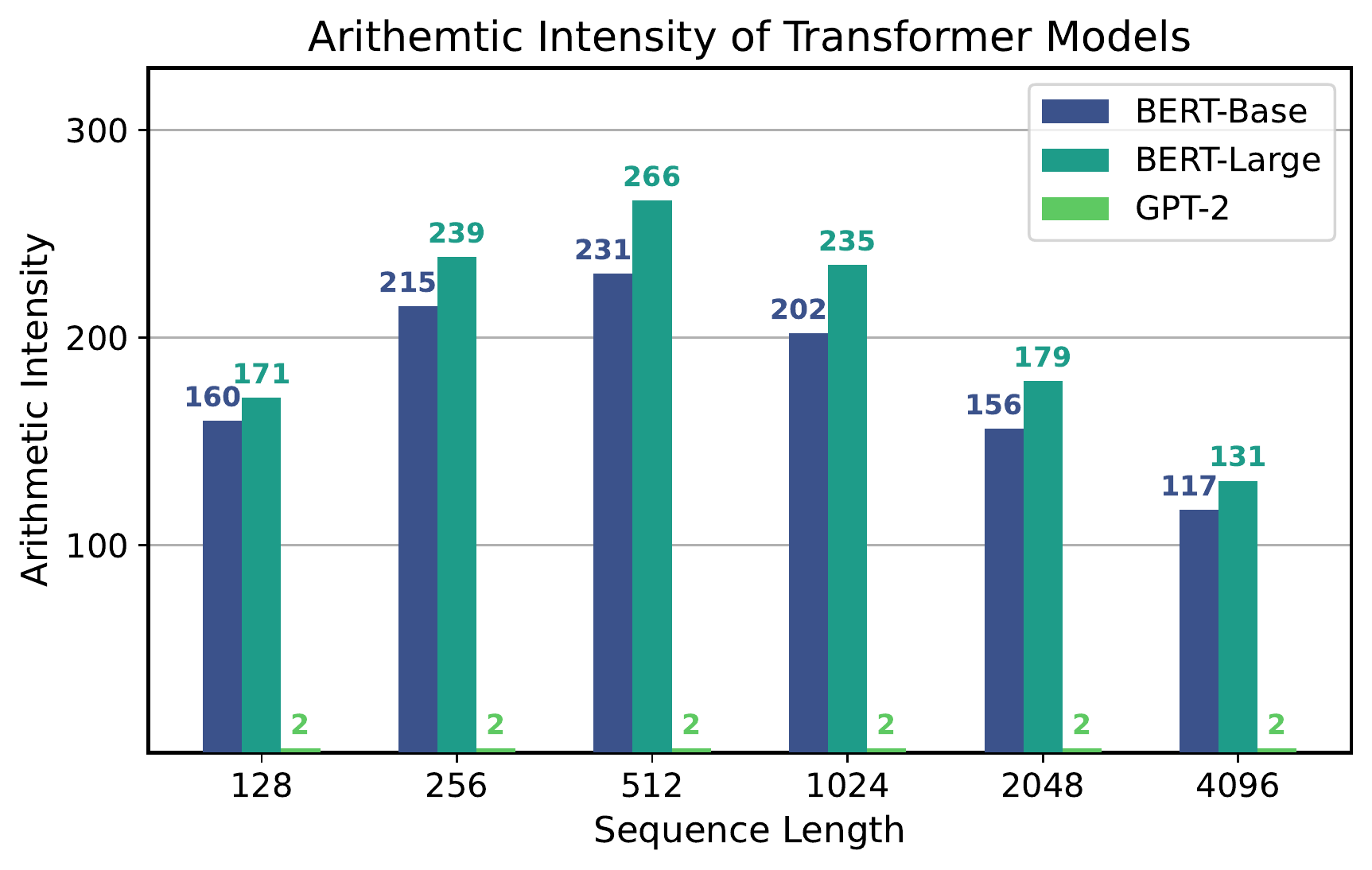}
  \caption{Plot of the arithmetic intensity of the BERT-Base and BERT-Large encoders and the GPT-2 decoder across different sequence lengths. The arithmetic intensity initially increases since the larger matrix dimensions allow for more computations to be performed per parameter loaded. However, at higher sequence lengths the arithmetic intensity decreases.
  This is because, for the long sequence length, the  act-to-act matmuls, and Softmax computations of the MHA module begin to dominate.
  These have relatively lower arithmetic intensity compared to the projection layers in the FFN module.}
      \label{fig:ai-sequencelength}
\end{figure}

\begin{figure}[t!]
  \centering
  \includegraphics[width=\columnwidth]{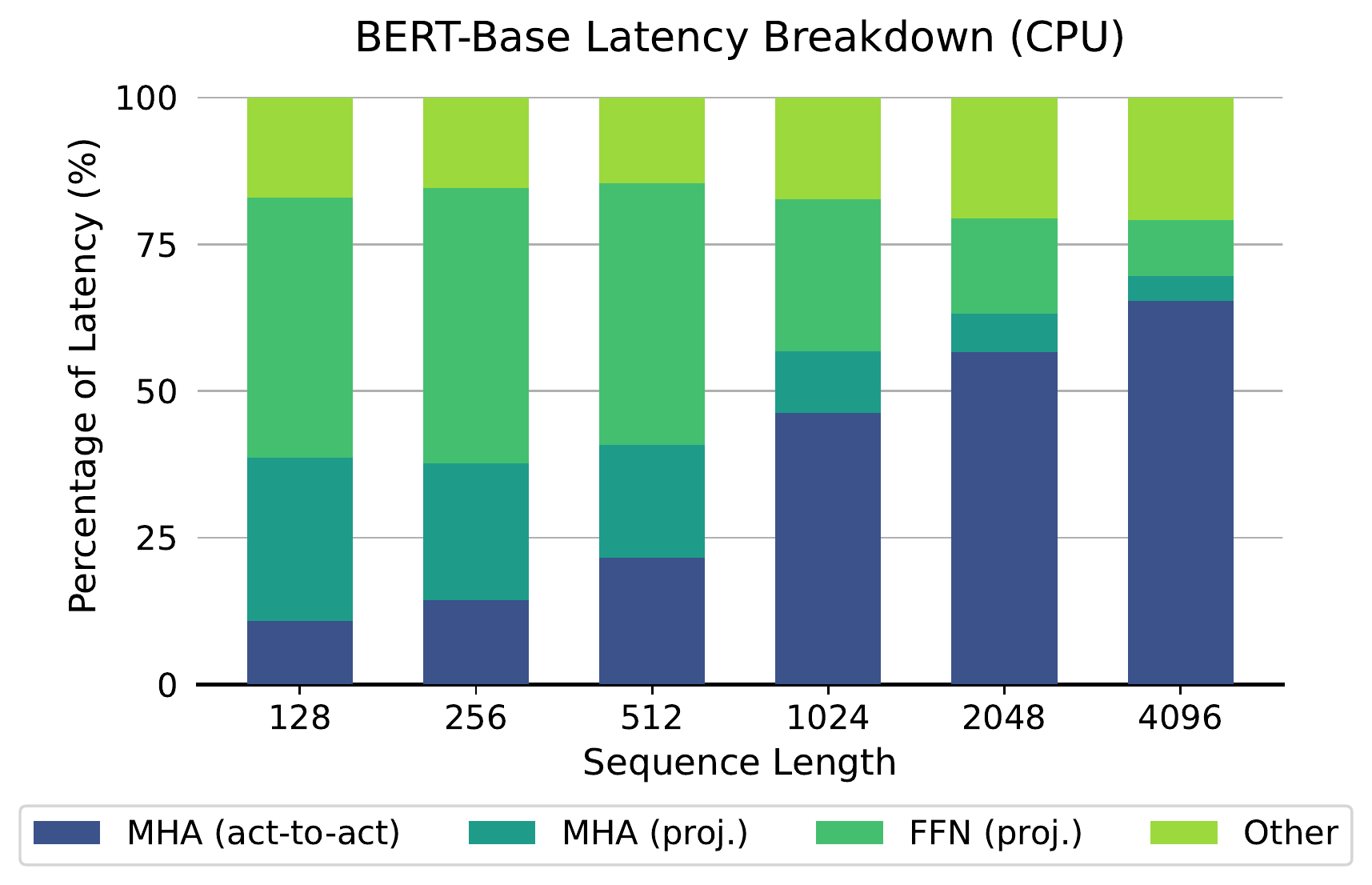}
  \caption{Plot of the computation breakdown in the BERT-Base encoder versus sequence length on a CPU. For smaller sequence lengths, the projection layers in the MHA and FFN modules dominate the model latency. However, for longer sequence lengths the  act-to-act matmuls begin to dominate. 
  \label{fig:breakdown-bert-base-cpu}
    }
  \vspace{-1em}
\end{figure}

\paragraph{\textbf{End-to-end Arithmetic Intensity.}}
We then modeled the arithmetic intensity by dividing the number of FLOPs required when inferring these models by the number of MOPs. 
The arithmetic intensity for BERT-Base, BERT-Large, and GPT-2 versus sequence length is shown in Fig.~\ref{fig:ai-sequencelength}.
For both BERT-Base and BERT-Large, the arithmetic intensity initially increases with sequence length until 512 and then decreases afterwards for larger sequence lengths. 
The reason for this is that, as will be analyzed in more detail in Sec.~\ref{sec:profiling},
the FFN module that has higher arithmetic intensity than the MHA module (Tab.~\ref{table:per-layer-encoder-12heads}) dominates the total FLOPs for small sequences (Fig.~\ref{fig:breakdown-bert-base-cpu}).
However, this trend reverses for larger sequence lengths, as the cost of act-to-act matmuls in the MHA module grow quadratically with the increase in sequence length, leading to a reduction in arithmetic intensity for the end-to-end model inference.

In comparison to encoder-only BERT inference, {decoder-only GPT-2 inference exhibits significantly lower arithmetic intensity}. 
This is due to the fact that the decoder is composed solely of matrix-vector operations, which limits the opportunities for data reuse.
That said, for a single matrix-vector operation, we perform roughly one multiplication and addition per parameter loaded since the loads cannot be shared across tokens.
This leads to performing roughly 2 operations per parameter loaded.
It is important to note that GPT-2 has fewer FLOPs than BERT-Base and BERT-Large as the sequence length is increased.
However, it is typically more challenging to run its inference efficiently due to its low arithmetic intensity.
This makes its performance \textit{memory bandwidth-bound}, as compared to  encoder-only BERT models. 
This behavior is also characterized in depth by~\cite{optimus}.

\begin{table*}[!t]
\caption{
Per-Layer FLOPs, memory operations (MOPs), and arithmetic intensity for the BERT-Base encoder with sequence lengths of 128, 512, and 4096 tokens. 
At low sequence lengths, the main contributors to both FLOPs and MOPs are the MHA and FFN projections. 
For longer sequence lengths, the act-to-act matmuls consume a greater proportion of FLOPs, and these operations along with Softmax consume the majority of MOPs. 
The act-to-act matmuls also have lower arithmetic intensity than the projection layers in the MHA and FFN for each sequence length. 
}
\begin{center}
\small{
\begin{tabular}{c|c|ccccc}
\toprule
Sequence Length & Operator & FLOPs ($\times$ $10^9$) & \% of total FLOPs & MOPs ($\times$ $10^9$) & \% of total MOPs & Arithmetic Intensity \\
\midrule
\multirow{5}{*}{128} & MHA (projections) & 7.25 & 32 & 0.04 & 27 & 192.00 \\
& MHA (act-to-act matmuls) & 0.60 & 3 & 0.006 & 7 & 63.62 \\
& FFN (projections) & 14.50 & 65 & 0.07 & 49 & 211.86 \\
& Other & 0.08 & 0.3 & 0.02 & 18 & 3.14 \\
& Total & 22.42 & 100 & 0.14 & 100 & 159.68 \\
\midrule
\multirow{5}{*}{512} & MHA (projections) & 28.99 & 30 & 0.07 & 16 & 438.86 \\
& MHA (act-to-act matmuls) & 9.62 & 10 & 0.09 & 20 & 101.95 \\
& FFN (projections) & 57.98 & 60 & 0.10 & 25 & 558.54 \\
& Other & 0.42 & 0.4 & 0.16 & 37 & 2.73 \\
& Total & 97.02 & 100 & 0.42 & 100 & 231.0 \\
\midrule
\multirow{5}{*}{4096} & MHA (projections) & 231.93 & 18 & 0.33 & 3 & 702.17 \\
& MHA (act-to-act matmuls) & 616.02 & 46 & 4.98 & 44 & 123.63 \\
& FFN (projections) & 463.86 & 35 & 0.43 & 4 & 1068.52 \\
& Other & 11.85 & 1 & 5.47 & 49 & 2.16 \\
& Total & 1323.66 & 100 & 11.22 & 100 & 117.96 \\
\bottomrule
\end{tabular}
}
\vspace{2mm}
\end{center}
\label{table:per-layer-encoder-12heads}
\end{table*}


\begin{table*}[!t]
\caption{
Per-Layer FLOPs, memory operations (MOPs), and arithmetic intensity for ResNet50.
Convolutions consume the dominant proportion of FLOPs, but BatchNorm, ReLU, and the other operations contribute a significant proportion of MOPs.
}
\begin{center}
\small{
\begin{tabular}{c|ccccc}
\toprule
Operator & FLOPs ($\times$ $10^9$) & \% of total FLOPs & MOPs ($\times$ $10^9$) & \% of total MOPs & Arithmetic Intensity \\
\midrule
Convolution & 7.26 & 99 & 0.04 & 36 & 183.36 \\
BatchNorm & 0.03 & 0.5 & 0.03 & 31 & 1.00 \\
ReLU & 0.008 & 0.1 & 0.02 & 15 & 0.50 \\
Other & 0.01 & 0.1 & 0.02 & 18 & 0.53 \\
Total (Unfused) & 7.31 & 100 & 0.11 & 100 & 66.94 \\
Total (Fused) & 7.28 & 100 & 0.06 & 100 & 121.36 \\
\bottomrule
\end{tabular}
}
\end{center}
\label{table:resnet50-flops}
\end{table*}



\paragraph{\textbf{Per-Layer FLOPs, MOPs, and Arithmetic Intensity.}}
\label{sec:per-layer-flops}

We then assessed the per-layer FLOPs, MOPs, and arithmetic intensity versus sequence length for the BERT-Base encoder 
(Tab. \ref{table:per-layer-encoder-12heads}). 
As shown in Tab. \ref{table:per-layer-encoder-12heads}, the proportion of FLOPs and MOPs consumed by the act-to-act matmuls increases with sequence length, and these operations have lower arithmetic intensity compared to the projection layers in the FFN and MHA modules. 
This explains the decrease in overall arithmetic intensity of encoder-only models for long sequence lengths, as observed in Fig. \ref{fig:ai-sequencelength}.

The low arithmetic intensity of the act-to-act matmuls relative to the projection layers is because the $d / h$ dimension in these two operations is small relative to the dimensions for the projection layers ($d$ and $d_{FFN}$) and also relative to $l$, as the sequence length is increased.
Small matrix dimensions lead to lower arithmetic intensity, as there are fewer operations to perform per element in the matrix, leading to reduced reuse.
The low arithmetic intensity is further exacerbated with large activation sizes that must be loaded and stored for the act-to-act matmuls. 
This activation size not only grows quadratically with the sequence length $l$, but it is further multiplied by the number of heads $h$ since each head has its own activation (attention score) in the multi-head scheme.
Therefore, as shown in Tab. \ref{table:per-layer-encoder-4heads} in Appendix~\ref{appendix:profiling}, a hypothetical BERT model with a smaller number of heads (thus with a larger $d/h$ dimension) would reduce the number of MOPs and improve the arithmetic intensity of the act-to-act attentions in the MHA module. 
This suggests that, when designing a Transformer architecture, the number of heads can entail a trade-off between accuracy versus performance metrics on hardware.

Additionally, Tab.~\ref{table:per-layer-encoder-12heads} illustrates that while the nonlinear operations (classified as ``Other'' in the table) consume a small number of overall FLOPs, they consume a significant proportion of MOPs, especially for longer sequence lengths. 
Similar to the case of the act-to-act matmuls, the large number of MOPs in the Softmax operations for long sequence lengths is primarily due to several $l$ $\times$ $l$ matrices which must be either written out or loaded per attention head. 
This also indicates that the nonlinear activations, when handled poorly, can become a noticeable contributor to the overall performance, even though they might be overlooked due to their insignificant contribution to total FLOPs.
We provide a similar per-layer analysis on the GPT-2 decoder in Tab.~\ref{table:per-layer-decoder} of Appendix~\ref{appendix:profiling},
which demonstrates the significantly reduced arithmetic intensity across all layers, compared to the encoder-only model, resulting from a large number of memory operations. 

\paragraph{\textbf{Comparison with ResNet50.}}
\label{sec:resnet50-flops}
To provide a baseline in terms of the FLOPs, MOPs, and arithmetic intensity for a typical CNN, we also included a corresponding analysis of ResNet50 (architectural details can be found in Appedix~\ref{appendix:cnn}).
Tab.~\ref{table:resnet50-flops} provides a breakdown of the FLOPs, MOPs, and arithmetic intensity for ResNet50.
Compared to the BERT-Base encoder with a sequence length of 128 (Tab.~\ref{table:per-layer-encoder-12heads}), ResNet50 without any operator fusion consumes 3.07 times fewer FLOPs and 1.28 times fewer MOPs,
resulting in lower end-to-end arithmetic intensity than that of BERT-Base across all sequence lengths in Tab.~\ref{table:per-layer-encoder-12heads}.
The low arithmetic intensity is  partially due to the nonlinear operations in ResNet50 that consume a negligible proportion of FLOPs yet a significant proportion of MOPs, similar to the BERT-Base encoder.
However, unlike the nonlinear operations in Transformers, 
these operations in ResNet50 can be fused with the preceding matmuls in a straightforward manner for inference.
In particular, the ReLU operations can be applied directly to the accumulated outputs, and the BatchNorm operations can actually be folded into the prior convolutions. 
Fusing ReLU eliminates the MOPs for this operation, and folding BatchNorm eliminates both the required FLOPs and MOPs for this operation.
Broadly speaking, \textit{operation fusion} refers to a methodology in which the output values from one operation (e.g., a matmul or convolution) are directly used as input to the subsequent operation (e.g., a ReLU or BatchNorm) without first writing the output values to off-chip memory.
Operator fusion eliminates the need for unnecessary memory loads and stores for the nonlinear operations, and therefore it further improves the end-to-end arithmetic intensity.
As shown in Tab.~\ref{table:resnet50-flops}, fusing these operations with the prior convolutions improves the overall arithmetic intensity for the ResNet-50 network from 66.9 to 121.4.
In Tab.~\ref{table:resnet50-convolution-flops} of Appendix \ref{appendix:resnet50}, we provide more detailed numbers for the FLOPs, MOPs, and arithmetic intensity of several convolutional layers in ResNet50 as a reference.

Note that arithmetic intensity provides a rough estimate of how much data reuse is possible for different models and operations in the \textit{ideal} case.
Later in Sec.~\ref{sec:sysmodel}, we will discuss that analytical modeling can provide a more accurate, non-ideal estimate by taking account of the hardware details.

\vspace{2mm}
\subsubsection{\textbf{Profiling}}
\label{sec:profiling}

To analyze the bottlenecks in Transformer workloads on commodity hardware, we profiled Transformer inference on an Intel Gold 6242 CPU.
We profiled the workload latency breakdown for both encoder-only BERT-Base and decoder-only GPT-2. 

\begin{figure}[t!]
  \centering
  \includegraphics[width=\columnwidth]{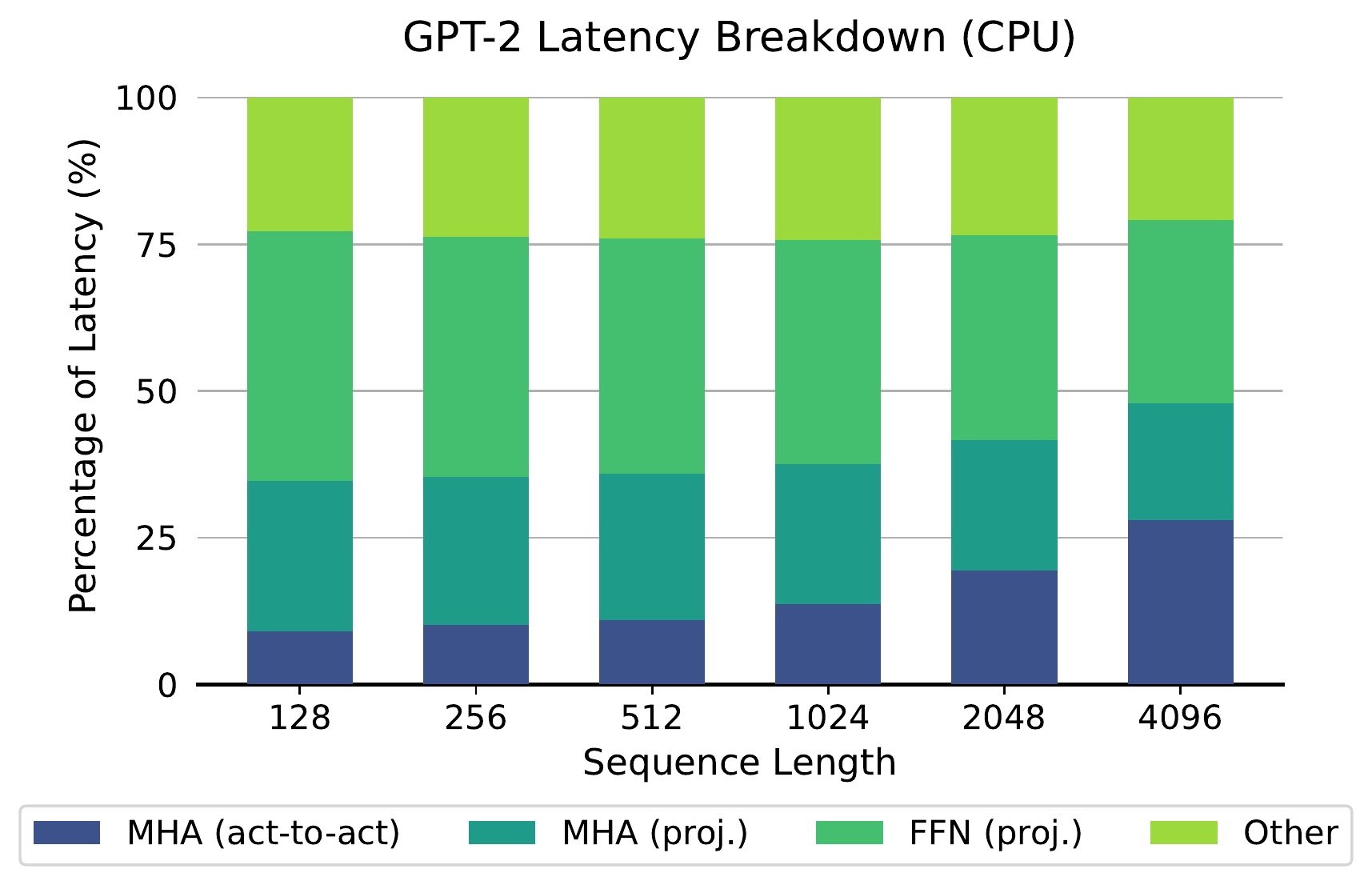}
  \caption{Plot of the computation breakdown in the GPT-2 decoder versus sequence length on a CPU. The projection layers in the MHA and FFN modules dominate latency for shorter sequence lengths, but for longer sequence lengths the  act-to-act matmuls become more significant. Note that nonlinear operations consume a more significant portion of latency than in encoder inference.}
      \label{fig:breakdown-gpt2-cpu}
      
\end{figure}

\begin{figure}[t!]
  \centering
  \includegraphics[width=\columnwidth]{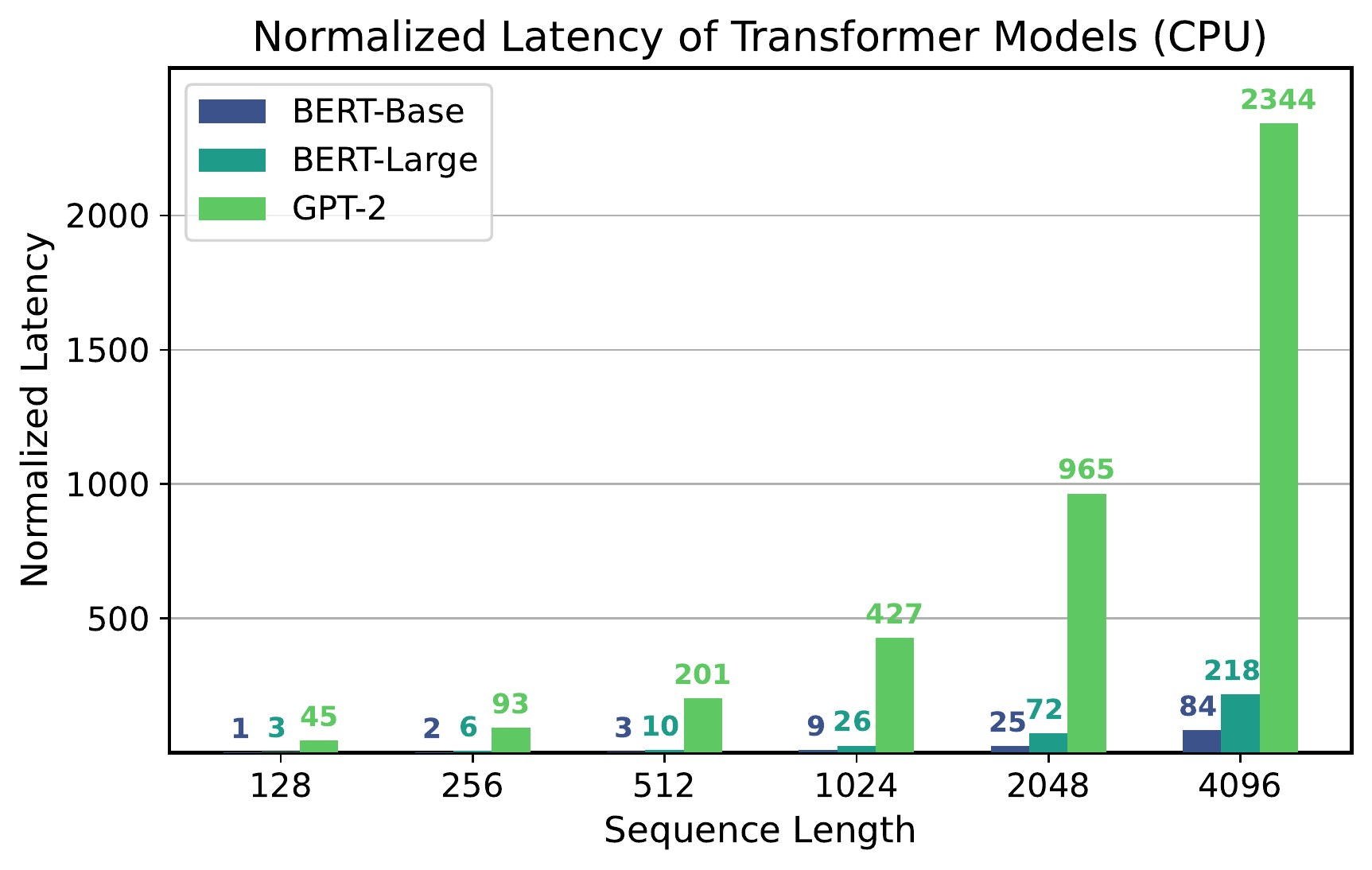}
  \caption{Plot of the normalized latency of the BERT-Base and BERT-Large encoders and the GPT-2 decoder versus sequence length on a CPU, normalized to the latency of BERT-Base with a sequence length of 128. The latency scales quadratically with respect to sequence length for both encoder-only and decoder-only networks. Additionally, for encoder-only and decoder-only networks with the same model architecture, the latency is significantly longer for the decoder-only network due to its reduced arithmetic intensity.}
      \label{fig:normalizedruntime-cpu}
\end{figure}

\paragraph{\textbf{Latency breakdown.}}
Fig.~\ref{fig:breakdown-bert-base-cpu} and \ref{fig:breakdown-gpt2-cpu} demonstrate how the latency breakdown changes with respect to sequence length on a CPU for BERT-Base and GPT-2, respectively. 
These breakdowns illustrate that for short sequence lengths (e.g., 128-512), the majority of computations are in the projection layers of the FFN module, and that the majority of the MHA computation is in the projection layers.
However, as sequence length increases, the  act-to-act matmuls begin to dominate, as they both scale quadratically with sequence~length. 

\paragraph{\textbf{End-to-end Latency.}}
Fig. \ref{fig:normalizedruntime-cpu} shows the normalized latency for different sequence lengths for BERT-Base, BERT-Large, and GPT-2. 
It is evident that the GPT-2 latency is far longer than the latency for either BERT-Base or BERT-Large for each sequence length, even though BERT-Base and GPT-2 have largely the same model configuration and end-to-end FLOPs (as was depicted in Fig.~\ref{fig:flops}).
This is mostly due to the lower arithmetic intensity of matrix-vector operations, which was highlighted in Fig.~\ref{fig:ai-sequencelength}.
A model with higher arithmetic intensity can run faster with the same (or possibly even more) FLOPs than a model with lower arithmetic intensity. These observations confirm our findings that decoder inference is a memory-bound problem and not a compute-bound problem.
We revisit this issue at Sec.~\ref{subsec:accelerating_decodin} to discuss some of the existing methodologies to speed up the decoding process.

\begin{boxA}
\textbf{Summary (Sec.~\ref{subsec:model-analysis}. Model Analysis):} 
Here are the high-level takeaways from this section.
\begin{itemize}[leftmargin=5mm]
\item Both FLOPs and normalized latency scale super-linearly with sequence length for all Transformer models due to the quadratic complexity of the act-to-act matmuls. 
However, this trend is less obvious with small sequence lengths, where the main contributor to the overall computation is the FFN, which scales linearly with sequence length, rather than the MHA module (Fig.~\ref{fig:flops} and \ref{fig:normalizedruntime-cpu}).
\item For encoder-only models, arithmetic intensity initially increases as the sequence length increases. However, it decreases for larger sequences since the MHA module (in particular, the act-to-act matmuls with lower arithmetic intensity) becomes the dominant contributor to total compute (Fig.~\ref{fig:ai-sequencelength}).
\item The arithmetic intensity of decoder-only models is significantly lower than that of encoder-only models, leading to significantly longer end-to-end latency for the same sequence length. 
This is due to the fact that decoder models involve matrix-vector operations with limited data reuse, making them memory bandwidth-bound rather than compute-bound (Fig.~\ref{fig:ai-sequencelength} and~\ref{fig:normalizedruntime-cpu}).
\item Matmuls consume over 99\% of the FLOPs in both encoder-only and decoder-only models, and nonlinear operations are a relatively small portion of overall FLOPs.
However, the nonlinear operations have extremely low arithmetic intensity, especially for the large sequence length, due to the large volume of activations they need to load and store.
\end{itemize}

\end{boxA}

\section{Hardware Design}
\label{sec:hardware_design}
So far, in Sec.~\ref{sec:architecture_and_performance}, we have conducted an analysis of the run-time characteristics and bottlenecks of Transformer architectures.
We now shift our focus to full-stack solutions for efficient Transformer inference, beginning with the design of efficient hardware. 
Sec. \ref{sec:acceleratorarch} then outlines the rationale of using domain specific accelerators for DNNs as well as the basic architectures and dataflows that are used in most DNN accelerators. 
Sec. \ref{subsec:literature-transformer-hw} then highlights existing work on accelerating Transformers. 
Sec. \ref{sec:sysmodel} then provides analysis using an analytical model to assess how Transformers run on a typical accelerator. 
Finally, Sec.~\ref{sec:thiswork} provides a case study illustrating the process of building a typical accelerator for Transformers. 
Overall, this section gives relevant performance analysis and provides justification for the selected hardware decisions from a full-stack perspective. 
Note that we are concerned here only with efficiently inferring DNNs.
In particular, designing hardware for efficient model training is outside the scope of this paper.

\subsection{Overview of Typical DNN Accelerators}
\label{sec:acceleratorarch}

A typical deep learning accelerator has a few key components, as outlined in \cite{hardwaresoftwaresurvey}:

\begin{itemize}[leftmargin=5mm]
    \item \textit{Off-chip DRAM} used for holding the weights and activations of the full network, which needs to be large enough to hold all model weights and activations;
    \item Smaller on-chip memory, referred to here as the \textit{global buffer}, which needs to be large enough to hold a subset of the weights and inputs in order to feed the \textit{processing elements} (PEs);
    \item An array of {PEs}, each of which has the capability to perform MAC operations, and which often contains one or more small local memories called \textit{register files} (RFs) that can store data with lower per-access energy than the global buffer; and
    \item An internal \textit{network-on-chip} (NoC) that transfers data between PEs.
\end{itemize}

\begin{figure}[t!]
  \centering
  \includegraphics[width=\columnwidth]{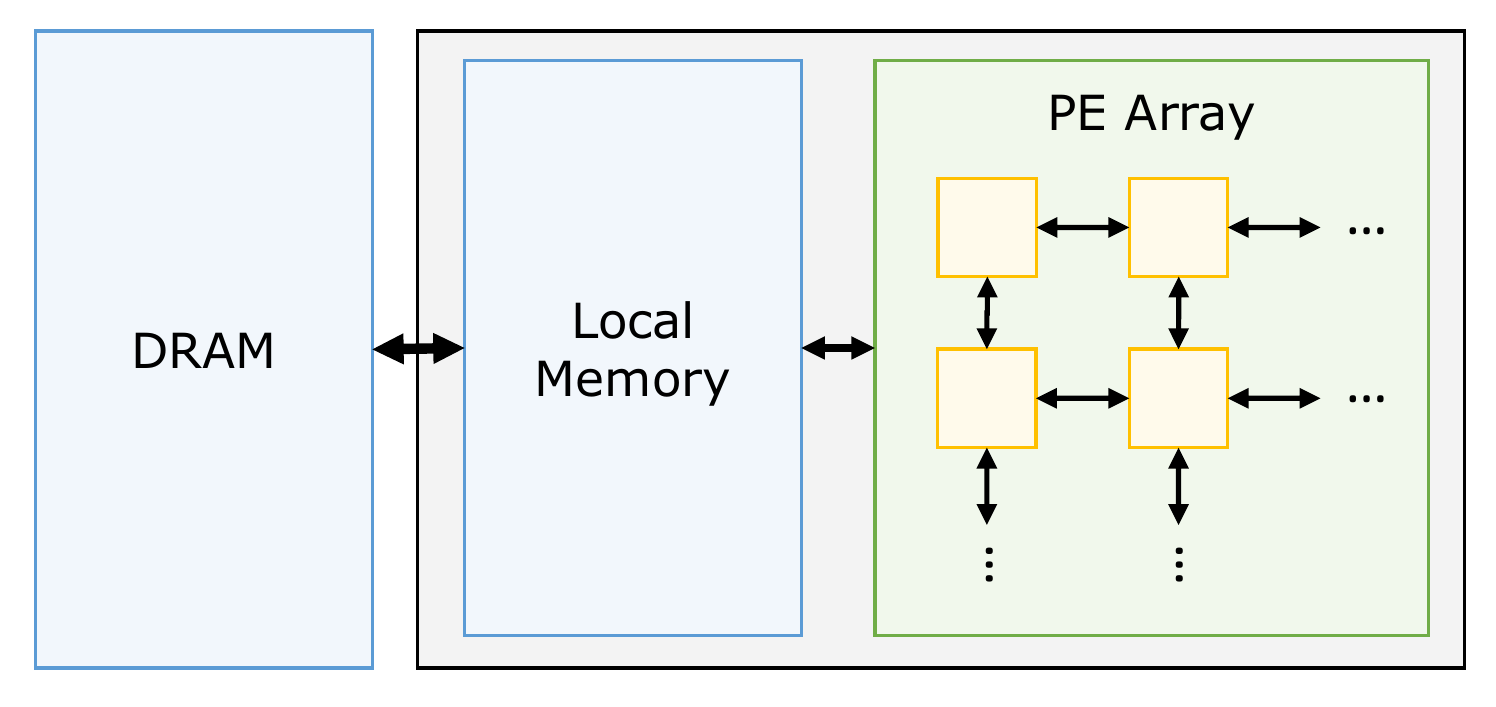}
  \caption{Basic structure of a DNN accelerator, assuming a spatial dataflow between the Processing Elements (PEs) (inspired by \cite{efficientprocessing}).}
      \label{fig:basic-structure}
\vspace{-3mm}
\end{figure}

Fig.~\ref{fig:basic-structure} shows the structure of a typical DNN accelerator. 
The global buffer is designed to be able to hold a sufficient number of the weights and activations in order to allow for data reuse and limit the number of transfers to and from the off-chip DRAM. The local memories in the PEs are used to provide local data reuse in order to reduce global buffer accesses whenever possible. 
Without reuse, MAC operation requires loading three parameters, the two input values that are being multiplied as well as the current partial sum (which is the partially accumulated value for a given location in the output matrix), and then storing the output value back to memory. This is important because memory reads and writes are orders of magnitude more expensive from an energy perspective \cite{energyproblem}. For example, for one particular technology, reads from a local buffer are roughly 6 times as expensive as a single MAC operation, and reads from external DRAM are roughly 200 times as expensive \cite{efficientprocessing}. Leveraging reuse opportunities is therefore critical in order to reduce the number of expensive memory accesses performed.

To maximize data reuse, there are two broad classes of dataflows that are widely adopted, which are referred to as temporal and spatial dataflows \cite{efficientprocessing,hardwaresoftwaresurvey,dataflows}. \textit{Temporal dataflows} contain an array of centrally-controlled PEs that load data from the global buffer and perform the requested ALU (Arithmetic Logic Unit) operations before writing the data back to the global buffer.
These PEs do not contain local memories, and there is no communication or data movement between PEs. Data reuse in this type of dataflow is only attainable through weight or partial sum reuse in the global buffer. 
Examples of temporal dataflows include Single-Instruction Multiple Data (SIMD) and Single-Instruction Multiple Thread (SIMT) execution units. These type of units are commonly used in both CPUs and GPUs for vector processing. In temporal architectures, fully-connected and convolutional layers are both mapped as matrix-matrix multiplication~operations. 

In \textit{spatial dataflows}, the PEs can communicate and data can be moved between PEs to leverage data reuse, without repeated reads from the global buffer. The PEs themselves often contain RFs to hold weights or partial sums locally to improve data reuse, and additional reuse can be attained through passing data between adjacent PEs. 
Spatial dataflows are commonly used in FPGA and ASIC-based accelerators, especially for convolutional networks \cite{dataflows}. These dataflows allow for data reuse across multiple dimensions in order to drastically reduce the required memory accesses. In order to maximize reuse in spatial dataflows, several different reuse schemes have been employed:

\begin{itemize}[leftmargin=5mm]
    \item \textit{Weight stationary} dataflows minimize the number of reads required for weight matrices by keeping weights in the local RFs in PEs and streaming through inputs \cite{240gops};
    \item \textit{Output stationary} dataflows minimize energy from reading and writing partial sums by accumulating the outputs in the local RFs in the PEs \cite{shidianno};
    \item \textit{No local reuse} dataflows have no RF in each PE, and use the area savings from having no RFs to allocate a larger global buffer~\cite{diannao}; and
    \item \textit{Row stationary} dataflows maximize reuse for both partial sums and weights by holding a row of the weights stationary in a row of the PEs, streaming in inputs, and streaming out partial sums \cite{eyeriss}.
\end{itemize}

Note that since DNNs consist of sequences of layers, it is also possible to fuse operations in order to further leverage data reuse across multiple layers. We encourage the reader to refer to Section V of \cite{efficientprocessing}, Sections IV-A and IV-B of \cite{compressionandacceleration}, and Sections III-A to III-C of \cite{hardwaresoftwaresurvey} for more comprehensive surveys and comparisons of the basic architecture of a DNN accelerator and typical accelerator~dataflows.

\begin{boxA}
\textbf{Summary (Sec.~\ref{sec:acceleratorarch}. Accelerating Neural Networks):} Typical DNN accelerators consist of on-chip memory for holding a subset of model weights and inputs and an array of processing elements (PEs) which can perform MAC operations.
Off-chip DRAM is used for holding weights and activations for the full network, and an internal network-on-chip (NoC) can be used for transferring data between PEs. 
DNN accelerators typically aim to leverage either temporal dataflows (by performing the same operation in parallel on several datapoints) or spatial dataflows (where data can be transferred between PEs to leverage additional reuse opportunities). 
Spatial dataflow reuse schemes include weight stationary dataflows, which hold weights in local memories in the PEs to improve reuse.
\end{boxA}

\subsection{Adapting DNN Accelerators for Transformers}
\label{subsec:literature-transformer-hw}

There are several key considerations when designing DNN accelerators for Transformers or adapting existing CNN accelerators. 
One difference between accelerators for CNNs and for Transformers is that due to differences in terms of arithmetic intensity and matrix dimensions, these models have different optimal sizes for each level of the memory hierarchy as well as different memory bandwidth~requirements. 

Another consideration is how the nonlinear functions are computed during inference, which imposes an additional challenge in hardware design. 
These operations require either specialized support for on-chip computation, or else they must be offloaded to the CPU. 
In Sec.~\ref{sec:thiswork}, we will outline how the nonlinear operations can bottleneck inference even though they compose a small proportion of model FLOPs, especially if they must be offloaded to the CPU. 
Several accelerators for Transformer inference contain specialized post-processing units for nonlinear functions~\cite{edgebert,pervectorquant,optimus, lu2020hardware}.
However, adding an additional unit to support these operations also increases the area of the accelerator.
This tradeoff between supporting these operations on-chip and accelerator area will be also explored in Sec.~\ref{sec:thiswork}. 
Additionally, it can be challenging to design the hardware both to efficiently support the required nonlinear operations (e.g., Softmax and LayerNorm) and to support new nonlinear operations in future DNNs. 

There are also considerations around datapath design, depending on whether the accelerator is being designed for the MHA module or for end-to-end Transformer inference. 
Accelerators specialized for the MHA module are designed to match the dataflow of the MHA module, where all the operations are ``fused'' together, thus having less flexibility but better performance by reducing the number of required memory accesses~\cite{a3,elsa,energon,wang2021spatten,adaptivebutterfly,dta-trans}. 
Recall that operation fusion refers to using the output values from one operation (e.g., a matmul) directly as input to the following operation (e.g., a Softmax layer) without writing the intermediate values to off-chip memory. 
Several accelerators for the MHA module develop dedicated datapaths with separate units for the query $\times$ key, Softmax, and attention score $\times$ value operations in order to better leverage operator-level fusion. 
In contrast, accelerators for end-to-end Transformer inference typically employ a similar structure to Gemmini \cite{gemmini-dac} (which is outlined in more detail in Sec.~\ref{sec:thiswork}) where they are designed to be more flexible by performing individual operations separately in a more general matmul engine~\cite{edgebert,pervectorquant,optimus, lu2020hardware}.
These accelerators also aim to fuse operations whenever possible to improve performance (for example, by applying Softmax directly to the accumulated outputs of a matmul before writing them out).
However, the entire graph-level dataflow is not hardcoded in hardware as in MHA-specific accelerators.

In both cases, there are dataflow considerations for nonlinear function unit placement. 
This is because, as we have demonstrated in~\ref{subsec:model-analysis}, non-linear operations generally have a high number of MOPs despite their small FLOPs count, and therefore  the overall arithmetic intensity can be improved upon through operator fusion (as in the ResNet50 case). 
In the case of accelerators for the MHA module, in order to leverage operator-level fusion in the MHA module, the Softmax unit must be placed appropriately such that it can be computed after the query $\times$ key multiplication and before the attention score $\times$ value multiplication.
For example, \cite{adaptivebutterfly} places the Softmax unit in between specialized units for the query $\times$ key and attention score $\times$ value multiplications, and it computes LayerNorm in a separate hardware module.
Placing functional units to support operator fusion provides higher efficiency, but this comes at a cost of less flexibility since the architecture now makes assumptions about the operator-level dataflow.

\begin{boxA}
\textbf{Summary (Sec.~\ref{subsec:literature-transformer-hw}. Adapting DNN Accelerators for Transformers):} 
Accelerators for Transformers and CNNs have different optimal sizes for the memory hierarchy as well as different memory bandwidth requirements. 
Accelerators for the MHA module tend to design hardened datapaths to exploit operator fusion.
End-to-end Transformer accelerators tend not to design their datapath around the graph-level dataflow in the MHA module.
Transformer accelerators tend to incorporate a post-processing unit to compute nonlinear functions efficiently on-chip.
\end{boxA}

\begin{table*}[!t]
\caption{
Non-ideal arithmetic intensity for the BERT-Base encoder with sequence lengths of 128, 512, and 4096 tokens.
The non-ideal arithmetic intensity is lower than the ideal arithmetic intensities (provided in Tab.~\ref{table:per-layer-encoder-12heads}) due to using 32-bit output precision before nonlinear operations as well as constraints from the memory sizes.
Note that the differences in non-ideal arithmetic intensity between the $W_Q$, $W_K$, $W_V$ projections and the $W_{\text{out}}$ projection with the same operation dimensions are due to differences in output precision -- $W_{\text{out}}$ is followed by nonlinear operations, and therefore it uses 32-bit instead of 8-bit.
}
\begin{center}
\small{
\begin{tabular}{c|ccccccc}
\toprule
Sequence Length & $W_Q$, $W_K$, $W_V$ projections & Q $\times$ K & Attn. score $\times$ V & $W_{\text{out}}$ projection & $W_1$ projection & $W_2$ projection & Total \\
\midrule
128 & 170.670 & 25.400 & 63.750 & 128.000 & 130.723 & 186.182 & 106.110 \\
512 & 341.333 & 29.882 & 102.300 & 204.800 & 211.862 & 409.6 & 111.122 \\
4096 & 409.6 & 30.788 & 118.710 & 227.556 & 236.308 & 512.000 & 47.067 \\
\bottomrule
\end{tabular}
}
\end{center}
\label{table:achievable-ai}
\end{table*}

\vspace{-2mm}
\subsection{Analytical Modelling}
\label{sec:sysmodel}
Analytic modeling is a useful tool for identifying bottlenecks when inferring DNN benchmarks, as it provides a quick estimate of runtime behaviors on the target hardware platform.
At design time, it can be quite difficult to analyze the runtime behaviors of benchmark workloads as well as the potential impacts of hardware architectural changes on performance.
This contrasts with the case in Sec.~\ref{sec:profiling} where profiling can be conducted directly on actual hardware (e.g., CPUs).
In cases where profiling is difficult or infeasible, analytical modeling can provide estimates to quickly guide design decisions. 

\begin{figure}[t!]
  \centering
  \includegraphics[width=\columnwidth]{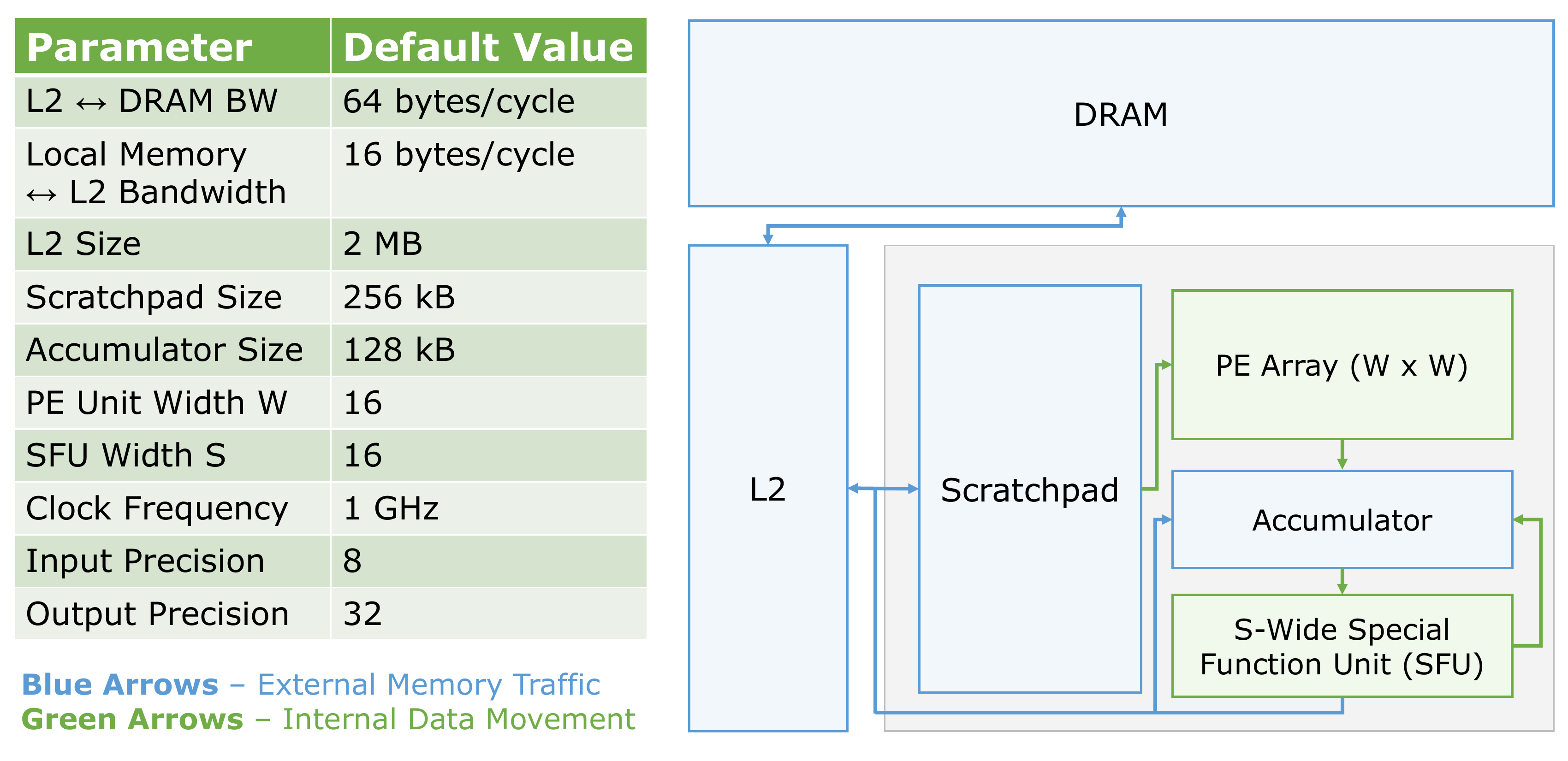}
  \caption{Diagram of the structure of the basic analytical performance model, as well as the parameters that were varied in this analysis.
  }
      \label{fig:model-structure}
\end{figure}

Here, we developed an analytical model to demonstrate how it can be useful in understanding the performance breakdown of Transformer inference on hardware accelerators.
Our analytic model is based off of the Gemmini-driven architecture~\cite{gemmini-dac}, which will be outlined in more detail in Section \ref{sec:gemmini}.
Its structure is illustrated in Fig.~\ref{fig:model-structure}, along with the tunable parameters. 
The model includes local memories, a PE array for computing tiled matrix-matrix multiplications, 
and it relies on external memory for storing all model parameters. 
The performance estimates assume that compute time and memory operation time can be overlapped perfectly, and that the total for each operation is the maximum of these two.
Note that double buffering was assumed in the scratchpad to ensure that compute could be overlapped with memory reads/writes wherever possible.
The model structure is comparable to typical DNN accelerators, with the notable assumption that the included special function unit (SFU) is able to compute all required nonlinear operations, and thus none of these operations have to be computed off-chip. 
The model also assumes $W$-cycle latency for the PE array, where $W$ is the width of the PE array, and 1-cycle latency per vector for the SFU.

\paragraph{\textbf{Latency Breakdown and End-to-end Latency.}}
One useful scenario of analytical modeling is to obtain the estimated latency breakdown and end-to-end runtime latency.  
As an example, we applied analytic modeling to the BERT-Base and BERT-Large encoders as well as the GPT-2 decoder, under the assumption of square tiling for all matrix operations and no operation fusion (i.e., each operation required inputs to be read from external memory and outputs to be flushed out).
In Appendix~\ref{appendix:modeling}, we provide the latency breakdowns for BERT-Base and GPT-2 (Fig.~\ref{fig:breakdown-bert-base} and \ref{fig:breakdown-gpt2}, respectively) as well as the end-to-end runtime latency of all models with different sequence lengths (Fig.~\ref{fig:normalizedruntime}).
In general, the results of the analytical model show similar trends in runtime latency scaling and breakdowns as compared with the profiling results on the CPU in Sec.~\ref{sec:profiling}, only with slight differences in details. 
Note that the analytical model was designed assuming a hardware architecture that was different from the CPU architecture, and therefore the runtime behaviors would not necessarily be identical for different hardware platforms.
More details can be found in Appendix~\ref{appendix:modeling}, including a comparison with the analytic modeling results on ResNet50.

\paragraph{\textbf{Non-ideal Arithmetic Intensity}}
As with the analysis in Sec.~\ref{subsec:model-analysis}, arithmetic intensity provides a rough estimate of how much data reuse is possible for different operations in the ideal case.
However, in real-world scenarios, such as when tiling operations are required due to the size of the matrices exceeding the capacity of the local scratchpad,  the arithmetic intensity will be further reduced.
In such a case, analytic modeling can provide a more accurate estimate, namely \textit{non-ideal} arithmetic intensity, by taking into account the hardware details.
To take the tiling effect into account, we counted DRAM to L2 memory traffic in our analytical modeling, but not L2 to Scratchpad, in order to avoid double counting.
Furthermore, we assume 32-bit output precision before the nonlinear operations, since it is known that low input precision (e.g., 8-bit) to those operations can result in a considerable accuracy degradation~\cite{ibert}.
The non-ideal arithmetic intensities for different operations in the BERT-Base encoder are provided in Tab.~\ref{table:achievable-ai} for sequence lengths of 128, 512, and 4096. 

Compared to the ideal arithmetic intensity that we have discussed in Sec.~\ref{sec:workload-analysis} (Fig.~\ref{fig:ai-sequencelength}), which is 160, 231, and 118 for each sequence length, we observe significant reductions in the non-ideal arithmetic intensity.
This is due to the effects of tiling as well as the large 32-bit output activations which must be loaded and stored before the nonlinear operations.
The gap becomes even larger with a large sequence length (up to 2.5$\times$ reduction for sequence length 4096), where the effect of loading and storing intermediate values are more pronounced. 
This is different from the case of ResNet50, whose non-ideal arithmetic intensity of 121.312 does not diverge a lot from the ideal arithmetic intensity of 122.172.
This also demonstrates how even though the ideal arithmetic intensity of Transformers was generally higher than that of ResNet50, the overall achieved arithmetic intensity is lower for Transformers across all sequence lengths.

\begin{boxA}
\textbf{Summary (Sec.~\ref{sec:sysmodel}. Analytical Modeling):} 
Analytical modeling is a useful tool for identifying bottlenecks and runtime characteristics of DNN inference on a target hardware platform.
This technique can be especially useful during the design phase, where profiling on actual hardware can be difficult, yet the analysis is necessary in order to make design decisions.
We provided examples of using analytic modeling to obtain latency breakdown and non-ideal arithmetic intensity.
In detail, we have demonstrated that the non-ideal arithmetic intensity of the Transformer can be further reduced (up to 2.5$\times$) compared to the ideal case when the hardware constraints and implementation details are taken into account.
\end{boxA}

\subsection{Case Study: Building a Transformer Accelerator}
\label{sec:thiswork}

We now illustrate with a more ``hands-on'' example how architects familiar with mainstream accelerators for convolutional, vision-based workloads can design state-of-the-art transformer accelerators.
Although the analytical model in Sec.~\ref{sec:sysmodel} presents ideal latency and runtime predictions for Transformer inferences, approaching the ideal performance and efficiency in real-world hardware accelerators can take considerable engineering effort, which we explore here.
We start with a fairly typical CNN accelerator generated by the Gemmini~\cite{gemmini-dac} accelerator-generator, optimized primarily for ResNet50-like workloads, and we discuss changes we made to this accelerator and it's software stack to efficiently support transformer workloads such as BERT. 
Several accelerators for end-to-end Transformer inference employ a similar structure to Gemmini and to our analytical model and also contain specialized post-processing units for nonlinear functions~\cite{edgebert,pervectorquant,optimus, lu2020hardware}. 

\subsubsection{\textbf{Baseline Accelerator}}
\label{sec:gemmini}

\begin{figure}[t!]
    \centering
    \includegraphics[width=0.9\linewidth]{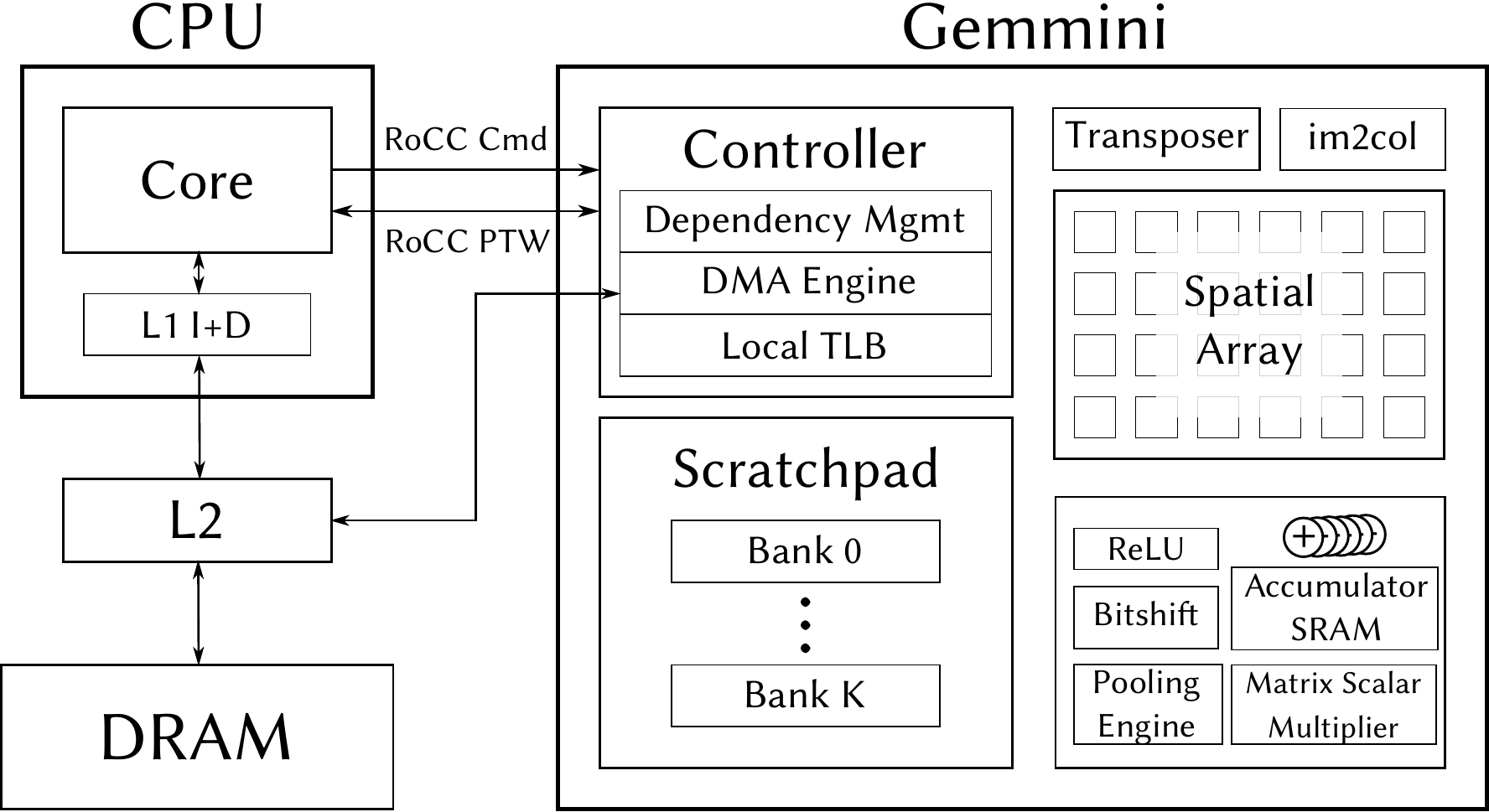}
    \caption{Baseline accelerator's hardware architectural overview.}
    \label{fig:gemmini-arch}
\end{figure}

We first generate a fairly typical CNN accelerator, which is illustrated in Fig.~\ref{fig:gemmini-arch}, using the Gemmini accelerator-generator.
The accelerator performs matmuls using a 16$\times$16 systolic array, which implements the weight-stationary dataflow.
When performing convolutions, the dimensions of the output-channels and input-channels are spatially unrolled.
The 8-bit integer weights and inputs are stored in a 256 kB local scratchpad memory, and the 32-bit partial sums are stored in a dual-ported 64 kB accumulator SRAM which performs matrix additions.
When DNN layers are too large to fit into the local scratchpad, they fall back onto an external L2 cache and DRAM which are shared with CPUs and other accelerators on the system-on-chip (SoC).
A host CPU tiles such layers to compute the full outputs.

Although most of a CNN's FLOPs are used to compute matmuls or convolutions, our baseline Gemmini-generated accelerator also contains peripheral circuitry to execute ReLU and max-pool operations, as well as integer-float multipliers to scale 32-bit partial sums to 8-bit inputs that can be fed into the next layer in a CNN.
Native support for these operations is important, as it eliminates the need for costly transfers back and forth between DRAM or outer caches (where the CPU can perform these operations) and the local scratchpad (where Gemmini stores its matrix operands).

Finally, note that this baseline CNN accelerator does not include any Transformer-specific features.
In particular, there is no support for non-linear normalization operations such as LayerNorm or Softmax.
Neither is there support for GELU, which is a relatively expensive non-linear activation function often implemented with costly lookup tables.
Instead, this baseline design is a typical example of an accelerator designed and optimized for quantized integer CNN inference.
It achieves real-time or near-real-time performance on end-to-end CNN workloads such as ResNet50~\cite{he2016deep}, SqueezeNet~\cite{iandola2016squeezenet}, or MobileNetV2~\cite{sandler2018mobilenetv2}, but (we will see that) the performance on Transformer workloads such as BERT is severely limited due to the need to perform operations such as GELU, LayerNorm, and Softmax on the CPU.

\subsubsection{\textbf{Performance Bottlenecks}}
\label{subsec:hardware_performance_bottlenecks}

\begin{figure}[t!]
    \centering
    \includegraphics[width=0.6\linewidth]{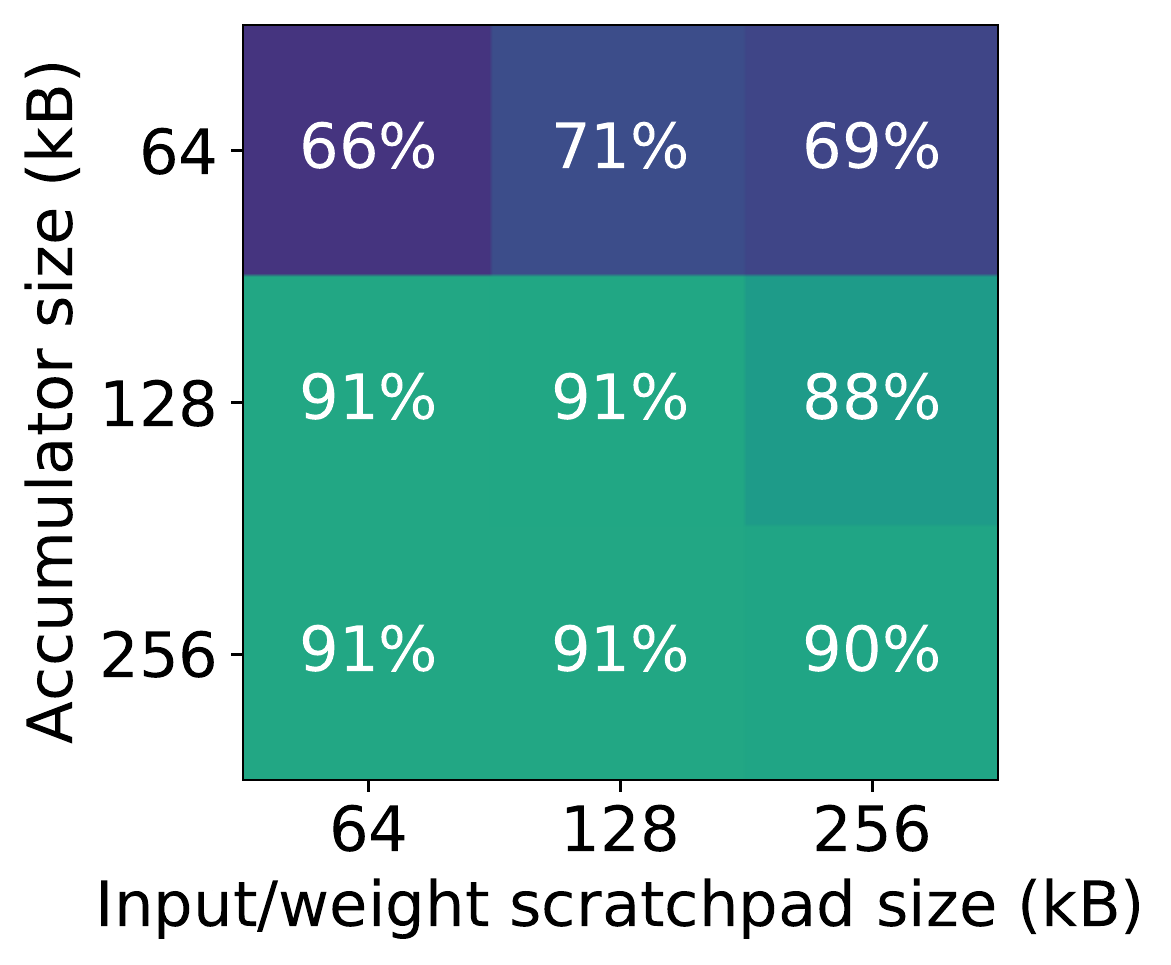}
    \caption{The matmul utilization while performing a BERT-base inference on our baseline CNN accelerator, with different scratchpad and accumulator sizes.}
    \label{fig:scratchpad-heatmap}
\end{figure}

\begin{figure*}[t]
    \centering
    \includegraphics[width=0.29\linewidth,align=c]{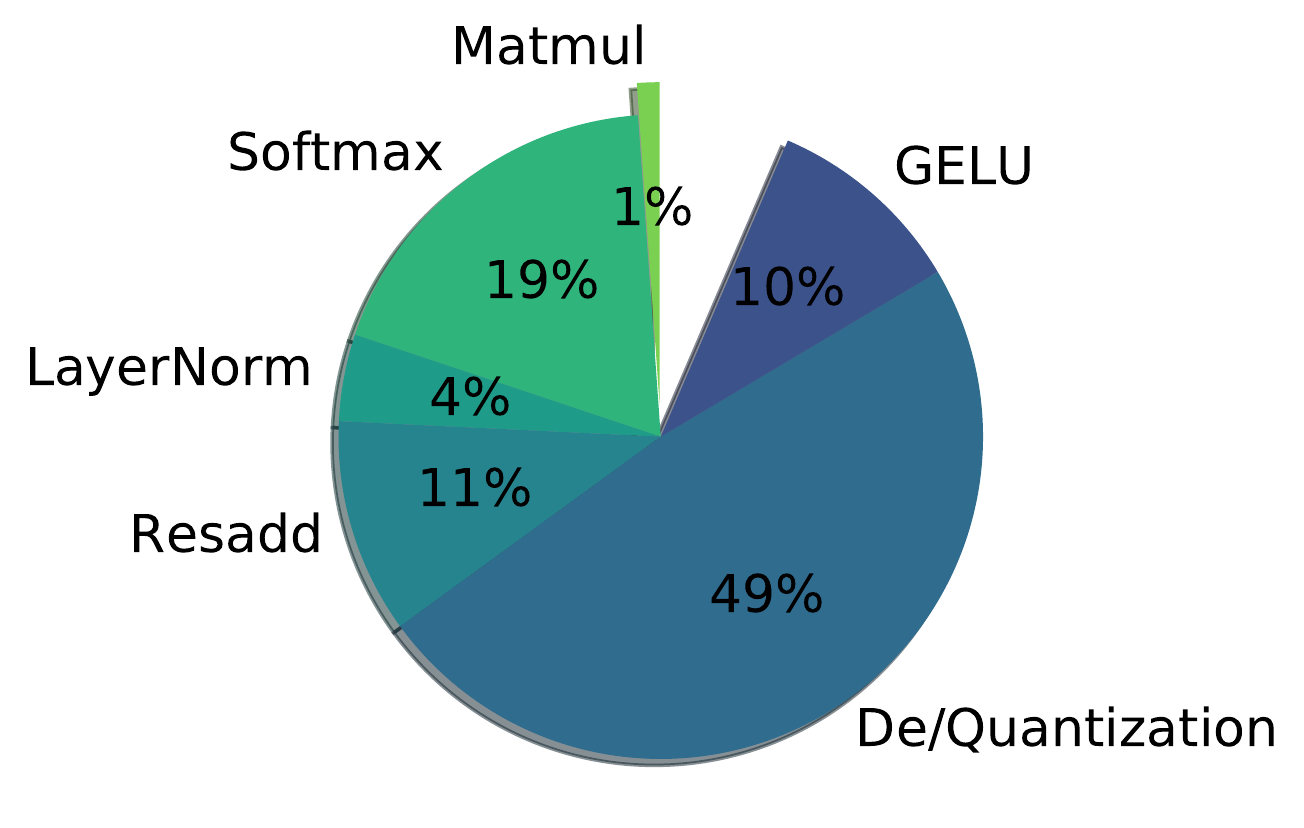}%
    \includegraphics[width=0.29\linewidth,align=c]{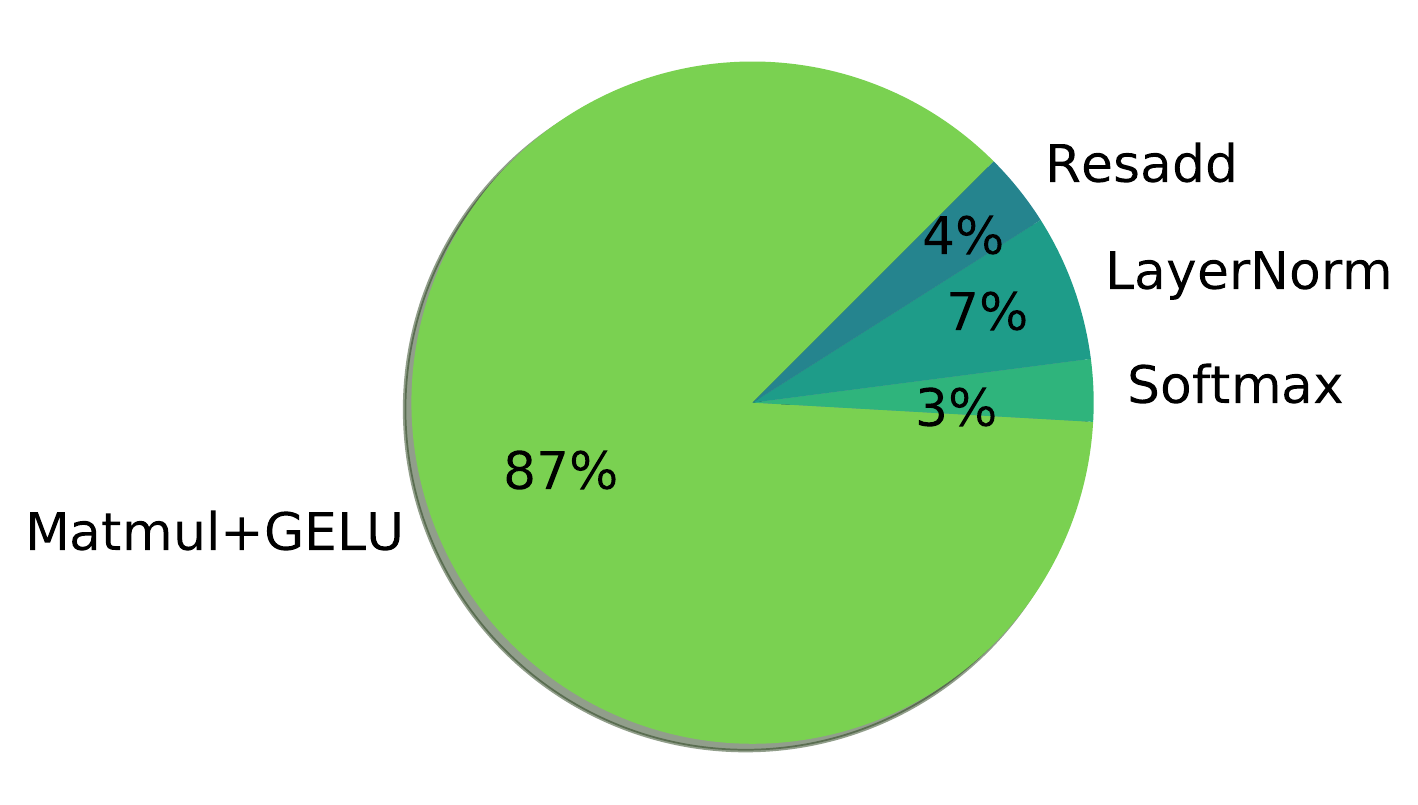}%
    \includegraphics[width=0.4\linewidth,align=c]{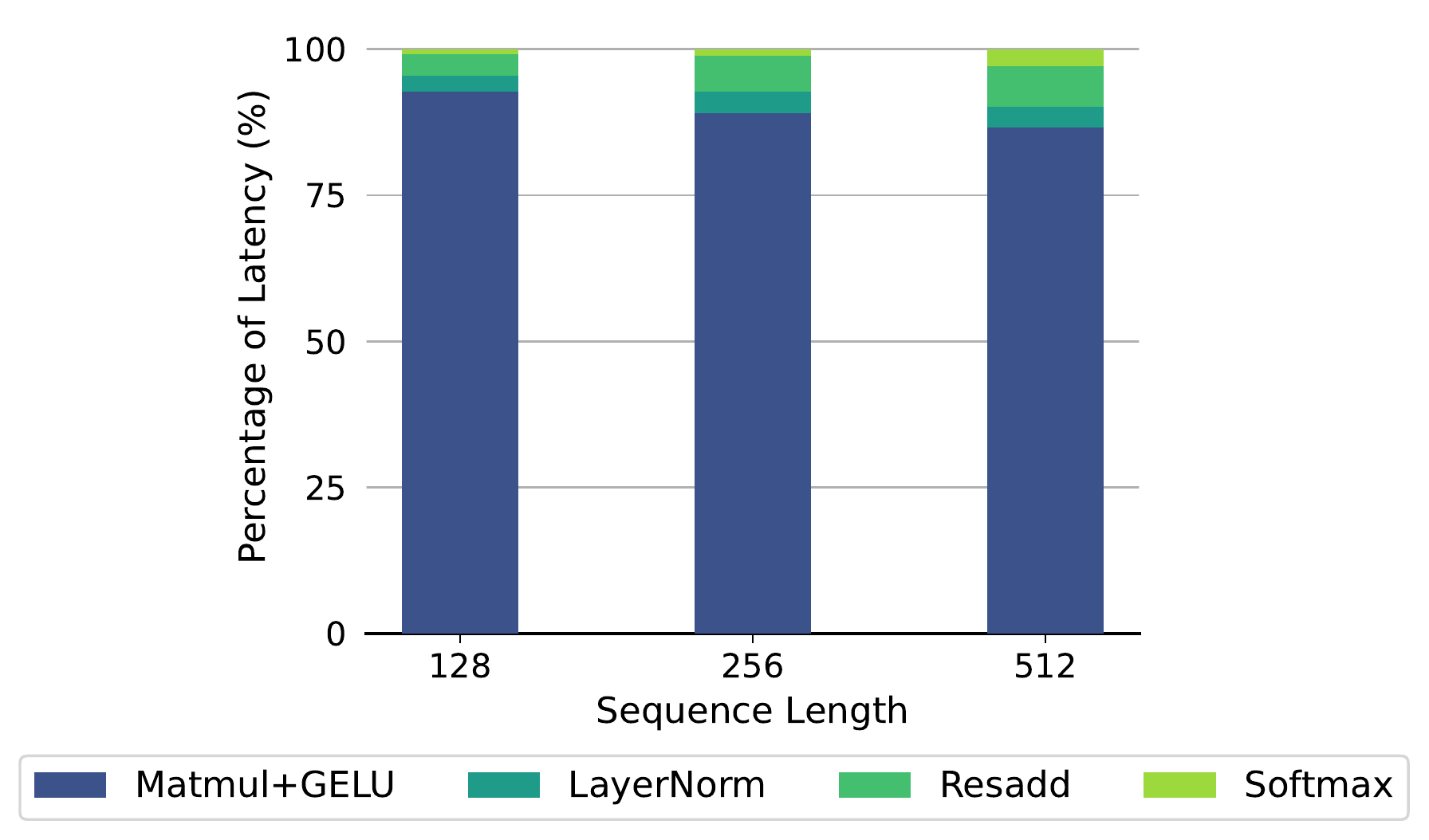}%

    \caption{The time spent on different operations during a BERT inference with a sequence-length of 512, when running on (Left) the baseline CNN accelerator, and (Middle) the accelerator after it has been extended with I-BERT's hardware/software features for Transformers. Note that with I-BERT support, quantization and dequantization operations are no longer required, because all operations happen in the integer format.
    (Right) The time spent on different operations with different sequence lengths after the change.
    For all sequence lengths, the total execution time is dominated by matmuls.
    }
    \label{fig:bert-dac-pie}
\end{figure*}

Our baseline CNN accelerator achieves far less than 1\% utilization of its functional units when performing BERT inferences.
Although individual matmuls achieve 74\% utilization, operations that the accelerator doesn't support natively, such as LayerNorm, significantly reduce performance as they must be performed by the CPU instead.
In fact, Fig.~\ref{fig:bert-dac-pie} shows that 96\% of execution is spent on non-matmul operations.
Note that over 99\% of FLOPs in our Transformer inference are MACs for matmuls, so the time consumed by each operation in the baseline accelerator's run is far from the theoretical ideal, unless further optimizations are made.

Furthermore, our baseline accelerator offloads GELU and Softmax operations to the host CPU, which performs them with floating-point units.
As shown in Fig.~\ref{fig:bit_precision_comparison}, floating-point adders or multipliers consume orders of magnitude more energy than the integer counterparts.
In our baseline CNN accelerator, matmuls are performed with INT8 inputs, but these must be dequantized and requantized in between matmul operations for floating-point activations to be performed on the CPU, further contributing to the energy and latency~overhead.

Finally, a specialized hardware accelerator's memory hierarchy must often be carefully tuned based on the workloads running on it.
CNNs primarily perform convolutions,\footnote{Note that some CNN operations, such as ``depthwise convolutions'' in models such as MobileNet~\cite{sandler2018mobilenetv2}, may also suffer from lower arithmetic intensities, but these operations are found in only a subset of state-of-the-art CNNs, and often constitute only a small portion of the total runtime of a vision model.}
which have very high arithmetic intensity, while Transformers primarily perform matmuls, often with small and/or rectangular matrices, with significantly lower arithmetic intensities and different optimal tiling strategies.
For example, we observe the low arithmetic intensities of the MHA module in Tab.~\ref{table:per-layer-encoder-12heads}.
This suggests that the memory hierarchy and memory bandwidth of our baseline CNN accelerator should be re-tuned for more efficient Transformer inference.

\subsubsection{\textbf{Memory Hierarchy}}
\label{sec:hardware:memory-hierarchy}

Transformer matmuls (in particular, the act-to-act matmuls) often have very different shapes and arithmetic intensities than the convolutional layers in CNNs, as also illustrated in Tab.~\ref{table:per-layer-encoder-12heads} and \ref{table:resnet50-flops}. 
As illustrated in Fig.~\ref{fig:scratchpad-heatmap}, simply adjusting the sizes of the input/weight scratchpad and 32-bit partial accumulator significantly improves the performance of BERT's matmul operations.
Larger accumulators enable higher output-reuse, which is more suited for several of the matmuls in Transformers. The query $\times$ key matmuls in particular have $l \times l$ output activation matrices, which for long sequence lengths are much larger than the $l \times d/h$ input query and key matrices.
Increasing the accumulation buffer size therefore allows for improved output reuse with these operations.

Given this observation, we reduce the size of our baseline accelerator's shared input/weight scratchpad to 64 kB from 256kB, and we increase the size of the partial-sum accumulator to 256 kB from 64kB.
This involves no increase in the total SRAM capacity and virtually no change to the total area of our accelerator. However, these changes yield a much more substantial 36\% reduction in total matmul latency.

\subsubsection{\textbf{Hardware-Software Co-Design}}
\label{subsec:hw-sw-codesign}

As described in Sec.~\ref{sec:sysmodel}, matmuls are the dominant kernel in Transformer workloads, but even after maximizing matmul performance on our baseline CNN accelerator, it still fails to achieve above 1\% utilization.
This is due to the overhead of CPU-offloaded non-linear operations.
Fig.~\ref{fig:bert-dac-pie} demonstrates that this is because only 1\% of time is actually spent on matmuls.
The rest is spent on floating-point non-linear activation, normalizations, or on quantization and dequantization operations, since they are offloaded to the CPU.

To alleviate the overhead of runtime quantization and dequantization, we switched our baseline Transformer workload from a naive BERT implementation, where only matmuls are quantized, to an integer-only BERT variant called I-BERT~\cite{ibert}.
More details on quantization and I-BERT will be revisited in Sec.~\ref{subsec:quantization} and~\ref{sec:transformer-specific-optim}, but the main idea of I-BERT is to replace floating-point nonlinear operations such as GELU and Softmax with integer polynomial approximations such that they are both faster and cheaper to implement in specialized hardware accelerators.

To incorporate I-BERT, we added new integer implementations of I-BERT's GELU, LayerNorm, and Softmax variants to our baseline CNN accelerator.
The 32-bit matmul results resident in the accumulator are fed into a newly added ``normalization unit'' which computes sums, sums-of-squares, maxes, and other reductions which are used by LayerNorm and Softmax.
Multiple passes of accumulator-reads are required to compute all the reductions in these operations.
For example, a sum is computed first before a variance is computed using that sum.
Afterwards, the matmul results in the accumulators are read one final time to be fed into a set of 16 activation units which compute I-BERT's GELU, LayerNorm, or Softmax variants in parallel.

With these new features, overall end-to-end BERT inference performance improved by 39.6$\times$ over the baseline accelerator's initial performance.
As Fig.~\ref{fig:bert-dac-pie} illustrates, the computational bottleneck once again became the matmuls rather than normalization or activation functions; and this trend persists across different sequence lengths.
Quantization and dequantization no longer become necessary, since the non-linear floating-point operations are replaced with I-BERT's integer polynomial approximations.
Also note that GELU operations can now be trivially fused with the preceding matmuls, so that they become one pipelined operation.
When synthesized with the ASAP7 PDK~\cite{asap7}, the new hardware units increased the total area consumption of the accelerator by only 14\%, and the GELU, LayerNorm, and Softmax operations increased the power consumption of a BERT inference by only 9.3\%.\footnote{Note that the ASAP7 PDK does not include energy models for SRAMs, so to derive the total energy consumption of our accelerator, we used the SRAM energy estimates in Accelergy's~\cite{wu2019accelergy} CACTI~\cite{cacti} plugin, and scaled them for 7nm.}

To summarize, as shown in Sec.~\ref{sec:sysmodel}, the nonlinear operations do not necessarily add much to the total FLOPs, area, or power consumption of Transformer accelerators in the ideal case. 
However, this may not be the case in practice, especially if the computation is offloaded to a CPU, leading to a non-trivial latency impact. 
We demonstrated that this can be addressed using the I-BERT implementations of LayerNorm, Softmax, and GELU, which only increases the area of a Transformer accelerator by 5-15\%, and adds 8\% to the total latency.

\begin{boxA}
\textbf{Summary (Sec.~\ref{sec:thiswork}. Accelerator for Transformers Case Study):} The baseline Gemmini accelerator designed for CNN architectures uses 8-bit integer weights, and it has an accumulator for partial sum storage as well as a small scratchpad that overflows to an external L2 cache. The performance of running Transformers on the baseline accelerator suffers for a number of reasons.
    \begin{itemize}[leftmargin=5mm]
        \item The bottleneck non-matmul operations running on the CPU takes 96\% of total runtime;
        \item Activation functions performed in floating point require repeated dequantization and requantization; and
        \item The lower arithmetic intensity nature of transformer inference is more sensitive to non-optimized memory hierarchy.
    \end{itemize}
To address these issues, we:
    \begin{itemize}[leftmargin=5mm]
        \item Reduced scratchpad capacity in favor of an increase in accumulator size, which enabled higher output reuse and improved memory efficiency;
        \item Switched to I-BERT, an integer version of BERT that approximates floating point activations, eliminating quantization overhead; and
        \item Added special normalizer units and activation units that offload GELU, LayerNorm and Softmax computations from the CPU.
    \end{itemize}
These changes mitigate the bottleneck on non-matmul operations, and they trade a 14\% area increase for a 39.6$\times$ performance improvement.
An important takeaway is that even though  the nonlinear operations in Transformers have little contribution to the overall FLOPs, area, or power, they can still have a non-trivial impact on latency.
\end{boxA}

\section{Model Optimization}
\label{sec:optimization}
Given a DNN model that has already been designed and trained, one important question is whether it is still possible to \textit{algorithmically} improve the efficiency of the model on the target hardware platform, through the adaptation of the model into a more hardware-friendly format. 
In this section, we discuss popular off-the-shelf model optimization methods, quantization and sparsity (i.e., pruning), in Sec.~\ref{subsec:quantization} and~\ref{subsec:sparsity}, respectively.
Then, in Sec.~\ref{sec:transformer-specific-optim}, we outline Transformer-specific optimization methods to improve the performance of Transformer-specific features such as attentions and nonlinear operations.

\subsection{Quantization} 
\label{subsec:quantization}

\begin{figure*}[!t]
    \centering
    \includegraphics[width=0.41\textwidth]{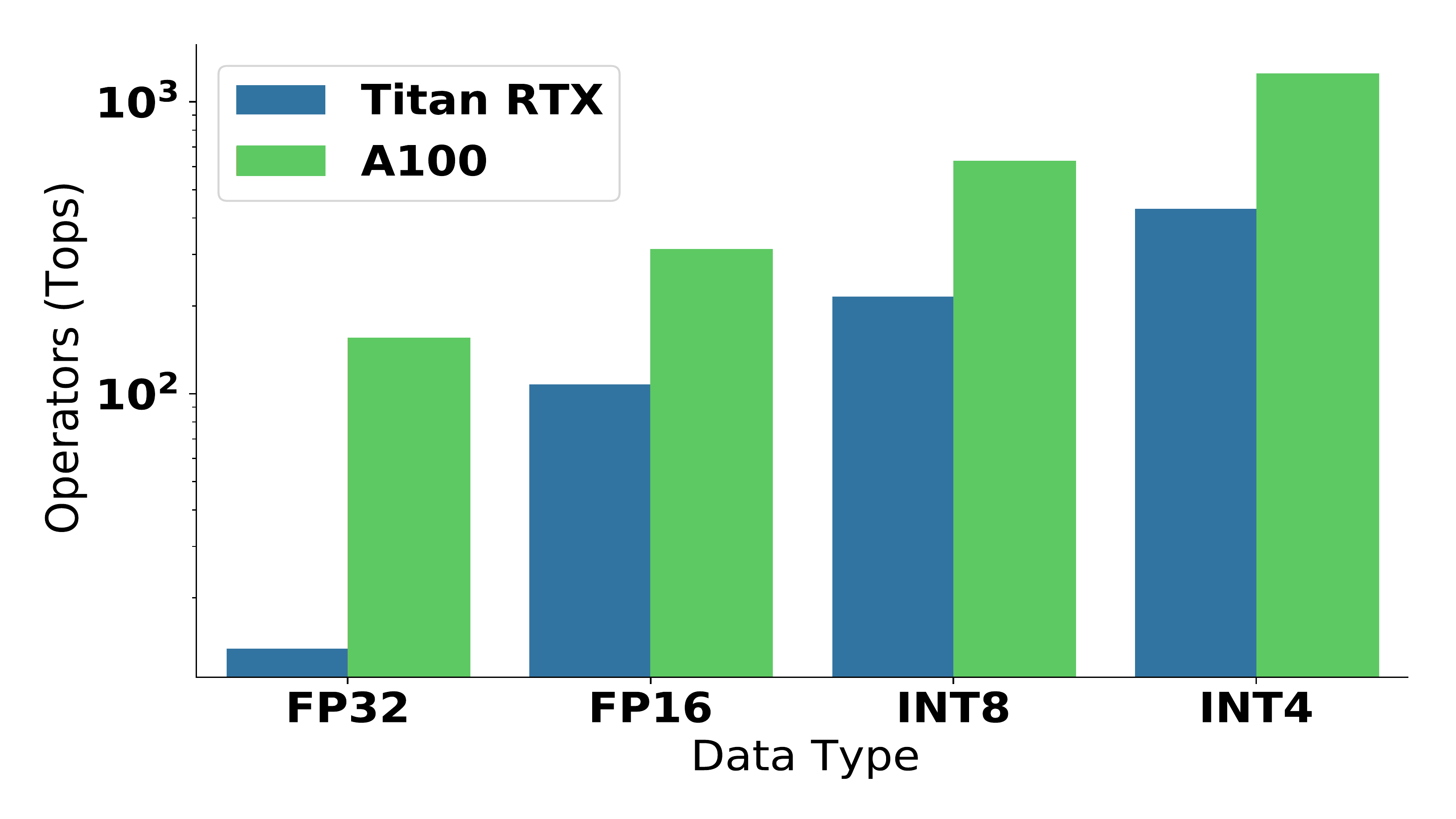}
    \includegraphics[trim=0 72 50 10, clip, width=0.58\textwidth]{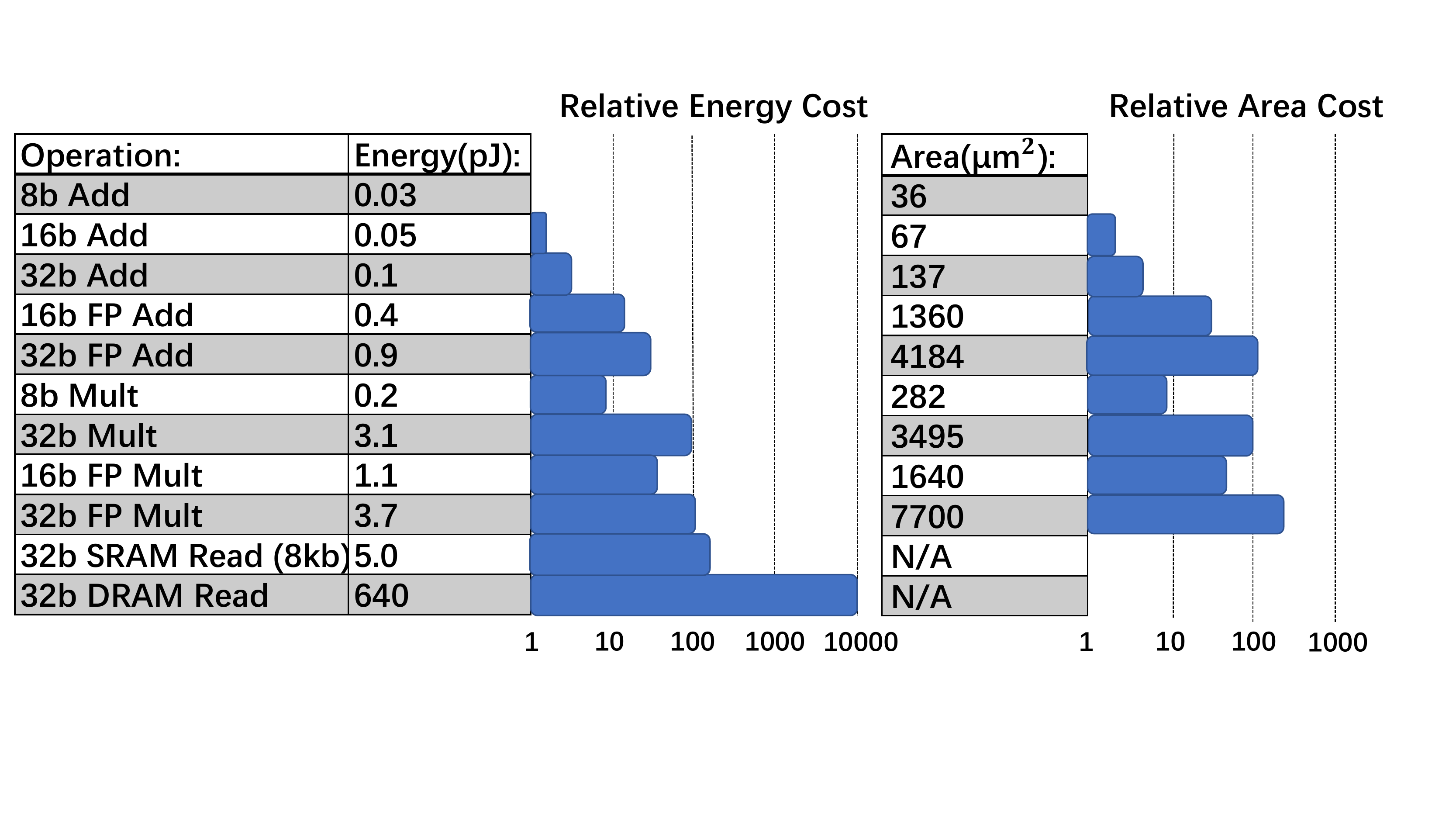}
    \caption{(Left) Comparison between peak throughput for different bit-precision logic on Titan RTX and A100 GPU. 
    (Right) Comparison of the corresponding energy cost and relative area cost for different precision for 45nm technology~\cite{energyproblem}.
    As one can see, lower precision provides exponentially better energy efficiency and higher throughput.
    }
    \label{fig:bit_precision_comparison}
\end{figure*}

DNN models are typically trained using high-precision floating-point computations.
However, high-precision arithmetic is often unnecessary for inference.
\textit{Quantization} is a procedure for compressing DNN models by representing parameters and/or activations with a lower-bit, typically (but not necessarily) fixed-point representation such as 8-bit integer (INT8), 
instead of 32-bit or 16-bit floating~point (FP32 or FP16).

Quantization offers multiple benefits for efficient DNN inference. One obvious advantage of reduced precision is the reduction in memory consumption. For example, quantizing model weights from FP32 to INT8 leads to a 4$\times$ smaller model size. This leads to reduced off-chip storage and bandwidth without any modifications to a DNN accelerator. 
Additionally, quantizing activations further allows for reduced memory traffic and storage for intermediate partial results. 
The memory hierarchy can also be restructured accounting for the precision difference, either by allowing for greater local reuse by storing a larger number of parameters (since each parameter now consumes less storage space), or else by using smaller internal buffers while maintaining the same amount of local data~reuse. 

A second advantage of quantizing model weights and activations is the reduced size, latency, and energy consumption of the ALUs and the corresponding PEs.
In general, floating point ALUs tend to be less efficient than integer ALUs in terms of area, latency, and energy consumption.
This is because floating-point PEs need to multiply mantissas, add exponents, and perform a left shift using the exponent to get the final result when performing a single multiplication operation, whereas fixed-point PEs only require a multiplication unit. 
For this reason, modern GPUs and TPUs often contain INT8 processing paths \cite{gpu,Jouppi2017},
which can significantly benefit from quantization.
For example, as illustrated in Fig.~\ref{fig:bit_precision_comparison}, performing INT8 addition can be $\sim$30$\times$ more energy efficient and $\sim$120$\times$ more area efficient, as compared to the FP32 counterpart.

Another critical application for quantization is model deployment on integer-only hardware.
Some edge processors for low-cost and power-efficient embedded devices such as ARM Cortex-M cores~\cite{armcortexm} and GAP-8~\cite{flamand2018gap} do not include dedicated floating point units.
When deploying models on these processors, not only must the model weights and activations be quantized, but also all computations must be conducted using only integer arithmetic.
Otherwise, deployment is impossible or results in considerable overhead due to the need to process non-integer operations off-chip.
This would lead to additional latency and energy consumption for data transfer to a general-purpose host processor.
This quantization technique of carrying out the entire inference using integer arithmetic is known as \textit{integer-only quantization}~\cite{jacob2018quantization,ibert,yao2020hawqv3,i-vit,kim2022integer}.
We have discussed in Sec.~\ref{subsec:hw-sw-codesign} that integer-only quantization reduces the end-to-end inference latency by $39.6\times$ on Gemmini.

Quantization methods can broadly be categorized into uniform and non-uniform quantization, depending on how they map the values.
\textit{Uniform quantization} splits the floating-point domain into evenly spaced intervals and maps each interval into a single fixed point value.
This can be obtained from a simple arithmetic rule:
\begin{equation}
    Q(r) = \mathrm{Int}(r / s) + Z,
\end{equation}
where $Q$ is the quantization operation, $r$ is the floating point value, $S$ is a scaling factor, and $Z$ is a shift factor.
\textit{Non-uniform quantization}, on the other hand, does not require the intervals to be evenly spaced.
By assigning more quantization bins to important regions, generally resulting in improved compression rates, non-uniform quantization can more accurately capture the original data distribution in the floating point domain than uniform quantization.
However, it is typically more challenging to efficiently deploy non-uniformly quantized models on general computation hardware.
As a result, uniform quantization is currently the de-facto method for its simplicity and efficient mapping to hardware.

While lower bit quantization can lead to a better compression rate, reducing the precision too aggressively can significantly degrade the model accuracy.
It is therefore crucial to achieve a balance between performance gains through reduced precision and maintaining model accuracy.
One promising strategy for alleviating this issue is \textit{mixed-precision quantization}.
It is known from previous work~\cite{shen2020q,wang2018haq,dong2019hawq,wu2018mixed} that different layers in a model exhibit different sensitivity to quantization, and that it is critical to assign higher bit precision to the more sensitive layers.
Notable works for quantizing Transformers with mixed-precision include Q-BERT~\cite{shen2020q} that uses the Hessian information (i.e., curvature) as a proxy for sensitivity, and HAT~\cite{wang2020hat} that applies reinforcement learning (RL) to learn the appropriate bit precision per layer.

Another challenge with quantizing pre-trained Transformer models is the presence of outliers in activations~\cite{kovaleva2021bert}. 
Uniform quantization, which attempts to divide the range from the minimum possible value to the maximum possible value into multiple bins, can result in significant performance degradation. 
This is because more values are mapped to the same quantized value (i.e., resolution degradation) due to the outliers that extend the interval of each quantization bin.
While non-uniform quantization can be a solution to circumvent the outlier issue~\cite{gobo}, a uniform quantization scheme that assigns larger bit precisions to activations containing outliers has been proposed as well~\cite{dettmersgpt3}. 
Furthermore, the recently introduced FP8 precision~\cite{micikevicius2022fp8}, which provides extra degrees of freedom in setting the exponent bit precision, has been found to be a suitable solution for quantizing models whose integer quantization results in reduced accuracy due to the presence of outliers~\cite{kuzmin2022fp8}.

For more information about this topic, see Section III-F of \cite{hardwaresoftwaresurvey}, Section IV-C-3 of \cite{compressionandacceleration}, and Section 3.1 of \cite{approximatecomputing}, as well as \cite{quantizationmethods} for a more comprehensive survey of software-level approaches.

\begin{boxA}
\textbf{Summary (Sec.~\ref{subsec:quantization}. Quantization):}
Quantization is a way of compressing DNN models by reducing the precision of model parameters and/or activations. 
The immediate benefit of quantization is reduced memory consumption, which allows reduced off-chip storage and bandwidth, and a more efficient memory hierarchy design. 
Furthermore, quantization can reduce the size, latency, and energy consumption of the ALUs and the corresponding PE via low-bit precision arithmetic. 
In some cases, quantization also makes it possible to deploy DNN models in integer-only hardware units, which otherwise may be impossible or may incur considerable overhead for offloading non-integer operations off-chip. 
While many DNN models are robust to a certain level of quantization noise, certain algorithm-level advancements are necessary to prevent accuracy degradation with lower-bit precision (e.g., INT4 or even less).
In particular, special considerations must be taken for quantizing pre-trained Transformers without accuracy degradation as they are known to have outlier activations.
\end{boxA}

\subsection{Sparsity}
\label{subsec:sparsity}

Another common avenue for reducing the overall number of computations required for deep learning inference is through introducing sparsity. 
Sparsity (also known as pruning) is a procedure of making DNN models sparse by removing those redundant/insensitive parameters.
While it has been observed that having a dense model may be necessary to successfully train a model, it is also possible to remove many of the parameters after the model has been trained, without any quality degradation.
It is known that training large models and then compressing them via pruning achieves better accuracy then training a compressed model from scratch~\cite{li2020train}.
This may be due to the fact that having redundant parameters from the beginning of the training may make the loss landscape easier to optimize~\cite{liu2022loss}; or it may be related to the increase in the likelihood of obtaining a ``lottery ticket''~\cite{frankle2018lottery}.

Broadly speaking, pruning can be categorized into two branches: unstructured pruning; and structured pruning.
\textit{Unstructured pruning} allows arbitrary patterns of sparsification for parameters and feature maps.
It can, in theory, produce significant computational
savings without accuracy degradation~\cite{ZCM20_quantized_TR}.
However, unstructured pruning can be challenging to leverage effectively in hardware. 
In order to store the data effectively without storing the null (i.e., zero) parameters, a compressed memory format is necessary. 
Additionally, the computation units must be adjusted to be able to operate directly on the compressed data.
Otherwise, the parameters must be decompressed before computations and then re-compressed afterward, leading to additional overhead.
For these reasons, commodity DNN accelerators might not efficiently exploit unstructured sparsity patterns.

\textit{Structured pruning} circumvents these limitations by strictly removing structured sets of parameters. 
For instance, in Transformers, rows and columns in linear layers, attention heads~\cite{michel2019sixteen}, or even entire layers~\cite{sajjad2020poor,fan2019reducing} can be structurally pruned.
Recent work has further integrated the structured pruning of different architectural components into a single framework  (e.g., pruning attention heads in MHA modules and filters in FFN modules together)~\cite{hou2020dynabert,liu2021ebert,kwon2022fast,xia2022structured}.
Such structured pruning methodologies immediately lead to dense matmuls that are smaller than the original, 
eliminating the need for a compressed memory format or special hardware support to gain memory reduction and latency improvement.
However, the compression rate might not be as good as with unstructured pruning. 
It has been shown that a state-of-the-art unstructured pruning method~\cite{kurtic2022optimal} can prune up to 90\% of the parameters in BERT~\cite{devlin2018bert} without any performance drop on the MNLI benchmark~\cite{williams2017broad}, whereas the same performance can only be achieved with a state-of-the-art structured pruning method~\cite{xia2022structured} by pruning up to 70\% of the parameters.

While the aforementioned pruning methods belong to \textit{weight pruning},\textit{ activation pruning} (i.e., \textit{dynamic pruning}) can also be applied to dynamically detect and zero out unimportant activations at run-time.
In Transformer inference, a popular branch of activation pruning is token pruning~\cite{goyal2020power,wang2021spatten,kim2020length,kim2021learned,marin2021token}, which detects and drops less important tokens in each Transformer layer from the rest of the inference.
The underlying rationale is that not all tokens (e.g., words in NLP tasks) are necessary to understand the meaning of the input sequence.
By reducing the sequence length that Transformers need to process, these methods have demonstrated a reduction of up to $\sim30-50\%$ in the total number of computations required, without causing a noticeable drop in accuracy in NLP benchmarks~\cite{rajpurkar2016squad,wang2018glue}.
However, accelerating such dynamic sparsity patterns can be a challenge, as it requires detection logic to determine the location of nonzeros on-the-fly.
Therefore, in many cases, dynamic sparsity requires designing algorithm and hardware~together.

Regardless of the pruning methodologies used, the primary concern is determining which weights should be preserved and which should be removed in order to improve the efficiency of the neural network without sacrificing its performance.
Common methodologies for pruning Transformers include the following:
\begin{itemize}[leftmargin=5mm]
    \item \textit{Magnitude pruning}~\cite{gale2019state} is a technique that uses the absolute value of each weight as a proxy for its importance.
    It prunes the weights with the smallest magnitudes during the training process. 
    The rationale behind this approach is that smaller weights contribute less to the model's final outcome.
    \item \textit{Movement pruning}~\cite{sanh2020movement,lagunas2021block} is a technique that takes into account the changes in weights during fine-tuning, assigning a larger importance score to the weights that move further away from zero as the fine-tuning process progresses.
    This technique has been found to be more effective than magnitude pruning for models that are trained using the pre-training and fine-tuning scheme (e.g., BERT~\cite{devlin2018bert}), as it better captures the importance of weights as the fine-tuning process progresses.
    \item \textit{First-order pruning}~\cite{michel2019sixteen} uses gradients with respect to the loss that flow into the weights or a group of weights as a proxy for evaluating the importance of the model accuracy. 
    This approach considers the gradient to be an indicator of the impact of zeroing out a parameter on the loss. 
    This scheme was further improved~\cite{hou2020dynabert}, where the product of weight magnitude and gradient was used as a proxy for importance, as it may be a more accurate estimate of the impact of zeroing out weights.
    \item \textit{Second-order pruning}~\cite{kurtic2022optimal,kwon2022fast,yu2022hessian} uses the Hessian matrix of the weights or a group of weights with respect to the loss as a proxy importance metric. 
    Compared to the first-order information, the second-order information is generally known to be a more accurate indicator of the effect of removing weights.
    However, due to the large size of the Hessian matrix, which grows quadratically with the number of weights, it is necessary to employ an appropriate and scalable approximation, often with algorithms from randomized numerical linear algebra~\cite{RandNLA_PCMIchapter_chapter,yao2019pyhessian}.
\end{itemize}

\vspace{2mm}
One of the main advantages of pruning is the reduction in memory footprint. 
The gain in memory efficiency is straightforward with structured pruning, which directly reduces the size and/or number of matrix multiplications.
In contrast, unstructured pruning often requires the use of sparse encodings (also known as sparse storage formats) to compress and store sparse data.
These methods use less memory by employing metadata to encode the positions of the nonzero entries in the matrices~\cite{hardwaresoftwaresurvey,sparseandirregular}.
Sparse encodings can reduce off-chip memory consumption and the corresponding required memory traffic. 
They can also reduce the required storage size on chip, thereby allowing for smaller buffers or, alternatively, increased reuse.
This is because, although the same amount of data can be stored in a buffer, the encoded data corresponds to a greater proportion of the full-sized input tensor.

Pruning can also lead to reduced energy consumption and latency due to the elimination of unnecessary computations. 
Similar to what we described above, this is relatively straightforward to achieve through structured pruning, but unstructured pruning requires special techniques for identifying and bypassing calculations involving null elements~\cite{eyeriss,cambricons,cambriconx,cnvlutin,snapea,scnn}. 
This can involve identifying and skipping individual elements or entire null vectors.
Some detection and skipping methods only save energy by not performing the operation involving the null element.
That is, the PE doesn't have to be used for the null computation, in which case it avoids energy consumption. 
Other methods additionally seek to reduce latency by assigning a different effectual computation to the skipped PE, rather than having them idle for the ineffectual compute cycles.
Furthermore, in order to maintain PE utilization with unstructured sparse matmuls, it may also be necessary to perform load balancing.
Since the distribution of zeros can be unbalanced between PEs, some PEs may require a longer execution time than others, resulting in idle waiting periods of the others.
Several works have used load balancing for accelerating neural networks with unstructured sparsity \cite{eie,adaptivetiling,sanger}.

We refer interested readers to Section V of \cite{sparseandirregular} and Section III of \cite{hardwaresoftwaresurvey} for a more comprehensive overview of sparse encoding methods.  
Additionally, a general overview of hardware architectures that leverage various sparsity patterns is provided in~\cite{sparseandirregular}.

\begin{boxA}
\textbf{Summary (Sec.~\ref{subsec:sparsity}. Sparsity):}
Sparsity (or pruning) is another widely-used method of reducing the inference cost of overparameterized DNN models by removing redundant or less important weights and activations. 
Similar to quantization, pruning helps to reduce off-chip memory consumption and the corresponding memory traffic, as well as energy consumption and latency. 
Pruning can be broadly divided into weight pruning and activation pruning.
Weight pruning can be further divided into unstructured pruning, which allows any sparsity pattern, and structured pruning, which imposes an additional constraint on the sparsity pattern.
While structured pruning can provide benefits in terms of memory, energy consumption, and latency without additional hardware support, it is known to achieve less compression rate than unstructured pruning.
Activation pruning prunes redundant activations during inference, which can be especially effective for Transformer models.
However, this requires support to dynamically detect and zero out unimportant activations at run-time.
\end{boxA}

\subsection{Transformer-specific Optimization Methods}
\label{sec:transformer-specific-optim}

The use of off-the-shelf optimization methods such as quantization and pruning can lead to significant performance advantages. 
 Nevertheless, there are other optimization strategies that are tailored specifically for the Transformer architecture, e.g., by taking advantage of the features within it. 
Here, we review the significant Transformer-specific optimization techniques that can further optimize Transformer inference.

\subsubsection{\textbf{Accelerating Attention}}
\label{sec:accelerating_attention}

Several works aim to optimize the attention mechanism in the MHA module. 
Recall that the time spent performing the matrix multiplications in the MHA module grows quadratically with sequence length for long sequences, as outlined in Sec.~\ref{sec:profiling}.
Therefore, for long sequences, computing attention becomes the dominant portion of the overall runtime. 
One common route for more efficiently computing the attention network is \textit{token pruning}.
This involves removing unimportant tokens so as to reduce the effective sequence lengths, as was discussed in Section \ref{subsec:sparsity}.
The need to efficiently identify and drop unimportant tokens on-the-fly has led to several hardware-software co-design approaches.
In SpAtten~\cite{wang2021spatten}, tokens are ranked based on the amount of attention they are getting from other tokens in the input sentence, and the tokens that are receiving less attention are pruned out. 
This approach is based on the simple rationale that the more a word is attended, the more important it is in the inference process. 
To make this efficient, a top-$k$ hardware engine is employed to filter out the low-importance tokens based on their attention scores.
DTA-Trans~\cite{dta-trans} takes a step further by introducing the two-tiered scheme where in the first round, it determines which tokens should be pruned, and in the second round, it further determines the bit-precision to be allocated to each of the remaining tokens based on their significance.

Another approach to accelerate attention is to leverage the dynamic sparsity patterns of the attention score activations~\cite{a3,elsa,sanger,energon,dota,salo,gradientpruning,speformer}.
It is reasonable to assume that many combinations of query and key tokens are not semantically meaningful, and thus the attention score associated with this combination will be close to zero. 
By taking advantage of this sparsity, the inference accuracy can be preserved while the computational cost is reduced  by avoiding the associated act-to-act matmuls (i.e., query $\times$ key or attention score $\times$ value). 
However, this requires specialized hardware logic to detect and accelerate those dynamic sparsity patterns on-the-fly.
For instance, ELSA~\cite{elsa} proposes a datapath that approximates the angular similarity between key and query vectors, thus allowing the prediction of whether their dot product is likely to be zero. This approach enables the pruning of less important key vectors with respect to a given query vector in advance.
The Sanger framework \cite{sanger} suggests quantizing the query and key values prior to computing the attention score, as this will zero out insignificant entries in the resulting attention score that would have been close to zero if those values were not quantized.
Similarly, DOTA~\cite{dota} proposes to approximate the attention score entries to be zeroed out by employing the matrix multiplication of low-rank (and hence smaller) projections of the query and key values as a proxy.
LeOPArd \cite{gradientpruning} uses bit-serial computing for the query $\times$ key multiplications in order to terminate computation early if it will not reach the pre-determined threshold.

It is worth noting that hardware support is essential for accelerating attention mechanisms, as it enables operations such as top-$k$~\cite{wang2021spatten,dta-trans}, angle approximation~\cite{elsa}, clustering~\cite{speformer}, and multi-precision computation~\cite{sanger,energon,approximatecomputing,dota} that are necessary to detect the dynamic sparsity patterns of attention scores.
Furthermore, specialized hardware support is needed to take advantage of the (mostly unstructured) dynamic sparsity for skipping computations. 
For example, Sanger~\cite{sanger} uses load rebalancing through splitting and packing, and it is equipped with a custom datapath that provides support for both sampled dense-dense matmuls and sparse matmuls.

\vspace{2mm}
\subsubsection{\textbf{Nonlinear Operations}}
\label{subsec:nonlinear}

As discussed in Sec.~\ref{subsec:nonlinear_ops_intro}, the Transformer architecture contains multiple nonlinear functions that pose multiple challenges in efficient hardware design.
Incorporating a hardware module specialized for computing these operations may be a viable solution.
However, this may incur a considerable overhead for hardware design, particularly when targeting low-end devices.
Therefore, various solutions have been proposed to circumvent this issue, without constructing a dedicated hardware~module.

One popular solution is \textit{function approximation}~\cite{pervectorquant,softermax,ibert,wang2021spatten,i-vit}, which seeks to approximate the exact value of the nonlinear function, in order to obtain a good yet computationally efficient approximation.
For instance, Keller et al.~\cite{pervectorquant} uses the Softermax~\cite{softermax} function, which uses a base-2 approximation that switches the base used in the exponential  calculation of the Softmax operation from $e$ to 2, allowing for simplified hardware implementations. 
Softermax ~\cite{softermax} also incorporates online normalization~\cite{onlinenormalizer}, thus reducing the number of passes required for the numerically stable Softmax computation from 3 to 2. 
I-BERT~\cite{ibert} provides a more general approximation algorithm that approximates the nonlinear operations with 2nd-order polynomials.
This not only simplifies the operations, but it also enables them to be performed using only integer arithmetic.
SpAtten~\cite{wang2021spatten} takes a similar approach to use a 5th-order Taylor approximation for computing Softmax, as described in~\cite{taylorseries}.
I-ViT~\cite{i-vit} further extends this idea to use hardware-friendly bit shifting operation to efficiently compute the nonlinear operations for vision Transformer inference.
While the major focus has been approximating the exponential operation for the Softmax, other works \cite{edgebert,logsumexp-and-softmax,highspeedlowcomplexity} have also exploited the log sum-exp trick to avoid the division operation, another operation that can be complicated to implement in hardware~\cite{logsumexp}. 

Another widely-adopted approach is \textit{lookup tables}, which store pre-calculated output values for a given range of inputs.
In this case, if the input is given, the corresponding value stored in the lookup table is outputted, eliminating the need for evaluating the function.
The use of lookup tables to accelerate the nonlinear function is by no means a new concept, with its root predating the advent of Transformer or DNN architectures~\cite{thomas2004libm,detrey2005parameterized}.
Recent approaches, therefore, have  more focused on reducing the size of the lookup table to save area and latency. 
For instance, $A^3$~\cite{a3} decomposes the exponential operation into a multiplication of two smaller-precision exponential operations, allowing one larger lookup table to be replaced with two smaller ones.
NN-LUT~\cite{nnlut} approximates the nonlinear operation using a single-hidden layer network and stores a numerical approximation of the network in a lookup table, thereby avoiding the need for executing the network.

\subsubsection{\textbf{Accelerating Decoding}}
\label{subsec:accelerating_decodin}
As discussed in Sec.~\ref{subsec:model-analysis}, Transformer decoding for generative inference can entail a significant inference latency due to the low hardware utilization and arithmetic intensity.
Due to the growing interest in generative tasks due to the recent advancements of Large Language Models~\cite{openai2019chatgpt,brown2020language,chowdhery2022palm,thoppilan2022lamda}, it is critical to optimize the latency for the decoding process.
One avenue to reduce inference latency is to skip unnecessary computation through \textit{early exiting}.
This method dynamically adjusts the depth of the decoder for each token generation by terminating the inference at a mid-layer and making a prediction using the intermediate hidden states, rather than waiting until the end layer.
While being a well-explored technique for encoder models~\cite{xin2020deebert,schuster2021consistent},
CALM~\cite{schuster2022confident} has only recently extended this methodology to decoder models.
A major challenge in decoding tasks is that, unlike in encoding tasks, the generation of one token relies on the activations of all previous tokens, due to the attention mechanism.
If a previous token is exited early, then there is nothing to attend for the skipped layers. 
To address this issue, CALM proposes ``state propagation,'' which copies the activations of the final layer before exiting to all the skipped layers.
This had a minimal impact on generation quality.

Another recent attempt is to collaboratively use multiple models with different sizes~\cite{kim2023big,chen2023accelerating}. 
The underlying motivation is that the majority of simple word generation can be offloaded to a faster, less accurate model with a smaller size. 
Once in a while, when the small model is unable to predict a word accurately, it switches the control to the larger model for more accurate prediction.
This approach not only enables the execution of the large model to be carried out less frequently, but it also enables its non-autoregressive (i.e., token-level parallel) execution since it can consume all tokens generated from the small model and process them in parallel, thereby utilizing hardware more efficiently.
Big Little Decoder~\cite{kim2023big} has shown $\sim$2$\times$ inference latency reduction across various models and generative tasks without compromising generation quality.

\subsubsection{\textbf{Selecting Which Optimization Methods to Use}}

So far, we have discussed various optimization techniques that can be applied to the Transformer architecture.
It is important to note that a significant portion of these optimization methods depends upon the underlying hardware support. 
Thus, when selecting which optimization techniques to employ, it is essential to adopt a holistic view of both the hardware and software stack, taking into account the characteristics of the underlying hardware. 
In particular, whether the accelerator supports MHA and FFN modules in the \textit{same datapath} versus containing \textit{separate datapaths} for each of these modules can have a significant impact on the optimizations that can be performed.

Accelerators with a unified datapath tend to pursue more general optimizations that can either be applied to both MHA and FFN modules, or at least those that do not require altering the datapath such that it can no longer compute the other modules. 
For example, several accelerators that support both MHA and FFN modules employ general static pruning methods for weight matrices \cite{edgebert,optimus,cooptimizedframework}, but do not aim to exploit attention-specific pruning methods such as dynamic sparsity.
However, more exotic optimizations can be pursued separately for the MHA and FFN modules, if they are computed in separate datapaths or if the PEs can be reconfigured. 
For instance, FABNet~\cite{adaptivebutterfly} exploits static sparsity patterns that can only be applied to the FFN module by adopting separate datapaths for the MHA and FFN modules.
FTRANS~\cite{ftrans} also applies different optimization methods for the MHA and FFN modules by incorporating reconfigurable PEs that can handle both workloads without having two separate datapaths.
However, employing separate datapaths or reconfigurable PEs can incur an additional overhead, as compared to using a general, unified datapath. 
Consequently, there is a trade-off to consider between the area overhead and the performance gain derived from the use of more aggressive optimizations.

\begin{boxA}
\textbf{Summary (Sec.~\ref{sec:transformer-specific-optim}. Transformer-specific Optimizations):}
While general off-the-shelf optimization methods can also benefit efficient Transformer inference, 
a great deal of research has been conducted to devise optimization strategies that take advantage of Transformer-specific characteristics. 
One opportunity is to optimize the attention mechanism in the MHA module, whose  runtime cost increases quadratically with sequence length. 
For instance, dynamic pruning has been widely applied to take advantage of the sparse nature of attention score activations. 
Additionally, efficient computation of the nonlinear operations should also be taken into account.
In order to reduce the hardware costs associated with the implementation of dedicated hardware units for nonlinear operations, function approximation, and lookup table methods have been proposed as viable alternatives.
Finally, the model optimization methods should also be aware of the underlying hardware architectures and datapaths.
The use of separate datapaths for the MHA and FFN modules can have higher area overhead, but can enable more aggressive optimization as compared to using a single datapath for both modules.

\end{boxA}
\section{Mapping Transformers to Hardware}
\label{sec:scheduling}
\begin{figure*}[htp]
\centering{
  \includegraphics[trim=0 30 0 0, clip, width=0.97\linewidth]{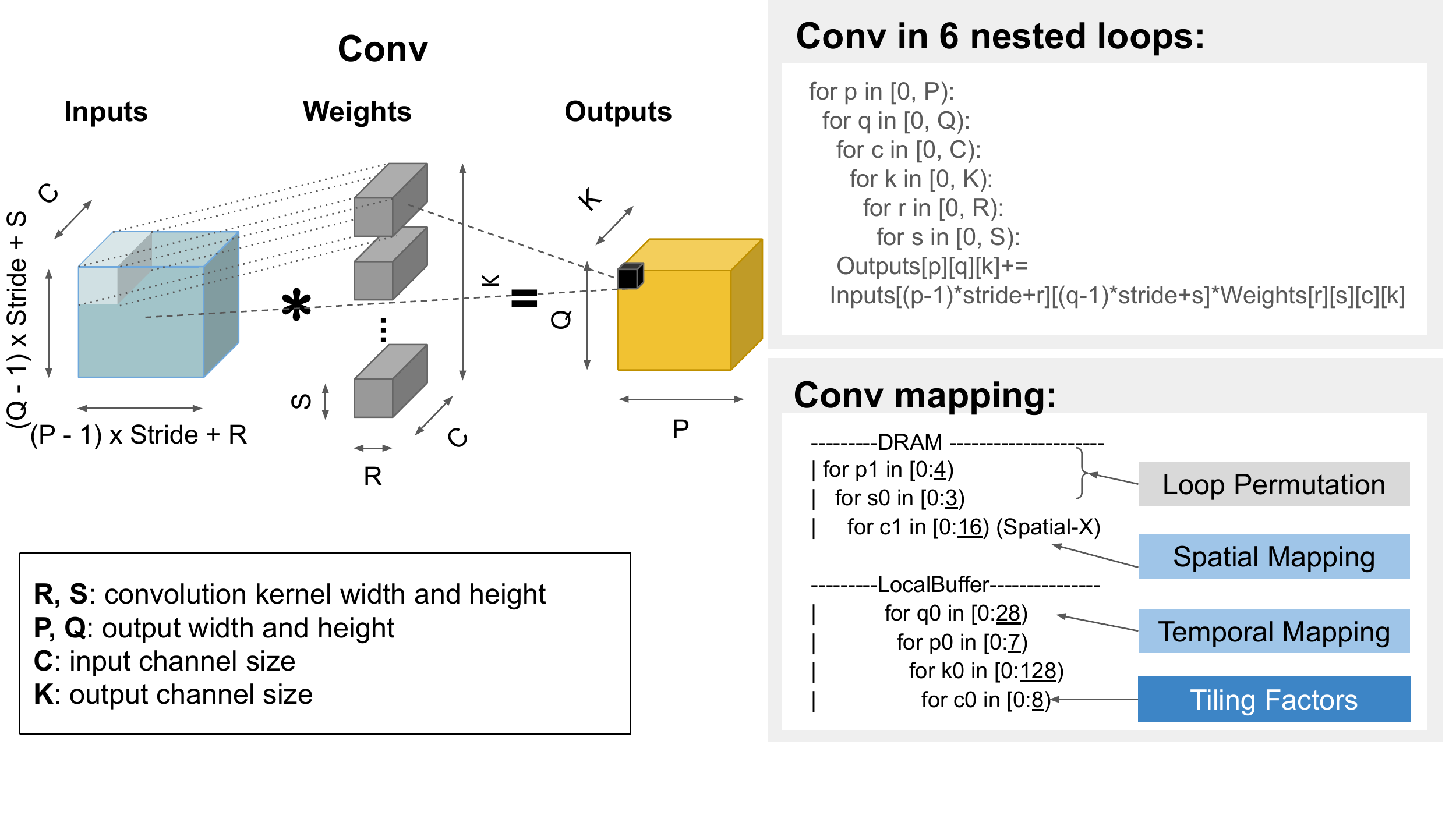}
  \caption{Visualization of mapping for the convolution operation in CNNs onto a typical DNN accelerator. Convolution is represented as a six-nested loop excluding the batch dimension. Loop permutation concerns the order in which each loop level should be executed, with memory accesses to and from either the accelerator's local memory or off-chip DRAM. Spatio-temporal mapping determines which loop level should be executed in parallel using accelerator hardware resources. Tiling factors are the loop bounds of each loop level, where each dimension can be broken down with tiles into several sub-loops. As shown in the example, the input channel size dimension $(C)$ is tiled with a tiling factor of 8, hence into two sub-loops with loop variables $c_0$ and $c_1$. }
  \label{fig:conv_mapping}
  }
\end{figure*}

\begin{figure*}[htp]
\centering{
  \includegraphics[trim=0 100 0 0, clip,width=0.97\linewidth]{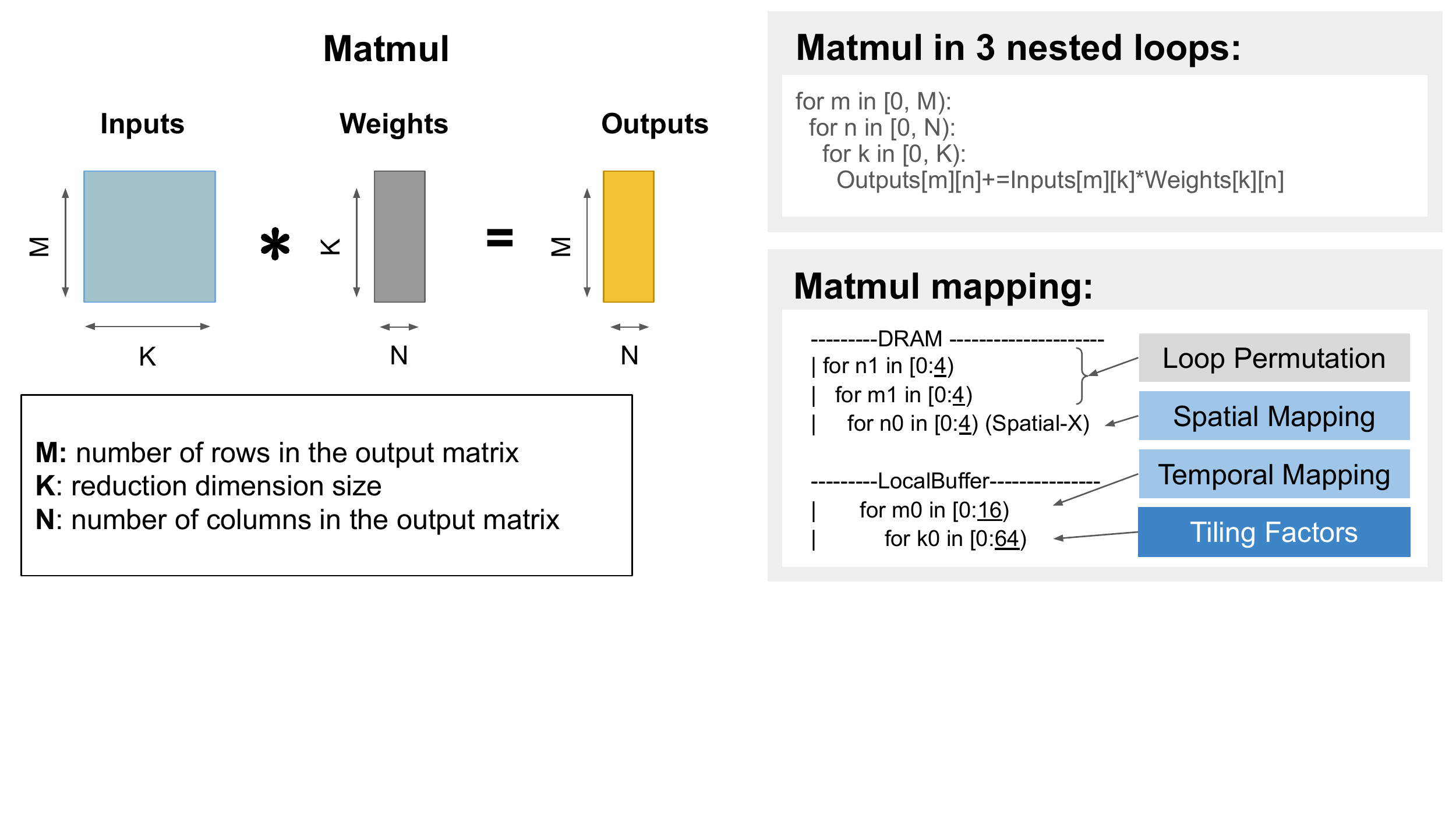}
  \caption{Visualization of mapping for the matmul operation in Transformer encoder/decoders onto a typical DNN accelerator. Matrix multiplication is represented as a three-nested loop. Loop permutation concerns the order in which each loop level should be executed, with memory accesses to and from either the accelerator's local memory or off-chip DRAM. Spatio-temporal mapping determines which loop level should be executed in parallel using accelerator hardware resources. Tiling factors are the loop bounds of each loop level, where each dimension can be broken down with tiles into several sub-loops. As shown in the example, the output column dimension $(N)$ is tiled with a tiling factor of 4, hence into two sub-loops with loop variables $n_0$ and $n_1$. As we will discuss in Sec.~\ref{subsec:mapspace-transformers}, even though matmuls have 3 nested loops instead of 6 as in the convolutions, finding an optimal mapping could still be as challenging.}
  \label{fig:matmul_mapping}
  }
\end{figure*}

In order to execute a Transformer block on a target hardware architecture, we must \emph{map} it into hardware instructions that carry out both the required computation and communication. The choices made during the mapping process play a significant role in performance.
However, the size of the space of possible mappings makes finding the optimal mappings difficult, and it requires the use of carefully considered exploration, heuristic, or learning-based~approaches.

In this section, we provide an introduction to the mapping problem in Sec.~\ref{mapping_intro}; and we discuss key mapping decisions for efficient execution of Transformers in Sec.~\ref{mapping_decisions}. We overview the taxonomy of existing mapping techniques in Sec.~\ref{previous_work_mapping} and similarly for techniques to model the performance of different mappings in Sec.~\ref{performance-modeling-mappings}. Finally, we end with Transformer-specific considerations for mappers in Sec.~\ref{transformer_scheduling}.

\subsection{What are Mappings?}
\label{mapping_intro}

A \textit{mapping} or \textit{schedule} is defined as a sequence of hardware instructions to execute a set of operations on the specific target hardware architecture. 
In the case of a systolic array accelerator such as Gemmini, such hardware instructions might include dense matmuls under a specific dataflow and load/stores to move data between off-chip DRAM and local SRAMs. A mapping will list the complete sequence of data and memory instructions, with the end goal of producing source code or compiled binaries that can be executed on hardware.

For some operations, there may exist multiple \textit{valid} mappings, where the validity refers to the guarantee of correctness from executing each mapping. Specifically, different sets of mapping decisions applied to the same problem may result in valid yet different mappings. We refer to the total space of mapping decisions and their resulting mappings as the \textit{mapspace}. Details about individual mapping decisions are discussed in the following Sec. \ref{mapping_decisions}.

It is not surprising that two different valid mappings may exhibit differences in end-to-end performance, when measured with respect to latency, bandwidth, and energy consumption. 
Hence, it is often the goal of a mapping or scheduling framework to obtain Pareto-optimal, valid mappings for a given software operation and desired performance metrics. 
For some operations, finding a good mapping is unnecessary, either because the problem is trivial, as the mapspace is small, or because the operation itself is not a performance bottleneck and does not warrant the effort of judicious scheduling. 
However, in the case of core computational operators in DNNs, including Transformers, the mapping problem is both \emph{challenging} due to large mapspace and \emph{rewarding} due to the potential gains in overall model execution speedup. 

Fig.~\ref{fig:conv_mapping} and \ref{fig:matmul_mapping} illustrate examples of key operators and their possible mappings, for CNNs and Transformers, respectively. 
As shown in the example mappings, each level of the nested loops must be: 
(1) assigned to be executed either with data from DRAM or from local accelerator memory; 
(2) assigned to be executed \emph{spatially} (i.e., in parallel) or \emph{temporally} (i.e., sequentially), if the accelerator contains spatially parallelized compute resources; and
(3) assigned to be executed as one loop or tiled into multiple subloops (and if so, with which tiling factors). In particular, for the case of Gemmini, spatial mapping concerns the decision to assign which loop levels to be executed on the N-by-N systolic array mesh of PEs.

\subsection{What Are the Key Mapping Decisions?}
\label{mapping_decisions}

Mapping occurs in two steps. First, the graph is transformed at a \emph{graph level} into a set of tensor operations. 
This may involve \emph{fusing} successive operations, \emph{sparsifying} tensors, and deciding on appropriate quantization strategies. Then, each resulting tensor operation is \emph{scheduled} in order to transform it into hardware instructions of the appropriate size.

\subsubsection{\textbf{Graph-level}}
\label{subsec:graph-level}

Graph-level scheduling involves decisions that change the structure of the computational graph, rather than simply the execution schedule of tensor operations represented by individual nodes within the graph. Typical changes include the following:

\begin{itemize}[leftmargin=5mm]
    \item \emph{Layer fusion} or \textit{operation fusion} refers to combining multiple layers (e.g., a matmul followed by a normalization layer) into a single tensor operation to be scheduled and run on the accelerator. This reduces \emph{interlayer} communication, as the results of one layer can remain on the chip as input without being written to and later read from main memory, at the cost of \emph{intralayer} communication. 
    As we will see in Sec.~\ref{transformer_scheduling}, layer fusion may not provide
    as much latency improvement, as with CNN architectures, since static fusion opportunities are not
    as straightforward as fusing convolutions with BatchNorm layers.
    For Transformers, it is possible to combine several operations in
    the same kernel, but this may increase  intralayer communication to an extent that renders such approaches infeasible.
    Furthermore, this can also be dependent on the target hardware~platform.
    \item 
    \emph{Dynamic sparsification} of tensors happens when the pruning decisions are made based on the activation maps. 
    Common methods for dynamic sparsification
    includes locality-sensitive hashing to zero out dot products likely to be small.
    This can significantly reduce the number of arithmetic operations required in the operation, as was also discussed in~\ref{subsec:sparsity}.
    Such optimizations are heavily data-dependent as they require access to the activations and, as a result, cannot always be estimated a priori~\cite{kao2022demystifyingNpu}. 
    As a result, relatively few results on sparsity-aware mapping exist, and those that do largely cover operation-level mappings for a given amount of sparsity.
    
    \item \emph{Static sparsification} of tensors happens when the pruning decisions are independent of the activations and are determined statically.
    As was discussed in Sec~\ref{subsec:sparsity}, there are various methods used for static sparsification. 
    In general, structured sparsity results in high speedup, but it also often results in non-trivial accuracy degradation, whereas unstructured sparsity is able to retain accuracy even with extreme sparsity levels, but it is hard to accelerate. Nevertheless, the latter is 
    going to become increasingly more important since it reduces the memory traffic, which is becoming a major bottleneck for power consumption.
\end{itemize}

\subsubsection{\textbf{Operation-level}}
\label{subsec:operation_level}
The \emph{operation-level scheduling} step decomposes tensor operations into a set of tasks to be run on a given architecture. This consists of several different steps, each of which presents a programmer with a decision problem. These include:
\begin{itemize}[leftmargin=5mm]
    \item 
    Dividing the operation into \emph{tiles} that can fit onto different layers of the memory hierarchy; the dimensions of the tiles are a choice (e.g., tile sizes in Fig.~\ref{fig:matmul_mapping}).
    \item 
    Determining the \emph{dataflow} of the computation, i.e., the order that the tiles are executed in and the tensors that are held stationary or moved across the processor. 
    This can be encoded as a loop ordering problem, with the innermost loops corresponding to axes of tensors being held stationary (e.g., any loop permutation in Fig.~\ref{fig:conv_mapping}). 
    \item 
    Deciding which axes to parallelize, and which to run serially, which we refer to as \emph{spatio-temporal mapping}.
    \item 
    Deciding how to \emph{interleave} communication and computation in order to minimize latency.
    For instance, \emph{double-buffering} may divide the scratchpad into two halves, with one half being used by the processor for computation while the other is loaded with data from memory. 
    \item 
    Mapping arithmetic instructions onto \textit{hardware instructions}. For some architectures, this may be as simple as replacing a matmul operation of the appropriate size (achieved by a tiling) with a call to the appropriate ISA (Instruction Set Architecture) instruction. 
    For others, it may involve selecting between different vector instructions, which may affect the decision of which axes to vectorize, and the resulting spatio-temporal mapping. 
\end{itemize}
A more complete description can be found in \cite{li21deepLearningCompilerSurvey}. 

 The choice of points in the mapspace heavily affects performance, by up to several orders of magnitude, as we will discuss in Sec.~\ref{subsec:mapspace-transformers}. 
 For this reason, the goal of a hardware mapper is to select a point in this space to minimize some cost such as energy, energy-delay product (EDP), latency, etc., on a given hardware target.
However, the size of the mapspace renders exploration difficult. For example, considering only tiling, spatio-temporal mapping, and loop ordering (dataflow), the number of possible mappings for a BERT attention layer can exceed $10^{12}$. As a result, the design and selection of mappers have been the subject of significant attention in both theory and practice.

Furthermore, the optimal mapping can significantly differ depending on hardware architecture, and mappings that work well for one set of hardware parameters often perform poorly on others \cite{kao2022demystifyingNpu}. This significantly increases the difficulty of mapping within a codesign context, as one must be computed for every pair of neural network and hardware architecture.

\begin{boxA}
\textbf{Summary (Sec.~\ref{mapping_decisions}. Key Mapping Decisions):} Mapping Transformers to hardware require decisions to be made both at the \emph{graph} and \emph{operator} levels. These decisions range from choosing simple \emph{numerical} or \emph{categorical} parameters to structural modifications to the program being run. The space of decisions required is enormous, growing combinatorially with each possible decision, but selecting a good point in the space can significantly affect performance. 

\end{boxA}

\begin{table*}[t]
\begin{tabular}{cp{0.7\textwidth}}
\toprule
Search Strategy &
Mappers \\ 
\midrule
\midrule
\textit{Brute-force \& Random Approaches:} &
Timeloop~\cite{parashar2019timeloop}, 
dMazeRunner~\cite{dave2019dmazerunner},
Flexflow~\cite{jia2019beyond},
Triton~\cite{tillet2019triton},
Interstellar~\cite{yang2020interstellar},
Marvel~\cite{chatarasi2020marvel}\\
\midrule
\textit{Feedback-based Approaches:} &
AutoTVM ~\cite{chen2018tvm} (\textit{XGBoost}), Ansor~\cite{ZJS+20Ansor} (\textit{beam search}), 
Halide~\cite{ragan2013halide} (\textit{beam search}~\cite{adams2019learning},
\textit{OpenTuner}~\cite{ansel2014opentuner, mullapudi2016automatically}),
FlexFlow~\cite{jia2019beyond} (\textit{MCMC}),
ConfuciuX~\cite{confuciuX-micro2020} (\textit{RL}), 
Gamma~\cite{gamma-iccad2020} (\textit{genetic algorithm}),
Mind Mapping~\cite{hegde2021mind} (\textit{gradient-based search}) \\
\midrule
\textit{Constrained Optimization Approaches:} &
Polly+Pluto~\cite{grosser2011polly, bondhugula2008practical, bondhugula2016pluto+},
Tensor Comprehension~\cite{vasilache2018tensor}, 
Tiramisu~\cite{bagehadi2019tiramisu}, 
IOOpt~\cite{olivry2021ioopt}, 
Analytical characterization~\cite{li2021analytical},
CoSA~\cite{huang2021cosa} \\
\bottomrule
\end{tabular}
\caption{State-of-the-art DNN schedulers for heterogeneous accelerators.
}\label{table:mapper}
\end{table*}

\subsection{Finding Performant Mappings}
\label{previous_work_mapping}
In order to deal with the size of the search space, many 
accelerator-aware mapping techniques~\cite{parashar2019timeloop, yang2020interstellar, gamma-iccad2020, hegde2021mind, huang2021cosa,li2021analytical} and fully-fledged compilers frameworks~\cite{paszke2019pytorch, abadi2016tensorflow, tensorrt, chen2015mxnet, chen2018tvm, sabne2020xla, kjolstad2017tensor} have been developed.
These are briefly discussed below.

\subsubsection{\textbf{Mapping Strategies}}
To deal with the size of the search space, mapping algorithms focus on a subspace of the mapspace, only making decisions about how to perform a subset of steps required to map the network onto the architecture.

\paragraph{\textbf{Graph-level Schedulers}}
Most of existing ML compilation frameworks (e.g., XLA~\cite{sabne2020xla}, TensorRT~\cite{tensorrt}, TVM~\cite{chen2018tvm}, Glow~\cite{rotem2018glow} and CuDNN~\cite{chetlur2014cudnn}) target graph-level optimizations such as operation fusion, resource allocation, graph partitioning, graph rewriting, etc. A large number of operation fusion techniques~\cite{zhang2022full, zhou2020transferable, alwani2016fused} have been developed to optimize the mapping for more data reuse across DNN layers. 
Among these, a few Transformer-specific operation fusion techniques have been proposed~\cite{choiaccelerating, pati2021demystifying}.
In particular, \cite{choiaccelerating} decomposes the Softmax layer and dynamically fuses the GPU kernels for decomposed layers with the proceeding and succeeding matmul layers in the MHA block. 
Relatedly, \cite{pati2021demystifying} shows that fusing LayerNorm layers and composing big matmul from small matmuls are beneficial to transform performance on GPUs. 
In~\cite{kao2021optimized}, an optimized dataflow for DNN accelerators is introduced to efficiently fuse the key matmul and Softmax layers in MHA. To learn about the operation fusion tradeoffs in the Gemmini accelerator, we have performed a case study and included the analysis in Sec.~\ref{transformer_scheduling}.

\paragraph{\textbf{Operation-level Mappers}}
In Sec.~\ref{subsec:operation_level}, 
we discussed that the decisions around tiling, dataflow, and spatio-temporal mapping can result in an enormous search space, and that selecting a good point in the space is key to achieving high efficiency and utilization in ML accelerators.
Within the scope of a given subspace, mappers can generally be divided into three general categories, based on how they make their decisions:
brute-force search; 
feedback-based search; and 
constrained optimization.  
Tab.~\ref{table:mapper} summarizes existing mappers that leverage different techniques to navigate the mapspace. 

\emph{Brute-force methods}~\cite{parashar2019timeloop,
dave2019dmazerunner,yang2020interstellar,chatarasi2020marvel} entail various sampling strategies that either exhaustively explore or randomly sample a large number of points from the mapspace.  
To lower the exhaustive search cost, mappers in this category typically rely on developer heuristics to prune the mapspace and lightweight performance models to compare all valid mappings to find the best mapping in a reasonable amount of time. 
The disadvantages of this approach are two-fold: 
not only does a brute-force search tend to be exceedingly expensive, especially for more complex target workloads and hardware architectures; 
but also this costly process repeats for any target workload or accelerator architecture changes, without leveraging any prior knowledge. 

\textit{Feedback-driven} approaches use ML algorithms or
other statistical methods ~\cite{chen2018tvm, ragan2013halide,
adams2019learning, jia2019beyond} either to improve the accuracy of the cost model or to directly search for the solution using blackbox-tuning.
Although such approaches can potentially learn the scheduling space accurately, their computational cost is significant due to both 
the cost of evaluating enough schedules to learn a model as well as 
the cost of learning based algorithms. 
As a result, these approaches typically apply to existing hardware or analytical models where
large-scale measurement is feasible.

\emph{Constrained-optimization} approaches contrast with exhaustive search and learning-based algorithms, in that they formulate scheduling problems as a numerical optimization problem to determine variable
assignments subject to a given set of constraints and objective functions.
Popular techniques, such as Mixed Integer Programming (MIP), have demonstrated their applicability to solve large-scale and complex problems.  
In particular, polyhedral transformation has leveraged constrained-optimization based approaches for auto-vectorization and loop
tiling~\cite{bondhugula2008practical, grosser2011polly,
kong2013polyhedral,park2013predictive,
baghdadi2015pencil, acharya2018polyhedral, acharya2018approach}.
These polyhedral optimizations focus on testing the feasibility of a transform and offering information to guide iterative searches. 
On the other hand,~\cite{li2021analytical, huang2021cosa} leverage the regularities in the ML workloads and hardware to formulate the mapping as optimization problems, which can then be directly solved by off-the-shelf solvers.

\begin{boxA}
\textbf{Summary (Sec.~\ref{previous_work_mapping}. Finding Performant Mappings): }
A comprehensive set of strategies has been developed to address the challenge of mapping DNNs on accelerators and general-purpose processors. 
The techniques originally developed to target CNNs can be applied to Transformers as the key operations are also tensor algebra operations.
 At the graph level, operator fusion is an important optimization technique that encodes a vast mapping space to decide how the execution of layers are overlapped. 
At the operation level, mapping strategies  can be broadly categorized as either \emph{search} strategies - either random or feedback-driven - or \emph{optimization} or \emph{heuristic} strategies.
\end{boxA}

\subsection{Performance Modeling of Mappings}
\label{performance-modeling-mappings}

Performance models can provide the mappers with performance feedback for different mappings without executing mappings on real hardware or running simulations on accelerators under development. 
They can significantly reduce the evaluation costs for mappers and be used as performance proxies to optimize mappings. 
Different performance models offer different levels of fidelity, runtime costs, target workload scopes, and compatibility for various mapping algorithms. 
The selection of the performance model is both mapper and target workload dependent.  

For Transformers, the mappers can use domain-specific polynomial~\cite{huang2021cosa} and analytical models~\cite{parashar2019timeloop, kwon2020maestro, lu2021tenet, mei2021zigzag} to provide fast comparisons among mappings.  
These models leverage known iteration space bounds in tensor algebra workloads, as well as statically analyzable data access patterns, to estimate performance. The polynomial models expressed in mathematical forms can also be used directly as the objectives in optimization-based mappers. 

Another class of popular performance models involves data-driven ML models~\cite{chen2018tvm, hegde2021mind, kaufman2021learned}. Instead of building the performance model analytically to express known relations between mapping decisions and performance, these models use statistical techniques to iteratively fit a model to the mapping performance data collected over time. They typically require large amounts of data in order to learn and provide accurate predictions. Once trained, they can be easily integrated with ML-based mappers~\cite{snoek2012practical, hegde2021mind}. 

The major drawback of prior models is that the generated mappings might not perform optimally (or even well) on the actual accelerator since the models can fail to capture the implementation differences in the hardware accurately. 
Cycle-exact software models based on real hardware implementation can achieve higher fidelity~\cite{samajdar2020systematic, venkatesan2019magnet}, as can FPGA emulation using platforms such as Firesim~\cite{karandikar2018firesim}, which can be used to model hardware in development. However, such platforms require more than a set of problem dimensions and a description of mappings; they require a stream of explicit instructions.

Generating this stream of instructions requires that one account for a large number of edge cases. 
For example, a simple tiling operation for a matmul $-$ representable as a single line of tile sizes in Timeloop~\cite{parashar2019timeloop} $-$ requires both the insertion of instructions specifying memory movement between different levels of the memory hierarchy as well as the generation code for edge cases that appear when matrix dimensions are not evenly divisible by the tile size. Furthermore, the codesign process requires this process to be automatable.
In other words, each mapping must be translatable to code automatically.

As a result, \emph{code generation}~\cite{lattner2004llvm, halide2013-pldi, chen2018tvm} 
tools are used to actually implement mappings onto hardware (or simulators). 
Many of these tools integrate not only a specification of the hardware backend but also mapping decision algorithms, often tuned for that hardware target. 
Such tools can also be useful for neural architecture search (NAS) in order to obtain accurate performance numbers for a given hardware architecture to guide automated DNN architecture design (see Sec.~\ref{sec:nas} for details).
However, these tools are difficult to adapt for a codesign framework where both the mapspace and the hardware target can vary. 

In order to address this problem, \emph{user-schedulable languages} such as Halide~\cite{ragan2013halide}, TVM~\cite{chen2018tvm}, Rise/Elevate~\cite{HLK+20RiseElevate}, and Exo~\cite{YBR+22Exo} have been developed. 
These tools take as input a description of the computation to be performed and a point in the mapspace.
They are generally defined by a set of \emph{rewrite rules}, representing certain transformations such as splitting and rearranging loops, replacing appropriate loops with ISA instructions, and fusing loops. 
These languages also allow the user to specify and customize the hardware instruction set and seamlessly convert found schedules into executable code by representing these points as a series of rewrite rules \cite{mullapudi2016automatically, ZJS+20Ansor}.

\begin{figure*}[t!]
\centering{
\begin{subfigure}{.5\textwidth}
  \centering
  \includegraphics[width=0.95\linewidth]{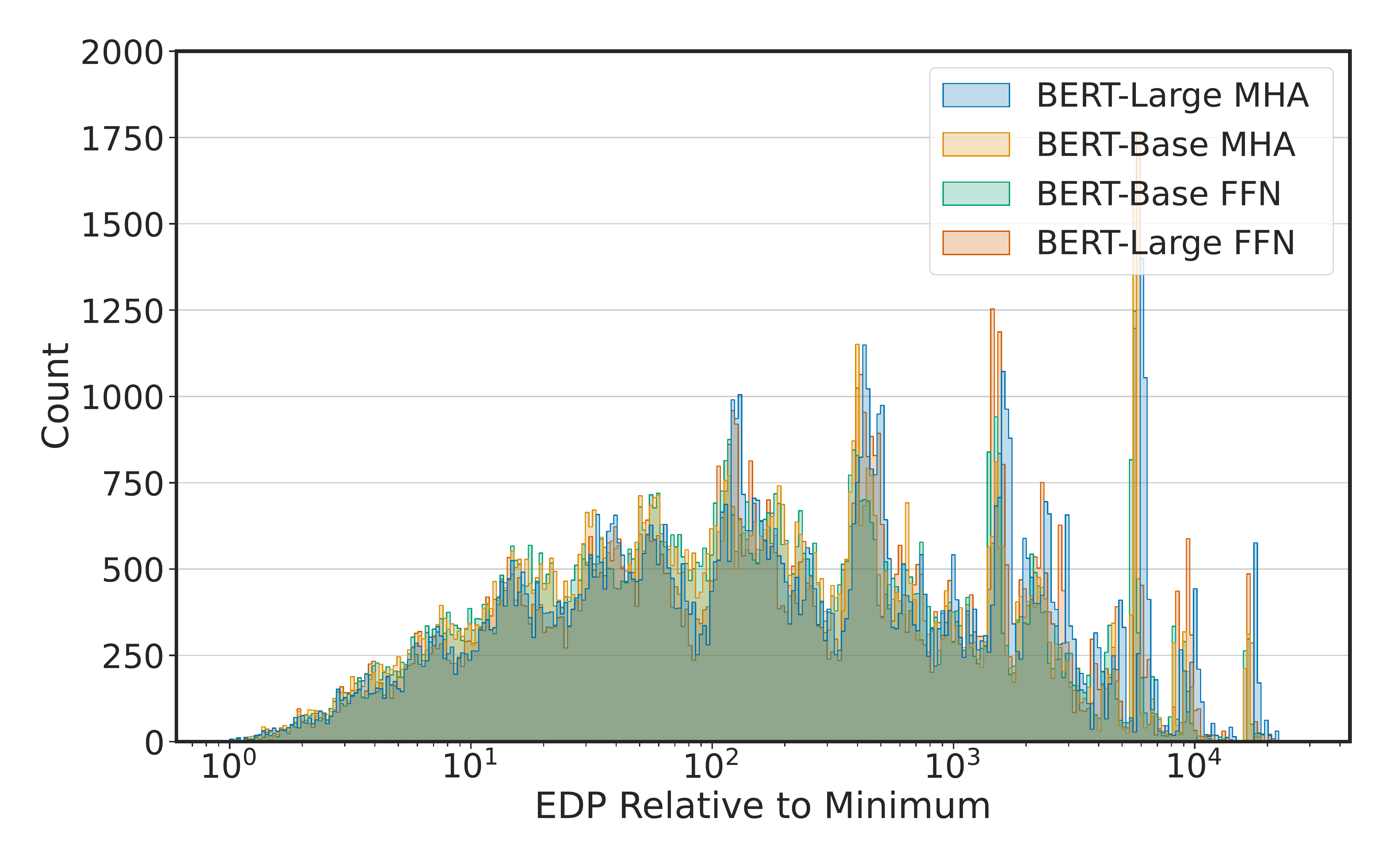}
  \label{fig:sub1}
\end{subfigure}%
\begin{subfigure}{.5\textwidth}
  \centering
  \includegraphics[width=0.95\linewidth]{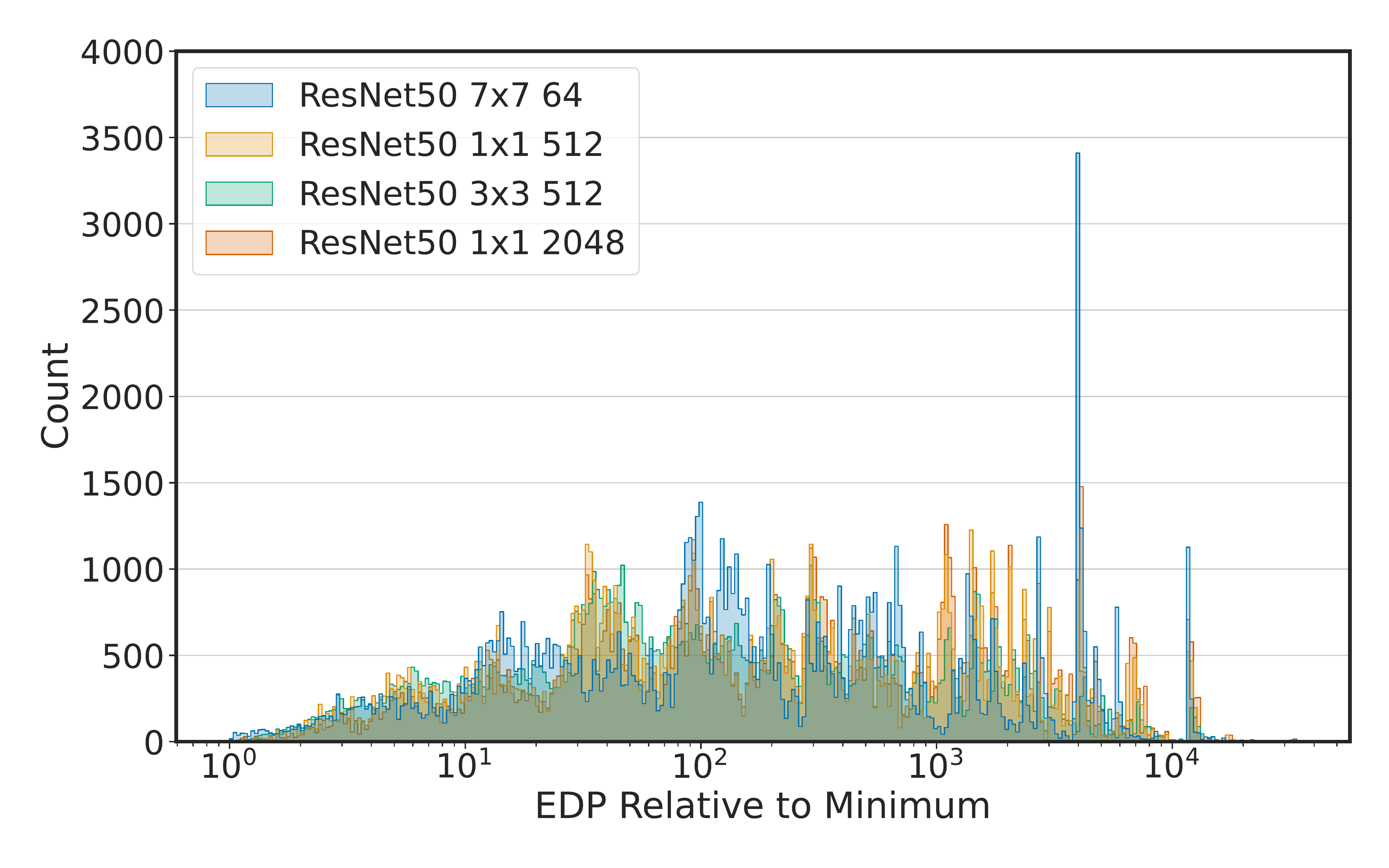}
  \label{fig:sub1}
\end{subfigure}%
  \caption{A comparison of the mapspace of (Left) BERT and (Right) ResNet50. Distributions of 100K randomly sampled valid mappings are shown. 
  Both distributions show a similar range of EDP of up to four degrees of magnitude difference with respect to the best (minimum) observed value.
  Neither distribution is significantly more skewed towards lower or higher relative EDP. 
  Overall, we find that mapspaces for BERT matmuls and ResNet50 convolutions are similarly vast in size with no significant difference in the shapes of their distribution. 
  This indicates that brute-force or random search for BERT matmul scheduling is equally as challenging as in the case with ResNet50 operators. 
  \label{fig:bert_mapspace}
  }
  }
\end{figure*}
\begin{boxA}
\textbf{Summary (Sec.~\ref{performance-modeling-mappings}. Performance Modeling of Mappings):}
Performance estimation of Transformers running on novel architecture is essential for finding an optimal algorithm, mapping, and hardware combination. There are various open-source performance models available to estimate the mapping performance on hardware, ranging from domain-specific analytical models and data-driven ML models to cycle-exact models. The selection of the performance model for Transformer depends on the target workload size, hardware complexity, and development stage. Additionally, there are many mature code generation tools one can leverage to optimize the Transformer design for off-the-shelf~hardware.
\end{boxA}

\subsection{Transformer vs CNN Mapping}
\label{transformer_scheduling}

\smallskip
Prior work discussed in Sec.~\ref{previous_work_mapping} for finding good mapping strategies largely focuses on mapping CNNs onto accelerators or general-purpose hardware. 
As with CNNs, the vast majority of cycles for Transformers are spent on matmuls from the MHA and FFN modules. 
In essence, existing mappers for CNNs can easily extend to scheduling Transformer matmuls. 
However, as we have discussed in Sec~\ref{subsec:hardware_performance_bottlenecks}, Transformer blocks include LayerNorm and Softmax operations, which can be computationally non-trivial in certain realistic scenarios (which was also observed by~\cite{choiaccelerating, pati2021demystifying}). 
In turn, the presence of these operations imposes constraints on scheduling the preceding and succeeding matmuls.
This leads to a much more complex problem for scheduling optimizations for Transformers overall. In this subsection: 
\begin{itemize}[leftmargin=5mm]
    \item 
    We characterize the mapspace of Transformer blocks in comparison to that of CNNs (Sec.~\ref{subsec:mapspace-transformers}).
    \item 
    We take a deeper dive into the issue of increased scheduling complexity of Transformer matrix operations due to the presence of LayerNorm and Softmax (Sec.~\ref{subsec:scheduling_complexity_nonlinear}).
\end{itemize}


\vspace{2mm}
\subsubsection{\textbf{Mapspace Characterization of Transformers.}}
\label{subsec:mapspace-transformers}
We empirically characterize the search space of legal mappings for representative Transformer and CNN architectures.
To do so, we chose BERT \cite{devlin2018bert} and ResNet50 \cite{he2016deep}, respectively.
A total of 100K random valid mappings were searched via the Timeloop mapper \cite{parashar2019timeloop}, and the estimated latency and energy were measured by the Timeloop model. 
The target spatial hardware architecture is the Gemmini systolic generator~\cite{gemmini-dac}. Both the BERT and ResNet50 models are assumed to have been 8-bit integer quantized. We assume an input sequence length of 512, which is a typical assumption for BERT-Base and BERT-Large models.

\begin{figure}[t!]
\centering{
\centerline{
  \includegraphics[width=\linewidth]{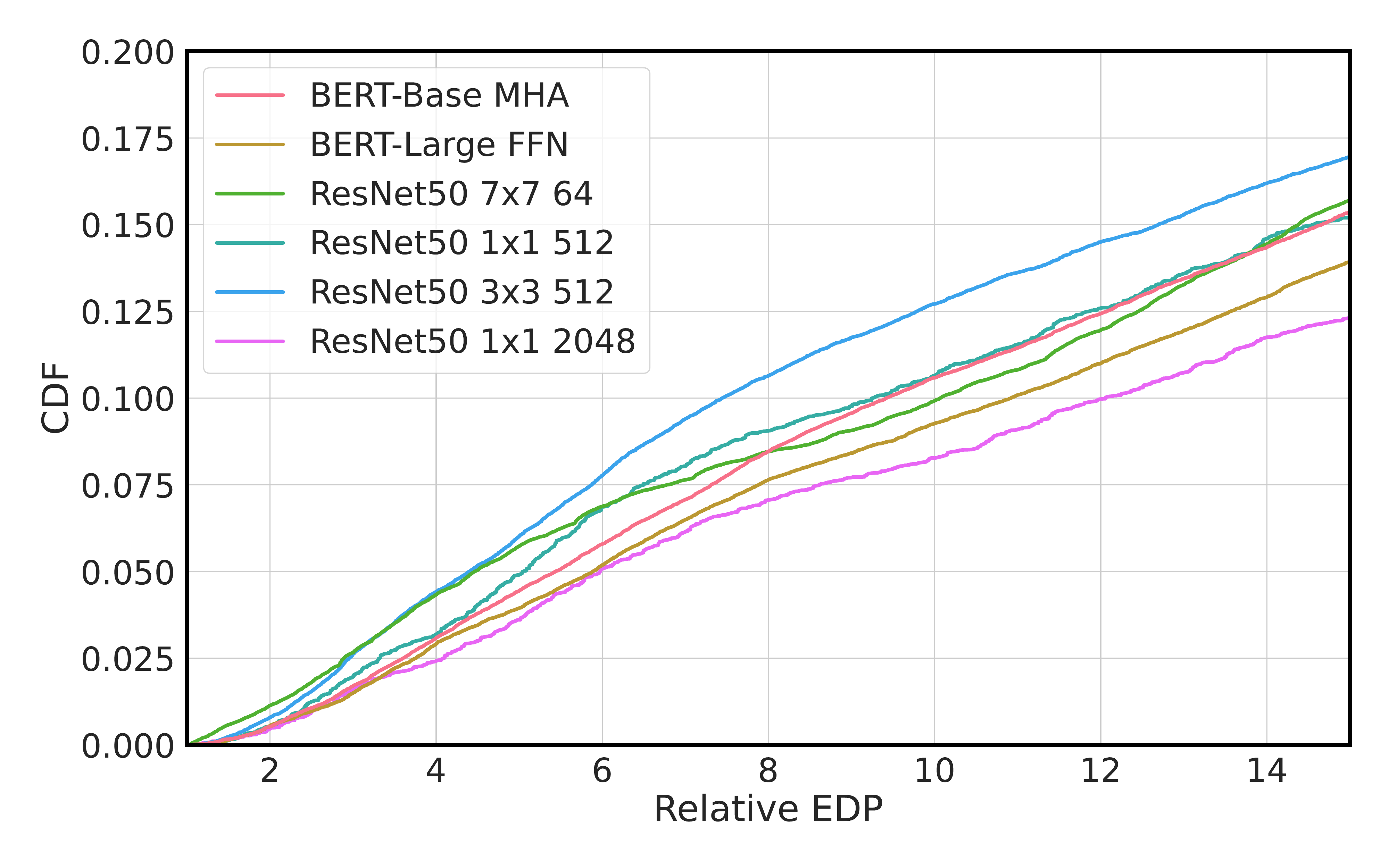}
}
  \vspace{-2mm}
  \caption{Comparison of empirical cumulative distribution functions (CDFs) for Transformer matmuls and ResNet50 convolutional operations around the regime where near-optimal mappings are found. The 10th percentile values for each relative EDP distribution are: 7.44 for the $3\times3, 512$ convolution kernel; 12.06 for the $1\times1, 2048$ kernel; 9.42 for the BERT MHA matmul; and 9.84 for the BERT FFN matmul. Results show that the percentage of near-optimal mappings for BERT matmuls are similar if not smaller than that of ResNet convolution kernels. This indicates that the search problem of finding optimal mappings can be as challenging for BERT matmuls. 
  }
  \label{fig:bert_resnet_cdf}
  }
\end{figure}
\vspace{2mm}
\paragraph{\textbf{EDP Distribution Analysis.}}
Fig.~\ref{fig:bert_mapspace} demonstrates the comparison between the mapspaces for BERT (Left) and ResNet50 (Right). 
For the BERT MHA module, results in the figure correspond to mappings for the matmul performing each of the query, key, and value projections. 
For the FFN, we take the matmul in the $W_1$ projection, which takes part in expanding the hidden dimension to four times its value. 

From ResNet50, we choose convolution operations of varying kernel sizes and shapes, spanning $1\times1$, $3\times3$, and $7\times7$ convolution kernels. Specifically, we use the 7$\times$7 kernel convolution with output channel size 64 and stride 2 in the conv1 layer of ResNet50, the 3$\times$3 kernel convolution with 512 channels and 1$\times$1 kernel convolutions with output channel sizes 512 and 2048 that belong to the final convolution layer conv5\_x. These four particular convolutions vary in channel and kernel sizes and reasonably represent convolutions found in ResNet50.  

From our mapspace analysis, we observe that both BERT and ResNet50 mapspace have a similar range of potential energy-delay product values from randomly sampled mappings. Accounting for variations between operators for each architecture, the distribution of EDP values for themselves are also largely similar, in that the portion of Pareto-optimal mappings with lower EDP values are small for both BERT and ResNet50 operators. 

As an alternative visualization, Fig.~\ref{fig:bert_resnet_cdf} compares the empirical cumulative distribution functions (CDFs) of the same set of 100K random mappings. Here, we closely examine the difference in the CDFs near the regime where near-optimal mappings are found. We observe that the relative EDP value corresponding to the tenth percentile is 7.44 for the $3\times3, 512$ convolution and 12.06 for the $1\times1, 2048$ convolution kernels. BERT matmuls for MHA and FFN projection had tenth percentile values of 9.42 and 9.84, respectively. 

Alternatively, we also examine the percentage of mappings with relative EDP values less than 3 times the observed minimum. This percentage represents a rough upper bound on the number of mappings that can be safely labeled as near-optimal, and the difference in percentages signify relative difficulties in searching for optimal mapping for different operators. 
We find 1.58\% for the $3\times3, 512$ kernel and 2.62\% for the $1\times1, 2048$ kernel with mappings within this range of EDP. In the case of BERT, the MHA projection matmul is shown to have 1.70\% of mappings with relative EDP less than 3 with 1.48\% of mappings in this range for the FFN matmul.

{Overall, the analysis of EDP distribution from randomly sampled valid mappings indicates that BERT matmuls, despite having fewer loop levels for tiling and re-ordering compared to convolutions, are as challenging to schedule as CNNs}. 
As much as graph and operator-level scheduling had a significant impact on end-to-end performance and efficiency of CNN inference, the same importance of appropriate scheduling also applies to Transformer matrix operations.

\begin{figure}[t!]
\centering{
  \includegraphics[width=\linewidth]{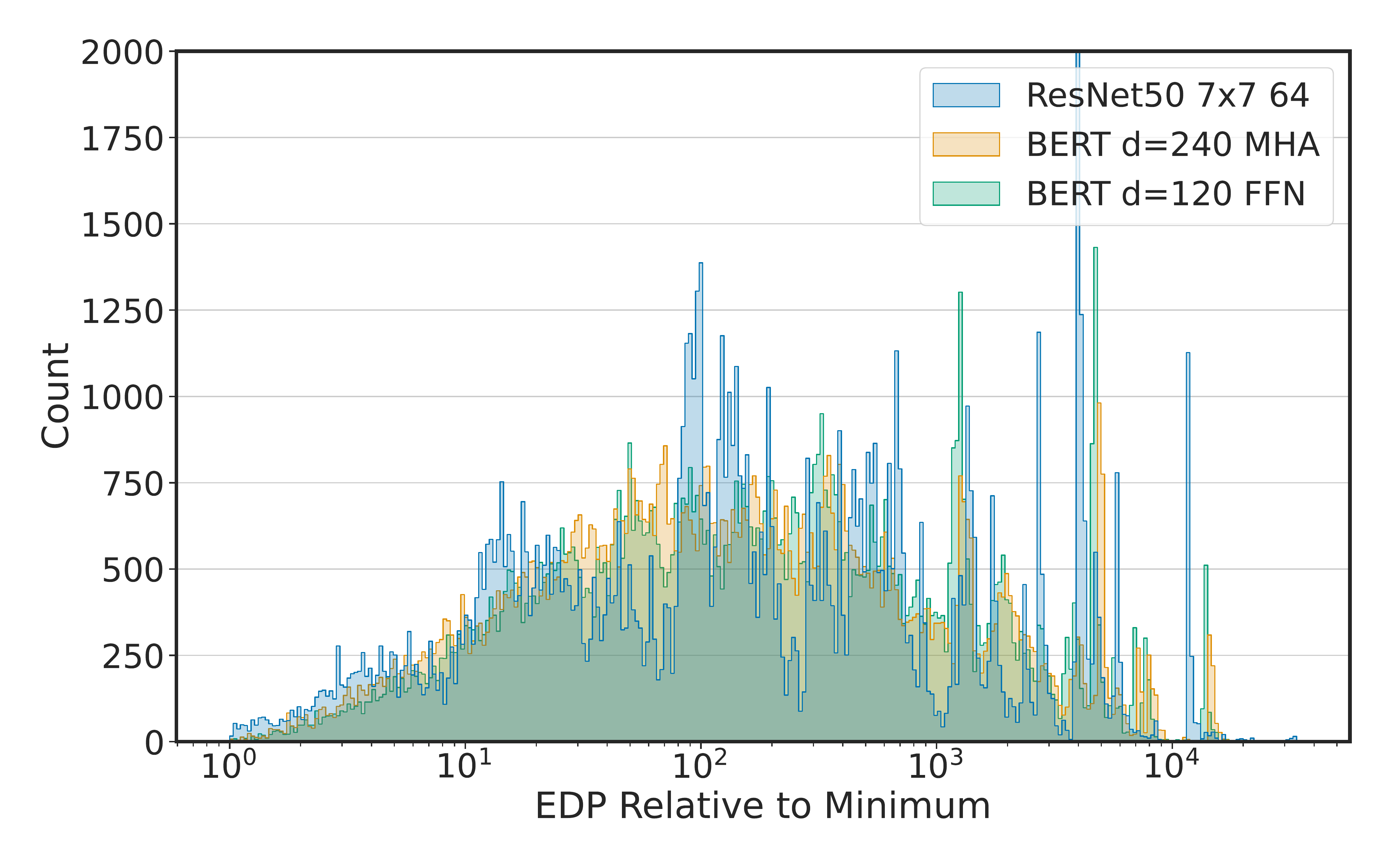}
  \vspace{-5mm}
  \caption{Distributions of random, valid mappings for ResNet50 operators, compared against distributions for Transformer MHA and FFN matrix multiplications, where the size of matmuls are tuned so that total MACs are equivalent. The MHA matmul dimensions were calibrated to $240 \times 240 \times 512$ and similarly for the FFN $W_1$ projection matmul, set to $480 \times 120 \times 512$ for $d_{\text{FFN}} = 4 \times d$. Comparing against Fig.~\ref{fig:bert_mapspace}, we see that the EDP distributions are still similar for the ResNet convolution and BERT matmuls. This implies that, even after accounting for differences in the total number of MACs, the mapspace of BERT matmuls exhibit as vast a range of relative EDPs as the mapspace of CNN convolution kernels. }
  \label{fig:bert_synthetic}
  }
\end{figure}
\paragraph{\textbf{EDP Distribution Analysis with Fixed Total Number of MACs.}}
As an additional analysis on mapspace characterization, we further force the total number of MACs to be fixed.
This enables an even fairer comparison between the distributions of mapping results for Transformer and ResNet50 operators. 
We continue to assume the Transformer input sequence length to be 512 and the feed-forward network expansion ratio of 4 times the hidden dimension size. 
To keep the number of MACs equal, we calculate the hidden dimension size that would yield the same total MACs as for the ResNet50 conv1 layer's 7$\times$7 kernel with output channel dimension 64. For the matmul in the query projection of MHA, the corresponding hidden dimension size was 240.
Similarly, for the matmul in $W_1$ projection of the FFN block, the corresponding hidden dimension size was 120. 
To elucidate the comparison between synthetic BERT layers and actual ResNet50 convolutions, we plot the corresponding pairs mapping distributions in Fig.~\ref{fig:bert_synthetic}. 
Even after forcing equivalent numbers of MACs, we see that the range of relative EDP values are similar between BERT matmuls and ResNet50 convolutions. 
{{This finding further accentuates how complex the scheduling problem can be for matmuls found in Transformer models}}.

\medskip
\subsubsection{\textbf{Scheduling Complexities of LayerNorm and Softmax.}}
\label{subsec:scheduling_complexity_nonlinear}

While we find that matmuls in Transformers are already non-trivial targets for which to obtain efficient execution schedules to execute on DNN accelerators, the problem is further complicated by the presence of several non-linear operations, including LayerNorm and Softmax that are interposed between different matrix operations. 
When pursuing more aggressive optimizations, an enticing strategy is to fuse relatively high-arithmetic-intensity matmuls with the low-arithmetic-intensity normalization operations following them, such as LayerNorm and Softmax.
This can be especially enticing in handling quantized workloads, where partial sums awaiting normalization are often of much higher bitwidth than the final normalized outputs.
Architects familiar with CNN-type accelerators may find this especially intuitive, since convolutions are often fused with ReLU or max-pool operations. 

Similarly, for Transformer Encoders, we could overlap the execution of normalization operation and the previous matmul, yet this is possible only with additional hardware support and appropriate constraints on the matmul execution schedule. 
To enable complete latency-hiding of nonlinear operations, the tiling factor size of either output dimension of the matmul must be maximized, so that rows/columns are immediately ready and stored at the Gemmini accumulator scratchpad for computing the mean and standard deviation.
We refer to this alternate scheduling approach as \textit{fusion-optimized scheduling}. 

On the other hand, in memory-constrained edge devices, the strategy is (somewhat unintuitively) counter-productive.
The normalization operations found in Transformers often require long vectors of data to be resident in on-chip local memory before any normalized output element can be produced.
Furthermore, when fusing matmuls with these normalization operations, awkward matmul tile shapes are typically required.
These awkward tile shapes are often much larger in either dimension, as opposed to being square-shaped, and such skewed tile shapes tend to yield far worse arithmetic intensity.  
This greatly reduces the performance of the matmuls, and it may \textit{increase} the total memory traffic, even accounting for the high bitwidths of the unnormalized partial sums which must be sent to and from outer memory when fusion is not enabled.

\begin{figure*}[t!]
\centering{
\begin{subfigure}{.5\textwidth}
  \centering
  \includegraphics[width=0.95\linewidth]{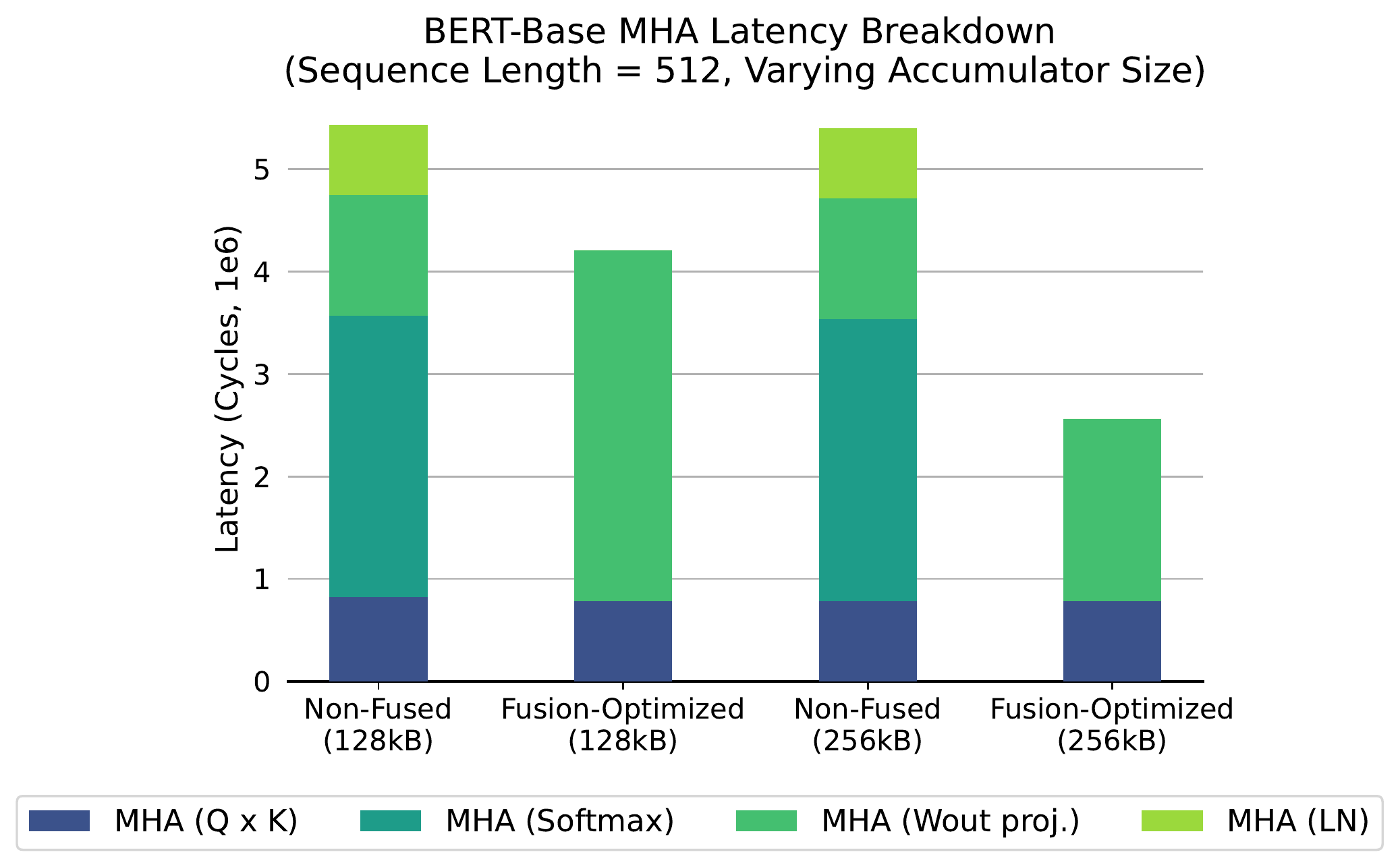}
  \label{fig:sub1}
\end{subfigure}%
\begin{subfigure}{.5\textwidth}
  \centering
  \includegraphics[width=0.95\linewidth]{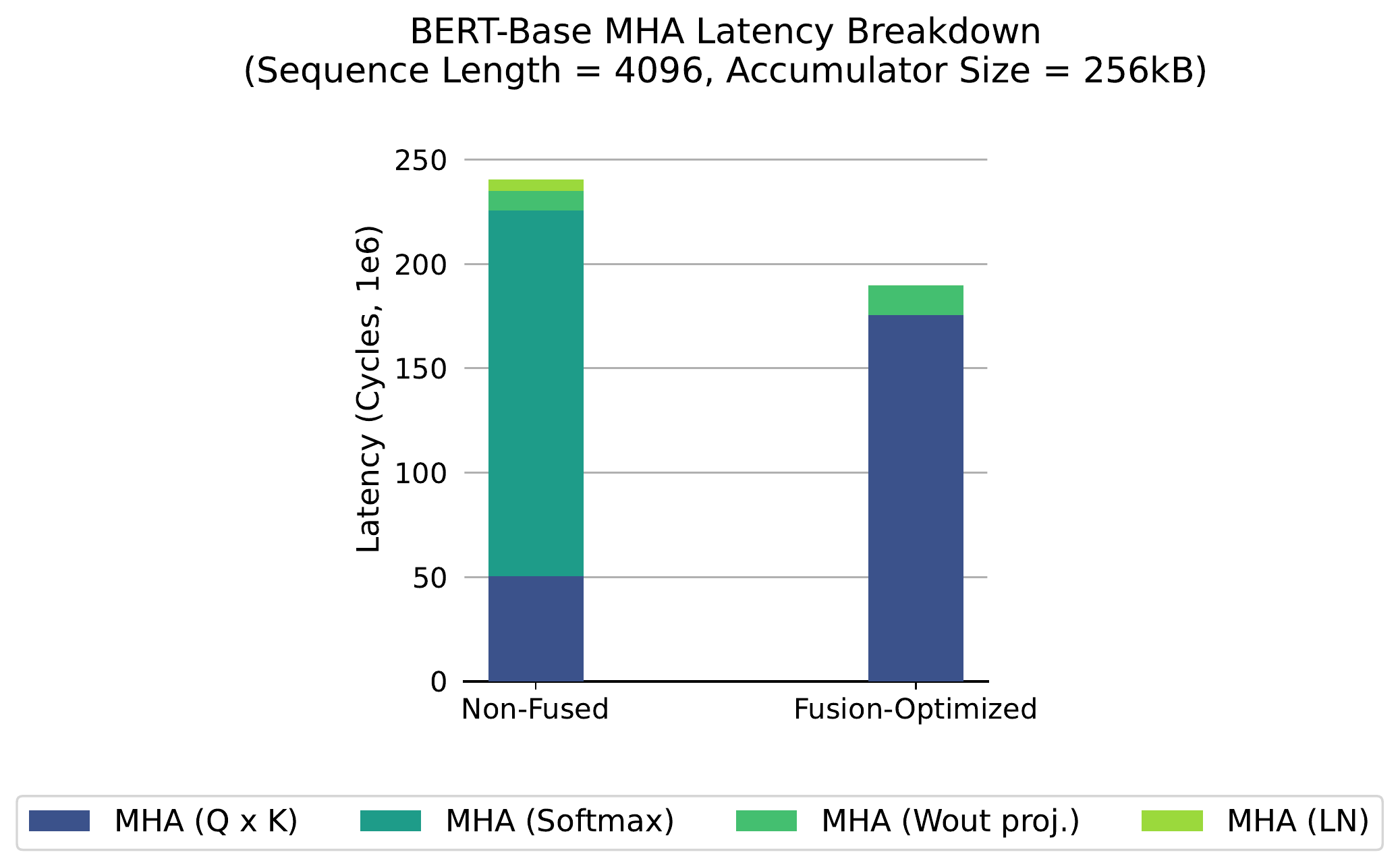}
  \label{fig:sub1}
\end{subfigure}%
  \vspace{-2mm}
  \caption{Impact of fusion-optimized scheduling for BERT MHA that enables latency hiding of LayerNorm and Softmax. Results are based on the BERT-Base architecture and the Gemmini accelerator. 
  (Left) Input sequence length is  assumed to be 512, and the accumulator SRAM size is increased from 128kB to 256kB. 
  Hiding the Softmax latency improves combined matmul and Softmax latency by 78\%. 
  However, overlapping $W_{\text{out}}$ projection with LayerNorm can either hurt or improve total latency, depending on the accumulator size. 
  Overall, fusion-optimized scheduling for both matmuls in MHA yields 23\% and 52\% latency improvements for accumulator sizes 128kB and 256kB, respectively. 
  (Right) The input sequence length is increased to 4096. Again, we see that overlapping the query $\times$ key matmul with Softmax improves latency by 22\%. 
  Overall, fusion of both MHA matmuls with nonlinear operation yields a 21\% latency improvement. 
  \label{fig:mha_fusion}}
}
\end{figure*}

\begin{figure}[t!]
\centering{
  \includegraphics[width=0.85\linewidth]{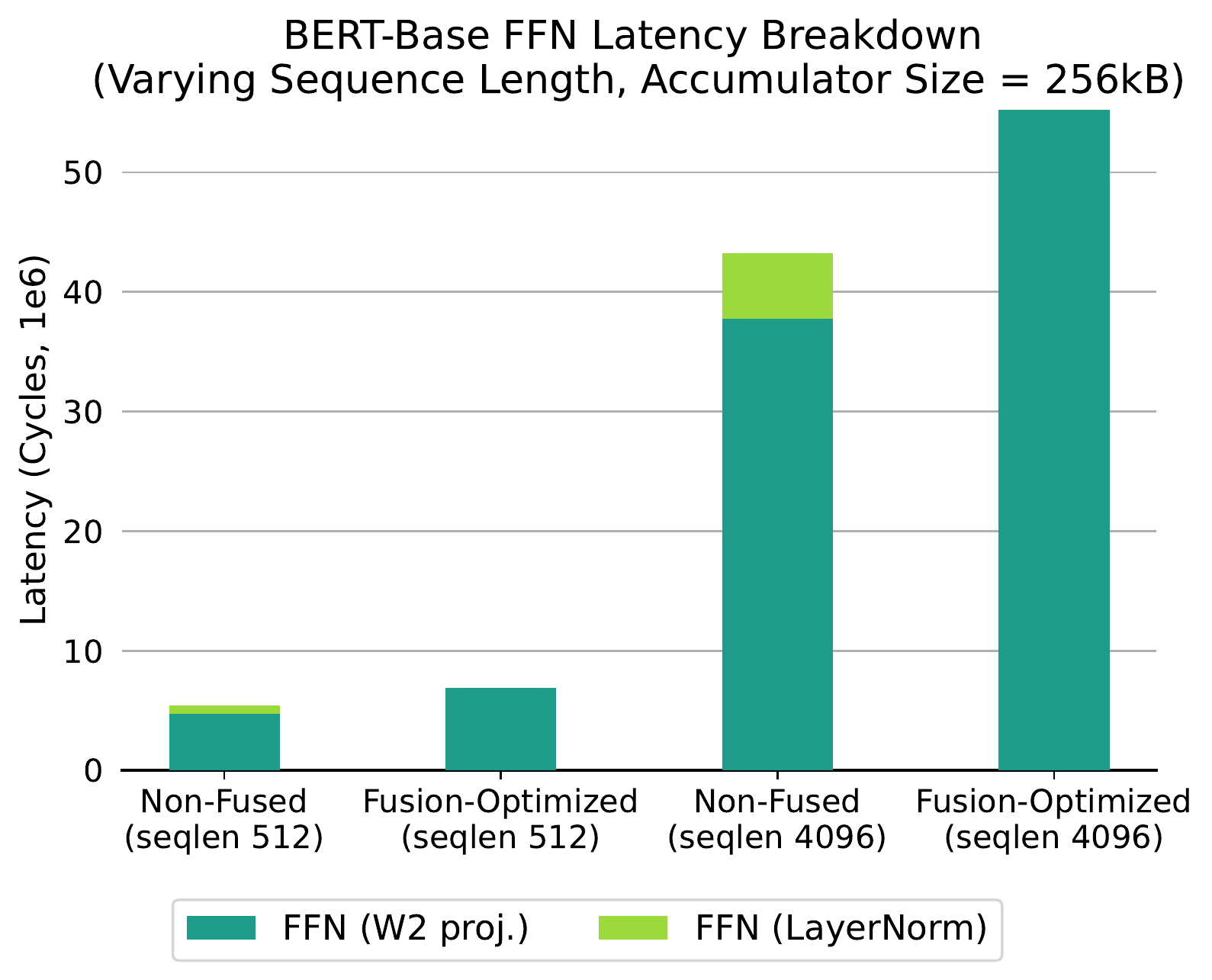}
  \vspace{-2mm}
  \caption{Impact of fusion-optimized scheduling for BERT FFN matmul that enables latency hiding of the LayerNorm operation. Input sequence length is varied from 512 to 4096. We observe that fusion-optimized scheduling hurts total latency by 27\% in both cases. This motivates the need to carefully evaluate the impact of chaining matmul and LayerNorm execution on systolic arrays since the impact of mapping constraints may outweigh the gains from latency hiding of nonlinear operations. 
  \label{fig:ffn_fusion}}
}
\end{figure}

In Fig.~\ref{fig:mha_fusion}, we take a deeper look at the performance implications of fusion-optimized scheduling for BERT matmuls.
We consider the BERT-Base encoder with hidden dimension 768 and 12 attention heads. By default we assume a sequence length of 512. 
As the target hardware, we take the 16$\times$16 weight-stationary systolic array Gemmini with custom hardware units for I-BERT implementations of nonlinear operations (activation, normalization, etc.), as described in Sec. ~\ref{subsec:hw-sw-codesign}.
The total latency of each adjacent pair of matmul and LayerNorm/Softmax operations is estimated via Timeloop\cite{parashar2019timeloop}. 
Opportunities for overlapping computations include: (1) the MHA query $\times$ key matmul and following Softmax; (2) MHA $W_{\text{out}}$ projection and following LayerNorm; and (3) FFN $W_2$ projection and following~LayerNorm. 

We compare two scheduling strategies.
In the first strategy, we use  Gemmini's default heuristic-based scheduler, which greedily maximizes loop tile factors at the local SRAM level for each of the three matmul dimensions. 
In this approach, we do not attempt to overlap the matmul computation with the following nonlinear operation, meaning that the matmul is scheduled independently as if it were executed on its own.
In Fig.~\ref{fig:mha_fusion}, we denote this approach as \emph{non-fused} scheduling. 
The second strategy is the aforementioned fusion-optimized~scheduling. 

The left plot of Fig.~\ref{fig:mha_fusion} summarizes how matmul and nonlinear operation fusion within the MHA block can be influenced by the accumulator SRAM size. In this experiment, while the on-chip scratchpad for input activation and weights is held fixed at 256kB, the output activation accumulator size is increased from 128kB to 256kB. 
We note two findings: first, within MHA, fusing query $\times$ key matmuls with Softmax for each attention head reduces latency regardless of accumulator size.
In particular, we see that Softmax latency is significant compared to the matmul, taking up around 78\% of the total cycles, and hiding this latency significantly reduces total latency.
At the same time, the query $\times$ key matmul latency is relatively unchanged by the additional scheduling constraints, mainly because inner dimension of the matmul is small ($d/l = 64$ for BERT-Base). 
On the other hand, the mapping constraints from fusion-optimized scheduling significantly harm the execution latency of the $W_{\text{out}}$ projection matmul after fusing with the following LayerNorm, resulting in 83\% worse latency than the non-fused schedule. 
However, once the accumulator size is doubled, the performance hit on matmul scheduling is alleviated. 
Increased accumulator SRAM size of 256kB allows more partial sums to be stored in the buffer instead of spilling to DRAM, thereby reducing total latency by 4\%.

In the right plot of Fig.~\ref{fig:mha_fusion}, we further investigate the impact of sequence length on fusion-optimized scheduling for the MHA block.
Here, the sequence length is increased from 512 to 4096, which impacts the ratio of cycles from matmuls, Softmax, and LayerNorm in the MHA block. 
In particular, note that the size of the query $\times$ key matmul and the Softmax computation depends quadratically on the sequence length, while the other matmul and LayerNorm exhibit a linear dependence. 
When fusing the query $\times$ key matmul with the subsequent Softmax, the mapping constraints worsen matmul performance despite a larger (256kB) accumulator size.
This is because with increased dimensions of the query $\times$ key matmul and forced tiling factors, the scheduler can no longer avoid tiling at the DRAM level.
However, by overlapping the Softmax operation and thereby eliminating the need to load and store intermediate activations (which quadratically scales with the sequence length),
the latency increase from the query $\times$ key matrix can be offset, resulting in an overall 22\% reduction in latency.

On the other hand, Fig.~\ref{fig:ffn_fusion} shows the results on matmul and LayerNorm overlapping in the FFN $W_2$ projection. Even with a larger accumulator size and in both sequence lengths, we consistently observe that fusion-optimized scheduling worsens total latency by 27\%. 
Together with previous findings, we see that {latency improvements of fusion-optimized scheduling are dependent on the accumulator SRAM size and sequence length}. 
Furthermore, we find that, in the BERT-Base scale, it is {consistently favorable to overlap the MHA query $\times$ key with the ensuing Softmax but consistently disadvantageous to chain the FFN $W_2$ projection matmul with LayerNorm}. This is in contrast with previous studies on GPU kernel fusion for Transformers ~\cite{choiaccelerating, pati2021demystifying}, and it highlights how scheduling for Transformer matmuls becomes more complex when targeting different styles of custom hardware designs, including the Gemmini accelerator.

\begin{boxA}
\textbf{Summary (Sec.~\ref{transformer_scheduling}. Transformer vs. CNN Mapping)}: 
Here are the high-level takeaways from this section.
\begin{itemize}[leftmargin=5mm]
\item Scheduling for Transformer matmuls is as challenging as scheduling CNN convolution operators.
Both mapspaces have similar distributions of relative EDPs and similar percentages of near-optimal mappings. Brute-force or random scheduling is not simpler for Transformer matmuls, despite them having fewer loop levels than convolutions. 
\item The presence of nonlinear operations such as LayerNorm and Softmax present additional complexities to the scheduling problem for Transformer matmuls. Latency of these nonlinear operations can be hidden by fusing its computation with the preceding matmul. This requires additional hardware support, as noted in Sec. ~\ref{subsec:hw-sw-codesign}, and it imposes constraints to the matmul scheduling. 
\item Whether this fusion-optimized scheduling yields end-to-end latency improvements depends on the Transformer and underlying hardware parameters. In particular, we observe that: (1)  size of the on-chip SRAM buffer for output activation; and (2) sequence length matter.
\item We consistently observe that overlapping the execution of query $\times$ key matmul with Softmax in the MHA block reduces latency up to 78\%, compared to executing the two operations separately on a systolic array accelerator. On the other hand, scheduling to overlap the FFN $W_2$ projection with the following LayerNorm  hurts performance by 27\%. 
\end{itemize}

\end{boxA}
\section{Adapting Transformer Architecture with NAS}
\label{sec:adapting_nas}
So far, we have conducted an in-depth exploration of the full-stack aspect of DNN inferencing, with a focus on the Transformer architecture, from the hardware level to optimization and scheduling strategies to improve their inference performance.
Another important avenue in full stack optimization of DNNs is obviously to optimize DNN architecture itself and to tailor it for a specific hardware platform.

In this section, we will primarily focus on automated neural architecture search (NAS) as a method for designing DNNs.
Sec.~\ref{sec:nas} will provide a general overview of NAS, and then Sec.~\ref{sec:hwnas} will explore hardware-aware NAS methods. 
These two subsections will be mainly focused on NAS techniques for CNNs, as NAS was initially introduced and extensively researched from the pre-Transformer era. 
However, we believe it is helpful to provide a comprehensive overview and background to understand NAS.
In Sec.~\ref{sec:transformer-nas}, NAS methods specific to Transformer architectures will be discussed.
Finally, in Sec.~\ref{subsection:nas_case_study}, a case study of applying NAS method in the scenario of optimizing Transformer inference on a target hardware architecture will be provided.

\subsection{Neural Architecture Search}
\label{sec:nas}
Typically, DNN architectures are designed and trained to achieve the maximum accuracy for a given task, without necessarily considering the target hardware or inference latency, memory, and power requirements.
However, often there exist several different variations of the DNN architecture which result in the same accuracy but have better hardware performance. 

There is a rich literature in this area. Notable works here include
MobileBERT~\cite{sun2020mobilebert}, which is one of the earliest attempts, and which adopts the bottleneck structure to design a thinner version of Transformer,
as well as Lite Transformer~\cite{wu2020lite}, which proposes the Long-Short Range Attention, in which a group of heads are replaced with convolution operations to capture short-range contexts more efficiently.
SqueezeBERT~\cite{iandola2020squeezebert} is another work that incorporates grouped convolutions into the Transformer architecture to reduce the model size and latency.
This approach is not limited to NLP, and similar models have been proposed in computer vision (CV)~\cite{mehta2021mobilevit,chen2022mobile,li2022efficientformer,cai2022efficientvit}
and speech recognition~\cite{gulati2020conformer,burchi2021efficient,kim2022squeezeformer}, to name just a few.

It is often very difficult to find these DNN architectures since the search space is exponentially large,
even without considering the underlying hardware platform. 
Even for those with expertise in DNN architecture design, the impact of an architectural change on accuracy and runtime performance can be nontrivial to predict.
As such, automated NAS methods
have been proposed to adapt a DNN architecture for a given constraint.
However, it is critical to note that NAS methods often require prohibitive amounts of compute and trials before finding a candidate architecture. 
For instance, in one of the early NAS works~\cite{zoph2018learning}, finding an optimized CNN took 22,400 GPU-hours. 
Moreover, NAS methods are not yet fully automated, and they often require hand-tuning the search space.

\begin{figure}[t!]
\centering{
\centerline{
  \includegraphics[width=0.5\textwidth]{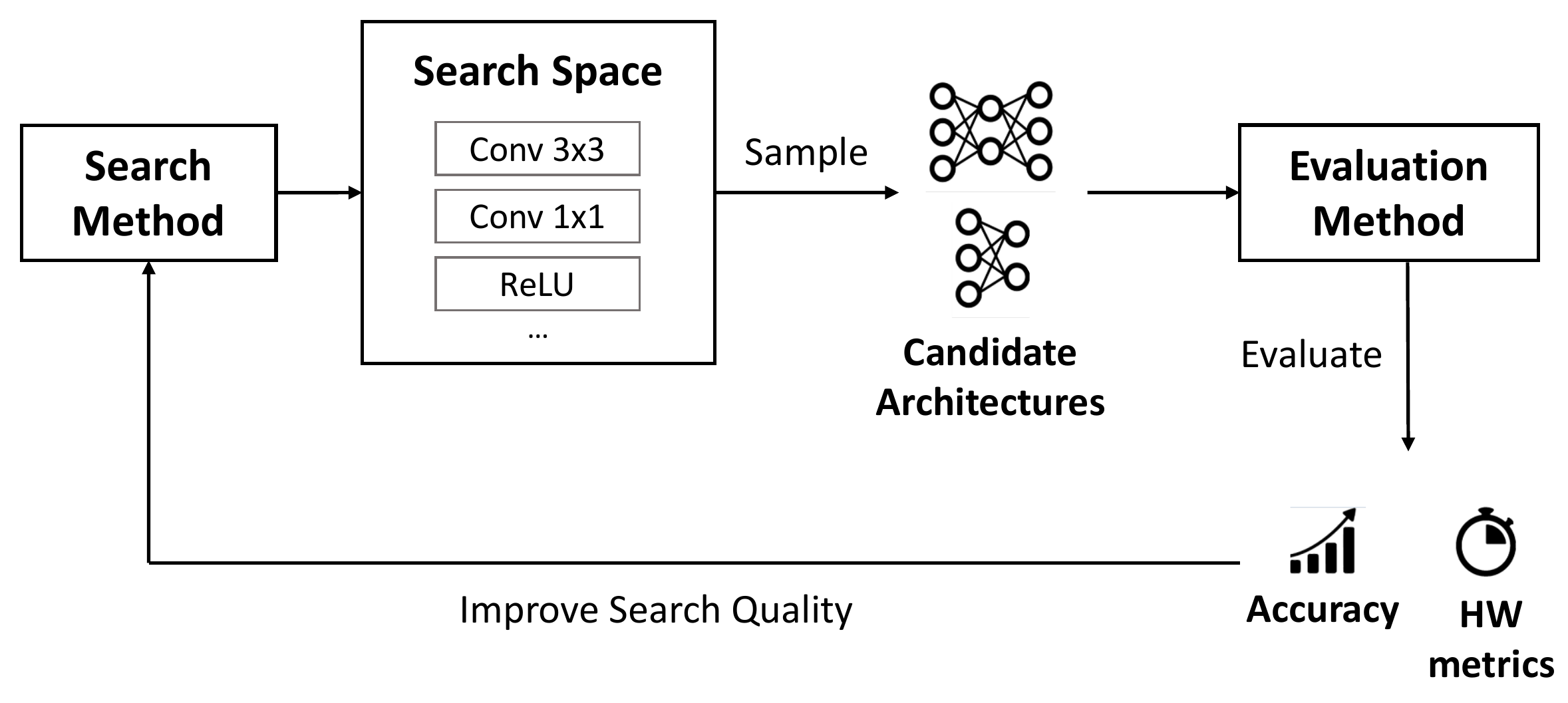}
  }
  \vspace{2mm}
  \caption{
  Illustration of the general structure of NAS frameworks.
  Candidate DNN architectures are sampled from the search space according to the search method, and then they are evaluated. 
  The evaluation result is then used by the search method to guide better exploration of architectures in the search space.
  }
  \label{fig:nas_overview}
  }
\end{figure}

Broadly speaking, a NAS framework consists of three main components: search space; search method; and evaluation method~\cite{elsken2019neural,benmeziane2021comprehensive}.
The \textit{search space} consists of a set of valid operations (e.g., convolution, pooling, activation, etc.) and their connectivity that define valid DNN architectures, from which a candidate model can be drawn.
Prior knowledge and human intuition regarding good DNN designs is often necessary in order to  restrict the search space and improve the search efficiency.
The \textit{search method} defines how to explore the search space. 
Exhaustive search is obviously intractable.
Therefore, it is critical to have methods for quickly exploring the search space and sampling candidate architectures.
The \textit{evaluation method} is a way of assessing how well candidate architectures perform on unseen data.
The most naive method is to evaluate all candidate architectures after the full training process is complete. 
However, this incurs a large overhead, and more efficient methods of estimating performance are 
often used as a proxy for the final accuracy.
Fig.~\ref{fig:nas_overview} schematically shows these different components.

Below, we discuss each of these components in more detail.
Note that the main purpose of this section is not to conduct a thorough survey of existing works, but instead to provide a broader overview on various methodologies for improving NAS from a practitioner's standpoint.
We refer readers to \cite{elsken2019neural,ren2021comprehensive,benmeziane2021comprehensive,sekanina2021neural} for more comprehensive survey on NAS.

\vspace{2mm}
\subsubsection{\textbf{Search Space}}
The search space for NAS defines a set of valid DNN architectures over which the NAS framework can search. 
Designing a proper search space is critical, as its size and coverage can directly affect the final outcome of the NAS framework.
One naive principle of designing a search space is the \textit{layer-wise search}~\cite{zoph2016neural,cai2018proxylessnas,guo2020single} where each layer (or operation) can be searched independently from other layers.
For instance, in~\cite{zoph2016neural}, the RNN controller model produces a description of individual layer in a sequence to construct a candidate DNN architecture.

However, the layer-wise search space often suffers from the large search space size that grows exponentially with the depth of candidate architectures, and this could degrade the search efficiency and the final performance.
The \textit{cell-wise search}~\cite{zoph2018learning,pham2018efficient,dong2018dpp,zhong2018practical,liu2017hierarchical} can alleviate this shortcoming by searching cells (i.e., blocks or modules that consist of multiple layers) rather than an entire architecture, which can later be stacked up repeatedly to compose an architecture.
This is motivated by many successful hand-designed DNN architectures that consist of repeating cells or blocks of a similar structure~\cite{he2016deep,howard2017mobilenets}.
NASNet~\cite{zoph2018learning} is one of the earliest works that proposes to search two types of cells: the normal cell, which stacks up multiple times without changing spatial resolution; and the reduction cell, which is inserted once every fixed number of repeated normal cells, in order to reduce the spatial dimension towards the output layers.
This significantly reduces the search time by 7$\times$ compared to the previous layer-wise search method proposed by the same authors~\cite{zoph2016neural}.
Likewise, the cell-wise search space substantially reduces the search space (as cells are much smaller than the whole network) by imposing an additional structural constraint in valid DNN architecture,
and therefore it has been widely adopted in follow-up works~\cite{liu2018darts,real2019regularized}

\vspace{2mm}
\subsubsection{\textbf{Search Method}}
\label{sec:searchmethod}

\begin{figure*}[t!]
    \centering
    \includegraphics[width=\linewidth]{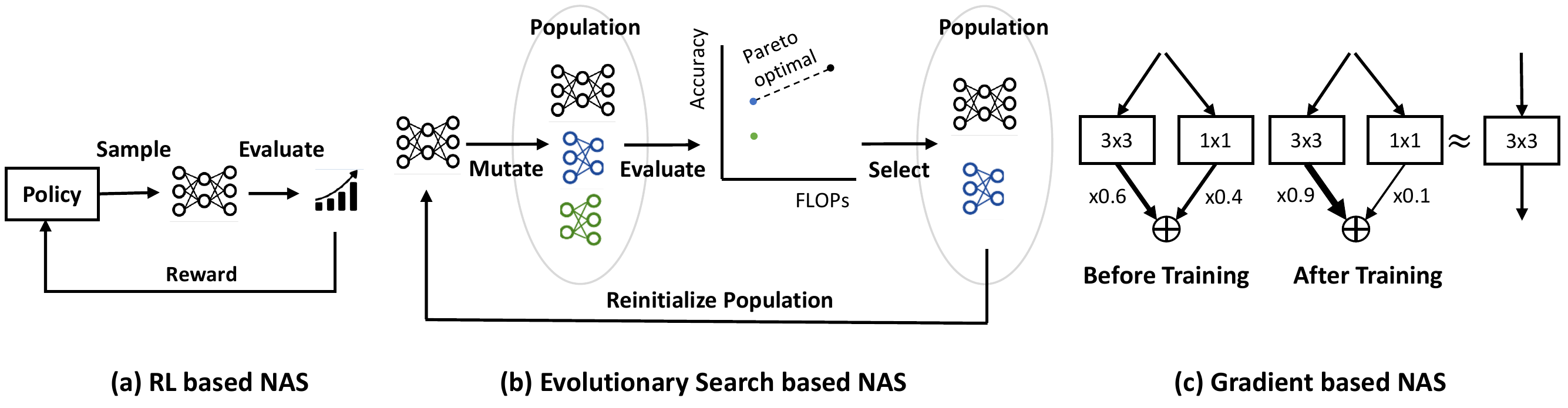}
    \caption{Comparison of different NAS search methods.
    (a) RL-based methods employ a controller that samples architectures based on a policy, which is reinforced by the evaluation results of the sampled architecture as a reward. 
    (b) Evolutionary search-based methods initialize a population, sample them based on the evaluation results, and then generate a next-round population by mutating the remaining architectures. 
    (c) Gradient-based methods (e.g., continuous relaxation) train weights along with model parameters that are multiplied to each operation choice. After the training, the weights are converged to favor a particular operation over the others, thus approximating the sampled architecture. 
    }
    \label{fig:nas_search}
\end{figure*}

Since the NAS search space is usually too large for an exhaustive search, efficient search methods are necessary to ensure overall performance.
In early work on NAS, \textit{RL-based} methods were used as the search method
~\cite{zoph2016neural, zoph2018learning,pham2018efficient,baker2016designing,zhong2018practical} (Fig.~\ref{fig:nas_search}, a).
At a high level, RL-based NAS frameworks contain the controller (i.e., RL agent) that takes an action of sampling DNN architectures, whose evaluation accuracy after training is fed into the controller as a reward signal
to refine its sampling policy.
The controller can be trained using different RL algorithms such as policy gradient~\cite{zoph2016neural} or Q-learning~\cite{baker2016designing}.

An alternative search strategy for NAS is \textit{evolutionary search}~\cite{real2019regularized,liu2017hierarchical} (Fig.~\ref{fig:nas_search}, b).
Here one initializes a population of different DNN architectures, 
which are then mutated (e.g., by adding, removing, or changing layers), evaluated, and selected based on their validation accuracy in every evolution step.
This generates a new population for the subsequent evolution step.
The search cost for evolutionary search can be quite expensive, as it requires validating all DNNs in the population for every evolution step. 
Therefore, it is often coupled with various methods to reduce validation costs such as weight sharing.
These will be discussed in more detail in Sec.~\ref{section:weight_sharing}.

The aforementioned methods can be regarded as a black-box optimization problem over a discrete search space. 
Due to the discrete nature of the search space with a large number of tunable knobs, the search cost can become prohibitively large.
This is further exacerbated by the long evaluation time of a single RL or evolution iteration, which often requires training from scratch.
For instance, RL-based NASNet~\cite{zoph2018learning} and evolutionary search-based AmoebaNet~\cite{real2019regularized} require a few thousands of GPU hours for end-to-end search~\cite{ren2021comprehensive}.
In contrast, DARTS~\cite{liu2018darts} proposes the \textit{continuous relaxation} of the search space, which allows them to efficiently explore and optimize the search space through gradient-based optimization methods (Fig.~\ref{fig:nas_search}, c).
In essence, DARTS introduces a trainable weight to allow for a weighted average of multiple operations, instead of requiring a selection of a single operation.
This weight can be trained alongside other model parameters during training, and it can eventually converge to favor a particular operation over the others.
This method reduces the search cost from thousands of GPU hours in the preceding RL or evolutionary search based methods to a few hours. 
Due to the search efficiency, the \textit{gradient based search} has become a popular choice for many NAS frameworks~\cite{wu2019fbnet,wan2020fbnetv2}.

\vspace{2mm}
\subsubsection{\textbf{Weight Sharing and Supernetwork}}
\label{section:weight_sharing}
One of the main challenges with NAS methods is the prohibitive training cost.
To address this, ENAS~\cite{pham2018efficient} proposed \textit{weight sharing}.
ENAS views a DNN model as a directed acyclic graph, 
where the nodes represent the computation with their own trainable weights 
and the edges represent the information flow from one node to another. 
Then, an individual candidate DNN can be regarded as a sub-network of a larger, over-parameterized super-network (supernet).
This redefines NAS as a process of searching for good sub-networks sampled from the supernet whose weights are shared across all sub-networks. 
Once the supernet is trained, its sub-networks can be sampled and evaluated without the need to train the models from scratch.
This significantly reduces the overall search cost.

This method, also known as the \textit{supernet-based} NAS, was picked up by several subsequent algorithms~\cite{bender2018understanding,cai2018proxylessnas,wu2019fbnet,liu2018darts,cai2019once,guo2020single,yu2020bignas}.
In particular, Single Path One-Shot NAS~\cite{guo2020single} constructs a supernet by stacking the choice blocks.
The choice block consists of multiple operation choices (e.g., convolution with different kernel sizes or skip operation) from which a single operation can be selected at a time.
For every training step, a different sub-network is obtained and trained by uniformly sampling one operation for each choice block, expecting all sub-networks with different permutations of choices to be trained fully and equally.
After training, an evolutionary algorithm is applied to search optimal sub-networks from the supernet, without paying the expensive costs of from-scratch training.

However, the accuracy of sub-networks obtained from a fully-trained supernet is typically inferior to the same model architectures trained from scratch in a stand-alone fashion~\cite{bender2018understanding}.
Therefore, the discovered sub-network architectures often need to be re-trained.
To address this, Once-For-All~\cite{cai2019once} proposes the progressive shrinking algorithm, and BigNAS~\cite{yu2020bignas} proposes the sandwich rule and in-place distillation.
Both aim to train a supernet in a way that its sub-networks can achieve good accuracy (i.e., comparable accuracy to the from-scratch trained counterparts) without an additional training process.
These methods can have a high value from a practical standpoint as sub-networks can be sampled (e.g., via evolutionary search) and immediately deployed.
 
\vspace{2mm}
\subsubsection{\textbf{Evaluation Method}}
One needs a metric to evaluate sampled architectures on a validation dataset to rank the ``goodness'' of candidate architectures.
The early NAS algorithms~\cite{zoph2016neural,baker2016designing} fully trained sampled architectures until convergence, which is not feasible for large datasets.
A widely adopted strategy for applying NAS to larger-scale tasks is to discover an accurate cell architecture using a smaller dataset (e.g., CIFAR-10 in computer vision) and then apply it to building a larger model for a larger dataset (e.g., ImageNet)~\cite{liu2018progressive,liu2018darts,real2019regularized,tan2019mnasnet,luo2018neural}.
The premise here is that a DNN architecture optimized for one task can be transferred well to other tasks in a similar domain.
This premise has been challenged by some of the recent NAS work~\cite{cai2018proxylessnas}.
Supernet-based NAS algorithms can be a good alternative to avoiding the use of proxy tasks~\cite{bender2018understanding,cai2018proxylessnas,wu2019fbnet,liu2018darts,cai2019once,guo2020single,yu2020bignas}.
These algorithms require only a single iteration of supernet training, which can be performed directly on large-scale datasets without prohibitive compute requirements.

\begin{boxA}
\textbf{Summary (Sec.~\ref{sec:nas}. NAS):} Neural architecture search (NAS) is a promising alternative to hand-designing efficient DNNs.
NAS consists of: (1) a search space that defines valid candidate architectures; (2) a search method that defines how to efficiently explore the search space; and (3) an evaluation method for evaluating the goodness of candidate architectures.
Despite its potential, NAS presents its own set of challenges, which often necessitate manual tuning of the search space, and which can be prohibitively expensive in terms of time and resources.
To address this, many recent advances in the NAS community have focused on improving search efficiency.
Notable methodologies include: 
(1) the cell-based search that confines the search space size;
(2) the continuous relaxation of the search space that allows efficient gradient-based optimization methods;
(3) the weight sharing scheme across candidate architectures;
and (4) faster evaluation methods for the candidate architecture performance.
\end{boxA}

\subsection{Hardware-aware NAS}
\label{sec:hwnas}
Hardware-aware NAS aims to optimize not only the accuracy of DNNs but also the various performance metrics (such as latency, energy consumption, or memory usage) on target hardware platforms.
One key question here is how to incorporate these metrics into learning. It is often difficult to quickly measure the latency or energy consumption of a candidate model.
As such, most works in the literature only consider FLOPs or total number of parameters.
However, as also discussed above in Sec.~\ref{subsec:model-analysis}, FLOPs does not necessarily have a correlation with latency or energy.
Therefore, multiple hardware-aware NAS frameworks have been introduced to directly consider latency instead, or to use approximate metrics for measuring it (e.g., measuring latency of individual layers and accumulating them to approximate total latency, as opposed to measuring the end-to-end runtime).
Here, we discuss popular strategies to incorporate hardware performance into NAS frameworks.
For a more exhaustive survey on hardware-aware NAS techniques and their algorithmic details, see~\cite{benmeziane2021comprehensive}.

The most straightforward way is to \textit{directly measure} hardware performance and bring it as an additional optimization objective for NAS frameworks~\cite{tan2019mnasnet,lin2020mcunet}.
For instance, MNasNet~\cite{tan2019mnasnet} extends the existing RL-based NAS framework to the multi-objective optimization setting.
It aims to maximize accuracy, while limiting latency on the target platform to less than a certain target latency.
Instead of solving the multi-purpose optimization problem, it combines the two objectives (accuracy and latency) into a single objective, by taking a weighted product.
This modified objective is then provided as a reward for updating the controller.
By directly optimizing latency on the target platform, MNasNet finds DNN architectures that are $\sim2\times$ faster than MobileNetV2~\cite{sandler2018mobilenetv2} and NASNet~\cite{zoph2018learning} with a comparable ImageNet classification accuracy on a Pixel phone.

Another notable work is MCUNet~\cite{lin2020mcunet} that targets searching DNNs for resource-constrained microcontrollers.
Unlike GPUs or mobile devices, microcontrollers for tiny IoT devices lack large memory and storage. 
As a result, it is critical to design a model that fits their tight memory budgets.
MCUNet incorporates its supernet-based NAS framework~\cite{guo2020single,bender2018understanding} with TinyEngine, a lightweight inference engine that the authors have developed as part of the project. 
In this way, MCUNet samples sub-networks for every evolutionary step and feeds them to TinyEngine to optimize the memory scheduling and measure the optimal memory usage.

However, due to the limited number of available devices, measuring hardware performance directly can be slow and not parallelizable~\cite{yang2018netadapt}.
Furthermore, it is not possible to pre-measure the hardware performance for all possible DNNs in the search space~\cite{wu2019fbnet}.
To overcome this issue, some of the hardware-aware NAS methods incorporate operation-wise \textit{lookup tables}~\cite{yang2018netadapt,wu2019fbnet,wan2020fbnetv2}.
Rather than storing the end-to-end hardware performance, the lookup table only contains pre-measured performance numbers of individual operations which can be summed up to estimate the overall performance of a given DNN.
In FBNet~\cite{wu2019fbnet}, for instance, the latency number estimated from a lookup table is used as a regularizer term in its gradient based NAS framework to penalize operations that would be slow on the target hardware device.

Finally, some hardware-aware NAS frameworks rely on lightweight \textit{prediction models} that can quickly predict hardware performance numbers for a given DNN.
For instance, ProxylessNAS~\cite{cai2018proxylessnas} has trained a model that takes as inputs a DNN configuration (e.g., operation types, input and output shapes, and other operation attributes like kernel sizes) and outputs the estimated latency on the target hardware platform.

\begin{boxA}
\textbf{Summary (Sec.~\ref{sec:hwnas}. HW-Aware NAS): }
Hardware efficiency metrics can be coupled with NAS loss function to find an architecture that considers both accuracy as well as latency (or similar metrics).  
While directly measuring performance metrics on a real hardware environment is the most accurate method, it can be slow and poorly parallelizable.
Instead, the hardware performance can be estimated in high accuracy using an operation-wise lookup table or by training a simple prediction model.
\end{boxA}

\begin{table}[!ht]
\caption{
Summary of existing literature on Transformer-specific NAS techniques. SPOS and OFA stand for Single Path One-Shot~\cite{guo2020single} and Once-for-All~\cite{cai2019once}, respectively.
}
\begin{center}
\small{
\setlength{\tabcolsep}{5pt}{
\begin{tabular}{c|c|ccc}
\toprule
\multirow{2}{*}{\textbf{Name}} & \multirow{2}{*}{\textbf{Domain}} & \textbf{Search} & \textbf{Search} & \textbf{Weight}  \\
 & & \textbf{Space} & \textbf{Method} & \textbf{sharing} \\
\midrule
Evolved Tfm.~\cite{so2019evolved} & \multirow{4}{*}{NLP} & Cell & EA & $\times$ \\ 
HAT~\cite{wang2020hat} &  & Layer & EA  & OFA~\cite{cai2019once} \\ 
NAS-BERT~\cite{xu2021bert} &  & Layer & EA  & SPOS~\cite{guo2020single} \\ 
Primer~\cite{so2021searching} &  & Cell & EA & $\times$ \\
\midrule
Autoformer~\cite{wu2021autoformer} & \multirow{5}{*}{CV} & Layer & EA & SPOS~\cite{guo2020single}\\ 
GLiT~\cite{chen2021glit} &  & Layer & EA & SPOS~\cite{guo2020single} \\ 
ViT-ResNAS~\cite{liao2021searching} &  & Layer & EA  & SPOS~\cite{guo2020single} \\  
NAS-ViT~\cite{gong2021nasvit} &  & Layer & EA  & BigNAS~\cite{yu2020bignas} \\ 
BurgerFormer~\cite{yang2022searching} &  & Layer & EA & SPOS~\cite{guo2020single}\\

\bottomrule
\end{tabular}
}
}
\end{center}
\label{table:nas_transformer}
\end{table}

\subsection{Transformer-specific NAS}
\label{sec:transformer-nas}
Early work on NAS focused on CNN models mostly for computer vision applications.
However, after the Transformer architecture was introduced and matured enough to achieve state-of-the-art results not just for NLP tasks but also for other tasks, several works started to explore
NAS methods to find more efficient alternatives.
With the introduction and maturation of the Transformer architecture, which allowed for state-of-the-art results on a variety of tasks, a number of recent works have begun to explore the use of NAS methods to find more efficient alternatives.
As  the Transformer architecture was initially developed for NLP tasks, the earliest NAS works for Transformers were primarily in this domain.

Evolved Transformer~\cite{so2019evolved} was one of the earliest attempts to apply NAS for searching better Transformer architectures, and it did so via an evolutionary search algorithm.
Inspired by NASNet~\cite{zoph2018learning}, Evolved Transformer adopts the cell-wise search space to search two cell structures.
Each of these cell structures can be stacked for multiple times to form the encoder and decoder of the encoder-decoder Transformer architecture.
The cell structure contains a stack of multiple blocks, and each block has its own hyperparameters such as operation type, normalization type,  and dimensions which can be searched. 
The main challenge here is that NLP tasks require a much longer time to train and evaluate (e.g., the popular WMT 2014 En-De translation benchmark contains over 3 million sentence pairs). 
Furthermore, unlike the previous works~\cite{liu2018progressive,liu2018darts,real2019regularized,tan2019mnasnet,luo2018neural} for CNNs that found CIFAR-10 to be a reasonable proxy for much larger ImageNet, these NLP tasks  do not typically have good smaller proxy tasks.
To address this, Evolved Transformer proposes to dynamically allocate resources to more promising architectures
by early stopping those who fail to achieve the hurdle fitness within a small number of steps. 

Due to the large computational cost of training Transformers on NLP tasks, weight sharing and supernet based NAS have become popular options.
HAT~\cite{wang2020hat} extends the Once-for-All~\cite{cai2019once} scheme to Transformer architectures to train a single supernet from which sub-networks with different depths, number of heads, and dimensions can be sampled.
Furthermore, HAT is hardware-aware, in that it directly optimizes for latency along with accuracy using a multi-layer latency prediction model.
HAT shares the benefits of Once-for-All, which allows sub-networks to be sampled through evolutionary search and deployed immediately to target hardware devices without retraining.

NAS-BERT~\cite{xu2021bert} is another supernet based NAS for Transformers that extends Single Path One-Shot~\cite{guo2020single}.
Different from the aforementioned methods, NAS-BERT proposes a NAS method that can be applied at the pre-training stage of encoder-only BERT so as to be agnostic to downstream tasks.
In order to avoid the prohibitive cost of directly performing architecture search in a big supernet on the heavy pre-training task, NAS-BERT employs two novel techniques: 
(1) \textit{block-wise training}, that splits the entire supernet into multiple blocks of successive Transformer layers which are then trained separately; and 
(2) \textit{progressive shrinking}, that dynamically prunes less promising sub-networks based on their validation loss.

Primer~\cite{so2021searching} searches for a more efficient decoder-only Transformer for auto-regressive language modeling.
Unlike the majority of NAS methods that view a model as a connection of multiple operations selected from a NAS search space, Primer views it as a single valid Tensorflow (TF) program comprised of fine-grained TF primitive operations like addition, exponential, convolution, and many others.
Using evolutionary search, it targets to search a TF program defining a decoder block that can be stacked multiple times to form an auto-regressive language model. 
The hope is that this minimizes inductive bias when designing the search space, as the possible set of operations and their connectivity are no longer pre-determined by human experts. 
In order to reduce the heavy computational cost of auto-regressive pre-training, Primer brings the idea of hurdles of Evolved Transformer~\cite{so2019evolved}.
Additionally, it uses relatively small LM1B dataset as a proxy task  to discover model architecture,
which is then transferred to much larger target tasks such as PG-19~\cite{rae2019compressive} and C4~\cite{raffel2020exploring}.

The Transformer architecture, initially developed for NLP tasks, has been adapted for use in the field of CV.
Referred to as Vision Transformers (ViTs)~\cite{dosovitskiy2020image, touvron2021training, liu2021swin}, these models have been demonstrated to outperform popular CNN architectures in various CV applications, thus driving research towards the development of NAS techniques to automatically design better ViT architectures. 
 However, due to the architectural similarities, these works have much in common with the NAS methodologies for NLP-targeted Transformers.
For instance, Autoformer~\cite{wu2021autoformer} and ViT-ResNAS~\cite{liao2021searching} are extensions of Single Path One-Shot~\cite{guo2020single} to the ViT  search space, including depth, hidden dimensions, and the number of heads of each Transformer layer.
Burgerformer~\cite{yang2022searching} takes a step further to take into account the micro design, i.e., the type of operations, activations, and normalization, as well.
NASViT extends BigNAS~\cite{yu2020bignas} and AlphaNet~\cite{wang2021alphanet} to apply  the sandwich sampling rule to train a supernet. 
GLiT~\cite{chen2021glit} proposes a hierarchical NAS scheme for searching hybrid convolution-attention architectures. 
It determines the number of convolutional and multi-head attention heads in each layer in the first stage of NAS, as well as the detailed hyperparameters such as dimensions in the second stage.

One noticeable characteristic of most of the NAS methods introduced above for Transformer architectures (both for NLP and CV applications) is their use of supernet-based, weight-shared methodologies, which is summarized in Tab.~\ref{table:nas_transformer}.
This is presumably due to the immense computational cost associated with training Transformer architectures.
The use of supernet-based NAS can limit the range of architectures that can be discovered, due to the large constraints it imposes on the search space, 
which may prevent the discovery of unique or innovative architectures. 
Therefore, there is a need to explore better ways to balance the flexibility and efficiency of NAS techniques.

\begin{boxA}
\textbf{Summary (Sec.~\ref{sec:transformer-nas}. Transformer-specific NAS):} The existing NAS frameworks have been extended to design more efficient Transformer architectures.
Due to the large computational cost for training Transformer models, which is even further exacerbated when combined with unsupervised pre-training methodologies, most of the existing methods heavily rely on the weight-sharing scheme followed by evolutionary search.
A key challenge in Transformer-specific NAS is that existing work is primarily limited to tuning relatively trivial hyperparameters such as hidden dimensions, depth, and number of heads.  
However, this is likely to preclude the discovery of novel Transformer variants.
\end{boxA}

\subsection{Case Study: Running NAS and Co-design on the Transformer}
\label{subsection:nas_case_study}

So far,  we have discussed the general concept of NAS, its application to hardware-aware scenarios, and its extension into the Transformer architecture.
Here, we conduct a case study to demonstrate the performance gains of applying NAS to Transformer inference on Gemmini, with the goal of optimizing not only accuracy, but also hardware costs such as latency and energy.

\begin{table}[!t]
\caption{NAS Search Space.}
\begin{center}
\begin{tabular}{c|c}
\toprule
\textbf{{Parameter}}   & \textbf{Range of Values} \\
\midrule
$N$    & $\{3,4,5,6\}$ \\
$h$    & $\{4,6,8,10,12\}$   \\
$d$   & ${384-768}$, step size=96    \\
$d_{\text{FFN}}$   & $768-3072$, step size=128  \\
\bottomrule
\end{tabular}
\end{center}
\label{tab:searchspace}
\end{table}

\subsubsection{\textbf{Experiment Setup.}}
As a baseline architecture, we use a 6-layer Transformer architecture with all other model configurations remaining the same as BERT-Base or GPT-2 (see the details in Tab.~\ref{table:symbols}).
We consider Language Modeling task, and we train a randomly initialized model on the WikiText-2~\cite{merity2016pointer} benchmark with 37k training examples and 4k validation examples using a language modeling training objective.
 To evaluate the model performance, we measured perplexity on the validation examples, excluding empty strings, where lower scores indicate better performance.
 The stand-alone baseline model was trained for 50 epochs with the Adam optimizer and a linear learning rate scheduling with a peak learning rate in the range $\{5, 2, 1, 0.5\}\times10^{-5}$.
The training examples are concatenated to reach a maximum sequence length of 512 and batched using a batch size of 16.

For NAS, we adopt the BigNAS-style~\cite{yu2020bignas} strategy to train a supernet, and then we used an evolutionary algorithm to search sub-networks out of the fully trained supernet.
The NAS search space is comprised of various combinations of the number of layers $l$, number of heads $h$, hidden dimension $d$, and FFN dimension $d_{\text{FFN}}$ (see Tab.~\ref{tab:searchspace} for details). 
For supernet training, we use the same training hyperparameters as the stand-alone training, except that in each training iteration, we sample four sub-networks: the largest possible; the smallest possible; and two randomly sampled sub-networks.
The model parameter update is then performed using the sandwich rule, which involves taking the average of the gradients collected from the backward paths of these four sub-networks.

For the evolutionary search, we initialize a population of 40 sub-networks and perform 40 rounds of evolution iterations.
In each iteration, the validation perplexity and energy-delay-product (EDP) of each sub-network on the target hardware are collected, and only the sub-networks that are Pareto-optimal are retained.
Here, we use EDP as a single hardware cost metric, as it allows for the conversion of a multi-objective optimization problem into a single-objective optimization problem, by combining both latency and energy into one metric.
The retained sub-networks are then mutated with a mutation probability of 0.2 to refill the population for the next iteration.
To measure the hardware cost, we use a lookup table-based method for quickly assessing the latency and energy consumption of each sub-network on the target hardware, instead of using RTL (Register Transfer Logic) simulation, which can be time-consuming.
The entries in the lookup table are obtained from Timeloop~\cite{parashar2019timeloop} simulations, which provide simulated latency and energy numbers for each operation. 
The end-to-end latency and energy are then estimated by summing the per-operation costs.
After the evolutionary search, the Pareto-optimal sub-networks are then evaluated with an RTL simulator to obtain a more precise estimation of the latency.
For the energy measure, we continue to use the numbers from Timeloop, as it is technically challenging to measure the energy consumption via RTL.

For the target hardware, we use Gemmini with the optimizations applied in Sec.~\ref{sec:thiswork} with the dedicated normalization units for running non-linear operations on-chip. We configure Gemmini with a scratchpad size of 64 kB and accumulator size of 256 kB based on the insights in Sec.~\ref{sec:hardware:memory-hierarchy} to maximize the accumulator size.

\subsubsection{\textbf{Experiment Results.}}
\label{sec:experimentresults}

\begin{figure*}[ht]
     \centering
     \begin{subfigure}[b]{0.45\textwidth}
         \centering
         \includegraphics[width=\textwidth]{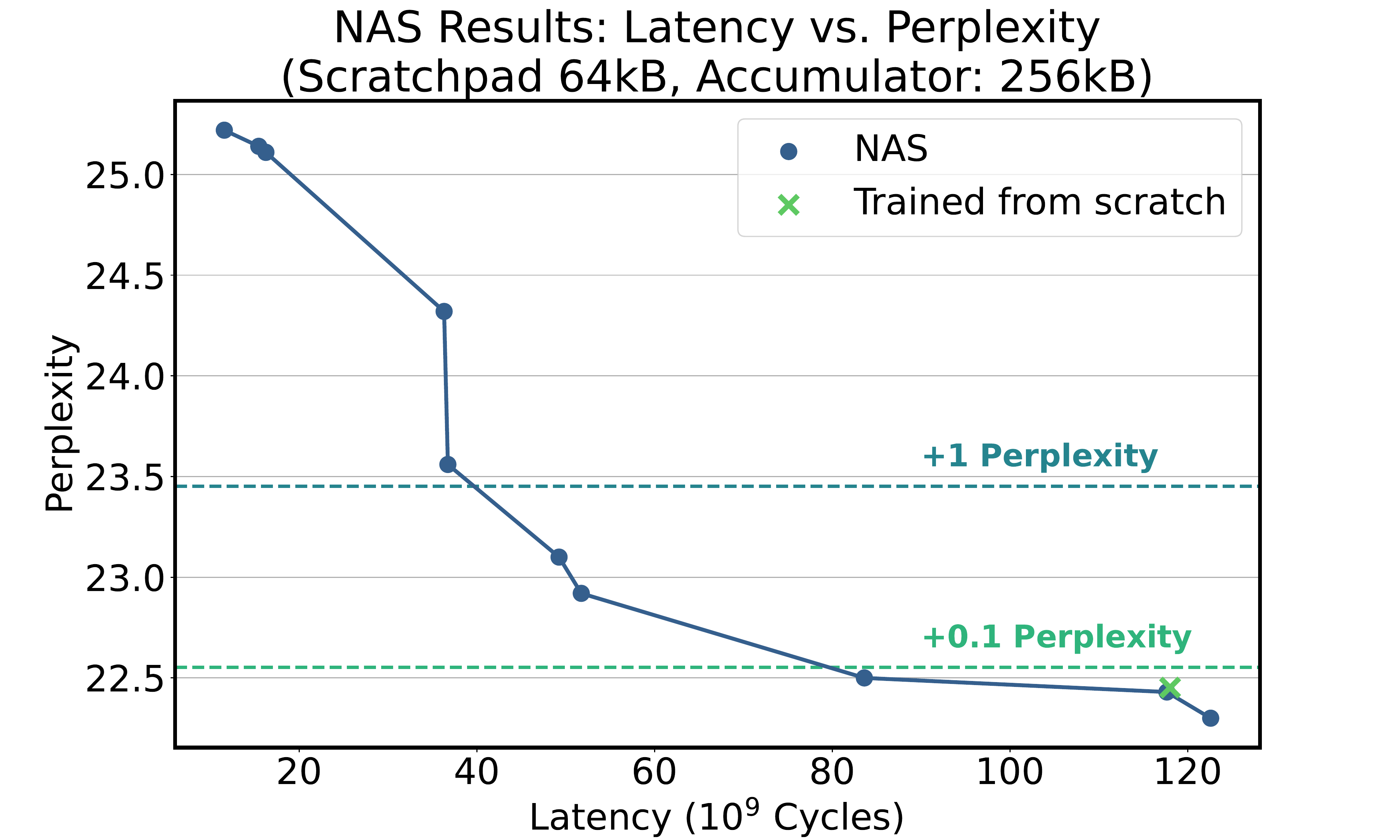}
     \end{subfigure}
     \hfill
     \begin{subfigure}[b]{0.45\textwidth}
         \centering
         \includegraphics[width=\textwidth]{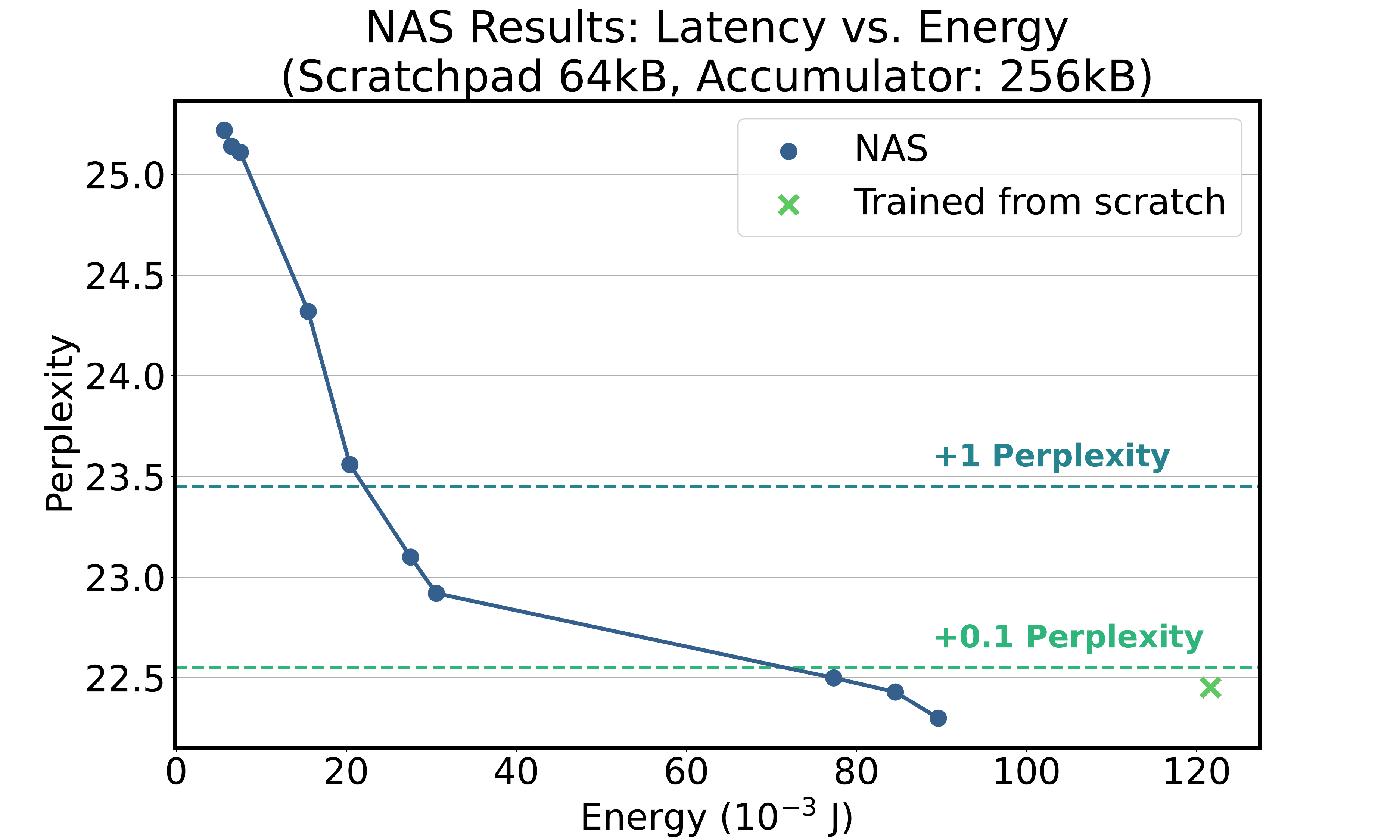}
     \end{subfigure}
     
  \caption{(Left) Latency-perplexity and (Right) Energy-perplexity plots of the Transformer architectures found via evolutionary search on our optimal Gemmini hardware configuration.
  Similar to Fig.~\ref{fig:nas_edp}, lower perplexity indicates better performance, and we plot lines to illustrate +0.1 and +1 point perplexity degradation.
  }
  \label{fig:evolution}
\end{figure*}

We show the NAS Pareto-frontier results for both latency and energy in Fig.~\ref{fig:evolution} (blue curves) where each point corresponds to a different Transformer architecture that has been found from the evolutionary search algorithm discussed above. Additionally, we plot the baseline 6-layer Transformer model trained from scratch as a reference ($\times$ mark).
All the EDP values are normalized with the baseline EDP.
Note that the baseline model corresponds to the largest Transformer architecture in our search space in Tab.~\ref{tab:searchspace}.

\begin{table}[!t]
\caption{Sample Architecture found using NAS with $3.6\times10^9$ EDP and $22.51$ perplexity.}
\begin{center}
\begin{tabular}{c|c}
\toprule
\textbf{{Parameter}}   & \textbf{Values} \\
\midrule
$N$    & $6$ \\
$h$    & $[12, 6, 12, 8, 10, 6]$   \\
$d$   & $672$   \\
$d_{\text{FFN}}$   & $[1280, 1280, 2560, 768, 2048, 1024]$  \\
\bottomrule
\end{tabular}
\end{center}
\label{tab:sample_nas_arch}
\end{table}


\begin{figure}[!t]
\centering{
\centerline{
  \includegraphics[width=0.45\textwidth]{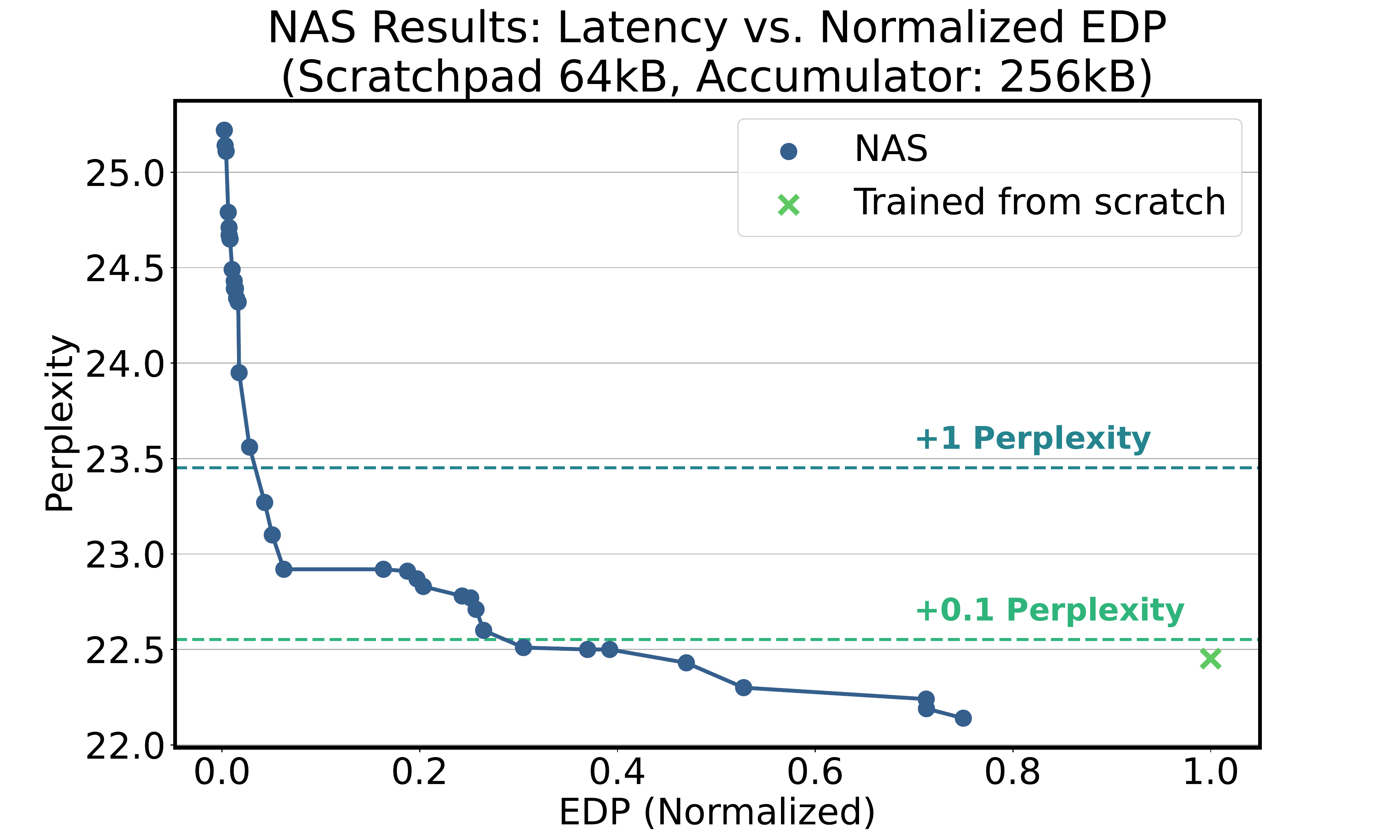}
  }
  \vspace{2mm}
  \caption{
  EDP-perplexity plots of the Transformer architectures found via evolutionary search on our Gemmini hardware configuration. Lower perplexity indicates better performance of the trained models. For better comparison, we additionally plot lines to illustrate +0.1 and +1 point perplexity degradation.
  }
  \label{fig:nas_edp}
  }
\end{figure}

We first present results from the evolutionary search process over EDP in Fig.~\ref{fig:nas_edp}.
As can be seen in the plot, the NAS framework allows us to obtain multiple Transformer architectures with better hardware cost to perplexity trade-offs.
That is, it finds architectures with similar or even better perplexity, as compared to the baseline with smaller hardware costs.
As an example, we select the architecture with the lowest EDP while having less than +0.1 perplexity loss, whose EDP is $3.6\times10^9$ and perplexity is $22.51$.
The architecture parameters are listed in Table~\ref{tab:sample_nas_arch}.
This architecture illustrates the importance of a diverse search space, as the number of attention heads varies from 6 to 12 in each layer, and as the fully connected layer dimensions vary from 768 to 2560. By being able to change these parameters on a per-layer basis, one may discover more Pareto-optimal architectures compared to if these parameters were fixed for every layer.

In Fig.~\ref{fig:evolution}, we separate out latency and energy, and substitute in RTL values for the latency. As one can see, it is possible to attain a 1.4$\times$ reduction in latency versus the baseline Transformer trained from scratch with 0.1 point perplexity degradation. 
If one could tolerate about one point degradation in perplexity, latency can be reduced by 2.4$\times$, and possibly even further with more advanced architecture search techniques. 
With regards to energy, one can attain a 1.6$\times$ improvement considering 0.1 point perplexity degradation, 
and 4.4$\times$ if perplexity is allowed to increase by 1 point. 
Taking both together, it is possible to reduce EDP by 2.2$\times$ with just 0.1 point perplexity degradation, and 10.6$\times$ with 1 point perplexity degradation.
These examples illustrate the power of co-design in allowing practitioners to choose a combination that best matches their needs. It is important to note that this represents a single run of our co-design methodology on a specific hardware platform, and results may vary depending on the target hardware and optimization goals.

\begin{boxA}
\textbf{Summary (Sec.~\ref{subsection:nas_case_study}. NAS for Transformers Case Study):} This case study used a supernet-based NAS to sample diverse architectures followed by an evolutionary search to discover Pareto-optimal architectures that trade-off between perplexity and energy-delay-product, a measure of runtime efficiency. Many discovered architectures have significant latency and energy improvements compared to the baseline trained from scratch when running on an optimized Gemmini hardware accelerator. The importance of using NAS to explore the search space is underscored by the fact that many well-performing architectures use diverse layer configurations. When trained on WikiText-2 language modeling benchmark, these techniques found Transformer architectures with a 2.2$\times$ EDP reduction while tolerating a 0.1 point perplexity degradation, and 10.6$\times$ with a 1 point degradation over the baseline.
\end{boxA}
\section{Conclusion}
The Transformer architecture~\cite{vaswani2017attention} has revolutionized the field of natural language understanding~\cite{devlin2018bert,radford2018improving,radford2019language,liu2019roberta,yang2019xlnet,lan2019albert,raffel2019exploring}, which has been further accelerated with the recent development of large language models with hundreds of billions of parameters~\cite{brown2020language,scao2022bloom,du2022glam,rae2021scaling,smith2022using,hoffmann2022training,chowdhery2022palm}. 
This architecture has also been extended to a wide range of fields, including computer vision~\cite{mehta2021mobilevit,chen2022mobile,li2022efficientformer,cai2022efficientvit}, and speech recognition~\cite{gulati2020conformer,burchi2021efficient,kim2022squeezeformer}. 
While Transformer models have shown significant performance improvements, their growing size and run-time complexity present a critical challenge in efficient inference. 
While DNN accelerators that enable fast and efficient deep learning computation can be a  viable solution,
there is still limited understanding of the run-time characteristics and bottlenecks of Transformer workloads, as well as the design principles necessary for effectively running these models, in comparison to CNN architectures.

In this paper, 
we have conducted a comprehensive analysis of Transformer workloads, in order to better understand run-time characteristics and identify performance bottlenecks of Transformers running on commodity hardware and accelerators (Sec.~\ref{sec:architecture_and_performance}).
Furthermore, we have performed an extensive survey of the current hardware and software solutions, with the goal of identifying potential
optimization opportunities in the full-stack deployment of Transformers.
Specifically, our survey covered the following topics: 
\begin{itemize}[leftmargin=5mm]
   \item The design of hardware architectures for Transformer inference, including the impact of the non-linear operations on designing hardware accelerators (Sec.~\ref{sec:hardware_design});
\item Optimization strategies such as pruning and quantization that can be applied to a fixed Transformer architecture for better performance (Sec.~\ref{sec:optimization});
\item Mapping and scheduling of operations in the Transformer architecture and the associated challenges (Sec.~\ref{sec:scheduling}); and
\item The use of automated NAS for designing more efficient Transformer architectures and adapting them to target hardware (Sec.~\ref{sec:nas}).
\end{itemize}

\noindent
The key findings from this study include:
\begin{itemize}[leftmargin=5mm]
    \item Despite the small FLOPs count, the nonlinear operations in Transformers can be highly influential on overall performance, if they are not taken into account properly when designing domain-specific accelerators. The computation of nonlinear operations such as Softmax and LayerNorm also requires computing runtime statistics, whereas the BatchNorm operations in CNNs can be absorbed into prior convolutional layers during~inference.
    \item Hardware design for CNNs may not necessarily be the same as that for Transformers. 
    For instance, increasing the accumulator sizes to enable higher output reuse yielded significant performance improvement in Gemmini for Transformer applications.
    \item It appears less complex to schedule matmuls in Transformers than convolutions in CNNs, due to the fact that scheduling matmuls involves 3 loops, as compared to 6 for convolutions. 
    However, we observed that scheduling matmuls involve similar amounts of decision points and a wide range of performance outcomes, making it as challenging as scheduling convolutions.
    \item Fusing LayerNorms with the preceding matmuls in the Transformer architecture imposes several constraints on the mapping, particularly related to tile sizes.
    As a result, careful consideration must be taken when deciding whether to fuse operations, contrary to the common belief that operator fusion is generally~beneficial.
    \end{itemize}

Finally, throughout the paper, we conducted case studies to quantify the advantages of co-design and co-optimization techniques across the stack on full-stack Transformer inference. 
Overall, the result exhibited 88.7$\times$ EDP improvement without a noticeable performance drop compared to a naive implementation without full-stack considerations.
\begin{itemize}[leftmargin=5mm]
    \item In Sec.~\ref{sec:thiswork}, we applied hardware design techniques in order to avoid the high communication overhead associated with offloading unsupported operations to the host CPU. Gemmini was originally designed for CNN workloads, and making it perform Softmax, LayerNorm, and GELU on-chip required additional changes.
    Our implementation of  dedicated normalization units to support Softmax and LayerNorm had an associated 5$-$15\% area cost and 8\% latency increase.
    Nonetheless, this extra overhead was compensated by the gain achieved by running the nonlinear operations on-chip using the polynomial approximation proposed in I-BERT~\cite{ibert}. 
    Combined with memory hierarchy re-balancing, this provided a net 39.6$\times$ latency reduction.   
    \item In Section~\ref{subsection:nas_case_study}, we ran NAS to search for Pareto-optimal Transformer architectures given the tradeoff between EDP and perplexity on a popular language modeling task. 
    We used Timeloop simulated numbers to estimate the cost of various architectures within a large search space and guide the automated NAS process.
    The total contribution as shown in Fig.~\ref{fig:evolution} is 2.24$\times$ EDP reduction without a noticeable perplexity drop, and 10.56$\times$ EDP reduction with 1 perplexity drop.
\end{itemize}

We anticipate that our in-depth analysis and results, along with the comprehensive survey presented in this paper will facilitate further advencements in understanding Transformer inference and optimizing its inference efficiency from various angles.
We believe that this will enable Transformers to reach their full potential and expand their application to a much wider range of areas than what they have achieved so far.

\section{Acknowledgements}
We acknowledge gracious support from Meta and in particular
Michael Anderson, Satish Nadathur and Summer Deng, as well as
Google Cloud, Google TRC team, and specifically Jonathan Caton, Prof. David Patterson, and Jing Li.
Prof. Keutzer's lab is sponsored by Intel corporation, Intel VLAB team, Intel One-API center of
excellence, as well as funding through BDD and BAIR.
Sehoon Kim would like to acknowledge the support from Korea Foundation for Advanced Studies (KFAS).
Amir Gholami was supported through funding from Samsung SAIT.
Michael W. Mahoney would also like to acknowledge
a J. P. Morgan Chase Faculty Research Award 
as well as 
the DOE, NSF, and ONR.
Our conclusions do not necessarily reflect the position or the policy of our sponsors, and no official endorsement should be~inferred.

\bibliographystyle{plain}
\bibliography{references}

\begin{thebibliography}{100}

\bibitem{tpu_edge}
{Edge TPU}.
\newblock \url{https://cloud.google.com/edge-tpu/}.
\newblock Accessed: 2018-12-05.

\bibitem{openai2019chatgpt}
Chatgpt: Optimizing language models for dialogue.
\newblock \url{https://openai.com/blog/chatgpt/}, 2022.

\bibitem{abadi2016tensorflow}
Mart{\'\i}n Abadi, Paul Barham, Jianmin Chen, Zhifeng Chen, Andy Davis, Jeffrey
  Dean, Matthieu Devin, Sanjay Ghemawat, Geoffrey Irving, Michael Isard, et~al.
\newblock $\{$TensorFlow$\}$: a system for $\{$Large-Scale$\}$ machine
  learning.
\newblock In {\em USENIX Symposium on Operating Systems Design and
  Implementation (OSDI)}, 2016.

\bibitem{abts2022groq}
Dennis Abts, John Kim, Garrin Kimmell, Matthew Boyd, Kris Kang, Sahil Parmar,
  Andrew Ling, Andrew Bitar, Ibrahim Ahmed, and Jonathan Ross.
\newblock The groq software-defined scale-out tensor streaming multiprocessor:
  From chips-to-systems architectural overview.
\newblock In {\em IEEE Hot Chips Symposium}, pages 1--69, 2022.

\bibitem{acharya2018approach}
Aravind Acharya, Uday Bondhugula, and Albert Cohen.
\newblock An approach for finding permutations quickly: Fusion and dimension
  matching.
\newblock {\em arXiv preprint arXiv:1803.10726}, 2018.

\bibitem{acharya2018polyhedral}
Aravind Acharya, Uday Bondhugula, and Albert Cohen.
\newblock Polyhedral auto-transformation with no integer linear programming.
\newblock In {\em Proceedings of the ACM SIGPLAN Conference on Programming
  Language Design and Implementation (PLDI)}, 2018.

\bibitem{adams2019learning}
Andrew Adams, Karima Ma, Luke Anderson, Riyadh Baghdadi, Tzu-Mao Li, Michael
  Gharbi, Benoit Steiner, Steven Johnson, Kayvon Fatahalian, Fr\'{e}do Durand,
  and Jonathan Ragan-Kelley.
\newblock Learning to optimize halide with tree search and random programs.
\newblock {\em ACM Transactions on Graphics (TOG)}, 2019.

\bibitem{snapea}
Vahideh Akhlaghi, Amir Yazdanbakhsh, Kambiz Samadi, Rajesh~K. Gupta, and Hadi
  Esmaeilzadeh.
\newblock Snapea: Predictive early activation for reducing computation in deep
  convolutional neural networks.
\newblock In {\em Proceedings of the 45th Annual International Symposium on
  Computer Architecture}, page 662–673, 2018.

\bibitem{cnvlutin}
Jorge Albericio, Patrick Judd, Tayler Hetherington, Tor Aamodt, Natalie~Enright
  Jerger, and Andreas Moshovos.
\newblock Cnvlutin: Ineffectual-neuron-free deep neural network computing.
\newblock In {\em Proceedings of the 43rd International Symposium on Computer
  Architecture}, 2016.

\bibitem{alwani2016fused}
Manoj Alwani, Han Chen, Michael Ferdman, and Peter Milder.
\newblock Fused-layer cnn accelerators.
\newblock In {\em Proceedings of the International Symposium on
  Microarchitecture (MICRO)}, 2016.

\bibitem{ansel2014opentuner}
Jason Ansel, Shoaib Kamil, Kalyan Veeramachaneni, Jonathan Ragan-Kelley,
  Jeffrey Bosboom, Una-May O'Reilly, and Saman Amarasinghe.
\newblock Opentuner: An extensible framework for program autotuning.
\newblock In {\em Proceedings of the International Conference on Parallel
  Architectures and Compilation Techniques (PACT)}, 2014.

\bibitem{armcortexm}
{ARM}.
\newblock {Cortex-M},
  https://developer.arm.com/ip-products/processors/cortex-m, 2020.

\bibitem{approximatecomputing}
Giorgos Armeniakos, Georgios Zervakis, Dimitrios Soudris, and J\"{o}rg Henkel.
\newblock Hardware approximate techniques for deep neural network accelerators:
  A survey.
\newblock {\em ACM Comput. Surv.}, mar 2022.
\newblock Just Accepted.

\bibitem{bagehadi2019tiramisu}
R.~{Baghdadi}, J.~{Ray}, M.~B. {Romdhane}, E.~D. {Sozzo}, A.~{Akkas},
  Y.~{Zhang}, P.~{Suriana}, S.~{Kamil}, and S.~{Amarasinghe}.
\newblock Tiramisu: A polyhedral compiler for expressing fast and portable
  code.
\newblock In {\em International Symposium on Code Generation and Optimization
  (CGO)}, 2019.

\bibitem{baghdadi2015pencil}
Riyadh Baghdadi, Ulysse Beaugnon, Albert Cohen, Tobias Grosser, Michael Kruse,
  Chandan Reddy, Sven Verdoolaege, Adam Betts, Alastair~F. Donaldson, Jeroen
  Ketema, Javed Absar, Sven Van~Haastregt, Alexey Kravets, Anton Lokhmotov,
  Robert David, and Elnar Hajiyev.
\newblock Pencil: A platform-neutral compute intermediate language for
  accelerator programming.
\newblock In {\em Proceedings of the International Conference on Parallel
  Architectures and Compilation Techniques (PACT)}, 2015.

\bibitem{baker2016designing}
Bowen Baker, Otkrist Gupta, Nikhil Naik, and Ramesh Raskar.
\newblock Designing neural network architectures using reinforcement learning.
\newblock {\em arXiv preprint arXiv:1611.02167}, 2016.

\bibitem{bender2018understanding}
Gabriel Bender, Pieter-Jan Kindermans, Barret Zoph, Vijay Vasudevan, and Quoc
  Le.
\newblock Understanding and simplifying one-shot architecture search.
\newblock In {\em International conference on machine learning}, pages
  550--559. PMLR, 2018.

\bibitem{benmeziane2021comprehensive}
Hadjer Benmeziane, Kaoutar~El Maghraoui, Hamza Ouarnoughi, Smail Niar, Martin
  Wistuba, and Naigang Wang.
\newblock A comprehensive survey on hardware-aware neural architecture search.
\newblock {\em arXiv preprint arXiv:2101.09336}, 2021.

\bibitem{logsumexp-and-softmax}
Pierre Blanchard, Desmond~J Higham, and Nicholas~J Higham.
\newblock {Accurately computing the log-sum-exp and softmax functions}.
\newblock {\em IMA Journal of Numerical Analysis}, 41(4):2311--2330, 08 2020.

\bibitem{bondhugula2016pluto+}
Uday Bondhugula, Aravind Acharya, and Albert Cohen.
\newblock The pluto+ algorithm: A practical approach for parallelization and
  locality optimization of affine loop nests.
\newblock {\em ACM Transactions on Programming Languages and Systems (TOPLAS)},
  2016.

\bibitem{bondhugula2008practical}
Uday Bondhugula, Albert Hartono, Jagannathan Ramanujam, and Ponnuswamy
  Sadayappan.
\newblock A practical automatic polyhedral parallelizer and locality optimizer.
\newblock In {\em Proceedings of the ACM SIGPLAN Conference on Programming
  Language Design and Implementation (PLDI)}, 2008.

\bibitem{brown2020language}
Tom~B Brown, Benjamin Mann, Nick Ryder, Melanie Subbiah, Jared Kaplan, Prafulla
  Dhariwal, Arvind Neelakantan, Pranav Shyam, Girish Sastry, Amanda Askell,
  et~al.
\newblock Language models are few-shot learners.
\newblock {\em arXiv preprint arXiv:2005.14165}, 2020.

\bibitem{burchi2021efficient}
Maxime Burchi and Valentin Vielzeuf.
\newblock Efficient conformer: Progressive downsampling and grouped attention
  for automatic speech recognition.
\newblock In {\em 2021 IEEE Automatic Speech Recognition and Understanding
  Workshop (ASRU)}, pages 8--15. IEEE, 2021.

\bibitem{cai2022efficientvit}
Han Cai, Chuang Gan, and Song Han.
\newblock Efficientvit: Enhanced linear attention for high-resolution
  low-computation visual recognition.
\newblock {\em arXiv preprint arXiv:2205.14756}, 2022.

\bibitem{cai2019once}
Han Cai, Chuang Gan, Tianzhe Wang, Zhekai Zhang, and Song Han.
\newblock Once-for-all: Train one network and specialize it for efficient
  deployment.
\newblock {\em arXiv preprint arXiv:1908.09791}, 2019.

\bibitem{cai2018proxylessnas}
Han Cai, Ligeng Zhu, and Song Han.
\newblock Proxylessnas: Direct neural architecture search on target task and
  hardware.
\newblock {\em arXiv preprint arXiv:1812.00332}, 2018.

\bibitem{hardwaresoftwaresurvey}
Maurizio Capra, Beatrice Bussolino, Alberto Marchisio, Guido Masera, Maurizio
  Martina, and Muhammad Shafique.
\newblock Hardware and software optimizations for accelerating deep neural
  networks: Survey of current trends, challenges, and the road ahead.
\newblock {\em {IEEE} Access}, 8:225134--225180, 2020.

\bibitem{cer2017semeval}
Daniel Cer, Mona Diab, Eneko Agirre, Inigo Lopez-Gazpio, and Lucia Specia.
\newblock Semeval-2017 task 1: Semantic textual similarity-multilingual and
  cross-lingual focused evaluation.
\newblock {\em arXiv preprint arXiv:1708.00055}, 2017.

\bibitem{chatarasi2020marvel}
Prasanth Chatarasi, Hyoukjun Kwon, Natesh Raina, Saurabh Malik, Vaisakh
  Haridas, Angshuman Parashar, Michael Pellauer, Tushar Krishna, and Vivek
  Sarkar.
\newblock Marvel: A data-centric compiler for dnn operators on spatial
  accelerators, 2020.

\bibitem{chen2021glit}
Boyu Chen, Peixia Li, Chuming Li, Baopu Li, Lei Bai, Chen Lin, Ming Sun, Junjie
  Yan, and Wanli Ouyang.
\newblock Glit: Neural architecture search for global and local image
  transformer.
\newblock In {\em Proceedings of the IEEE/CVF International Conference on
  Computer Vision}, pages 12--21, 2021.

\bibitem{chen2023accelerating}
Charlie Chen, Sebastian Borgeaud, Geoffrey Irving, Jean-Baptiste Lespiau,
  Laurent Sifre, and John Jumper.
\newblock Accelerating large language model decoding with speculative sampling.
\newblock {\em arXiv preprint arXiv:2302.01318}, 2023.

\bibitem{chen2015mxnet}
Tianqi Chen, Mu~Li, Yutian Li, Min Lin, Naiyan Wang, Minjie Wang, Tianjun Xiao,
  Bing Xu, Chiyuan Zhang, and Zheng Zhang.
\newblock Mxnet: A flexible and efficient machine learning library for
  heterogeneous distributed systems.
\newblock {\em arXiv preprint arXiv:1512.01274}, 2015.

\bibitem{chen2018tvm}
Tianqi Chen, Thierry Moreau, Ziheng Jiang, Lianmin Zheng, Eddie Yan, Haichen
  Shen, Meghan Cowan, Leyuan Wang, Yuwei Hu, Luis Ceze, et~al.
\newblock $\{$TVM$\}$: An automated end-to-end optimizing compiler for deep
  learning.
\newblock In {\em 13th $\{$USENIX$\}$ Symposium on Operating Systems Design and
  Implementation ($\{$OSDI$\}$ 18)}, pages 578--594, 2018.

\bibitem{diannao}
Tianshi Chen, Zidong Du, Ninghui Sun, Jia Wang, Chengyong Wu, Yunji Chen, and
  Olivier Temam.
\newblock Diannao: A small-footprint high-throughput accelerator for ubiquitous
  machine-learning.
\newblock In {\em Proceedings of the 19th International Conference on
  Architectural Support for Programming Languages and Operating Systems},
  ASPLOS '14, pages 269--284, New York, NY, USA, 2014. ACM.

\bibitem{eyeriss}
Y.~{Chen}, J.~{Emer}, and V.~{Sze}.
\newblock Eyeriss: A spatial architecture for energy-efficient dataflow for
  convolutional neural networks.
\newblock In {\em 2016 ACM/IEEE 43rd Annual International Symposium on Computer
  Architecture (ISCA)}, pages 367--379, June 2016.

\bibitem{chen2022mobile}
Yinpeng Chen, Xiyang Dai, Dongdong Chen, Mengchen Liu, Xiaoyi Dong, Lu~Yuan,
  and Zicheng Liu.
\newblock Mobile-former: Bridging mobilenet and transformer.
\newblock In {\em Proceedings of the IEEE/CVF Conference on Computer Vision and
  Pattern Recognition}, pages 5270--5279, 2022.

\bibitem{eyeriss-isca2016}
Yu-Hsin Chen, Joel Emer, and Vivienne Sze.
\newblock {Eyeriss: A Spatial Architecture for Energy-efficient Dataflow for
  Convolutional Neural Networks}.
\newblock In {\em Proceedings of the International Symposium on Computer
  Architecture (ISCA)}, 2016.

\bibitem{dataflows}
Yu-Hsin Chen, Joel Emer, and Vivienne Sze.
\newblock Using dataflow to optimize energy efficiency of deep neural network
  accelerators.
\newblock {\em IEEE Micro}, 37(3):12--21, 2017.

\bibitem{chen2019eyeriss}
Yu-Hsin Chen, Tien-Ju Yang, Joel Emer, and Vivienne Sze.
\newblock Eyeriss v2: A flexible accelerator for emerging deep neural networks
  on mobile devices.
\newblock {\em IEEE Journal on Emerging and Selected Topics in Circuits and
  Systems}, 2019.

\bibitem{dadiannao}
Yunji Chen, Tao Luo, Shaoli Liu, Shijin Zhang, Liqiang He, Jia Wang, Ling Li,
  Tianshi Chen, Zhiwei Xu, Ninghui Sun, and Olivier Temam.
\newblock {DaDianNao: A Machine-learning Supercomputer}.
\newblock In {\em Proceedings of the International Symposium on
  Microarchitecture (MICRO)}, 2014.

\bibitem{chetlur2014cudnn}
Sharan Chetlur, Cliff Woolley, Philippe Vandermersch, Jonathan Cohen, John
  Tran, Bryan Catanzaro, and Evan Shelhamer.
\newblock cudnn: Efficient primitives for deep learning.
\newblock {\em arXiv preprint arXiv:1410.0759}, 2014.

\bibitem{choiaccelerating}
Jaewan Choi, Hailong Li, Byeongho Kim, Seunghwan Hwang, and Jung~Ho Ahn.
\newblock Accelerating transformer networks through recomposing softmax layers.
\newblock In {\em International Symposium on Workload Characterization
  (IISWC)}, 2021.

\bibitem{gpu}
Jack Choquette, Wishwesh Gandhi, Olivier Giroux, Nick Stam, and Ronny
  Krashinsky.
\newblock Nvidia a100 tensor core gpu: Performance and innovation.
\newblock {\em IEEE Micro}, 41(2):29--35, 2021.

\bibitem{chowdhery2022palm}
Aakanksha Chowdhery, Sharan Narang, Jacob Devlin, Maarten Bosma, Gaurav Mishra,
  Adam Roberts, Paul Barham, Hyung~Won Chung, Charles Sutton, Sebastian
  Gehrmann, et~al.
\newblock Palm: Scaling language modeling with pathways.
\newblock {\em arXiv preprint arXiv:2204.02311}, 2022.

\bibitem{asap7}
L.T. Clark, V.~Vashishtha, L.~Shifren, A.~Gujia, S.~Sinha, B.~Cline,
  C.~Ramamurthya, and G.~Yeric.
\newblock {ASAP7: A 7-nm FinFET Predictive Process Design Kit}.
\newblock {\em Microelectronics Journal}, 2016.

\bibitem{dai2019transformer}
Zihang Dai, Zhilin Yang, Yiming Yang, Jaime~G Carbonell, Quoc Le, and Ruslan
  Salakhutdinov.
\newblock Transformer-xl: Attentive language models beyond a fixed-length
  context.
\newblock In {\em Proceedings of the 57th Annual Meeting of the Association for
  Computational Linguistics}, pages 2978--2988, 2019.

\bibitem{sparseandirregular}
Shail Dave, Riyadh Baghdadi, Tony Nowatzki, Sasikanth Avancha, Aviral
  Shrivastava, and Baoxin Li.
\newblock Hardware acceleration of sparse and irregular tensor computations of
  ml models: A survey and insights.
\newblock {\em Proceedings of the IEEE}, 109(10):1706--1752, 2021.

\bibitem{dave2019dmazerunner}
Shail Dave, Youngbin Kim, Sasikanth Avancha, Kyoungwoo Lee, and Aviral
  Shrivastava.
\newblock {DMazeRunner}: Executing perfectly nested loops on dataflow
  accelerators.
\newblock {\em ACM Transactions on Embedded Computing Systems}, 2019.

\bibitem{compressionandacceleration}
Lei Deng, Guoqi Li, Song Han, Luping Shi, and Yuan Xie.
\newblock Model compression and hardware acceleration for neural networks: A
  comprehensive survey.
\newblock {\em Proceedings of the IEEE}, 108(4):485--532, 2020.

\bibitem{detrey2005parameterized}
J{\'e}r{\'e}mie Detrey and Florent de~Dinechin.
\newblock A parameterized floating-point exponential function for fpgas.
\newblock In {\em Proceedings. 2005 IEEE International Conference on
  Field-Programmable Technology, 2005.}, pages 27--34. IEEE, 2005.

\bibitem{dettmersgpt3}
Tim Dettmers, Mike Lewis, Younes Belkada, and Luke Zettlemoyer.
\newblock Gpt3. int8 (): 8-bit matrix multiplication for transformers at scale.
\newblock In {\em Advances in Neural Information Processing Systems}.

\bibitem{devlin2018bert}
Jacob Devlin, Ming-Wei Chang, Kenton Lee, and Kristina Toutanova.
\newblock {BERT}: Pre-training of deep bidirectional transformers for language
  understanding.
\newblock {\em arXiv preprint arXiv:1810.04805}, 2018.

\bibitem{dolan2005automatically}
William~B Dolan and Chris Brockett.
\newblock Automatically constructing a corpus of sentential paraphrases.
\newblock In {\em Proceedings of the Third International Workshop on
  Paraphrasing (IWP2005)}, 2005.

\bibitem{dong2018dpp}
Jin-Dong Dong, An-Chieh Cheng, Da-Cheng Juan, Wei Wei, and Min Sun.
\newblock Dpp-net: Device-aware progressive search for pareto-optimal neural
  architectures.
\newblock In {\em Proceedings of the European Conference on Computer Vision
  (ECCV)}, pages 517--531, 2018.

\bibitem{dong2019hawq}
Zhen Dong, Zhewei Yao, Amir Gholami, Michael~W Mahoney, and Kurt Keutzer.
\newblock {HAWQ}: Hessian aware quantization of neural networks with
  mixed-precision.
\newblock In {\em Proceedings of the IEEE International Conference on Computer
  Vision}, pages 293--302, 2019.

\bibitem{dosovitskiy2020image}
Alexey Dosovitskiy, Lucas Beyer, Alexander Kolesnikov, Dirk Weissenborn,
  Xiaohua Zhai, Thomas Unterthiner, Mostafa Dehghani, Matthias Minderer, Georg
  Heigold, Sylvain Gelly, et~al.
\newblock An image is worth 16x16 words: Transformers for image recognition at
  scale.
\newblock {\em arXiv preprint arXiv:2010.11929}, 2020.

\bibitem{RandNLA_PCMIchapter_chapter}
Petros Drineas and Michael~W Mahoney.
\newblock Lectures on randomized numerical linear algebra.
\newblock In {\em The Mathematics of Data}, IAS/Park City Mathematics Series,
  pages 1--48. AMS/IAS/SIAM, 2018.

\bibitem{du2022glam}
Nan Du, Yanping Huang, Andrew~M Dai, Simon Tong, Dmitry Lepikhin, Yuanzhong Xu,
  Maxim Krikun, Yanqi Zhou, Adams~Wei Yu, Orhan Firat, et~al.
\newblock Glam: Efficient scaling of language models with mixture-of-experts.
\newblock In {\em International Conference on Machine Learning}, pages
  5547--5569. PMLR, 2022.

\bibitem{shidianno}
Zidong Du, Robert Fasthuber, Tianshi Chen, Paolo Ienne, Ling Li, Tao Luo,
  Xiaobing Feng, Yunji Chen, and Olivier Temam.
\newblock Shidiannao: Shifting vision processing closer to the sensor.
\newblock In {\em 2015 ACM/IEEE 42nd Annual International Symposium on Computer
  Architecture (ISCA)}, pages 92--104, 2015.

\bibitem{logsumexp}
Robert Eisele.
\newblock The log-sum-exp trick in machine learning, 2016.

\bibitem{elsken2019neural}
Thomas Elsken, Jan~Hendrik Metzen, and Frank Hutter.
\newblock Neural architecture search: A survey.
\newblock {\em The Journal of Machine Learning Research}, 20(1):1997--2017,
  2019.

\bibitem{npu}
Hadi Esmaeilzadeh, Adrian Sampson, Luis Ceze, and Doug Burger.
\newblock {Neural Acceleration for General-Purpose Approximate Programs}.
\newblock In {\em Proceedings of the International Symposium on
  Microarchitecture (MICRO)}, 2012.

\bibitem{fan2019reducing}
Angela Fan, Edouard Grave, and Armand Joulin.
\newblock Reducing transformer depth on demand with structured dropout.
\newblock {\em arXiv preprint arXiv:1909.11556}, 2019.

\bibitem{adaptivebutterfly}
Hongxiang Fan, Thomas Chau, Stylianos~I. Venieris, Royson Lee, Alexandros
  Kouris, Wayne Luk, Nicholas~D. Lane, and Mohamed~S. Abdelfattah.
\newblock Adaptable butterfly accelerator for attention-based nns via hardware
  and algorithm co-design, 2022.

\bibitem{cooptimizedframework}
Chao Fang, Aojun Zhou, and Zhongfeng Wang.
\newblock An algorithm-hardware co-optimized framework for accelerating n:m
  sparse transformers.
\newblock {\em IEEE Transactions on Very Large Scale Integration (VLSI)
  Systems}, pages 1--14, 2022.

\bibitem{flamand2018gap}
Eric Flamand, Davide Rossi, Francesco Conti, Igor Loi, Antonio Pullini, Florent
  Rotenberg, and Luca Benini.
\newblock Gap-8: A risc-v soc for ai at the edge of the iot.
\newblock In {\em 2018 IEEE 29th International Conference on
  Application-specific Systems, Architectures and Processors (ASAP)}, pages
  1--4. IEEE, 2018.

\bibitem{frankle2018lottery}
Jonathan Frankle and Michael Carbin.
\newblock The lottery ticket hypothesis: Finding sparse, trainable neural
  networks.
\newblock {\em arXiv preprint arXiv:1803.03635}, 2018.

\bibitem{gale2019state}
Trevor Gale, Erich Elsen, and Sara Hooker.
\newblock The state of sparsity in deep neural networks.
\newblock {\em arXiv preprint arXiv:1902.09574}, 2019.

\bibitem{tetris-asplos17}
Mingyu Gao, Jing Pu, Xuan Yang, Mark Horowitz, and Christos Kozyrakis.
\newblock {Tetris: Scalable and Efficient Neural Network Acceleration with 3D
  Memory}.
\newblock In {\em Proceedings of the International Conference on Architectural
  Support for Programming Languages and Operation Systems (ASPLOS)}, 2017.

\bibitem{gemmini-dac}
Hasan Genc, Seah Kim, Alon Amid, Ameer Haj-Ali, Vighnesh Iyer, Pranav Prakash,
  Jerry Zhao, Daniel Grubb, Harrison Liew, Howard Mao, Albert Ou, Colin
  Schmidt, Samuel Steffl, John Wright, Ion Stoica, Jonathan Ragan-Kelley, Krste
  Asanovic, Borivoje Nikolic, and Yakun~Sophia Shao.
\newblock Gemmini: Enabling systematic deep-learning architecture evaluation
  via full-stack integration.
\newblock In {\em Proceedings of the 58th Annual Design Automation Conference
  (DAC)}, 2021.

\bibitem{quantizationmethods}
Amir Gholami, Sehoon Kim, Zhen Dong, Zhewei Yao, Michael~W. Mahoney, and Kurt
  Keutzer.
\newblock A survey of quantization methods for efficient neural network
  inference, 2021.

\bibitem{240gops}
Vinayak Gokhale, Jonghoon Jin, Aysegul Dundar, Berin Martini, and Eugenio
  Culurciello.
\newblock A 240 g-ops/s mobile coprocessor for deep neural networks.
\newblock In {\em 2014 IEEE Conference on Computer Vision and Pattern
  Recognition Workshops}, pages 696--701, 2014.

\bibitem{gong2021nasvit}
Chengyue Gong, Dilin Wang, Meng Li, Xinlei Chen, Zhicheng Yan, Yuandong Tian,
  Vikas Chandra, et~al.
\newblock Nasvit: Neural architecture search for efficient vision transformers
  with gradient conflict aware supernet training.
\newblock In {\em International Conference on Learning Representations}, 2021.

\bibitem{goyal2020power}
Saurabh Goyal, Anamitra~Roy Choudhury, Saurabh Raje, Venkatesan Chakaravarthy,
  Yogish Sabharwal, and Ashish Verma.
\newblock Power-bert: Accelerating bert inference via progressive word-vector
  elimination.
\newblock In {\em International Conference on Machine Learning}, pages
  3690--3699. PMLR, 2020.

\bibitem{grosser2011polly}
Tobias Grosser, Hongbin Zheng, Raghesh Aloor, Andreas Simb{\"u}rger, Armin
  Gr{\"o}{\ss}linger, and Louis-No{\"e}l Pouchet.
\newblock Polly-polyhedral optimization in llvm.
\newblock In {\em Proceedings of the First International Workshop on Polyhedral
  Compilation Techniques (IMPACT)}, 2011.

\bibitem{gulati2020conformer}
Anmol Gulati, James Qin, Chung-Cheng Chiu, Niki Parmar, Yu~Zhang, Jiahui Yu,
  Wei Han, Shibo Wang, Zhengdong Zhang, Yonghui Wu, et~al.
\newblock Conformer: Convolution-augmented transformer for speech recognition.
\newblock {\em arXiv preprint arXiv:2005.08100}, 2020.

\bibitem{guo2020single}
Zichao Guo, Xiangyu Zhang, Haoyuan Mu, Wen Heng, Zechun Liu, Yichen Wei, and
  Jian Sun.
\newblock Single path one-shot neural architecture search with uniform
  sampling.
\newblock In {\em European Conference on Computer Vision}, pages 544--560.
  Springer, 2020.

\bibitem{HLK+20RiseElevate}
Bastian Hagedorn, Johannes Lenfers, Thomas Kundefinedhler, Xueying Qin, Sergei
  Gorlatch, and Michel Steuwer.
\newblock Achieving high-performance the functional way: A functional pearl on
  expressing high-performance optimizations as rewrite strategies.
\newblock {\em Proc. ACM Program. Lang.}, 4(ICFP), aug 2020.

\bibitem{a3}
Tae~Jun Ham, S.~J. Jung, Seonghak Kim, Young~H. Oh, Yeonhong Park, Yongchan
  Song, Junghun Park, Sang-Hee Lee, K.~Park, J.~Lee, and Deog-Kyoon Jeong.
\newblock A$^3$: Accelerating attention mechanisms in neural networks with
  approximation.
\newblock {\em 2020 IEEE International Symposium on High Performance Computer
  Architecture (HPCA)}, pages 328--341, 2020.

\bibitem{elsa}
Tae~Jun Ham, Yejin Lee, Seong~Hoon Seo, Soosung Kim, Hyunji Choi, Sung~Jun
  Jung, and Jae~W. Lee.
\newblock Elsa: Hardware-software co-design for efficient, lightweight
  self-attention mechanism in neural networks.
\newblock In {\em 2021 ACM/IEEE 48th Annual International Symposium on Computer
  Architecture (ISCA)}, pages 692--705, 2021.

\bibitem{eie}
Song Han, Xingyu Liu, Huizi Mao, Jing Pu, Ardavan Pedram, Mark~A. Horowitz, and
  William~J. Dally.
\newblock Eie: Efficient inference engine on compressed deep neural network.
\newblock {\em SIGARCH Comput. Archit. News}, 44(3), June 2016.

\bibitem{he2016deep}
Kaiming He, Xiangyu Zhang, Shaoqing Ren, and Jian Sun.
\newblock Deep residual learning for image recognition.
\newblock In {\em Proceedings of the IEEE conference on computer vision and
  pattern recognition}, pages 770--778, 2016.

\bibitem{resnet50}
Kaiming He, Xiangyu Zhang, Shaoqing Ren, and Jian Sun.
\newblock Deep residual learning for image recognition.
\newblock In {\em 2016 IEEE Conference on Computer Vision and Pattern
  Recognition (CVPR)}, pages 770--778, 2016.

\bibitem{hegde2021mind}
Kartik Hegde, Po-An Tsai, Sitao Huang, Vikas Chandra, Angshuman Parashar, and
  Christopher~W Fletcher.
\newblock Mind mappings: enabling efficient algorithm-accelerator mapping space
  search.
\newblock In {\em Proceedings of the International Conference on Architectural
  Support for Programming Languages and Operation Systems (ASPLOS)}, 2021.

\bibitem{hendrycks2016gaussian}
Dan Hendrycks and Kevin Gimpel.
\newblock Gaussian error linear units ({GELU}s).
\newblock {\em arXiv preprint arXiv:1606.08415}, 2016.

\bibitem{hoffmann2022training}
Jordan Hoffmann, Sebastian Borgeaud, Arthur Mensch, Elena Buchatskaya, Trevor
  Cai, Eliza Rutherford, Diego de~Las Casas, Lisa~Anne Hendricks, Johannes
  Welbl, Aidan Clark, et~al.
\newblock Training compute-optimal large language models.
\newblock {\em arXiv preprint arXiv:2203.15556}, 2022.

\bibitem{energyproblem}
Mark Horowitz.
\newblock 1.1 computing's energy problem (and what we can do about it).
\newblock In {\em 2014 IEEE International Solid-State Circuits Conference
  Digest of Technical Papers (ISSCC)}, pages 10--14, 2014.

\bibitem{hou2020dynabert}
Lu~Hou, Zhiqi Huang, Lifeng Shang, Xin Jiang, Xiao Chen, and Qun Liu.
\newblock Dynabert: Dynamic bert with adaptive width and depth.
\newblock {\em Advances in Neural Information Processing Systems},
  33:9782--9793, 2020.

\bibitem{howard2017mobilenets}
Andrew~G Howard, Menglong Zhu, Bo~Chen, Dmitry Kalenichenko, Weijun Wang,
  Tobias Weyand, Marco Andreetto, and Hartwig Adam.
\newblock Mobilenets: Efficient convolutional neural networks for mobile vision
  applications.
\newblock {\em arXiv preprint arXiv:1704.04861}, 2017.

\bibitem{mobilenet}
Andrew~G. Howard, Menglong Zhu, Bo~Chen, Dmitry Kalenichenko, Weijun Wang,
  Tobias Weyand, Marco Andreetto, and Hartwig Adam.
\newblock Mobilenets: Efficient convolutional neural networks for mobile vision
  applications, 2017.

\bibitem{hruska2017movidius}
J~Hruska.
\newblock New movidius myriad x vpu packs a custom neural compute engine, 2017.

\bibitem{huang2021cosa}
Qijing Huang, Minwoo Kang, Grace Dinh, Thomas Norell, Aravind Kalaiah, James
  Demmel, John Wawrzynek, and Yakun~Sophia Shao.
\newblock Cosa: Scheduling by constrained optimization for spatial
  accelerators.
\newblock In {\em 2021 ACM/IEEE 48th Annual International Symposium on Computer
  Architecture (ISCA)}, pages 554--566. IEEE, 2021.

\bibitem{iandola2016squeezenet}
Forrest~N Iandola, Song Han, Matthew~W Moskewicz, Khalid Ashraf, William~J
  Dally, and Kurt Keutzer.
\newblock {SqueezeNet}: Alexnet-level accuracy with 50x fewer parameters and<
  0.5 mb model size.
\newblock {\em arXiv preprint arXiv:1602.07360}, 2016.

\bibitem{iandola2020squeezebert}
Forrest~N Iandola, Albert~E Shaw, Ravi Krishna, and Kurt~W Keutzer.
\newblock Squeezebert: What can computer vision teach nlp about efficient
  neural networks?
\newblock {\em arXiv preprint arXiv:2006.11316}, 2020.

\bibitem{YBR+22Exo}
Yuka Ikarashi, Gilbert~Louis Bernstein, Alex Reinking, Hasan Genc, and Jonathan
  Ragan-Kelley.
\newblock Exocompilation for productive programming of hardware accelerators.
\newblock In {\em Proceedings of the 43rd ACM SIGPLAN International Conference
  on Programming Language Design and Implementation}, PLDI 2022, page
  703–718, New York, NY, USA, 2022. Association for Computing Machinery.

\bibitem{iyer2017first}
Shankar Iyer, Nikhil Dandekar, and Kornl Csernai.
\newblock First quora dataset release: Question pairs.(2017).
\newblock {\em URL https://data. quora.
  com/First-Quora-Dataset-Release-Question-Pairs}, 2017.

\bibitem{jacob2018quantization}
Benoit Jacob, Skirmantas Kligys, Bo~Chen, Menglong Zhu, Matthew Tang, Andrew
  Howard, Hartwig Adam, and Dmitry Kalenichenko.
\newblock Quantization and training of neural networks for efficient
  integer-arithmetic-only inference.
\newblock In {\em Proceedings of the IEEE Conference on Computer Vision and
  Pattern Recognition}, pages 2704--2713, 2018.

\bibitem{jia2014caffe}
Yangqing Jia, Evan Shelhamer, Jeff Donahue, Sergey Karayev, Jonathan Long,
  Ross~B. Girshick, Sergio Guadarrama, and Trevor Darrell.
\newblock {Caffe: Convolutional Architecture for Fast Feature Embedding}.
\newblock {\em CoRR}, abs/1408.5093, 2014.

\bibitem{jia2019beyond}
Zhihao Jia, Matei Zaharia, and Alex Aiken.
\newblock Beyond data and model parallelism for deep neural networks.
\newblock In {\em Proceedings of Machine Learning and Systems (MLSys)}, 2019.

\bibitem{Jouppi2017}
N.~P. {Jouppi}, C.~{Young}, N.~{Patil}, D.~{Patterson}, G.~{Agrawal},
  R.~{Bajwa}, S.~{Bates}, S.~{Bhatia}, N.~{Boden}, A.~{Borchers}, R.~{Boyle},
  P.~{Cantin}, C.~{Chao}, C.~{Clark}, J.~{Coriell}, M.~{Daley}, M.~{Dau},
  J.~{Dean}, B.~{Gelb}, T.~V. {Ghaemmaghami}, R.~{Gottipati}, W.~{Gulland},
  R.~{Hagmann}, C.~R. {Ho}, D.~{Hogberg}, J.~{Hu}, R.~{Hundt}, D.~{Hurt},
  J.~{Ibarz}, A.~{Jaffey}, A.~{Jaworski}, A.~{Kaplan}, H.~{Khaitan},
  D.~{Killebrew}, A.~{Koch}, N.~{Kumar}, S.~{Lacy}, J.~{Laudon}, J.~{Law},
  D.~{Le}, C.~{Leary}, Z.~{Liu}, K.~{Lucke}, A.~{Lundin}, G.~{MacKean},
  A.~{Maggiore}, M.~{Mahony}, K.~{Miller}, R.~{Nagarajan}, R.~{Narayanaswami},
  R.~{Ni}, K.~{Nix}, T.~{Norrie}, M.~{Omernick}, N.~{Penukonda}, A.~{Phelps},
  J.~{Ross}, M.~{Ross}, A.~{Salek}, E.~{Samadiani}, C.~{Severn}, G.~{Sizikov},
  M.~{Snelham}, J.~{Souter}, D.~{Steinberg}, A.~{Swing}, M.~{Tan},
  G.~{Thorson}, B.~{Tian}, H.~{Toma}, E.~{Tuttle}, V.~{Vasudevan}, R.~{Walter},
  W.~{Wang}, E.~{Wilcox}, and D.~H. {Yoon}.
\newblock In-datacenter performance analysis of a tensor processing unit.
\newblock In {\em 2017 ACM/IEEE 44th Annual International Symposium on Computer
  Architecture (ISCA)}, pages 1--12, June 2017.

\bibitem{confuciuX-micro2020}
Sheng-Chun Kao, Geonhwa Jeong, and Tushar Krishna.
\newblock {ConfuciuX: Autonomous Hardware Resource Assignment for DNN
  Accelerators using Reinforcement Learning}.
\newblock In {\em Proceedings of the International Symposium on
  Microarchitecture (MICRO)}, 2020.

\bibitem{gamma-iccad2020}
Sheng-Chun Kao and Tushar Krishna.
\newblock {GAMMA: Automating the HW Mapping of DNN Models on Accelerators via
  Genetic Algorithm}.
\newblock In {\em Proceedings of the International Conference on Computer-Aided
  Design (ICCAD)}, 2020.

\bibitem{kao2022demystifyingNpu}
Sheng-Chun Kao, Angshuman Parashar, Po-An Tsai, and Tushar Krishna.
\newblock Demystifying map space exploration for npus, 2022.

\bibitem{kao2021optimized}
Sheng-Chun Kao, Suvinay Subramanian, Gaurav Agrawal, and Tushar Krishna.
\newblock An optimized dataflow for mitigating attention performance
  bottlenecks.
\newblock In {\em Proceedings of the International Conference on Architectural
  Support for Programming Languages and Operation Systems (ASPLOS)}, 2022.

\bibitem{karandikar2018firesim}
S.~{Karandikar}, H.~{Mao}, D.~{Kim}, D.~{Biancolin}, A.~{Amid}, D.~{Lee},
  N.~{Pemberton}, E.~{Amaro}, C.~{Schmidt}, A.~{Chopra}, Q.~{Huang},
  K.~{Kovacs}, B.~{Nikolic}, R.~{Katz}, J.~{Bachrach}, and K.~{Asanovic}.
\newblock Firesim: Fpga-accelerated cycle-exact scale-out system simulation in
  the public cloud.
\newblock In {\em 2018 ACM/IEEE 45th Annual International Symposium on Computer
  Architecture (ISCA)}, pages 29--42, 2018.

\bibitem{kaufman2021learned}
Sam Kaufman, Phitchaya Phothilimthana, Yanqi Zhou, Charith Mendis, Sudip Roy,
  Amit Sabne, and Mike Burrows.
\newblock A learned performance model for tensor processing units.
\newblock In {\em Proceedings of Machine Learning and Systems (MLSys)}, 2021.

\bibitem{pervectorquant}
Ben Keller, Rangharajan Venkatesan, Steve Dai, Stephen~G. Tell, Brian Zimmer,
  William~J. Dally, C.~Thomas~Gray, and Brucek Khailany.
\newblock A 17–95.6 tops/w deep learning inference accelerator with
  per-vector scaled 4-bit quantization for transformers in 5nm.
\newblock In {\em 2022 IEEE Symposium on VLSI Technology and Circuits (VLSI
  Technology and Circuits)}, pages 16--17, 2022.

\bibitem{kim2020length}
Gyuwan Kim and Kyunghyun Cho.
\newblock Length-adaptive transformer: Train once with length drop, use anytime
  with search.
\newblock {\em arXiv preprint arXiv:2010.07003}, 2020.

\bibitem{kim2022squeezeformer}
Sehoon Kim, Amir Gholami, Albert Shaw, Nicholas Lee, Karttikeya Mangalam,
  Jitendra Malik, Michael~W Mahoney, and Kurt Keutzer.
\newblock Squeezeformer: An efficient transformer for automatic speech
  recognition.
\newblock {\em arXiv preprint arXiv:2206.00888}, 2022.

\bibitem{kim2022integer}
Sehoon Kim, Amir Gholami, Zhewei Yao, Nicholas Lee, Patrick Wang, Aniruddha
  Nrusimha, Bohan Zhai, Tianren Gao, Michael~W Mahoney, and Kurt Keutzer.
\newblock Integer-only zero-shot quantization for efficient speech recognition.
\newblock In {\em ICASSP 2022-2022 IEEE International Conference on Acoustics,
  Speech and Signal Processing (ICASSP)}, pages 4288--4292. IEEE, 2022.

\bibitem{ibert}
Sehoon Kim, Amir Gholami, Zhewei Yao, Michael~W Mahoney, and Kurt Keutzer.
\newblock I-bert: Integer-only bert quantization.
\newblock In {\em International conference on machine learning}, pages
  5506--5518. PMLR, 2021.

\bibitem{kim2023big}
Sehoon Kim, Karttikeya Mangalam, Jitendra Malik, Michael~W Mahoney, Amir
  Gholami, and Kurt Keutzer.
\newblock Big little transformer decoder.
\newblock {\em arXiv preprint arXiv:2302.07863}, 2023.

\bibitem{kim2021learned}
Sehoon Kim, Sheng Shen, David Thorsley, Amir Gholami, Woosuk Kwon, Joseph
  Hassoun, and Kurt Keutzer.
\newblock Learned token pruning for transformers.
\newblock {\em arXiv preprint arXiv:2107.00910}, 2021.

\bibitem{kjolstad2017tensor}
Fredrik Kjolstad, Shoaib Kamil, Stephen Chou, David Lugato, and Saman
  Amarasinghe.
\newblock The tensor algebra compiler.
\newblock In {\em Proceedings of the International Conference on Object
  Oriented Programming Systems Languages and Applications}. ACM New York, NY,
  USA, 2017.

\bibitem{knowles2021graphcore}
Simon Knowles.
\newblock Graphcore.
\newblock In {\em IEEE Hot Chips Symposium}, pages 1--25, 2021.

\bibitem{kong2013polyhedral}
Martin Kong, Richard Veras, Kevin Stock, Franz Franchetti, Louis-No{\"e}l
  Pouchet, and Ponnuswamy Sadayappan.
\newblock When polyhedral transformations meet simd code generation.
\newblock In {\em Proceedings of the ACM SIGPLAN Conference on Programming
  Language Design and Implementation (PLDI)}, 2013.

\bibitem{kovaleva2021bert}
Olga Kovaleva, Saurabh Kulshreshtha, Anna Rogers, and Anna Rumshisky.
\newblock Bert busters: Outlier dimensions that disrupt transformers.
\newblock {\em arXiv preprint arXiv:2105.06990}, 2021.

\bibitem{alexnet-2017}
Alex Krizhevsky, Ilya Sutskever, and Geoffrey~E. Hinton.
\newblock Imagenet classification with deep convolutional neural networks.
\newblock {\em Commun. ACM}, 60(6), 2017.

\bibitem{adaptivetiling}
H.~T. Kung, Bradley McDanel, and Sai~Qian Zhang.
\newblock Adaptive tiling: Applying fixed-size systolic arrays to sparse
  convolutional neural networks.
\newblock In {\em 2018 24th International Conference on Pattern Recognition
  (ICPR)}, pages 1006--1011, 2018.

\bibitem{kurtic2022optimal}
Eldar Kurtic, Daniel Campos, Tuan Nguyen, Elias Frantar, Mark Kurtz, Benjamin
  Fineran, Michael Goin, and Dan Alistarh.
\newblock The optimal bert surgeon: Scalable and accurate second-order pruning
  for large language models.
\newblock {\em arXiv preprint arXiv:2203.07259}, 2022.

\bibitem{kuzmin2022fp8}
Andrey Kuzmin, Mart Van~Baalen, Yuwei Ren, Markus Nagel, Jorn Peters, and
  Tijmen Blankevoort.
\newblock Fp8 quantization: The power of the exponent.
\newblock {\em arXiv preprint arXiv:2208.09225}, 2022.

\bibitem{kwon2020maestro}
Hyoukjun Kwon, Prasanth Chatarasi, Vivek Sarkar, Tushar Krishna, Michael
  Pellauer, and Angshuman Parashar.
\newblock Maestro: A data-centric approach to understand reuse, performance,
  and hardware cost of dnn mappings.
\newblock In {\em Proceedings of the International Symposium on
  Microarchitecture (MICRO)}, 2020.

\bibitem{kwon2022fast}
Woosuk Kwon, Sehoon Kim, Michael~W Mahoney, Joseph Hassoun, Kurt Keutzer, and
  Amir Gholami.
\newblock A fast post-training pruning framework for transformers.
\newblock {\em arXiv preprint arXiv:2204.09656}, 2022.

\bibitem{lagunas2021block}
Fran{\c{c}}ois Lagunas, Ella Charlaix, Victor Sanh, and Alexander~M Rush.
\newblock Block pruning for faster transformers.
\newblock {\em arXiv preprint arXiv:2109.04838}, 2021.

\bibitem{lan2019albert}
Zhenzhong Lan, Mingda Chen, Sebastian Goodman, Kevin Gimpel, Piyush Sharma, and
  Radu Soricut.
\newblock Albert: A lite bert for self-supervised learning of language
  representations.
\newblock {\em arXiv preprint arXiv:1909.11942}, 2019.

\bibitem{lattner2004llvm}
Chris Lattner and Vikram Adve.
\newblock Llvm: A compilation framework for lifelong program analysis \&
  transformation.
\newblock In {\em International Symposium on Code Generation and Optimization
  (CGO)}, 2004.

\bibitem{ftrans}
Bingbing Li, Santosh Pandey, Haowen Fang, Yanjun Lyv, Ji~Li, Jieyang Chen, Mimi
  Xie, Lipeng Wan, Hang Liu, and Caiwen Ding.
\newblock Ftrans: Energy-efficient acceleration of transformers using fpga.
\newblock In {\em Proceedings of the ACM/IEEE International Symposium on Low
  Power Electronics and Design}, ISLPED '20, page 175–180, New York, NY, USA,
  2020. Association for Computing Machinery.

\bibitem{li21deepLearningCompilerSurvey}
Mingzhen Li, Yi~Liu, Xiaoyan Liu, Qingxiao Sun, Xin You, Hailong Yang, Zhongzhi
  Luan, Lin Gan, Guangwen Yang, and Depei Qian.
\newblock The deep learning compiler: A comprehensive survey.
\newblock {\em IEEE Transactions on Parallel and Distributed Systems},
  32(3):708--727, 2021.

\bibitem{li2021analytical}
Rui Li, Yufan Xu, Aravind Sukumaran-Rajam, Atanas Rountev, and P~Sadayappan.
\newblock Analytical characterization and design space exploration for
  optimization of cnns.
\newblock In {\em Proceedings of the International Conference on Architectural
  Support for Programming Languages and Operation Systems (ASPLOS)}, 2021.

\bibitem{li2022efficientformer}
Yanyu Li, Geng Yuan, Yang Wen, Eric Hu, Georgios Evangelidis, Sergey Tulyakov,
  Yanzhi Wang, and Jian Ren.
\newblock Efficientformer: Vision transformers at mobilenet speed.
\newblock {\em arXiv preprint arXiv:2206.01191}, 2022.

\bibitem{gradientpruning}
Zheng Li, Soroush Ghodrati, Amir Yazdanbakhsh, Hadi Esmaeilzadeh, and Mingu
  Kang.
\newblock Accelerating attention through gradient-based learned runtime
  pruning.
\newblock In {\em Proceedings of the 49th Annual International Symposium on
  Computer Architecture}, ISCA '22, page 902–915, New York, NY, USA, 2022.
  Association for Computing Machinery.

\bibitem{i-vit}
Zhikai Li and Qingyi Gu.
\newblock I-vit: Integer-only quantization for efficient vision transformer
  inference, 2022.

\bibitem{li2020train}
Zhuohan Li, Eric Wallace, Sheng Shen, Kevin Lin, Kurt Keutzer, Dan Klein, and
  Joey Gonzalez.
\newblock Train big, then compress: Rethinking model size for efficient
  training and inference of transformers.
\newblock In {\em International Conference on machine learning}, pages
  5958--5968. PMLR, 2020.

\bibitem{liao2019davinci}
Heng Liao, Jiajin Tu, Jing Xia, and Xiping Zhou.
\newblock Davinci: A scalable architecture for neural network computing.
\newblock In {\em IEEE Hot Chips Symposium}, pages 1--44, 2019.

\bibitem{liao2021searching}
Yi-Lun Liao, Sertac Karaman, and Vivienne Sze.
\newblock Searching for efficient multi-stage vision transformers.
\newblock {\em arXiv preprint arXiv:2109.00642}, 2021.

\bibitem{ZCM20_quantized_TR}
Z.~Liao, R.~Couillet, and M.~W. Mahoney.
\newblock Sparse quantized spectral clustering.
\newblock Technical Report Preprint: arXiv:2010.01376, 2020.

\bibitem{lie2022cerebras}
Sean Lie.
\newblock Cerebras architecture deep dive: First look inside the hw/sw
  co-design for deep learning: Cerebras systems.
\newblock In {\em IEEE Hot Chips Symposium}, pages 1--34, 2022.

\bibitem{lin2020mcunet}
Ji~Lin, Wei-Ming Chen, Yujun Lin, Chuang Gan, Song Han, et~al.
\newblock Mcunet: Tiny deep learning on iot devices.
\newblock {\em Advances in Neural Information Processing Systems},
  33:11711--11722, 2020.

\bibitem{liu2022loss}
Chaoyue Liu, Libin Zhu, and Mikhail Belkin.
\newblock Loss landscapes and optimization in over-parameterized non-linear
  systems and neural networks.
\newblock {\em Applied and Computational Harmonic Analysis}, 59:85--116, 2022.

\bibitem{liu2018progressive}
Chenxi Liu, Barret Zoph, Maxim Neumann, Jonathon Shlens, Wei Hua, Li-Jia Li,
  Li~Fei-Fei, Alan Yuille, Jonathan Huang, and Kevin Murphy.
\newblock Progressive neural architecture search.
\newblock In {\em Proceedings of the European conference on computer vision
  (ECCV)}, pages 19--34, 2018.

\bibitem{liu2017hierarchical}
Hanxiao Liu, Karen Simonyan, Oriol Vinyals, Chrisantha Fernando, and Koray
  Kavukcuoglu.
\newblock Hierarchical representations for efficient architecture search.
\newblock {\em arXiv preprint arXiv:1711.00436}, 2017.

\bibitem{liu2018darts}
Hanxiao Liu, Karen Simonyan, and Yiming Yang.
\newblock Darts: Differentiable architecture search.
\newblock {\em arXiv preprint arXiv:1806.09055}, 2018.

\bibitem{liu2019roberta}
Yinhan Liu, Myle Ott, Naman Goyal, Jingfei Du, Mandar Joshi, Danqi Chen, Omer
  Levy, Mike Lewis, Luke Zettlemoyer, and Veselin Stoyanov.
\newblock {RoBERTa}: A robustly optimized bert pretraining approach.
\newblock {\em arXiv preprint arXiv:1907.11692}, 2019.

\bibitem{liu2021swin}
Ze~Liu, Yutong Lin, Yue Cao, Han Hu, Yixuan Wei, Zheng Zhang, Stephen Lin, and
  Baining Guo.
\newblock Swin transformer: Hierarchical vision transformer using shifted
  windows.
\newblock In {\em Proceedings of the IEEE/CVF International Conference on
  Computer Vision}, pages 10012--10022, 2021.

\bibitem{liu2021ebert}
Zejian Liu, Fanrong Li, Gang Li, and Jian Cheng.
\newblock Ebert: Efficient bert inference with dynamic structured pruning.
\newblock In {\em Findings of the Association for Computational Linguistics:
  ACL-IJCNLP 2021}, pages 4814--4823, 2021.

\bibitem{lu2021tenet}
Liqiang Lu, Naiqing Guan, Yuyue Wang, Liancheng Jia, Zizhang Luo, Jieming Yin,
  Jason Cong, and Yun Liang.
\newblock Tenet: A framework for modeling tensor dataflow based on
  relation-centric notation.
\newblock In {\em Proceedings of the International Symposium on Computer
  Architecture (ISCA)}, 2021.

\bibitem{sanger}
Liqiang Lu, Yicheng Jin, Hangrui Bi, Zizhang Luo, Peng Li, Tao Wang, and Yun
  Liang.
\newblock Sanger: A co-design framework for enabling sparse attention using
  reconfigurable architecture.
\newblock In {\em MICRO-54: 54th Annual IEEE/ACM International Symposium on
  Microarchitecture}, MICRO '21, page 977–991, New York, NY, USA, 2021.
  Association for Computing Machinery.

\bibitem{lu2020hardware}
Siyuan Lu, Meiqi Wang, Shuang Liang, Jun Lin, and Zhongfeng Wang.
\newblock {Hardware Accelerator for Multi-Head Attention and Position-Wise
  Feed-Forward in the Transformer}.
\newblock {\em arXiv preprint arXiv:2009.08605}, 2020.

\bibitem{luo2018neural}
Renqian Luo, Fei Tian, Tao Qin, Enhong Chen, and Tie-Yan Liu.
\newblock Neural architecture optimization.
\newblock {\em Advances in neural information processing systems}, 31, 2018.

\bibitem{marin2021token}
Dmitrii Marin, Jen-Hao~Rick Chang, Anurag Ranjan, Anish Prabhu, Mohammad
  Rastegari, and Oncel Tuzel.
\newblock Token pooling in vision transformers.
\newblock {\em arXiv preprint arXiv:2110.03860}, 2021.

\bibitem{max_trick}
James McCaffrey.
\newblock The max trick when computing softmax, 2016.

\bibitem{mehta2021mobilevit}
Sachin Mehta and Mohammad Rastegari.
\newblock Mobilevit: light-weight, general-purpose, and mobile-friendly vision
  transformer.
\newblock {\em arXiv preprint arXiv:2110.02178}, 2021.

\bibitem{mei2021zigzag}
Linyan Mei, Pouya Houshmand, Vikram Jain, Sebastian Giraldo, and Marian
  Verhelst.
\newblock Zigzag: Enlarging joint architecture-mapping design space exploration
  for dnn accelerators.
\newblock {\em IEEE Transactions on Computers}, 70(8), 2021.

\bibitem{merity2016pointer}
Stephen Merity, Caiming Xiong, James Bradbury, and Richard Socher.
\newblock Pointer sentinel mixture models, 2016.

\bibitem{michel2019sixteen}
Paul Michel, Omer Levy, and Graham Neubig.
\newblock Are sixteen heads really better than one?
\newblock {\em arXiv preprint arXiv:1905.10650}, 2019.

\bibitem{micikevicius2022fp8}
Paulius Micikevicius, Dusan Stosic, Neil Burgess, Marius Cornea, Pradeep Dubey,
  Richard Grisenthwaite, Sangwon Ha, Alexander Heinecke, Patrick Judd, John
  Kamalu, et~al.
\newblock Fp8 formats for deep learning.
\newblock {\em arXiv preprint arXiv:2209.05433}, 2022.

\bibitem{onlinenormalizer}
Maxim Milakov and Natalia Gimelshein.
\newblock Online normalizer calculation for softmax, 2018.

\bibitem{mullapudi2016automatically}
Ravi~Teja Mullapudi, Andrew Adams, Dillon Sharlet, Jonathan Ragan-Kelley, and
  Kayvon Fatahalian.
\newblock Automatically scheduling halide image processing pipelines.
\newblock {\em ACM Transactions on Graphics (TOG)}, 2016.

\bibitem{cacti}
Naveen Muralimanohar, Rajeev Balasubramonian, and Norman~P Jouppi.
\newblock Cacti 6.0: A tool to model large caches.

\bibitem{taylorseries}
Peter Nilsson, Ateeq Ur~Rahman Shaik, Rakesh Gangarajaiah, and Erik Hertz.
\newblock Hardware implementation of the exponential function using taylor
  series.
\newblock In {\em 2014 NORCHIP}, pages 1--4, 2014.

\bibitem{tensorrt}
{NVIDIA}.
\newblock Tensor{RT}: https://developer.nvidia.com/tensorrt, 2018.

\bibitem{olivry2021ioopt}
Auguste Olivry, Guillaume Iooss, Nicolas Tollenaere, Atanas Rountev,
  P~Sadayappan, and Fabrice Rastello.
\newblock Ioopt: automatic derivation of i/o complexity bounds for affine
  programs.
\newblock In {\em Proceedings of the ACM SIGPLAN Conference on Programming
  Language Design and Implementation (PLDI)}, 2021.

\bibitem{parashar2019timeloop}
Angshuman Parashar, Priyanka Raina, Yakun~Sophia Shao, Yu-Hsin Chen, Victor~A
  Ying, Anurag Mukkara, Rangharajan Venkatesan, Brucek Khailany, Stephen~W
  Keckler, and Joel Emer.
\newblock Timeloop: A systematic approach to dnn accelerator evaluation.
\newblock In {\em 2019 IEEE international symposium on performance analysis of
  systems and software (ISPASS)}, pages 304--315. IEEE, 2019.

\bibitem{scnn}
Angshuman Parashar, Minsoo Rhu, Anurag Mukkara, Antonio Puglielli, Rangharajan
  Venkatesan, Brucek Khailany, Joel Emer, Stephen~W. Keckler, and William~J.
  Dally.
\newblock Scnn: An accelerator for compressed-sparse convolutional neural
  networks.
\newblock In {\em Proceedings of the 44th Annual International Symposium on
  Computer Architecture}, ISCA '17, page 27–40, New York, NY, USA, 2017.
  Association for Computing Machinery.

\bibitem{park2013predictive}
Eunjung Park, John Cavazos, Louis-No{\"e}l Pouchet, C{\'e}dric Bastoul, Albert
  Cohen, and P~Sadayappan.
\newblock Predictive modeling in a polyhedral optimization space.
\newblock {\em International journal of parallel programming}, 2013.

\bibitem{optimus}
Junki Park, Hyunsung Yoon, Daehyun Ahn, Jungwook Choi, and Jae-Joon Kim.
\newblock Optimus: Optimized matrix multiplication structure for transformer
  neural network accelerator.
\newblock In I.~Dhillon, D.~Papailiopoulos, and V.~Sze, editors, {\em
  Proceedings of Machine Learning and Systems}, volume~2, pages 363--378. 2020.

\bibitem{paszke2019pytorch}
Adam Paszke, Sam Gross, Francisco Massa, Adam Lerer, James Bradbury, Gregory
  Chanan, Trevor Killeen, Zeming Lin, Natalia Gimelshein, Luca Antiga, et~al.
\newblock Pytorch: An imperative style, high-performance deep learning library.
\newblock {\em Advances in neural information processing systems}, 32, 2019.

\bibitem{pati2021demystifying}
Suchita Pati, Shaizeen Aga, Nuwan Jayasena, and Matthew~D Sinclair.
\newblock Demystifying bert: Implications for accelerator design.
\newblock In {\em International Symposium on Workload Characterization
  (IISWC)}, 2021.

\bibitem{pei2019tianjic}
Jing Pei, Lei Deng, Sen Song, Mingguo Zhao, Youhui Zhang, Shuang Wu, Guanrui
  Wang, Zhe Zou, Zhenzhi Wu, Wei He, et~al.
\newblock Towards artificial general intelligence with hybrid tianjic chip
  architecture.
\newblock {\em Nature}, 572(7767):106--111, 2019.

\bibitem{pham2018efficient}
Hieu Pham, Melody Guan, Barret Zoph, Quoc Le, and Jeff Dean.
\newblock Efficient neural architecture search via parameters sharing.
\newblock In {\em International conference on machine learning}, pages
  4095--4104. PMLR, 2018.

\bibitem{prabhakar2021sambanova}
Raghu Prabhakar and Sumti Jairath.
\newblock Sambanova sn10 rdu: Accelerating software 2.0 with dataflow.
\newblock In {\em IEEE Hot Chips Symposium}, pages 1--37, 2021.

\bibitem{dota}
Zheng Qu, Liu Liu, Fengbin Tu, Zhaodong Chen, Yufei Ding, and Yuan Xie.
\newblock Dota: Detect and omit weak attentions for scalable transformer
  acceleration.
\newblock In {\em Proceedings of the 27th ACM International Conference on
  Architectural Support for Programming Languages and Operating Systems},
  ASPLOS '22, page 14–26, New York, NY, USA, 2022. Association for Computing
  Machinery.

\bibitem{radford2018improving}
Alec Radford, Karthik Narasimhan, Tim Salimans, and Ilya Sutskever.
\newblock Improving language understanding by generative pre-training, 2018.

\bibitem{radford2019language}
Alec Radford, Jeffrey Wu, Rewon Child, David Luan, Dario Amodei, and Ilya
  Sutskever.
\newblock Language models are unsupervised multitask learners.
\newblock {\em OpenAI blog}, 1(8):9, 2019.

\bibitem{rae2021scaling}
Jack~W Rae, Sebastian Borgeaud, Trevor Cai, Katie Millican, Jordan Hoffmann,
  Francis Song, John Aslanides, Sarah Henderson, Roman Ring, Susannah Young,
  et~al.
\newblock Scaling language models: Methods, analysis \& insights from training
  gopher.
\newblock {\em arXiv preprint arXiv:2112.11446}, 2021.

\bibitem{rae2019compressive}
Jack~W Rae, Anna Potapenko, Siddhant~M Jayakumar, and Timothy~P Lillicrap.
\newblock Compressive transformers for long-range sequence modelling.
\newblock {\em arXiv preprint arXiv:1911.05507}, 2019.

\bibitem{raffel2019exploring}
Colin Raffel, Noam Shazeer, Adam Roberts, Katherine Lee, Sharan Narang, Michael
  Matena, Yanqi Zhou, Wei Li, and Peter~J Liu.
\newblock Exploring the limits of transfer learning with a unified text-to-text
  transformer.
\newblock {\em arXiv preprint arXiv:1910.10683}, 2019.

\bibitem{raffel2020exploring}
Colin Raffel, Noam Shazeer, Adam Roberts, Katherine Lee, Sharan Narang, Michael
  Matena, Yanqi Zhou, Wei Li, Peter~J Liu, et~al.
\newblock Exploring the limits of transfer learning with a unified text-to-text
  transformer.
\newblock {\em J. Mach. Learn. Res.}, 21(140):1--67, 2020.

\bibitem{ragan2013halide}
Jonathan Ragan-Kelley, Connelly Barnes, Andrew Adams, Sylvain Paris, Fredo
  Durand, and Saman Amarasinghe.
\newblock Halide: a language and compiler for optimizing parallelism, locality,
  and recomputation in image processing pipelines.
\newblock {\em Acm Sigplan Notices}, 2013.

\bibitem{halide2013-pldi}
Jonathan Ragan-Kelley, Connelly Barnes, Andrew Adams, Sylvain Paris, Fr{\'e}do
  Durand, and Saman Amarasinghe.
\newblock Halide: A language and compiler for optimizing parallelism, locality,
  and recomputation in image processing pipelines.
\newblock In {\em Proceedings of the ACM SIGPLAN Conference on Programming
  Language Design and Implementation (PLDI)}, 2013.

\bibitem{rajpurkar2016squad}
Pranav Rajpurkar, Jian Zhang, Konstantin Lopyrev, and Percy Liang.
\newblock {SQuAD}: 100,000+ questions for machine comprehension of text.
\newblock {\em arXiv preprint arXiv:1606.05250}, 2016.

\bibitem{real2019regularized}
Esteban Real, Alok Aggarwal, Yanping Huang, and Quoc~V Le.
\newblock Regularized evolution for image classifier architecture search.
\newblock In {\em Proceedings of the aaai conference on artificial
  intelligence}, volume~33, pages 4780--4789, 2019.

\bibitem{ren2021comprehensive}
Pengzhen Ren, Yun Xiao, Xiaojun Chang, Po-Yao Huang, Zhihui Li, Xiaojiang Chen,
  and Xin Wang.
\newblock A comprehensive survey of neural architecture search: Challenges and
  solutions.
\newblock {\em ACM Computing Surveys (CSUR)}, 54(4):1--34, 2021.

\bibitem{rotem2018glow}
Nadav Rotem, Jordan Fix, Saleem Abdulrasool, Garret Catron, Summer Deng, Roman
  Dzhabarov, Nick Gibson, James Hegeman, Meghan Lele, Roman Levenstein, et~al.
\newblock Glow: Graph lowering compiler techniques for neural networks.
\newblock {\em arXiv preprint arXiv:1805.00907}, 2018.

\bibitem{sabne2020xla}
Amit Sabne.
\newblock Xla: Compiling machine learning for peak performance.
\newblock 2020.

\bibitem{sajjad2020poor}
Hassan Sajjad, Fahim Dalvi, Nadir Durrani, and Preslav Nakov.
\newblock Poor man’s bert: Smaller and faster transformer models.
\newblock {\em arXiv preprint arXiv:2004.03844}, 2020.

\bibitem{samajdar2020systematic}
Ananda Samajdar, Jan~Moritz Joseph, Yuhao Zhu, Paul Whatmough, Matthew Mattina,
  and Tushar Krishna.
\newblock A systematic methodology for characterizing scalability of dnn
  accelerators using scale-sim.
\newblock In {\em Proceedings of the International Symposium on Performance
  Analysis of Systems and Software (ISPASS)}, 2020.

\bibitem{sandler2018mobilenetv2}
Mark Sandler, Andrew Howard, Menglong Zhu, Andrey Zhmoginov, and Liang-Chieh
  Chen.
\newblock {MobilenetV2}: Inverted residuals and linear bottlenecks.
\newblock In {\em Proceedings of the IEEE Conference on Computer Vision and
  Pattern Recognition}, pages 4510--4520, 2018.

\bibitem{sanh2020movement}
Victor Sanh, Thomas Wolf, and Alexander Rush.
\newblock Movement pruning: Adaptive sparsity by fine-tuning.
\newblock {\em Advances in Neural Information Processing Systems},
  33:20378--20389, 2020.

\bibitem{scao2022bloom}
Teven~Le Scao, Angela Fan, Christopher Akiki, Ellie Pavlick, Suzana Ili{\'c},
  Daniel Hesslow, Roman Castagn{\'e}, Alexandra~Sasha Luccioni, Fran{\c{c}}ois
  Yvon, Matthias Gall{\'e}, et~al.
\newblock Bloom: A 176b-parameter open-access multilingual language model.
\newblock {\em arXiv preprint arXiv:2211.05100}, 2022.

\bibitem{schuster2022confident}
Tal Schuster, Adam Fisch, Jai Gupta, Mostafa Dehghani, Dara Bahri, Vinh~Q Tran,
  Yi~Tay, and Donald Metzler.
\newblock Confident adaptive language modeling.
\newblock {\em arXiv preprint arXiv:2207.07061}, 2022.

\bibitem{schuster2021consistent}
Tal Schuster, Adam Fisch, Tommi Jaakkola, and Regina Barzilay.
\newblock Consistent accelerated inference via confident adaptive transformers.
\newblock {\em arXiv preprint arXiv:2104.08803}, 2021.

\bibitem{sekanina2021neural}
Lukas Sekanina.
\newblock Neural architecture search and hardware accelerator co-search: A
  survey.
\newblock {\em IEEE Access}, 9:151337--151362, 2021.

\bibitem{salo}
Guan Shen, Jieru Zhao, Quan Chen, Jingwen Leng, Chao Li, and Minyi Guo.
\newblock Salo: An efficient spatial accelerator enabling hybrid sparse
  attention mechanisms for long sequences.
\newblock In {\em Proceedings of the 59th ACM/IEEE Design Automation
  Conference}, DAC '22, page 571–576, New York, NY, USA, 2022. Association
  for Computing Machinery.

\bibitem{shen2020q}
Sheng Shen, Zhen Dong, Jiayu Ye, Linjian Ma, Zhewei Yao, Amir Gholami,
  Michael~W Mahoney, and Kurt Keutzer.
\newblock {Q-BERT}: Hessian based ultra low precision quantization of bert.
\newblock In {\em AAAI}, pages 8815--8821, 2020.

\bibitem{nvdla-hotchips}
Frans Sijstermans.
\newblock {The NVIDIA Deep Learning Accelerator}.
\newblock In {\em Hot Chips}, 2018.

\bibitem{vgg}
Karen Simonyan and Andrew Zisserman.
\newblock {Very Deep Convolutional Networks for Large-scale Image Recognition}.
\newblock {\em CoRR}, abs/1408.1556, 2014.

\bibitem{smith2022using}
Shaden Smith, Mostofa Patwary, Brandon Norick, Patrick LeGresley, Samyam
  Rajbhandari, Jared Casper, Zhun Liu, Shrimai Prabhumoye, George Zerveas,
  Vijay Korthikanti, et~al.
\newblock Using deepspeed and megatron to train megatron-turing nlg 530b, a
  large-scale generative language model.
\newblock {\em arXiv preprint arXiv:2201.11990}, 2022.

\bibitem{snoek2012practical}
Jasper Snoek, Hugo Larochelle, and Ryan~P Adams.
\newblock Practical bayesian optimization of machine learning algorithms.
\newblock {\em Advances in neural information processing systems}, 25, 2012.

\bibitem{so2019evolved}
David So, Quoc Le, and Chen Liang.
\newblock The evolved transformer.
\newblock In {\em International Conference on Machine Learning}, pages
  5877--5886. PMLR, 2019.

\bibitem{so2021searching}
David So, Wojciech Ma{\'n}ke, Hanxiao Liu, Zihang Dai, Noam Shazeer, and Quoc~V
  Le.
\newblock Searching for efficient transformers for language modeling.
\newblock {\em Advances in Neural Information Processing Systems},
  34:6010--6022, 2021.

\bibitem{socher2013recursive}
Richard Socher, Alex Perelygin, Jean Wu, Jason Chuang, Christopher~D Manning,
  Andrew~Y Ng, and Christopher Potts.
\newblock Recursive deep models for semantic compositionality over a sentiment
  treebank.
\newblock In {\em Proceedings of the 2013 conference on empirical methods in
  natural language processing}, pages 1631--1642, 2013.

\bibitem{softermax}
Jacob~R. Stevens, Rangharajan Venkatesan, Steve Dai, Brucek Khailany, and Anand
  Raghunathan.
\newblock Softermax: Hardware/software co-design of an efficient softmax for
  transformers.
\newblock In {\em 2021 58th ACM/IEEE Design Automation Conference (DAC)}, pages
  469--474, 2021.

\bibitem{speformer}
Yang Sun, Wei Hu, Fang Liu, Min Jiang, FeiHu Huang, and Dian Xu.
\newblock Speformer: An efficient hardware-software cooperative solution for
  sparse spectral transformer.
\newblock In {\em 2022 IEEE 9th International Conference on Cyber Security and
  Cloud Computing (CSCloud)/2022 IEEE 8th International Conference on Edge
  Computing and Scalable Cloud (EdgeCom)}, pages 180--185, 2022.

\bibitem{sun2020mobilebert}
Zhiqing Sun, Hongkun Yu, Xiaodan Song, Renjie Liu, Yiming Yang, and Denny Zhou.
\newblock Mobilebert: a compact task-agnostic bert for resource-limited
  devices.
\newblock {\em arXiv preprint arXiv:2004.02984}, 2020.

\bibitem{efficientprocessing}
Vivienne Sze, Yu-Hsin Chen, Tien-Ju Yang, and Joel Emer.
\newblock Efficient processing of deep neural networks: A tutorial and survey,
  2017.

\bibitem{inception-v1}
Christian Szegedy, Wei Liu, Yangqing Jia, Pierre Sermanet, Scott Reed, Dragomir
  Anguelov, Dumitru Erhan, Vincent Vanhoucke, and Andrew Rabinovich.
\newblock Going deeper with convolutions.
\newblock In {\em 2015 IEEE Conference on Computer Vision and Pattern
  Recognition (CVPR)}, pages 1--9, 2015.

\bibitem{talpes2020compute}
Emil Talpes, Debjit~Das Sarma, Ganesh Venkataramanan, Peter Bannon, Bill McGee,
  Benjamin Floering, Ankit Jalote, Christopher Hsiong, Sahil Arora, Atchyuth
  Gorti, et~al.
\newblock Compute solution for tesla's full self-driving computer.
\newblock {\em IEEE Micro}, 40(2):25--35, 2020.

\bibitem{edgebert}
Thierry Tambe, Coleman Hooper, Lillian Pentecost, Tianyu Jia, En-Yu Yang, Marco
  Donato, Victor Sanh, Paul Whatmough, Alexander~M. Rush, David Brooks, and
  Gu-Yeon Wei.
\newblock {EdgeBERT: Sentence-Level Energy Optimizations for Latency-Aware
  Multi-Task NLP Inference}.
\newblock page 830–844, 2021.

\bibitem{tan2019mnasnet}
Mingxing Tan, Bo~Chen, Ruoming Pang, Vijay Vasudevan, Mark Sandler, Andrew
  Howard, and Quoc~V Le.
\newblock Mnasnet: Platform-aware neural architecture search for mobile.
\newblock In {\em Proceedings of the IEEE/CVF Conference on Computer Vision and
  Pattern Recognition}, pages 2820--2828, 2019.

\bibitem{efficientnet}
Mingxing Tan and Quoc Le.
\newblock {E}fficient{N}et: Rethinking model scaling for convolutional neural
  networks.
\newblock In Kamalika Chaudhuri and Ruslan Salakhutdinov, editors, {\em
  Proceedings of the 36th International Conference on Machine Learning},
  volume~97 of {\em Proceedings of Machine Learning Research}, pages
  6105--6114. PMLR, 09--15 Jun 2019.

\bibitem{thomas2004libm}
James~W Thomas, John~P Okada, Peter Markstein, and Ren-Chang Li.
\newblock The libm library and floatingpoint arithmetic in hp-ux for
  itanium-based systems.
\newblock Technical report, Technical report, Hewlett-Packard Company, Palo
  Alto, CA, USA, 2004.

\bibitem{thoppilan2022lamda}
Romal Thoppilan, Daniel De~Freitas, Jamie Hall, Noam Shazeer, Apoorv
  Kulshreshtha, Heng-Tze Cheng, Alicia Jin, Taylor Bos, Leslie Baker, Yu~Du,
  et~al.
\newblock Lamda: Language models for dialog applications.
\newblock {\em arXiv preprint arXiv:2201.08239}, 2022.

\bibitem{tillet2019triton}
Philippe Tillet, HT~Kung, and David Cox.
\newblock Triton: an intermediate language and compiler for tiled neural
  network computations.
\newblock In {\em Proceedings of the 3rd ACM SIGPLAN International Workshop on
  Machine Learning and Programming Languages}, 2019.

\bibitem{touvron2021training}
Hugo Touvron, Matthieu Cord, Matthijs Douze, Francisco Massa, Alexandre
  Sablayrolles, and Herv{\'e} J{\'e}gou.
\newblock Training data-efficient image transformers \& distillation through
  attention.
\newblock In {\em International Conference on Machine Learning}, pages
  10347--10357. PMLR, 2021.

\bibitem{vasilache2018tensor}
Nicolas Vasilache, Oleksandr Zinenko, Theodoros Theodoridis, Priya Goyal,
  Zachary DeVito, William~S. Moses, Sven Verdoolaege, Andrew Adams, and Albert
  Cohen.
\newblock Tensor comprehensions: Framework-agnostic high-performance machine
  learning abstractions, 2018.

\bibitem{vaswani2017attention}
Ashish Vaswani, Noam Shazeer, Niki Parmar, Jakob Uszkoreit, Llion Jones,
  Aidan~N Gomez, {\L}ukasz Kaiser, and Illia Polosukhin.
\newblock Attention is all you need.
\newblock In {\em Advances in neural information processing systems}, pages
  5998--6008, 2017.

\bibitem{venkatesan2019magnet}
Rangharajan Venkatesan, Yakun~Sophia Shao, Miaorong Wang, Jason Clemons, Steve
  Dai, Matthew Fojtik, Ben Keller, Alicia Klinefelter, Nathaniel Pinckney,
  Priyanka Raina, et~al.
\newblock Magnet: A modular accelerator generator for neural networks.
\newblock In {\em Proceedings of the International Conference on Computer-Aided
  Design (ICCAD)}, 2019.

\bibitem{wan2020fbnetv2}
Alvin Wan, Xiaoliang Dai, Peizhao Zhang, Zijian He, Yuandong Tian, Saining Xie,
  Bichen Wu, Matthew Yu, Tao Xu, Kan Chen, et~al.
\newblock Fbnetv2: Differentiable neural architecture search for spatial and
  channel dimensions.
\newblock In {\em Proceedings of the IEEE/CVF Conference on Computer Vision and
  Pattern Recognition}, pages 12965--12974, 2020.

\bibitem{wang2018glue}
Alex Wang, Amanpreet Singh, Julian Michael, Felix Hill, Omer Levy, and Samuel~R
  Bowman.
\newblock {GLUE}: A multi-task benchmark and analysis platform for natural
  language understanding.
\newblock {\em arXiv preprint arXiv:1804.07461}, 2018.

\bibitem{wang2021alphanet}
Dilin Wang, Chengyue Gong, Meng Li, Qiang Liu, and Vikas Chandra.
\newblock Alphanet: improved training of supernets with alpha-divergence.
\newblock In {\em International Conference on Machine Learning}, pages
  10760--10771. PMLR, 2021.

\bibitem{wang2020hat}
Hanrui Wang, Zhanghao Wu, Zhijian Liu, Han Cai, Ligeng Zhu, Chuang Gan, and
  Song Han.
\newblock Hat: Hardware-aware transformers for efficient natural language
  processing.
\newblock {\em arXiv preprint arXiv:2005.14187}, 2020.

\bibitem{wang2021spatten}
Hanrui Wang, Zhekai Zhang, and Song Han.
\newblock Spatten: Efficient sparse attention architecture with cascade token
  and head pruning.
\newblock In {\em 2021 IEEE International Symposium on High-Performance
  Computer Architecture (HPCA)}, pages 97--110. IEEE, 2021.

\bibitem{wang2018haq}
Kuan Wang, Zhijian Liu, Yujun Lin, Ji~Lin, and Song Han.
\newblock {HAQ}: Hardware-aware automated quantization.
\newblock {\em In Proceedings of the IEEE conference on computer vision and
  pattern recognition}, 2019.

\bibitem{highspeedlowcomplexity}
Meiqi Wang, Siyuan Lu, Danyang Zhu, Jun Lin, and Zhongfeng Wang.
\newblock A high-speed and low-complexity architecture for softmax function in
  deep learning.
\newblock In {\em 2018 IEEE Asia Pacific Conference on Circuits and Systems
  (APCCAS)}, pages 223--226, 2018.

\bibitem{williams2017broad}
Adina Williams, Nikita Nangia, and Samuel~R Bowman.
\newblock A broad-coverage challenge corpus for sentence understanding through
  inference.
\newblock {\em arXiv preprint arXiv:1704.05426}, 2017.

\bibitem{williams2009roofline}
Samuel Williams, Andrew Waterman, and David Patterson.
\newblock Roofline: an insightful visual performance model for multicore
  architectures.
\newblock {\em Communications of the ACM}, 52(4):65--76, 2009.

\bibitem{wu2019fbnet}
Bichen Wu, Xiaoliang Dai, Peizhao Zhang, Yanghan Wang, Fei Sun, Yiming Wu,
  Yuandong Tian, Peter Vajda, Yangqing Jia, and Kurt Keutzer.
\newblock Fbnet: Hardware-aware efficient convnet design via differentiable
  neural architecture search.
\newblock In {\em Proceedings of the IEEE/CVF Conference on Computer Vision and
  Pattern Recognition}, pages 10734--10742, 2019.

\bibitem{wu2018mixed}
Bichen Wu, Yanghan Wang, Peizhao Zhang, Yuandong Tian, Peter Vajda, and Kurt
  Keutzer.
\newblock Mixed precision quantization of convnets via differentiable neural
  architecture search.
\newblock {\em arXiv preprint arXiv:1812.00090}, 2018.

\bibitem{wu2021autoformer}
Haixu Wu, Jiehui Xu, Jianmin Wang, and Mingsheng Long.
\newblock Autoformer: Decomposition transformers with auto-correlation for
  long-term series forecasting.
\newblock {\em Advances in Neural Information Processing Systems},
  34:22419--22430, 2021.

\bibitem{wu2019accelergy}
Yannan~Nellie Wu, Joel~S Emer, and Vivienne Sze.
\newblock Accelergy: An architecture-level energy estimation methodology for
  accelerator designs.
\newblock In {\em 2019 IEEE/ACM International Conference on Computer-Aided
  Design (ICCAD)}, pages 1--8. IEEE, 2019.

\bibitem{wu2020lite}
Zhanghao Wu, Zhijian Liu, Ji~Lin, Yujun Lin, and Song Han.
\newblock Lite transformer with long-short range attention.
\newblock {\em arXiv preprint arXiv:2004.11886}, 2020.

\bibitem{xia2022structured}
Mengzhou Xia, Zexuan Zhong, and Danqi Chen.
\newblock Structured pruning learns compact and accurate models.
\newblock {\em arXiv preprint arXiv:2204.00408}, 2022.

\bibitem{xin2020deebert}
Ji~Xin, Raphael Tang, Jaejun Lee, Yaoliang Yu, and Jimmy Lin.
\newblock Deebert: Dynamic early exiting for accelerating bert inference.
\newblock {\em arXiv preprint arXiv:2004.12993}, 2020.

\bibitem{xu2021bert}
Jin Xu, Xu~Tan, Renqian Luo, Kaitao Song, Jian Li, Tao Qin, and Tie-Yan Liu.
\newblock Nas-bert: task-agnostic and adaptive-size bert compression with
  neural architecture search.
\newblock In {\em Proceedings of the 27th ACM SIGKDD Conference on Knowledge
  Discovery \& Data Mining}, pages 1933--1943, 2021.

\bibitem{yang2022searching}
Longxing Yang, Yu~Hu, Shun Lu, Zihao Sun, Jilin Mei, Yinhe Han, and Xiaowei Li.
\newblock Searching for burgerformer with micro-meso-macro space design.
\newblock In {\em International Conference on Machine Learning}, pages
  25055--25069. PMLR, 2022.

\bibitem{dta-trans}
Tao Yang, Hui Ma, Xiaoling Li, Fangxin Liu, Yilong Zhao, Zhezhi He, and
  Li~Jiang.
\newblock Dtatrans: Leveraging dynamic token-based quantization with accuracy
  compensation mechanism for efficient transformer architecture.
\newblock {\em IEEE Transactions on Computer-Aided Design of Integrated
  Circuits and Systems}, pages 1--1, 2022.

\bibitem{yang2018netadapt}
Tien-Ju Yang, Andrew Howard, Bo~Chen, Xiao Zhang, Alec Go, Mark Sandler,
  Vivienne Sze, and Hartwig Adam.
\newblock Netadapt: Platform-aware neural network adaptation for mobile
  applications.
\newblock In {\em Proceedings of the European Conference on Computer Vision
  (ECCV)}, pages 285--300, 2018.

\bibitem{yang2020interstellar}
Xuan Yang, Mingyu Gao, Qiaoyi Liu, Jeff Setter, Jing Pu, Ankita Nayak, Steven
  Bell, Kaidi Cao, Heonjae Ha, Priyanka Raina, et~al.
\newblock Interstellar: Using halide's scheduling language to analyze dnn
  accelerators.
\newblock In {\em Proceedings of the Twenty-Fifth International Conference on
  Architectural Support for Programming Languages and Operating Systems}, pages
  369--383, 2020.

\bibitem{yang2019xlnet}
Zhilin Yang, Zihang Dai, Yiming Yang, Jaime Carbonell, Russ~R Salakhutdinov,
  and Quoc~V Le.
\newblock {XLNet}: Generalized autoregressive pretraining for language
  understanding.
\newblock In {\em Advances in neural information processing systems}, pages
  5753--5763, 2019.

\bibitem{yao2020hawqv3}
Zhewei Yao, Zhen Dong, Zhangcheng Zheng, Amir Gholami, Jiali Yu, Eric Tan,
  Leyuan Wang, Qijing Huang, Yida Wang, Michael~W Mahoney, and Kurt Keutzer.
\newblock {HAWQV3}: Dyadic neural network quantization.
\newblock {\em arXiv preprint arXiv:2011.10680}, 2020.

\bibitem{yao2019pyhessian}
Zhewei Yao, Amir Gholami, Kurt Keutzer, and Michael~W. Mahoney.
\newblock {PyHessian}: Neural networks through the lens of the {H}essian.
\newblock {\em arXiv preprint arXiv:1912.07145}, 2019.

\bibitem{yu2020bignas}
Jiahui Yu, Pengchong Jin, Hanxiao Liu, Gabriel Bender, Pieter-Jan Kindermans,
  Mingxing Tan, Thomas Huang, Xiaodan Song, Ruoming Pang, and Quoc Le.
\newblock Bignas: Scaling up neural architecture search with big single-stage
  models.
\newblock In {\em European Conference on Computer Vision}, pages 702--717.
  Springer, 2020.

\bibitem{nnlut}
Joonsang Yu, Junki Park, Seongmin Park, Minsoo Kim, Sihwa Lee, Dong~Hyun Lee,
  and Jungwook Choi.
\newblock Nn-lut: Neural approximation of non-linear operations for efficient
  transformer inference.
\newblock In {\em Proceedings of the 59th ACM/IEEE Design Automation
  Conference}, DAC '22, page 577–582, New York, NY, USA, 2022. Association
  for Computing Machinery.

\bibitem{yu2022hessian}
Shixing Yu, Zhewei Yao, Amir Gholami, Zhen Dong, Sehoon Kim, Michael~W Mahoney,
  and Kurt Keutzer.
\newblock Hessian-aware pruning and optimal neural implant.
\newblock In {\em Proceedings of the IEEE/CVF Winter Conference on Applications
  of Computer Vision}, pages 3880--3891, 2022.

\bibitem{gobo}
Ali~Hadi Zadeh and A.~Moshovos.
\newblock Gobo: Quantizing attention-based nlp models for low latency and
  energy efficient inference.
\newblock In {\em 53rd IEEE/ACM International Symposium on Microarchitecture
  (MICRO)}, 2020.

\bibitem{zhang2022full}
Dan Zhang, Safeen Huda, Ebrahim Songhori, Kartik Prabhu, Quoc Le, Anna Goldie,
  and Azalia Mirhoseini.
\newblock A full-stack search technique for domain optimized deep learning
  accelerators.
\newblock In {\em Proceedings of the 27th ACM International Conference on
  Architectural Support for Programming Languages and Operating Systems}, pages
  27--42, 2022.

\bibitem{cambriconx}
Shijin Zhang, Zidong Du, Lei Zhang, Huiying Lan, Shaoli Liu, Ling Li, Qi~Guo,
  Tianshi Chen, and Yunji Chen.
\newblock Cambricon-x: An accelerator for sparse neural networks.
\newblock In {\em 2016 49th Annual IEEE/ACM International Symposium on
  Microarchitecture (MICRO)}, pages 1--12, 2016.

\bibitem{ZJS+20Ansor}
Lianmin Zheng, Chengfan Jia, Minmin Sun, Zhao Wu, Cody~Hao Yu, Ameer Haj-Ali,
  Yida Wang, Jun Yang, Danyang Zhuo, Koushik Sen, Joseph~E. Gonzalez, and Ion
  Stoica.
\newblock Ansor: {Generating} {High}-{Performance} {Tensor} {Programs} for
  {Deep} {Learning}.
\newblock Technical report, arXiv.
\newblock arXiv:2006.06762 [cs, stat] type: article.

\bibitem{zhong2018practical}
Zhao Zhong, Junjie Yan, Wei Wu, Jing Shao, and Cheng-Lin Liu.
\newblock Practical block-wise neural network architecture generation.
\newblock In {\em Proceedings of the IEEE conference on computer vision and
  pattern recognition}, pages 2423--2432, 2018.

\bibitem{cambricons}
Xuda Zhou, Zidong Du, Qi~Guo, Shaoli Liu, Chengsi Liu, Chao Wang, Xuehai Zhou,
  Ling Li, Tianshi Chen, and Yunji Chen.
\newblock Cambricon-s: Addressing irregularity in sparse neural networks
  through a cooperative software/hardware approach.
\newblock In {\em 2018 51st Annual IEEE/ACM International Symposium on
  Microarchitecture (MICRO)}, pages 15--28, 2018.

\bibitem{zhou2020transferable}
Yanqi Zhou, Sudip Roy, Amirali Abdolrashidi, Daniel Wong, Peter Ma, Qiumin Xu,
  Hanxiao Liu, Phitchaya Phothilimtha, Shen Wang, Anna Goldie, et~al.
\newblock Transferable graph optimizers for ml compilers.
\newblock In {\em Proceedings of the Conference on Neural Information
  Processing Systems (NeurIPS)}, 2020.

\bibitem{energon}
Zhe Zhou, Junlin Liu, Zhenyu Gu, and Guangyu Sun.
\newblock Energon: Towards efficient acceleration of transformers using dynamic
  sparse attention.
\newblock {\em IEEE Transactions on Computer-Aided Design of Integrated
  Circuits and Systems}, pages 1--1, 2022.

\bibitem{zoph2016neural}
Barret Zoph and Quoc~V Le.
\newblock Neural architecture search with reinforcement learning.
\newblock {\em arXiv preprint arXiv:1611.01578}, 2016.

\bibitem{zoph2018learning}
Barret Zoph, Vijay Vasudevan, Jonathon Shlens, and Quoc~V Le.
\newblock Learning transferable architectures for scalable image recognition.
\newblock In {\em Proceedings of the IEEE conference on computer vision and
  pattern recognition}, pages 8697--8710, 2018.

\end{thebibliography}
\clearpage
\onecolumn
\appendix
\section{Appendix}

\subsection{Decoder Model Architecture}
\label{appendix:decoder}
\begin{figure}[!h]
  \centering
  \includegraphics[width=0.6\columnwidth]{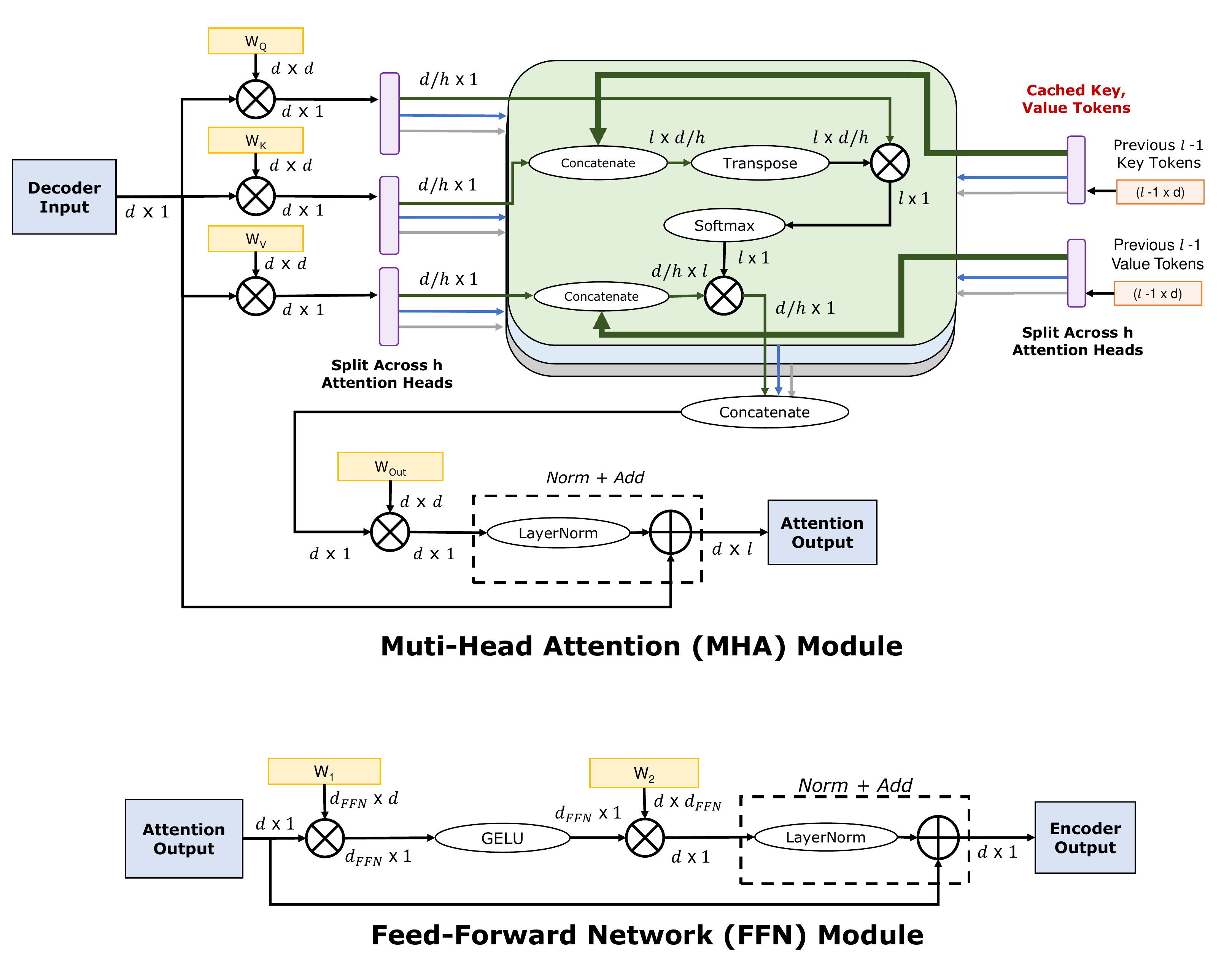}
  \caption{Map of the computations performed in the Transformer decoder. The decoder is primarily composed of matrix-vector operations. The diagram displays the computation for one decoder block and for the $l$th iteration.}
      \label{fig:decoder-comp-map}
\end{figure}

Fig.~\ref{fig:decoder-comp-map} illustrates the computations performed in the MHA and FFN modules of the Transformer decoder.
Compared to the Transformer encoder, the main differences are (1) that the majority of the matmuls are matrix-vector operations and (2) that the keys and values from the previous token generation iterations are cached.

\subsection{High-Level Overview of CNN Architecture}
\label{appendix:cnn}

\begin{figure}[!h]
  \centering
  \includegraphics[width=0.6\columnwidth]{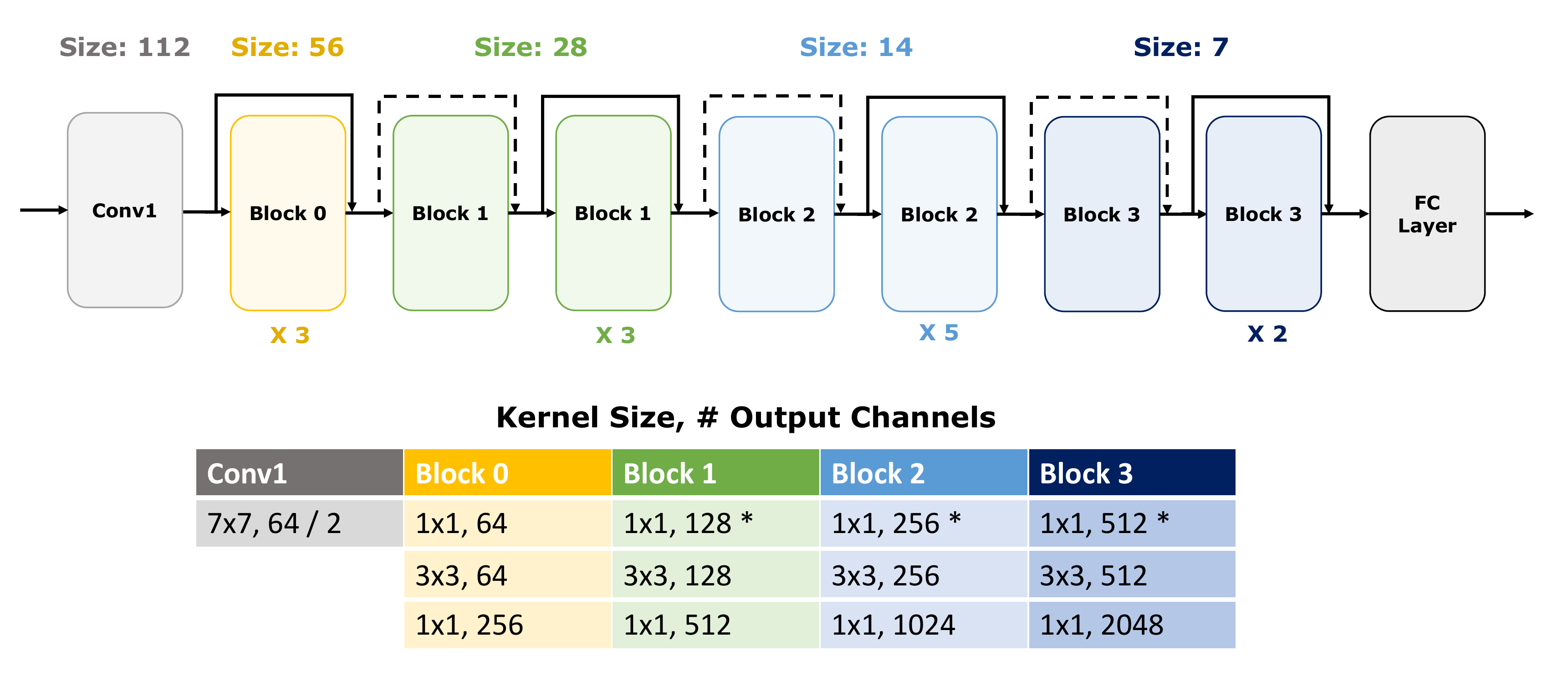}
  \caption{Diagram of the ResNet-50 model architecture. The operations with a star beside them have a stride of 2 for the first block of that type. The arrows correspond to residual additions. The dotted arrows correspond to additional 1$\times$1 convolutional layers that project the previous input to match the dimension of the output of the block. ReLU, BatchNorm, and Softmax layers are omitted for simplicity.}
      \label{fig:resnet50-architecture}
\end{figure}

Convolutional Neural Networks (CNNs) are a class of neural networks which were popularized by the release of AlexNet in 2012 and have seen widespread use across computer vision applications \cite{alexnet-2017,mobilenet,efficientnet,vgg,inception-v1}.
These networks leverage convolutions, which are operations that apply a set of weights (also referred to as a filter or kernel) to groups of elements in the input. The CNN model used for baseline comparisons in this paper is ResNet-50, which is a popular architecture for vision applications \cite{resnet50}. The ResNet-50 model architecture is outlined in Fig.~\ref{fig:resnet50-architecture}.

\begin{figure}[!h]
  \centering
  \includegraphics[width=\columnwidth]{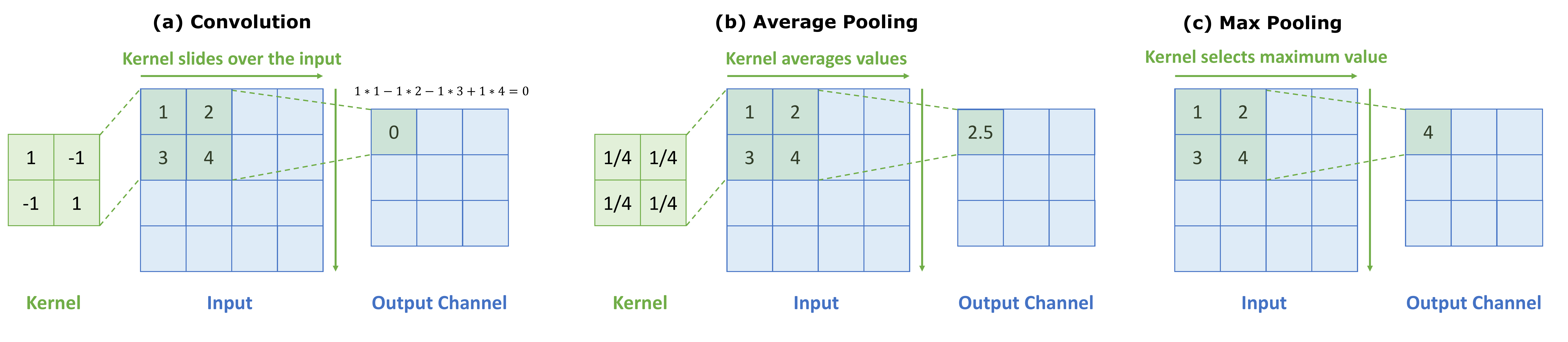}
  \caption{Diagrams outlining the Convolution, Average Pooling, and Max Pooling operations. 
  }
      \label{fig:convolution}
\end{figure}

The basic convolution operation is outlined in Fig.~\ref{fig:convolution}, assuming two-dimensional inputs.
In 2D, the convolution can be viewed as applying a sliding kernel across the input matrix in order to produce the output matrix.
When multiple kernels are applied to the image, each kernel produces a separate output channel.
The spacing between successive filter applications is termed the stride; for example, a stride of 2 means that the kernel is only applied to every second set of input pixels.

CNNs also contain several other operations, such as ReLU, Batch Normalization, and Pooling.
ReLU is a nonlinear activation function that can be expressed as $ ReLU(x) = max(0,x) $. Batch Normalization (or BatchNorm) is also used in CNNs instead of LayerNorm.
BatchNorm is outlined graphically in Fig.~\ref{fig:nonlinear-ops}.
As opposed to LayerNorm, BatchNorm normalizes the data per channel in the input tensor and uses statistics computed at training time.
This means that the BatchNorm operation can be fused with a prior convolution without impacting the requisite tiling dimensions; in fact, BatchNorm layers can be folded with convolutions to also eliminate the added FLOPs from these layers.

CNNs also contain pooling layers for downsampling. Pooling layers are similar to CNNs in that they apply a filter element-wise to the input.
However, these pooling filters use fixed patterns, such as using a filter made up of equal values in the case of Average Pooling, or selecting the maximum element in the group in the case of Max Pooling. These pooling operations are outlined graphically in Fig.~\ref{fig:convolution}. Note that some networks use strided convolutions for downsampling instead of incorporating pooling layers \cite{resnet50}. Finally, CNNs also often use one or more fully connected layers followed by the Softmax function for the output classifier \cite{alexnet-2017,mobilenet,efficientnet,vgg,inception-v1,resnet50}.

\subsection{Additional Profiling Results}
\label{appendix:profiling}

Tab.~\ref{table:per-layer-encoder-4heads} provides per-layer breakdowns of the FLOPs, MOPs, and arithmetic intensity for a hypothetical BERT model with only 4 attention heads  for sequence lengths of 128, 512, and~4096. 
Tab.~\ref{table:per-layer-decoder} provides per-layer breakdowns of the FLOPs, MOPs, and arithmetic intensity for GPT-2 for sequence lengths of 128, 512, and~4096. 

\begin{table*}[!h]
\caption{
Per-Layer FLOPs, memory operations (MOPs), and arithmetic intensity for the hypothetical BERT-Base encoder with 4 attention heads and with sequence lengths of 128, 512, and 4096 tokens.
The number of FLOPs consumed by each operation for each sequence length is similar to the BERT-Base encoder with 12 attention heads (Tab.~\ref{table:per-layer-encoder-12heads}).
However, the number of MOPs consumed by the activation-to-activation matmuls are significantly lower for each sequence length relative to the BERT-Base encoder with 12 attention heads. 
This leads to greater arithmetic intensity in the activation-to-activation matmuls and for end-to-end inference when using 4 attention heads rather than 12 attention heads.
}
\begin{center}
\small{
\begin{tabular}{c|c|ccccc}
\toprule
Sequence Length & Operator & FLOPs ($\times$ $10^9$) & \% of total FLOPs & MOPs ($\times$ $10^9$) & \% of total MOPs & Arithmetic Intensity \\
\midrule
\multirow{5}{*}{128} & MHA (projections) & 7.25 & 0.32 & 0.04 & 0.28 & 192.00 \\
& MHA (act-to-act matmuls) & 0.60 & 0.03 & 0.01 & 0.047 & 95.69 \\
& FFN (projections) & 14.47 & 0.65 & 0.07 & 0.51 & 211.86 \\
& Other & 0.07 & 0.003 & 0.02 & 0.16 & 3.30 \\
& Total & 22.42 & 1 & 0.13 & 1 & 167.14 \\
\midrule
\multirow{5}{*}{512} & MHA (projections) & 28.99 & 0.30 & 0.07 & 0.21 & 438.86 \\
& MHA (act-to-act matmuls) & 9.65 & 0.10 & 0.04 & 0.14 & 219.04 \\
& FFN (projections) & 57.98 & 0.60 & 0.10 & 0.32 & 558.54 \\
& Other & 0.32 & 0.003 & 0.10 & 0.33 & 3.07 \\
& Total & 96.94 & 1 & 0.32 & 1 & 303.59 \\
\midrule
\multirow{5}{*}{4096} & MHA (projections) & 231.93 & 0.18 & 0.33 & 0.07 & 702.17 \\
& MHA (act-to-act matmuls)& 617.63 & 0.47 & 1.76 & 0.37 & 350.61 \\
& FFN (projections) & 463.86 & 0.35 & 0.43 & 0.09 & 1068.52 \\
& Other & 5.41 & 0.004 & 2.25 & 0.47 & 2.40 \\
& Total & 1318.83 & 1 & 4.78 & 1 & 276.00 \\
\bottomrule
\end{tabular}
}
\end{center}
\label{table:per-layer-encoder-4heads}
\end{table*}

\begin{table*}[!h]
\caption{
Per-Layer FLOPs, MOPs, and arithmetic intensity for the GPT-2 decoder with sequence lengths of 128, 512, and 4096 tokens.
The number of FLOPs is similar to the BERT-Base encoder (provided in Tab.~\ref{table:per-layer-encoder-12heads}).
However, the number of MOPs is much larger than in the BERT-Base encoder. 
This results in lower arithmetic intensity in the GPT-2 decoder than the BERT-Base encoder.
}
\begin{center}
\small{
\begin{tabular}{c|c|ccccc}
\toprule
Sequence Length & Operator & FLOPs ($\times$ $10^9$) & \% of total FLOPs & MOPs ($\times$ $10^9$) & \% of total MOPs & Arithmetic Intensity \\
\midrule
\multirow{5}{*}{128} & MHA (projections) & 7.25 & 33 & 3.63 & 33 & 2.00 \\
& MHA (act-to-act matmuls) & 30 & 0.01 & 0.16 & 1 & 1.92 \\
& FFN (projections) & 14.50 & 66 & 7.26 & 66 & 2.00 \\
& Other & 0.07 & 0.3 & 0.03 & 0.3 & 2.58 \\
& Total & 22.12 & 100 & 11.08 & 100 & 2.0 \\
\midrule
\multirow{5}{*}{512} & MHA (projections) & 28.99 & 32 & 14.53 & 32 & 2.00 \\
& MHA (act-to-act matmuls) & 4.83 & 5 & 2.45 & 5 & 2.00 \\
& FFN (projections) & 57.98 & 63 & 29.04 & 63 & 2.00 \\
& Other & 0.35 & 0.4 & 0.14 & 0.3 & 2.47 \\
& Total & 92.15 & 100 & 46.17 & 100 & 2.00 \\
\midrule
\multirow{5}{*}{4096} & MHA (projections) & 231.93 & 23 & 116.27 & 23 & 2.00 \\
& MHA (act-to-act matmuls) & 309.24 & 31 & 155.98 & 31 & 1.98 \\
& FFN (projections) & 463.86 & 46 & 232.31 & 46 & 2.0 \\
& Other & 7.02 & 0.7 & 3.25 & 0.6 & 2.16 \\
& Total & 1012.04 & 100 & 507.80 & 100 & 1.99 \\
\bottomrule
\end{tabular}
}
\end{center}
\label{table:per-layer-decoder}
\end{table*}

\subsection{Additional ResNet-50 Workload Analysis}
\label{appendix:resnet50}

Tab.~\ref{table:resnet50-convolution-flops} provides detailed analysis of the FLOPs, MOPs, and arithmetic intensity for several convolutional layers in ResNet-50. 

\begin{table*}[!h]
\caption{
FLOPs, memory operations (MOPs), and arithmetic intensity for different convolutional layers in ResNet-50.
}
\begin{center}
\small{
\begin{tabular}{c|c|c|ccc}
\toprule
Kernel Size & Output Channels & Output Size & FLOPs ($\times$ $10^9$) & MOPs ($\times$ $10^9$) & Arithmetic Intensity \\
\midrule
1$\times$1 & 64 & 56$\times$56 & 0.31 & 0.0031 & 100.76 \\
3$\times$3 & 64 & 56$\times$56 & 0.69 & 0.0013 & 527.55 \\
1$\times$1 & 256 & 56$\times$56 & 0.31 & 0.0031 & 100.76 \\
1$\times$1 & 128 & 28$\times$28 & 0.31 & 0.0017 & 181.14 \\
3$\times$3 & 128 & 28$\times$28 & 0.92 & 0.0014 & 664.09 \\
1$\times$1 & 512 & 28$\times$28 & 0.41 & 0.0023 & 181.14 \\
1$\times$1 & 256 & 14$\times$14 & 0.51 & 0.0026 & 200.30 \\
3$\times$3 & 256 & 14$\times$14 & 1.39 & 0.0041 & 335.00 \\
1$\times$1 & 1024 & 14$\times$14 & 0.62 & 0.0031 & 200.30 \\
1$\times$1 & 512 & 7$\times$7 & 0.21 & 0.0023 & 87.53 \\
3$\times$3 & 512 & 7$\times$7 & 0.69 & 0.0072 & 95.96 \\
1$\times$1 & 2048 & 7$\times$7 & 0.31 & 0.0035 & 87.53 \\
\bottomrule
\end{tabular}
}
\end{center}
\label{table:resnet50-convolution-flops}
\end{table*}

\subsection{Additional Analytical Modeling Results}
\label{appendix:modeling}

\begin{figure}[!h]
  \centering
  \includegraphics[width=0.49\columnwidth]{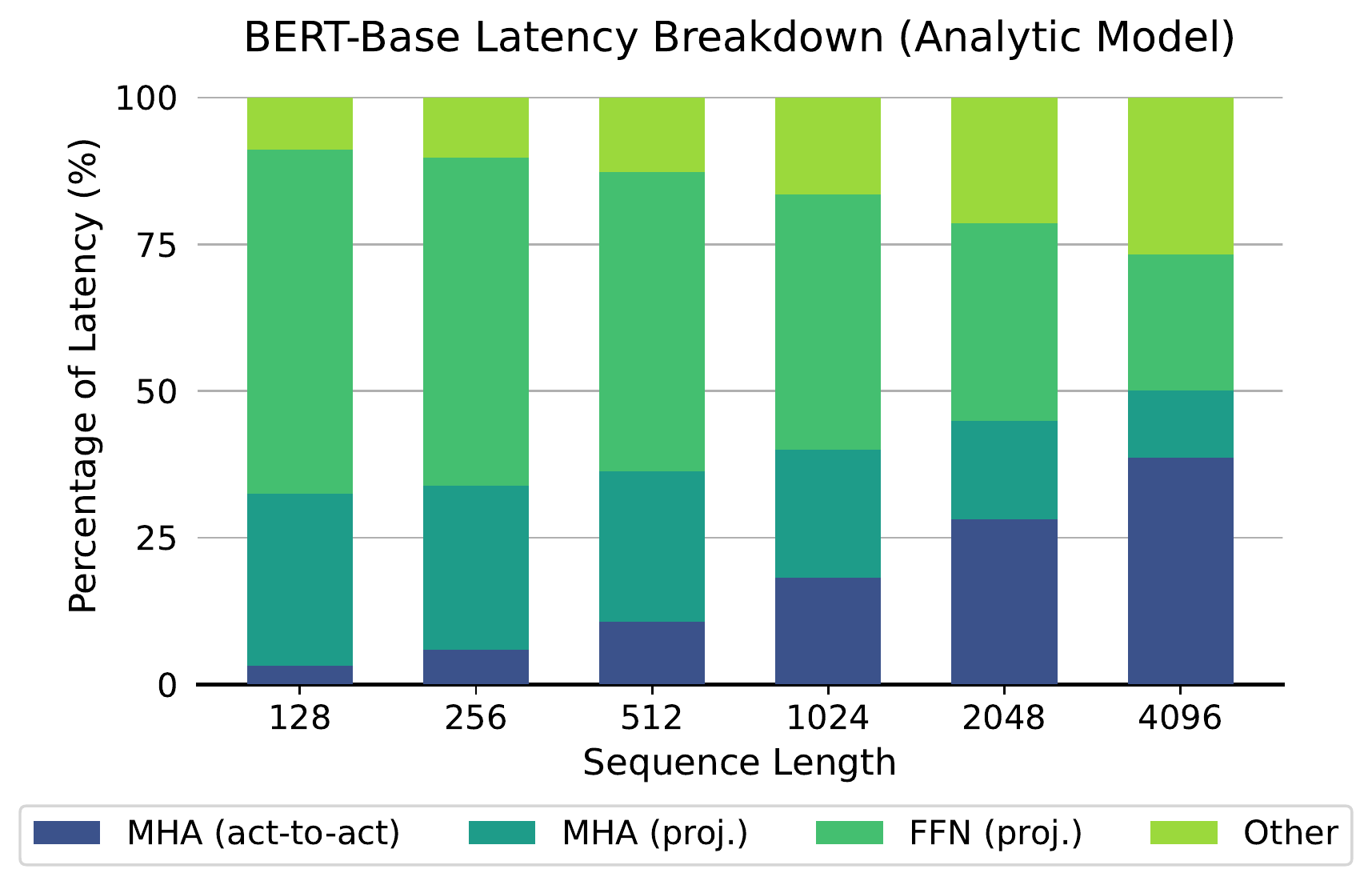}
  \caption{Plot of the computation breakdown in the BERT-Base encoder versus sequence length using our analytical model. Proj. and act-to-act refer to the projection operation (i.e., activation-to-weight matmul) and the activation-to-activation matmul, respectively. Other refers to the non-matmul operations.}
      \label{fig:breakdown-bert-base}
\end{figure}

\begin{figure}[!h]
  \centering
  \includegraphics[width=0.49\columnwidth]{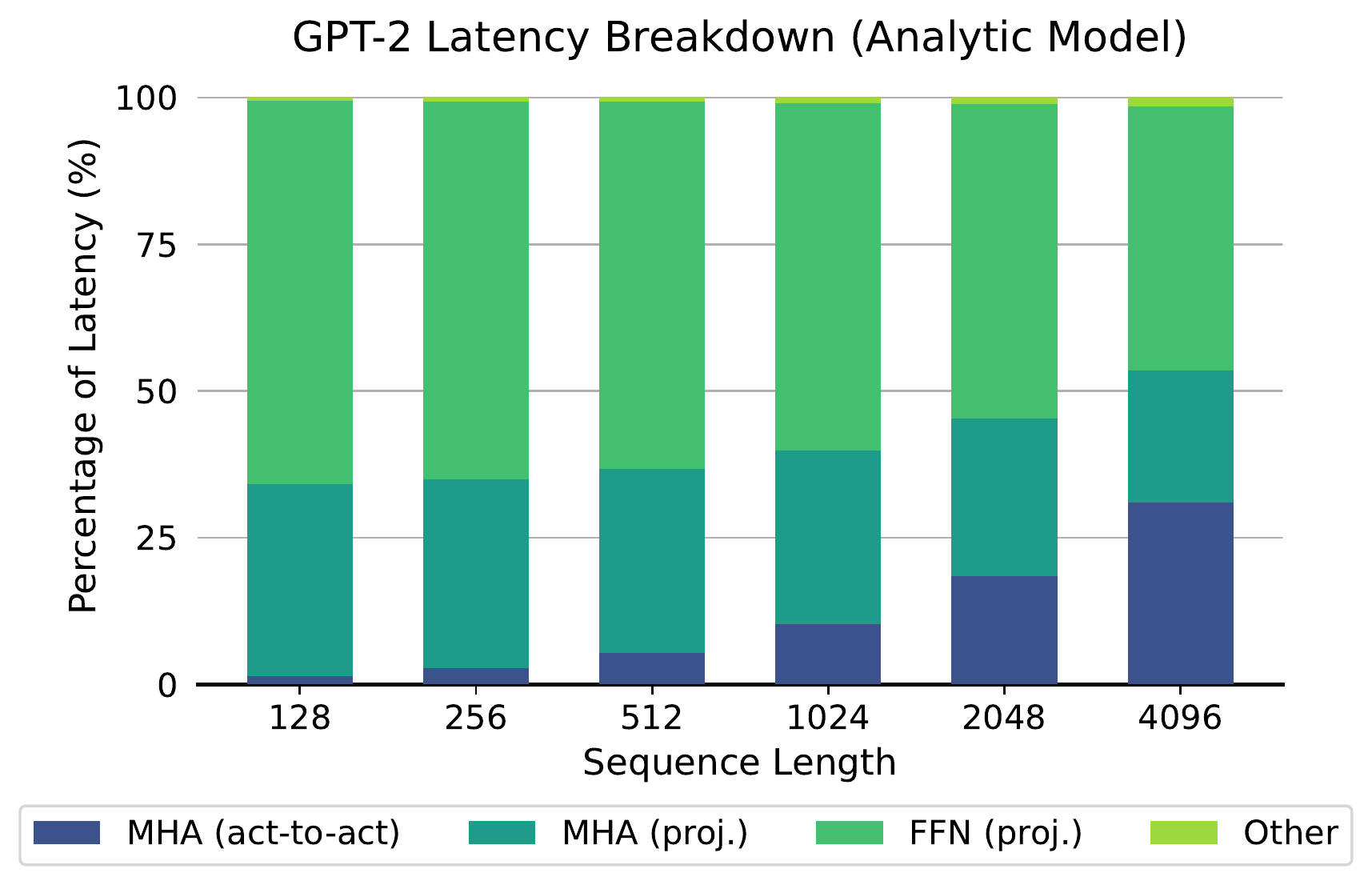}
  \caption{Plot of the computation breakdown in the GPT-2 decoder versus sequence length using our analytical model. Proj. and act-to-act refer to the projection operation (i.e., activation-to-weight matmul) and the activation-to-activation matmul, respectively. Other refers to the non-matmul operations.}
      \label{fig:breakdown-gpt2}
\end{figure}

\begin{figure}[!h]
  \centering
  \includegraphics[width=0.49\columnwidth]{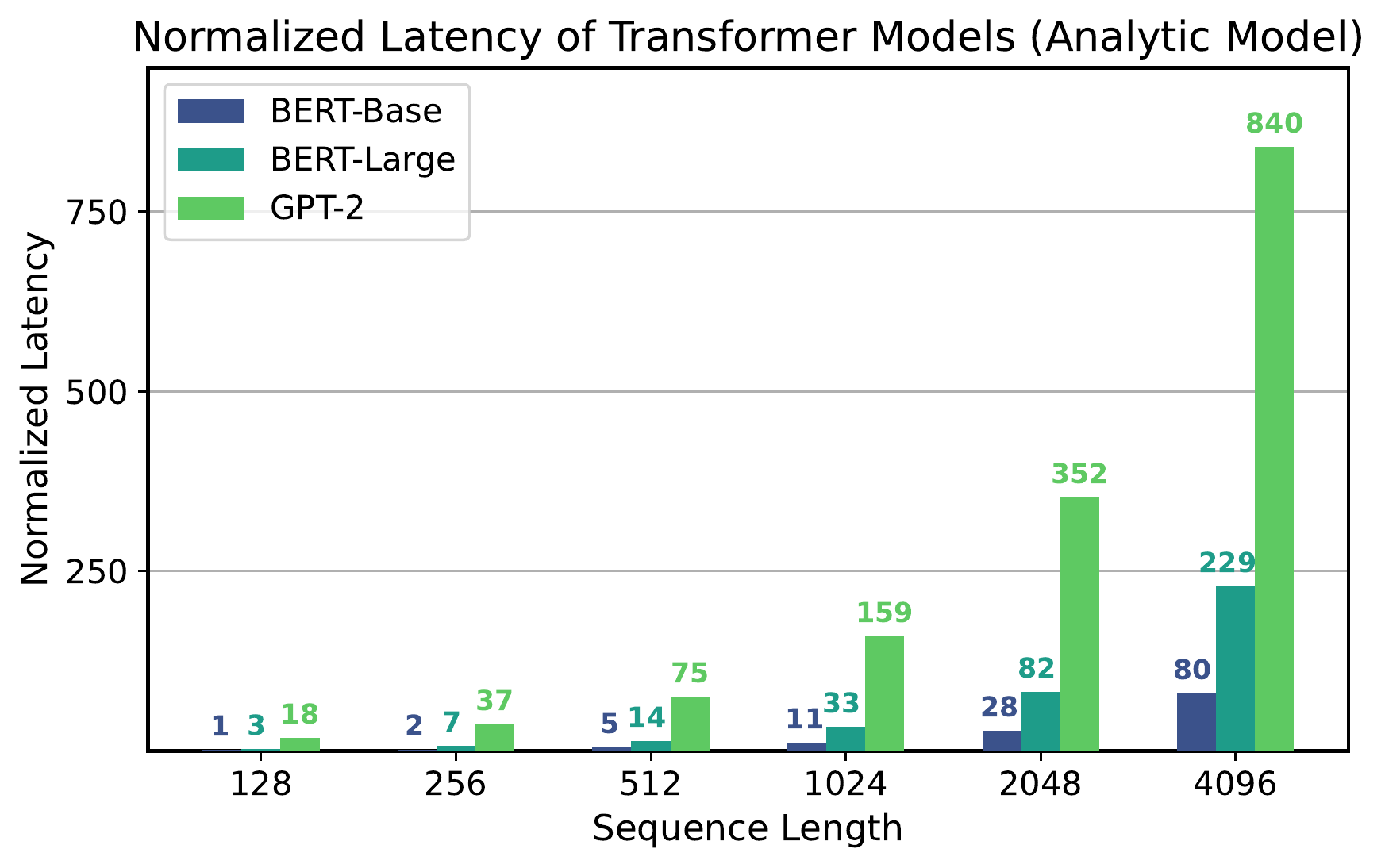}
  \caption{Plot of the runtime latency of the BERT-Base and BERT-Large encoders and the GPT-2 decoder versus sequence length using our analytical model, normalized to the runtime of BERT-Base with a sequence length of 128.}
      \label{fig:normalizedruntime}
\end{figure}

\paragraph{\textbf{Latency Breakdown and End-to-end Runtime.}}
We modeled the performance breakdown of the BERT-Base and BERT-Large encoders, under the assumption of square tiling for all matrix operations, and no operation fusion (i.e., each operation required inputs to be read from external memory and outputs to be flushed out).
We also modeled the performance breakdown of the GPT-2 decoder under the same assumption of no operation fusion. 
The latency breakdowns for BERT-Base and GPT-2 for different sequence lengths are provided in Fig.~\ref{fig:breakdown-bert-base} and \ref{fig:breakdown-gpt2}, respectively. Fig.~\ref{fig:normalizedruntime}  shows the runtime latency of BERT-Base, BERT-Large, and GPT-2, normalized to the runtime latency of BERT-Base with a sequence length of 128. 
The runtime scaling and breakdowns from the analytical model were similar to the trends observed when profiling inference on the CPU in Section \ref{sec:profiling}. 
However, for a fixed sequence length, the MHA computation takes up a greater proportion of the computation on the CPU. 
Note that the analytical model was designed assuming a hardware architecture that was different from the CPU architecture, and the relative breakdown between different operations would not necessarily be the same for different hardware platforms.

    \paragraph{\textbf{Comparison with ResNet-50.}}
We also modeled ResNet-50 to provide a baseline for our analysis. 
We first compared the runtime between BERT-Base and ResNet-50 without operation fusing. 
We found that the runtime of ResNet-50 was 1.28 times faster than the runtime of BERT-Base with a sequence length of 128.
As outlined in Section \ref{sec:workload-analysis}, ResNet-50 contains 3.07 times fewer FLOPs and 1.28 times fewer MOPs than BERT-Base with a sequence length of 128. 
This shows how differences in FLOPs between two DNN models don't necessarily represent the relationship between the runtime latency of these two models.

Additionally, we observe that, without operation fusion, nonlinear operations consume 32.4\% of overall runtime latency, even though convolutions consume 99.3\% of FLOPs, as outlined in Tab.~\ref{table:resnet50-flops}. 
We therefore assessed the latency of fusing the nonlinear operations with the prior convolutional layers. 
We found that a 1.32 times speedup can be obtained by fusing BatchNorm and ReLU with the prior convolutional layers, demonstrating how in the case of ResNet-50, the latencies from the nonlinear operations can be significantly reduced by fusing. 
However, in Sec.~\ref{subsec:scheduling_complexity_nonlinear}, we demonstrate how operation fusion can be non-trivial for the Transformer architecture.
For Transformers, fusing LayerNorm or Softmax with the prior matmuls may require changes in tiling dimension changes which can actually increase runtime latency.

\subsection{Acronyms and Abbreviations}
\label{appendix:decoder}
Tab.~\ref{table:acronyms} summarizes several acronyms/abbreviations used throughout this papers and their full names.

\begin{table*}[!h]
\caption{
Full names of the acronyms and abbreviations used in this paper.
}
\begin{center}
\small{
\begin{tabular}{c|c}
\toprule
Abbreviation & Full name \\
\midrule
act-to-act & activation-to-activation \\
ALU & arithmetic logic unit \\
attn. & attention \\
CDF & cumulative distribution function \\
CNN & convolutional neural network \\
CV & computer vision \\
DNN & deep neural network \\
EDP & energy-delay product \\
FFN & feed-forward network \\
FLOPs & floating-point operations \\
MAC & multiply-accumulate \\
matmul & matrix multiplication \\
MHA & multi-head attention \\
MOPs & memory operations \\
NAS & neural architecture search \\
NLP & natural language processing \\
PE & processing element \\
RF & register file \\
RL & reinforcement learning \\
RTL & register transfer logic \\
\bottomrule
\end{tabular}
}
\end{center}
\label{table:acronyms}
\end{table*}

\end{document}